\documentclass[numbers]{assets/matterlab}
\usepackage{microtype}
\usepackage{graphicx}
\usepackage{booktabs}
\usepackage{tabularx}
\usepackage{float}
\usepackage{xurl}
\usepackage{hyperref}
\usepackage{titlesec}

\usepackage{amsmath}
\usepackage{amssymb}
\usepackage{mathtools}
\usepackage{amsthm}
\usepackage{bm}

\usepackage[noabbrev,nameinlink]{cleveref}

\usepackage{amsmath,amsfonts,bm}

\def\eqref#1{equation~\ref{#1}}

\def\1{\bm{1}}

\newcommand{\valid}{\mathcal{D_{\mathrm{valid}}}}

\DeclareMathAlphabet{\mathsfit}{\encodingdefault}{\sfdefault}{m}{sl}
\SetMathAlphabet{\mathsfit}{bold}{\encodingdefault}{\sfdefault}{bx}{n}

\usepackage{orcidlink}
\usepackage{comment}
\usepackage[version=4]{mhchem}
\usepackage{longtable}
\usepackage{array}
\renewcommand{\arraystretch}{1.7}
\usepackage{multirow}           
\usepackage{amsfonts}           
\usepackage{nicefrac}
\usepackage{duckuments}         
\usepackage{thmtools,thm-restate}
\usepackage{enumitem}
\usepackage{xcolor,colortbl}
\usepackage{tikz}
\usetikzlibrary{arrows.meta,positioning,fit,backgrounds}
\usepackage{tikz-cd}
\usepackage{caption}
\usepackage{subcaption}
\usepackage{listings}
\usepackage{gensymb}
\usepackage{textcomp} 
\usepackage[inkscapelatex=false]{svg}

\providecommand{\code}[1]{\texttt{\small #1}}

\newcommand{\addressCHEM}{Department of Chemistry, University of Toronto,  80 St. George St., Toronto, ON M5S 3H6, Canada}
\newcommand{\addressAC}{Acceleration Consortium, 700 University Ave., Toronto, ON M7A 2S4, Canada}
\newcommand{\addressCS}{Department of Computer Science, University of Toronto, 40 St George St., Toronto, ON M5S 2E4, Canada}
\newcommand{\addressVECTOR}{Vector Institute for Artificial Intelligence, W1140-108 College St., Schwartz Reisman Innovation Campus, Toronto, ON M5G 0C6, Canada}
\newcommand{\addressMSE}{Department of Materials Science \& Engineering, University of Toronto, 184 College St., Toronto, ON M5S 3E4, Canada}
\newcommand{\addressCHEMENG}{Department of Chemical Engineering \& Applied Chemistry, University of Toronto, 200 College St., Toronto, ON M5S 3E5, Canada}
\newcommand{\addressMEDICALSCI}{Institute of Medical Science, 1 King's College Circle, Medical Sciences Building, Room 2374, Toronto, ON M5S 1A8, Canada}
\newcommand{\addressCIFAR}{Canadian Institute for Advanced Research (CIFAR), 661 University Ave., Toronto,
ON M5G 1M1, Canada}
\newcommand{\addressNVIDIA}{NVIDIA, 431 King St W \#6th, Toronto, ON M5V 1K4, Canada}

\usepackage[nolist]{acronym}
\usepackage{titletoc}
\usepackage{cinzel}
\usepackage{fontawesome5}
\usepackage{placeins}
\usepackage{subcaption}
\newcommand{\opt}{{\cinzel Óptima}}
\newcommand{\optima}{{\cinzel La Agente Óptima}}
\newcommand{\elagenteG}{{\cinzel El Agente Gr\'afico}}
\newcommand{\grafico}{{\cinzel Gr\'afico}}
\newcommand{\estructural}{{\cinzel El Agente Estructural}}
\newcommand{\archspecialist}{\texttt{Specialist-script}}
\newcommand{\archmainscript}{\texttt{Main-script}}
\newcommand{\archtoolloop}{\texttt{Tool-loop}}
\newcommand{\archlocalbo}{\texttt{Local-BO}}
\newcommand{\benchackley}{Ackley 6D}
\newcommand{\bencharylation}{Shields arylation}

\usepackage{cleveref}
\crefname{section}{Sec.}{Secs.}
\crefname{figure}{Fig.}{Figs.}
\crefname{table}{Tab.}{Tabs.}
\crefname{equation}{Eq.}{Eqs.}
\creflabelformat{equation}{#2\textup{#1}#3}
\renewcommand\affiliationformat[2][]{{$^{#1}$#2}}
\renewcommand\contributionformat[2][]{{$^{#1}$#2}}
\patchcmd{\mymaketitle}
  {\affiliationlist\par}
  {{\footnotesize\affiliationlist\par}}
  {}{\GenericError{}{Could not patch affiliation list formatting}{}{}}
\patchcmd{\mymaketitle}
  {\contributionlist\par}
  {{\footnotesize\contributionlist\par}}
  {}{\GenericError{}{Could not patch contribution list formatting}{}{}}

\title{La Agente Óptima: Towards Agentic Self-Driving Laboratories}

\author[1,7,\dagger]{Marcel~M\"uller}
\author[1,7,\dagger]{Jiaru~Bai}
\author[6,\dagger]{Willi~Gottstein}
\author[2,\ddagger]{Abhijoy~Mandal}
\author[10,\ddagger]{Mohammad~Nazeri}
\author[11,\ddagger]{Elia~Savino}
\author[2]{Yanlin~Fang}
\author[2]{Sujoy~Das}
\author[6,12]{Sergio~Pablo~Garc\'ia~Carrillo}
\author[1,7,13]{Yeonghun~Kang}
\author[1]{Juan~B.~P\'erez-S\'anchez}
\author[11]{Simone~Pilon}
\author[14]{Martin~Fitzner}
\author[11]{Timothy~Noël}
\author[4,6,10]{Frank~Gu}
\author[1,2,6,*]{Varinia~Bernales}
\author[1,2,3,4,5,6,7,8,9,*]{Al\'an~Aspuru-Guzik}

\affiliation[1]{\addressCHEM}
\affiliation[2]{\addressCS}
\affiliation[3]{\addressMSE}
\affiliation[4]{\addressCHEMENG}
\affiliation[5]{\addressMEDICALSCI}
\affiliation[6]{\addressAC}
\affiliation[7]{\addressVECTOR}
\affiliation[8]{\addressCIFAR}
\affiliation[9]{\addressNVIDIA}
\affiliation[10]{Institute of Biomedical Engineering, University of Toronto, 164 College St, Toronto, Canada}
\affiliation[11]{Flow Chemistry Group, van 't Hoff Institute for Molecular Sciences (HIMS), University of Amsterdam, Science Park 904, 1098 XH Amsterdam, Netherlands.}
\affiliation[12]{Instituto de Micro y Nanotecnolog\'ia, IMN-CNM, CSIC (CEI UAM+CSIC), Isaac Newton, 8, Tres Cantos, Madrid, Spain, 28760}
\affiliation[13]{Department of Chemistry, Sungkyunkwan University, 2066 Seobu-ro, Suwon-si, Gyeonggi, Republic of Korea, 16419}
\affiliation[14]{Merck KGaA, Frankfurter Str. 250, 64293 Darmstadt, Germany}

\contribution[\dagger]{These authors contributed equally}
\contribution[\ddagger]{these authors also contributed equally.}

\abstract{
Self-driving laboratories (SDLs) combine automated experimentation with adaptive decision-making to accelerate scientific discovery.
Their operation nevertheless often depends on human specialists who translate scientific objectives into executable closed-loop campaigns.
Specialists adjust them as data and operating conditions change.
Here, we present \optima{}, an agentic framework that constructs and supervises Bayesian optimization campaigns across computational and experimental systems while maintaining a persistent optimization state.
By separating large language model (LLM) reasoning from executed campaigns, \opt{} runs repetitive optimization loops consistently, returns control to the agent only when progress requires interpretation or campaign revision, and keeps every decision auditable.
We evaluate \opt{} across ablation studies, five digital discovery tasks, and two physical platforms.
Throughout, \opt{} maintained executable campaigns as both the scientific problem and execution environment evolved.
In a closed-loop contact angle optimization campaign, \opt{} identified and corrected a mid-run measurement failure, bringing the contact angle from 71.4\textdegree{} to 67.8\textdegree{}, just outside the 65\textdegree{} $\pm$ 1\textdegree{} target.
From this result, \opt{} correctly inferred that the target was likely unattainable with the available reagents and recommended changing the formulation.
In a five-day multi-objective flow-chemistry campaign, \opt{} increased the yield from 30\% to 59\% over 23 experiments.
Despite substantial inference costs, it cost less and used substantially less starting material than a human-directed campaign, while selecting a more mass-efficient operating point.
These results show that LLM-based agents can make rigorous, long-running optimization campaigns accessible to domain scientists without specialist setup, expanding the scope of SDLs.

}

\date{\today}
\correspondence{\email{varinia@bernales.org} and \email{alan@aspuru.com}}

\begin{document}

\acrodef{dft}[DFT]{density functional theory}
\acrodef{scf}[SCF]{self-consistent field}
\acrodef{dftb}[DFTB]{density functional tight-binding}
\acrodef{td-dft}[TD-DFT]{time-dependent density functional theory}
\acrodef{mlip}[MLIP]{machine-learned interatomic potential}
\acrodef{wft}[WFT]{wave function theory}
\acrodef{hf}[HF]{Hartree-Fock}
\acrodef{qcg}[QCG]{quantum cluster growth}
\acrodef{td}[TD]{time-dependent}
\acrodef{tddft}[TDDFT]{time-dependent density functional theory}
\acrodef{md}[MD]{molecular dynamics}
\acrodef{rmsd}[RMSD]{root mean square deviation}
\acrodef{qc}[QC]{quantum chemistry}
\acrodef{mof}[MOF]{metal-organic framework}
\acrodef{kg}[KG]{knowledge graph}
\acrodef{ogm}[OGM]{object graph mapper}
\acrodef{iri}[IRI]{internationalized resource identifier}
\acrodef{sdl}[SDL]{self-driving laboratory}
\acrodefplural{sdl}[SDLs]{self-driving laboratories}
\acrodef{doe}[DoE]{design of experiment}
\acrodef{osl}[OSL]{organic solid-state laser}
\acrodef{sds}[SDS]{sodium dodecyl sulfate}
\acrodef{rme}[RME]{reaction mass efficiency}
\acrodefplural{rme}[RMEs]{reaction mass efficiencies}
\acrodef{tfaa}[TFAA]{trifluoroacetic anhydride}
\acrodef{llm}[LLM]{large language model}
\acrodef{coala}[CoALA]{cognitive architectures for language agents}
\acrodef{ai}[AI]{artificial intelligence}
\acrodef{ttl}[TTL]{time-to-live}
\acrodef{bo}[BO]{Bayesian optimization}
\acrodef{gp}[GP]{Gaussian process}
\acrodefplural{gp}[GPs]{Gaussian processes}
\acrodef{a2a}[A2A]{agent-to-agent}
\acrodef{mcp}[MCP]{model context protocol}
\acrodef{api}[API]{application programming interface}
\acrodef{ui}[UI]{user interface}
\acrodef{mbsf}[MBSF]{mean best-so-far}
\acrodef{si}[SI]{Supporting Information}

\maketitle

\newpage
\section{Introduction}

Discovering a better catalyst, ligand, or emitter means searching a design space that is usually far larger than anyone could test exhaustively~\cite{zahrtPredictionHigherselectivityCatalysts2019,vogiatzisComputationalApproachMolecular2018,sanchez-lengelingInverseMolecularDesign2018}.
Navigating such a space effectively means deciding what to test next based on everything learned so far.
\Acp{sdl} \cite{tomSelfDrivingLaboratoriesChemistry2024} address this by closing the loop: they combine automated laboratory infrastructure, such as robotic arms or automated liquid-dispensing platforms, with automated experiment planning, so that each round of testing informs the decision of the next.
Among the methods available for this planning step, \ac{bo} has become the algorithm of choice for many parameter- and reaction-optimization tasks in chemistry~\cite{mockusBayesianApproachGlobal1989,balandatBoTorchFrameworkEfficient2020, Garnett2023,durholtBoFireBayesianOptimization2024,fitznerBayBEBayesianBack2025,hickmanAtlasBrainSelfdriving2025,desimpelBayesianOptimizationChemical2026}, enabling sequential, model-guided, adaptive decision-making that uses prior observations to select the next most informative or promising experiment.
This makes it particularly well suited to low-data, expensive-to-evaluate black-box optimization problems~\cite{frazierTutorialBayesianOptimization2018}.

In practice, however, deploying \ac{bo} remains difficult for many scientists.
Users typically rely either on bespoke \acp{ui} developed for a particular organization or experimental setup~\cite{torresMultiObjectiveActiveLearning2022}, or on direct access to a \ac{bo} package through its internal \acp{api}~\cite{fitznerBayBEBayesianBack2025}, embedded in a custom setup for a specific problem.
Dedicated optimization libraries can reduce the implementation burden.
General-purpose frameworks such as BoTorch and Ax support mixed parameter spaces, constraints, and parallel or asynchronous optimization, while Atlas provides capabilities specifically oriented toward experimental science and \acp{sdl}~\cite{balandatBoTorchFrameworkEfficient2020,olsonAxPlatformAdaptive2025,hickmanAtlasBrainSelfdriving2025}.
Nevertheless, connecting an optimizer to the system that evaluates each candidate, configuring and maintaining the resulting campaign, and interpreting its decision trajectory still typically require programming and \ac{bo} expertise that domain scientists may not have.
Although automated experimental discovery has already been demonstrated across a range of applications~\cite{strieth-kalthoffDelocalizedAsynchronousClosedloop2024,gongoraBayesianExperimentalAutonomous2020,burgerMobileRoboticChemist2020,bai2024dynamic}, making such workflows part of everyday scientific practice still depends heavily on specialist support because of these practical challenges.

At the same time, increasingly autonomous scientific workflows~\cite{gottweisAcceleratingScientificDiscovery2026} create a complementary need for scalable decision-making systems in which human intervention does not become the rate-limiting step as the scale of such campaigns increases.
A recent closed-loop catalysis campaign provides an example in which human reasoning accounted for 34.7 person-hours across 160 experiments, compared with about 1 hour of agent reasoning at twice the experimental throughput~\cite{cooperLargeLanguageModels2026}.
Ideally, such methods should therefore both empower experimentalists through accessible, general-purpose interfaces and support reliable semi-autonomous operation at scales that would be impractical to coordinate manually.

\Acp{llm} have shown substantial promise across scientific discovery~\cite{songEvaluatingLargeLanguage2025}.
When equipped with access to scientific tools~\cite{Ramos2025reviewLLMAgent}, \ac{llm}-based agents offer one way to reduce the aforementioned integration burden by operating those tools through their software interfaces.
Many reported applications address \textit{in-silico} research tasks, including quantum chemistry calculations~\cite{zouAgenteAutonomousAgent2025,Wang2025dreams,aitomia,Pham2026ChemGraph,perez-sanchezAgenteQunturResearch2026,baiAgenteGraficoStructured2026,kumarAgenteSolidoNew2026}, molecular structure investigations~\cite{chemcrow,Ding2025SciToolAgent,choiAgenteEstructuralArtificially2026}\footnote{
    As another example, Anthropic reported that Claude could perform \href{https://www.anthropic.com/research/making-claude-a-chemist}{NMR prediction and structure elucidation} without external scientific tools.
}, hypothesis generation~\cite{gottweisAcceleratingScientificDiscovery2026,Ghareeb2026multiagent}, as well as biochemistry tasks and molecular inverse design \cite{yousefiBioChemAIgentAIdrivenProtein2025,ClaudeScienceAI,worakulSolitariusAgenticArchitecture}.
Wet-lab studies have extended this approach to physical experimentation: Coscientist, a GPT-4-driven autonomous agent combining web search, documentation retrieval, code execution, and laboratory automation, designed, planned, and executed chemical experiments~\cite{boikoAutonomousChemicalResearch2023}, while \citet{Smith2026openaiSDL} coupled GPT-5 to a cloud laboratory for six rounds of cell-free protein synthesis optimization.
Although the latter demonstrates that an agent can sustain an optimization campaign over multiple experimental rounds, such demonstrations remain tied to execution environments tailored to particular tasks and platforms~\cite{tomSelfDrivingLaboratoriesChemistry2024}.

How agent reasoning should interact with dedicated optimization methods is a separate question, and one on which the current literature explores different options:
Several approaches build directly on suggestions generated by \acp{llm}~\cite{cooperLargeLanguageModels2026,macknightPretrainedKnowledgeElevates2025}.
Such approaches can benefit from the substantial chemical knowledge encoded during pretraining and have shown strong performance for reaction optimization, particularly in categorical search spaces that resemble well-represented chemistry~\cite{macknightPretrainedKnowledgeElevates2025}.
However, this advantage depends on the relevance of that prior knowledge: when the chemistry is under-studied, the search space is weakly represented in the literature, or optimization requires learning primarily from newly acquired experimental feedback, dedicated \ac{bo} methods provide a complementary and often preferable strategy~\cite{macknightPretrainedKnowledgeElevates2025}.
The same holds for the reliability of the search itself: in a 25-dimensional closed-loop catalysis campaign, an \ac{llm} concentrated on a narrow subset of formulations and made silent reasoning errors that were caught only by human inspection, whereas, within the same 160-experiment budget, a domain-expert team explored more diverse formulations and identified a catalyst that was more active on average than those found in two independent \ac{llm} runs~\cite{cooperLargeLanguageModels2026}.
Taken together, these studies suggest that \acp{llm} and \ac{bo} are best viewed as complementary components:
agent reasoning can contribute chemical knowledge, formulate objectives, and interact flexibly with experimental systems, whereas dedicated \ac{bo} provides a principled mechanism for data-driven exploration and optimization when prior knowledge is sparse, unreliable, or insufficient.
This complementarity is also supported by a recent study by~\citet{guptaLLMsBayesianOptimization2025}, who found that replacing measured outcomes with randomly permuted ones left agent performance essentially unchanged, indicating that candidate selection was driven by pretrained priors rather than by an updated posterior, and that combining these priors with an explicit acquisition step recovered the lost performance.

Other approaches have sought to enhance \ac{bo}'s performance with \acp{llm}.
For example,~\citet{mottafeghAdaptiveHumanintheLoopOptimization2026} used \ac{llm}-based agents to incorporate human knowledge-driven biases into structured priors for \acp{gp}.
Similarly,~\citet{cisseCanWeAutomate2026} showed that \ac{llm}/\ac{bo} hybrids outperform \ac{bo} alone in closed-loop optimization, in particular when early exploration is warm-started by \ac{llm}-generated hypotheses, although purely \ac{llm}-driven optimization matched or surpassed the hybrid in some settings.
In CICERO,~\citet{ritchhartAgenticWorkflowEnables2026} demonstrated an agentic workflow in which \acp{llm} formulate selective-precipitation campaigns and can optionally invoke \ac{bo} for batched experimental refinement.
However, in this case, \ac{bo} remains embedded in a domain-specific separations workflow with a tightly coupled experimental interface, target definition, and objective formulation.
\citet{brunzemaAgenticBayesianOptimization2026} placed an \ac{llm} agent at the centre of the optimization loop, delegating surrogate modelling and acquisition optimization to a Bayesian backend that the agent can reconfigure at run time, and reported that on reaction-optimization benchmarks neither an \ac{llm}-only optimizer nor standard \ac{bo} alone matched this combination.
Their evaluation, however, is carried out entirely on synthetic and analytical benchmark functions.
Across these examples, the optimizer remains private to a single agent or workflow and is exercised either on benchmark objectives or within a single experimental domain.
This motivates a general architecture in which the optimizer is a shared and persistent component that supports both \ac{bo}-driven search and direct experimental proposals from the agent across heterogeneous computational and experimental platforms.

Here, we present \optima{}, an agentic framework for autonomously constructing and managing \ac{bo} campaigns across computational and experimental systems.
It separates agent reasoning from programmatic execution and a persistent, typed optimizer state.
A user states a scientific objective in natural language; the agent configures the campaign, runs it, diagnoses and repairs problems as they arise, and keeps every decision it makes auditable afterward.
After outlining the architecture of \opt{} and its building blocks, we compare it against alternative variants and benchmark its performance.
We then demonstrate how \opt{} can be coupled to electronic structure theory experiments using \elagenteG{} and experimental platforms more generally by agentizing the liquid formulation platform RAISE~\cite{nazeriRAISESelfdrivingLaboratory2026} and flow chemistry platform RoboChem-Flex~\cite{pilonFlexibleAffordableSelfdriving2026}.

\section{Implementation}
\label{sec:implementation}

\subsection{Agentic framework}
\begin{figure}
    \centering
    \includegraphics[width=\linewidth]{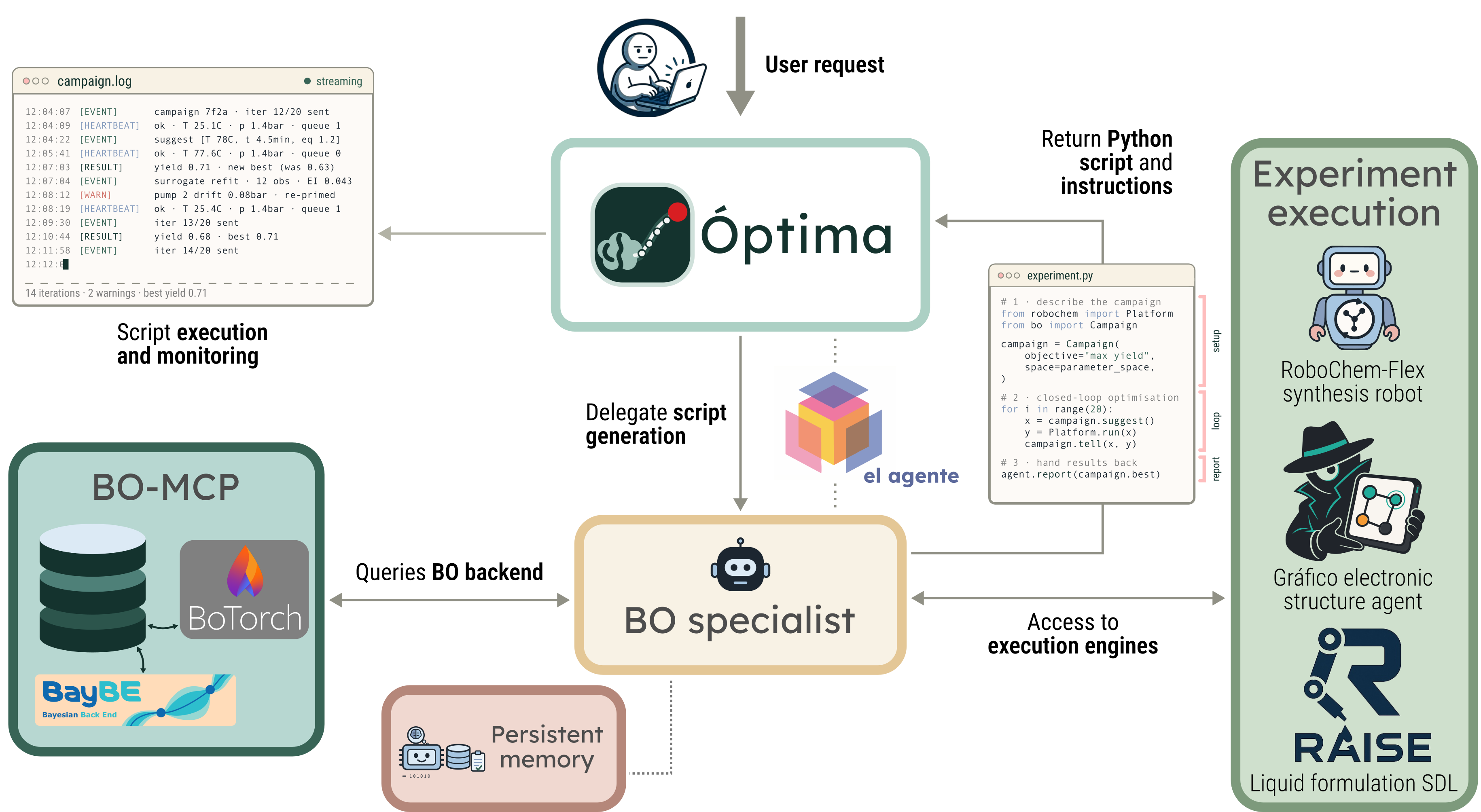}
    \caption{
        \textbf{Agentic architecture of \opt{}.}
        A user request is interpreted by \opt{}, which delegates campaign authoring to the \ac{bo} specialist.
        The specialist interacts with BO-MCP and the appropriate execution system to produce a Python campaign program and operating instructions.
        The specialist also maintains a persistent memory of reusable implementation notes acquired during campaign authoring and validation.
        The generated program is executed and monitored by \opt{}.
        BO-MCP stores optimizer state (see \cref{subsec:bo-mcp}), and computational or experimental platforms perform the evaluations.
        Campaign artifacts remain available in the shared room workspace for inspection by the agent and the user.
    }
    \label{fig:agentic-framework}
\end{figure}

\Cref{fig:agentic-framework} illustrates the overall agentic architecture.
\optima{} extends \elagenteG{}~\cite{baiAgenteGraficoStructured2026} by enabling it to construct and supervise \ac{bo} campaigns.
Each chat room provides a shared, filesystem-backed workspace that is accessible to the coordinating agent and its \ac{bo} specialist and exposed to the user through the frontend.
The coordinating agent interprets the scientific objective and campaign results, while the specialist implements and validates the campaign in a separate model context, keeping implementation details out of the coordinating agent's context.
The specialist maintains a persistent memory of implementation lessons acquired during campaign authoring and validation, allowing subsequent tasks to build on established solutions and design decisions.

For each task, the specialist writes a Python campaign program and its operating instructions to the shared workspace.
The program executes the repeated \ac{bo} loop with consistent settings, leaving the coordinating agent to interpret progress, revise the campaign, or propose experiments directly when scientific judgment is required.
The generated code and execution outputs remain inspectable by both agents and the user in the workspace, while staying outside the coordinating agent's context window until selectively inspected.

In the deployment used here, the coordinating agent and specialist run in the same application container, while BO-MCP runs as an independent service on the same Docker network (\cref{subsec:bo-mcp}).
This also separates the compute used for \ac{bo} campaign management, model fitting and acquisition optimization, from that used for agent reasoning, so each can be provisioned and scaled independently.
Evaluation systems retain their existing interfaces: electronic structure workflows are accessed through native \ac{llm} tool calls, RAISE through an MQTT bridge, and RoboChem-Flex through an authenticated HTTP \ac{api}.
The specialist adapts the campaign code to each evaluator, allowing existing systems to be integrated into \ac{bo} campaigns without requiring a dedicated agent protocol.

During long-running campaigns, \opt{} launches the campaign program as a monitored background task.
The monitoring system reads incremental process output and filters progress and heartbeat events.
The resulting steering messages enter an active \opt{} run at the next model invocation or wake the agent when it is idle, and the same events are mirrored to the frontend.
This event-driven path prompts \opt{} to notify the user and intervene when required without loading the complete polling log into the \ac{llm} context.
The same mechanism supervises computational workflows and experimental platforms.

\subsection{BO-MCP}
\label{subsec:bo-mcp}

BO-MCP exposes \ac{bo} as a service through \ac{mcp} and through a REST \ac{api} available to HTTP clients on an internal Docker network.
A thin tool layer forwards each request to a protocol-neutral operations layer that handles campaign lookup, state transitions, caching, and provenance.
The optimization engine sits behind a backend interface and is therefore replaceable:
BayBE~\cite{fitznerBayBEBayesianBack2025} serves as the default backend, a BoTorch-based engine is also available \cite{balandatBoTorchFrameworkEfficient2020}, and additional backends that implement the same interface can be registered as plugins and discovered through Python entry points.
For categorical chemical parameters, the BayBE backend supports both standard chemistry-aware representations, including Mordred and RDKit descriptors and Morgan fingerprints, and user-supplied numerical descriptors, as used several times throughout this work.
BO-MCP also maintains a queryable, per-campaign record of actions, providing a transparent account of what was done and when.
It also helps interpret the optimizer rather than treating it as a black box: the BoTorch backend reports kernel lengthscales for the model's computational input dimensions, whereas BayBE provides SHAP-based importance for either experimental parameters or their computational representations.
For each suggestion, the system also records how it was generated, including the acquisition function and value together with the model's predictions and uncertainty.
Together, these design choices make BO-MCP easy to use while still providing full transparency: an agent can run a full optimization campaign through simple tool calls or \ac{api} requests, while campaign actions and suggestions remain inspectable and explainable afterwards.

\subsection{PySCF execution graph}

\optima{} uses the PySCF execution graph from \elagenteG{} as an evaluator for quantum chemistry tasks that can be solved by PySCF.
The generated Python campaign code invokes the graph tool for each candidate with the same computational settings, ensuring that all evaluations follow a common protocol.
Typed nodes and admissible transitions coordinate the required calculations and validation steps, while scientific state passes directly between nodes via Python memory without entering the \ac{llm} context.
The graph returns a compact, structured result from which the campaign program derives objective values and submits valid observations to BO-MCP.
Independent candidates can be evaluated concurrently in isolated GPU processes, providing consistent execution and failure containment across the campaign.

\subsection{Agent-to-agent integration with \estructural{}}
\label{sec:a2a}

Molecular-structure operations are delegated over an \ac{a2a} interface to \estructural{}, an independently deployed agent that owns the generation and editing of molecular structures~\cite{choiAgenteEstructuralArtificially2026}.
Communication follows the \ac{a2a} protocol via HTTP: a request is submitted as a task, and the client polls that task until it reaches a terminal state.
As with BO-MCP, \estructural{} is reachable in two ways that share one endpoint: \optima{} calls it as a tool for single interactive requests, whereas the \ac{bo} specialist additionally constructs the same client directly inside generated campaign scripts, so that long optimization loops do not depend on \ac{llm} tool calls.
Either path passes the current chat room as the context identifier, which preserves conversational context across follow-up requests and directs \estructural{} to write into that room's workspace.
The protocol itself therefore carries only task state and compact textual results, while the structure files remain available to campaign scripts through the shared workspace, without ever serializing atomic coordinates into an \ac{llm} context.

\subsection{RAISE platform access}

The RAISE platform~\cite{nazeriRAISESelfdrivingLaboratory2026} is an \ac{sdl} built to link liquid formulation to interfacial property measurement in a closed loop; the present integration uses its first demonstrated capability, static contact angle measurement~\cite{nazeriRAISESelfdrivingLaboratory2026}.
RAISE was benchmarked against conventional goniometry and showed similar accuracy, while reducing measurement variability by 20--62\%, depending on the substrate.
It also achieved a throughput of about one contact angle measurement per minute.
It is exposed to \optima{} through a typed \texttt{run\_raise\_experiment} interface, which can be invoked as an agent tool or imported by generated Python campaign programs.
Each call sends a formulation to the physical platform through an MQTT bridge and waits for the returned static contact angle~\cite{nazeriRAISESelfdrivingLaboratory2026}.
The interface verifies that the returned formulation matches the submitted candidate before making the measurement available to the campaign program.
Because the current bridge uses shared request and response channels without identifiers, experiments are serialized to prevent responses from being assigned to the wrong candidate.
Validated measurements can then be submitted to BO-MCP as observations, allowing RAISE to serve as a physical evaluator.

\subsection{RoboChem-Flex platform access}

RoboChem-Flex is a low-cost, modular \ac{sdl} for synthetic organic chemistry that combines Cartesian-robot samplers, custom syringe pumps, and reconfigurable flow reactors with Python-based device control and inline analytical instruments~\cite{pilonFlexibleAffordableSelfdriving2026}.
In \optima{}, the platform is connected through \textit{RoBridge}, an authenticated and stateful HTTP service deployed on the robot computer alongside the platform control software and securely exposed online through Cloudflare.
The integration exposes tools that retrieve the platform manual and live OpenAPI description, inspect individual operations and schemas, and query read-only resources such as status and capabilities.
Before generating an executable request, the agent compares the proposed experimental conditions with the reactor configuration, analytical method, and parameter constraints reported by the live platform \ac{api}.
A dedicated \emph{bo-roboflex-specialist} is equipped with both RoboChem-Flex and BO-MCP \ac{api}-inspection toolsets that \opt{} can delegate campaign construction to.
\textit{RoBridge} provides the complementary deterministic safeguards: it admits only one active campaign, rejects requests inconsistent with its explicit robot state, validates conditions against live capabilities, and requires a locally authenticated technician to certify the physical vial layout before chemistry can begin.
Thus, \opt{} retains conceptual control over the complete closed loop while querying both services for their authoritative current state; optimization state remains in BO-MCP, platform and safety state remains in \textit{RoBridge}, and invalid actions are rejected on either side.
The technical implementation is detailed in \ac{si}, \cref{si:robridge}.

\section{Results}

We evaluate \optima{} in three settings of increasing interaction with the physical world: architectural benchmarks against controlled ablations, purely computational discovery campaigns, and closed-loop optimization on two live experimental platforms.
The benchmarks quantify optimization performance and resource use across repeated runs, whereas the case studies assess whether \opt{} turns a stated objective into a valid, reproducible campaign with a complete experimental record.

Overall, the benchmarks against controlled ablations support combining established \ac{bo} implementations with programmatic execution and provide qualified support for specialist-subagent delegation, particularly when campaign design materially affects the search.
They further show that optimization quality and operational robustness are distinct, task-dependent properties.
The remainder of the main text focuses on the applied showcases; complete benchmark results are reported in the \ac{si} (\cref{sec:framework-comparison-si}).

\subsection{Digital discovery campaigns}

\subsubsection{Combinatorial molecular discovery for organic solid-state lasers}

\begin{figure}
    \centering
    \includegraphics[width=1\linewidth]{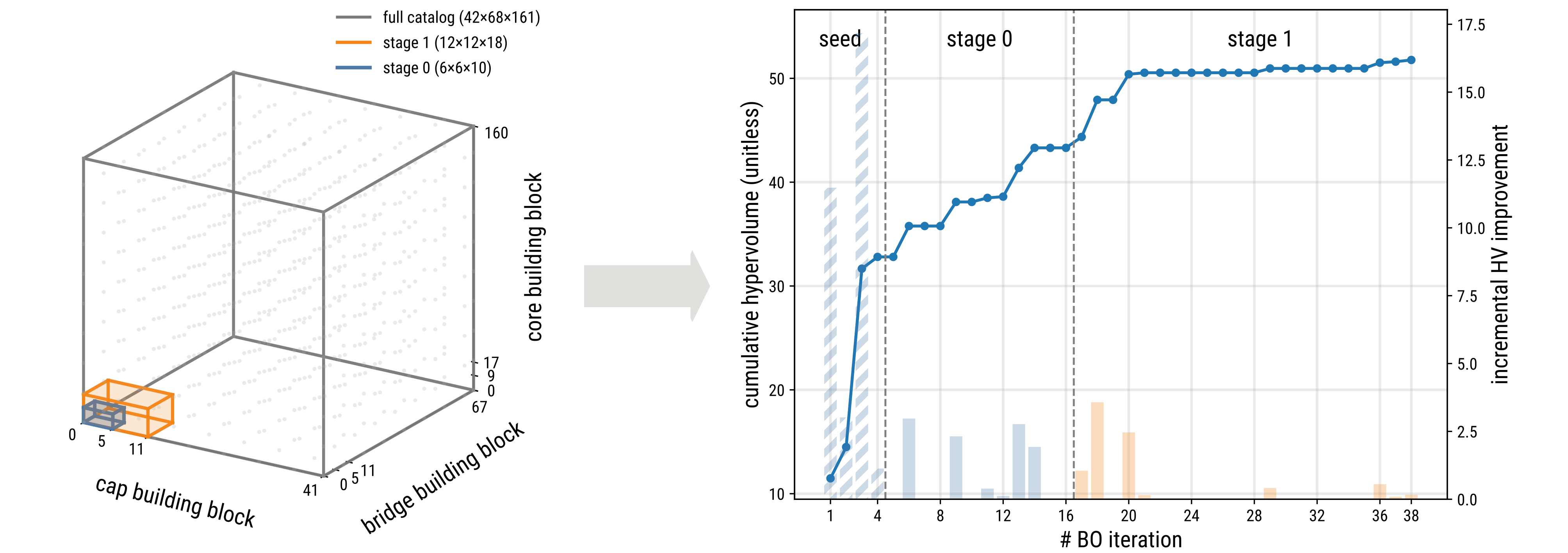}
    \caption{
        Autonomous digital discovery of \ac{osl} candidates with \optima{}.
        \textbf{(a)} Schematic representation of the explored search spaces relative to the full combinatorial design space spanned by the cap, bridge, and core building blocks.
        \textbf{(b)} Closed-loop \ac{bo} progress across the seed set, stage 0, and stage 1, shown as the cumulative best-so-far hypervolume together with the incremental hypervolume improvement at each iteration.
        Both panels were generated by \opt{} and required only minimal visual modification.
    }
    \label{fig:osl_campaign}
\end{figure}

\citet{strieth-kalthoffDelocalizedAsynchronousClosedloop2024} recently demonstrated the delocalized, asynchronous, closed-loop discovery of organic laser gain materials, in which \ac{bo} steered robotic synthesis and spectroscopic characterization of fragment-assembled emitter candidates across five laboratories.
Mirroring this discovery problem, \opt{} was tasked with running a closed-loop digital campaign over the same published fragment catalogues, exploring the fragment chemistry through an inexpensive digital proxy workflow and assuming the orchestration, diagnosis, and analysis roles that otherwise require bespoke infrastructure and human planners.
Rather than treating this as a static screening problem, \opt{} formulated it as a multi-objective \ac{bo} campaign coupled to molecular assembly, conformer generation, and quantum-chemical evaluation.

The molecular design space followed an \texttt{A--B--C--B--A} composition rule, where the decision variables corresponded to cap, bridge, and core fragments.
Before starting the optimization, \opt{} inferred the assembly rule from the fragment catalogues, including the reactive halogen sites used to join fragments.
The reconstructed products were validated against a user-provided reference subset of 1129 molecules, all reproducing the reference connectivity; this file served only as a correctness check and contributed neither candidates nor objective values.
Previously, constructing this design space required a domain expert to identify the assembly logic, write a dedicated Python script, and manually inspect the resulting structures.
\opt{} represented the resulting search space using three categorical parameters, \texttt{cap\_id}, \texttt{bridge\_id}, and \texttt{core\_id}, supplemented by a concise set of RDKit-derived molecular descriptors~\cite{rdkit} characterizing each fragment's size, composition, structure, and physicochemical properties.
These descriptors informed both the surrogate model, enabling it to exploit fragment similarity~\cite{fitznerBayBEBayesianBack2025}, and the distance metric used for frontier-aware search-space expansion.

Of the 462,672 theoretically accessible \texttt{A--B--C--B--A} combinations (42 caps, 68 bridges, and 162 cores in the published catalogues), the campaign began in a deliberately restricted and inexpensive search space of 360 candidates (Stage~0, \cref{fig:osl_campaign}a), selected by a small-fragment-first strategy.
The four initial observations were selected using fragment descriptors to provide balanced and representative coverage across all three fragment types.
The campaign optimized three objectives jointly: maximizing oscillator strength, minimizing the colour error relative to a target visible excitation energy, and minimizing a structural ambiguity penalty.
In this initial stage, the twelve \ac{bo}-selected candidates contributed more cumulative hypervolume (3.76) than the four seeds (2.78), indicating that the optimizer added value beyond the initial design (\cref{fig:osl_campaign}b).

After 16 successful observations, \opt{} analyzed the Pareto front and expanded the search space through a deterministic, frontier-aware rule:
the fragments occurring in Pareto-optimal candidates served as per-slot anchors, and additional caps, bridges, and cores were ranked by their z-scored descriptor distance to these anchors.
Stage~1 expanded the search space from 360 to 2592 candidates while retaining the original space, yet still covered only about 0.6\% of all theoretically accessible combinations.
The campaign then continued in the expanded Stage~1 space, with \ac{bo} building on the initial 16 observations.
The expanded campaign ultimately produced 38 successful observations in total and revealed recurring fragment motifs associated with favourable brightness--colour--robustness trade-offs.
Asked to interpret the most promising candidates, \opt{} identified several recurring chemical motifs.
These included strong dimethylamino--aryl donor caps, compact vinylene or thiophene $\pi$-bridges, and cores that tuned acceptor strength.
It also distinguished chemically promising Pareto points from formally non-dominated but impractical extremes, such as a rigid but essentially dark candidate with near-zero oscillator strength.
It also cautioned that these interpretations rest on an inexpensive digital proxy rather than converged photophysics.

The campaign also tested the resilience of the agentic workflow:
When a configuration mismatch disrupted Stage~1, \opt{} autonomously diagnosed and patched its evaluator, validated the fix, and resumed production without human intervention.
Throughout, \opt{} preserved campaign integrity by excluding failed evaluations and reconciling the optimizer state after timeouts.
Automatically generated diagnostics tracked optimization progress, search-space expansion, and Pareto trade-offs, providing an auditable record of the search.

Overall, the campaign demonstrates that \opt{} can autonomously run and recover an iterative molecular discovery workflow, while leaving strategic control over evaluation budgets and search-space expansion to the operator.
Over 47 top-level turns spanning about 3~d~22~h calendar time (16.0~h of agent-run wall-clock), this conversation made 397 \ac{llm} calls --- 137 in the main agent plus 260 subagent calls (240 across 4 gpt-5.4 subagents + 20 gpt-4.1 one-shots) --- consuming 35.4M input / 229k output tokens at a total cost of \$24.40.

\subsubsection{Finite-space discovery across chemistry and materials}
\label{sec:computational-showcases}

Four additional campaigns test \opt{} across finite candidate spaces spanning ligand electronics, excited-state screening, transition-metal complexes, and porous materials.
Their common quantitative record is collected in \cref{tab:computational-showcases}; full campaign details are provided in the corresponding \ac{si} sections.
The improvement column compares the seed baseline with the final campaign state.

\begin{table}[htbp]
    \centering
    \scriptsize
    \caption{Comparison of the four computational discovery campaigns.
        Activity reports main-orchestrator tool calls followed by all recorded \ac{llm} calls, including specialist and evaluator-routing calls.
        Tokens are input/output totals, and cost is the total recorded \ac{llm} cost.
        Evaluation counts distinguish attempted or historical observations from successful observations where relevant.}
    \label{tab:computational-showcases}
    \setlength{\tabcolsep}{3pt}
    \renewcommand{\arraystretch}{1.25}
    \begin{tabularx}{\linewidth}{@{}>{\raggedright\arraybackslash}p{0.16\linewidth}>{\raggedright\arraybackslash}p{0.12\linewidth}>{\raggedright\arraybackslash}p{0.14\linewidth}>{\raggedright\arraybackslash}X>{\raggedright\arraybackslash}p{0.13\linewidth}>{\raggedright\arraybackslash}p{0.12\linewidth}>{\raggedleft\arraybackslash}p{0.07\linewidth}@{}}
        \toprule
        \textbf{Showcase}     & \textbf{Search space}                          & \textbf{Evaluation budget}  & \textbf{Recorded BO improvement}                                  & \textbf{Tool / \ac{llm} calls} & \textbf{Tokens in / out} & \textbf{Cost (USD)} \\
        \midrule
        Phosphine ligands     & 364 ligands                                    & $8+40=48$; 48 successful    & Hypervolume $0.791\rightarrow1.038$; Pareto set $3\rightarrow17$  & 33 / 128                       & 6.18M / 50.4k            & \$7.08              \\
        Inverted-gap emitters & 1512 molecules                                 & 44 attempted; 39 successful & Best gap $0.227\rightarrow0.223$~eV                               & 34 / 279                       & 13.1M / 76.5k            & \$13.01             \\
        Co bisphosphines      & 144 ligands                                    & $4+10=14$; 6 feasible       & Feasible-only hypervolume $0.572\rightarrow1.000$                 & 94 / 557                       & 56.1M / 210k             & \$53.29             \\
        Xe/Kr MOFs            & 2800 nominal (420 valid); refined to 109 valid & $30+50=80$; 65 successful   & Desirability $0.487\rightarrow0.502$; Pareto set $7\rightarrow12$ & 25 / 162                       & 11.6M / 61.3k            & \$11.44             \\
        \bottomrule
    \end{tabularx}
\end{table}

\citet{laplazaGeneticOptimizationHomogeneous2022} optimized phosphine and carbene ligands for Ni-catalyzed aryl-ether cleavage with a genetic algorithm and a molecular-volcano fitness.
We instead exposed the underlying electronic and structural trade-offs by asking \opt{} to tune monodentate phosphines P(R$^1$)(R$^2$)(R$^3$) as a ligand-level proxy for Ni catalysis.
The campaign did not yield a single best ligand because its electronic objectives conflicted.
\opt{} recognized that \ac{bo} was instead finding smaller ligands with similar electronic properties, identified an uninformative steric objective, and recommended stopping as further evaluations produced diminishing returns (\ac{si}, \cref{si:phosphine}).

\citet{polliceOrganicMoleculesInverted2021} screened heptazine- and cyclazine-derived chromophores for inverted singlet--triplet gaps and reported excitation energies for the resulting library.
We used a size-filtered subset of 1512 molecules as a \ac{bo} benchmark with known published results.
The published values were hidden from the agent, allowing us to assess both its candidate selection and the accuracy of its inexpensive evaluator.
Before launching the campaign, \opt{} timed a trial evaluation and switched to a less expensive method and smaller budget when server data showed that its runtime estimate was too optimistic.
The small improvement in the best gap ($0.227\rightarrow0.223$~eV) reflects a fortunate random initialization: the first of five randomly sampled seed molecules was already near-optimal.
A post-campaign comparison with the published references showed that the inexpensive \ac{td-dft} evaluator ranked the evaluated molecules well but did not reproduce their inverted gaps.

\citet{hoodHighlyActiveCationic2020} reported cationic Co(II) bisphosphine hydroformylation catalysts whose activity depends sharply on the linker and phosphorus substituents.
We recast the ligand space as a finite multi-objective campaign in which \opt{} requested every three-dimensional [Co(acac)(P$_2$)]$^+$ structure from \estructural{} over the \ac{a2a} interface, left optimizer state to BO-MCP, and evaluated converged geometries through \grafico{}.
\opt{} therefore coordinated candidate selection, structure generation, and electronic-structure calculations across separate services.
Faced with eight failed evaluations, \opt{} distinguished computational failures from evidence about ligand performance.
It traced most failures to unconverged geometry optimizations and concluded that improving the generated starting structures and relaxation protocol, rather than extending the \ac{bo} campaign, was the appropriate next step (\ac{si}, \cref{si:cobalt}).

Screening \acp{mof} for Xe/Kr separation is conventionally approached by enumerating and scoring large hypothetical framework libraries one candidate at a time, commonly using pore-geometric filters such as the pore-limiting diameter as a first pass~\cite{zhouEfficientScreeningEnhanced2026}, or by inverse design against a user-specified selectivity target~\cite{limFinelyTunedInverse2021}.
Following the design goal of~\citet{limFinelyTunedInverse2021}, but with their selectivity objective replaced by a geometric proxy that requires no adsorption simulation, we posed the same design question as a small-budget \ac{bo} campaign to test the workflow.
\opt{} assembled candidates with PORMAKE~\cite{Lee2021pormake} and analyzed their pore geometry with Zeo++~\cite{Willems2012zeopp}.
Here, the salient capability emerged only after the first campaign: the agent recognized that independent topology, node, and edge variables described mostly unconstructible combinations, declined to spend more budget on the same representation, and rebuilt the problem as a finite set of connectivity-compatible triples while carrying prior successes forward as evidence.
The fully feasible follow-up improved only modestly, but \opt{} identified the search-space representation, rather than the optimizer, as the bottleneck and replaced it while retaining all prior observations (\ac{si}, \cref{si:mof}).

\subsection{Closed-loop formulation optimization with RAISE}

To evaluate \optima{} in a physical closed loop with RAISE~\cite{nazeriRAISESelfdrivingLaboratory2026}, we asked \opt{} to identify an ethanol and \ac{sds} mixture with a target contact angle of $65^{\circ}$, stopping when a measurement fell within $\pm 1^{\circ}$ and proceeding in small, explicitly approved increments.
Before campaign construction, \opt{} selected two literature-informed warm-start formulations.
\opt{} then executed a specialist-authored program that connected the RAISE evaluator to a BayBE-backed BO-MCP campaign.

\begin{figure}[htbp]
    \centering
    \renewcommand\thesubfigure{\Alph{subfigure}}
    \edef\raisephotoboxheight{\the\dimexpr0.606\linewidth-2\fboxsep-2\fboxrule\relax}
    \begin{subfigure}[t]{0.48\linewidth}
        \centering
        \vspace{0pt}
        \includegraphics[width=\linewidth]{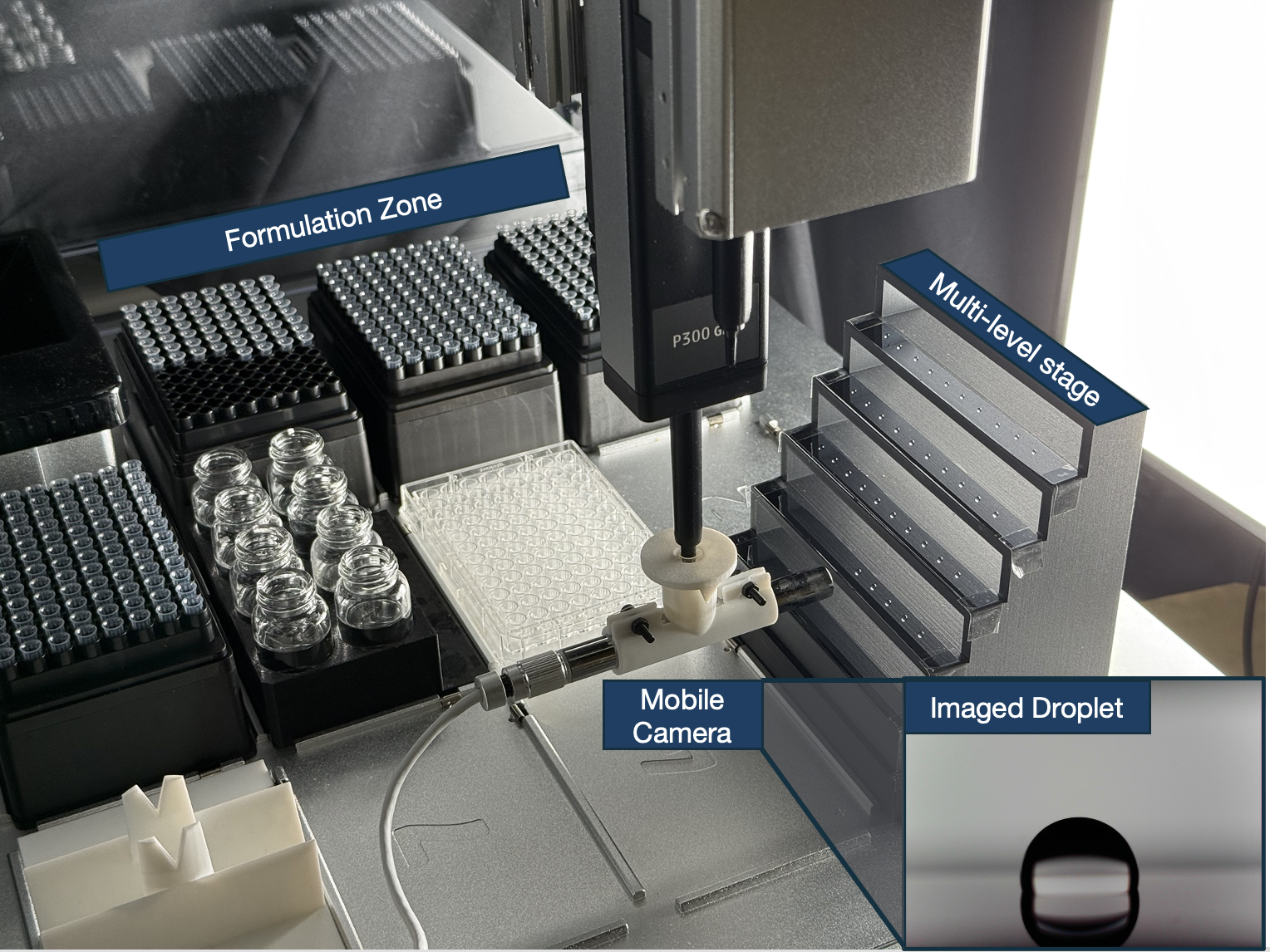}
        \caption{RAISE \ac{sdl} platform.}
        \label{fig:raise-platform}
    \end{subfigure}
    \hfill
    \begin{subfigure}[t]{0.48\linewidth}
        \centering
        \vspace{0pt}
        \includegraphics[width=\linewidth]{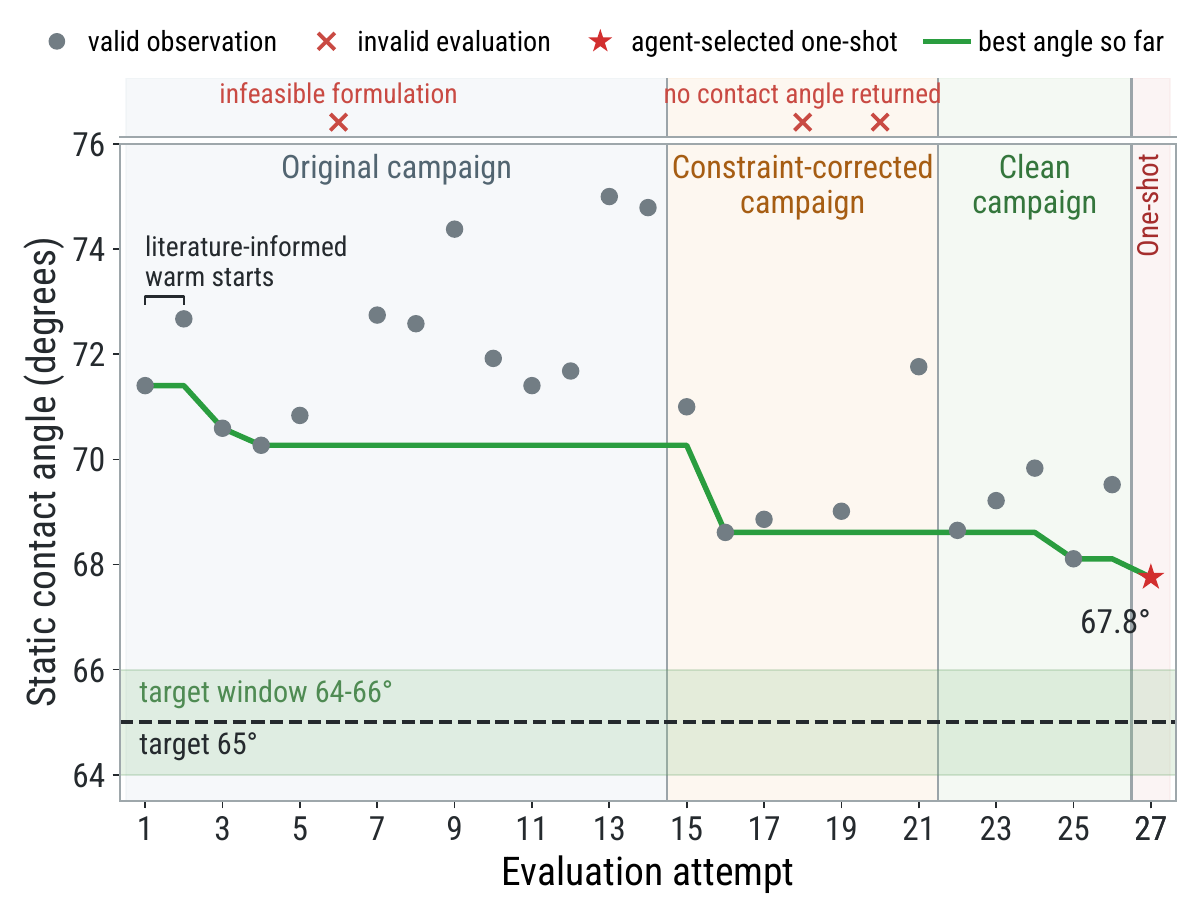}
        \caption{Contact-angle optimization trajectory.}
        \label{fig:raise-bo-trajectory}
    \end{subfigure}
    \caption{
        \textbf{Closed-loop formulation optimization with RAISE.} \textbf{(A)} The RAISE \ac{sdl} platform was used to prepare formulations and measure their static contact angles (reproduced from Ref.~\cite{nazeriRAISESelfdrivingLaboratory2026} under CC BY-NC 4.0 licence).
        \textbf{(B)} Contact-angle measurements and the best angle obtained so far across the original, constraint-corrected (ethanol upper bound reduced from 60 to 50~v/v\%), and clean campaigns, followed by the agent-selected one-shot experiment.
        Grey circles denote valid observations, red crosses mark one infeasible formulation and two evaluations for which RAISE returned no contact-angle value, and the red star marks the final one-shot result.
        The upper event strip places the invalid evaluations in sequence without assigning them contact-angle values; background shading identifies the campaign stages, and the green band denotes the $65 \pm 1^{\circ}$ target window.
        The panel shows the 27 evaluations in the reported campaigns and the final \opt{} one-shot.
    }
    \label{fig:liquid-handling-placeholder}
\end{figure}

The campaign exposed a consequential distinction when handling infeasible conditions and failed measurements.
The initial campaign script recorded every failed evaluation as a fallback observation of $180^{\circ}$.
This convention conflated a permanently infeasible formulation with two feasible formulations with missing measurements as equivalent outcomes.
The infeasible suggestion was 60~v/v\% ethanol, which exceeded the maximum ethanol concentration achievable under this campaign's well-volume and stock-concentration constraints, approximately 50~v/v\%.
Following operator approval, \opt{} reduced the upper bound to 50~v/v\% and transferred the 13 valid observations from the original campaign into a new campaign with corrected constraints.

The $180^{\circ}$ penalty assigned to the infeasible formulation led all eight subsequent suggestions to remain below 20~v/v\% ethanol.
The latter two measurement failures occurred at feasible formulations containing 30--36~v/v\% ethanol, near the best observations obtained at that stage.
Recording each failure as $180^{\circ}$ falsely associated those candidates with highly unfavourable contact angles and biased the surrogate against this promising region.
The experimentalist traced the measurement errors to a slight shift in the backlight position relative to the droplet, which disrupted contour detection in the image-processing pipeline.
This cause was not reported to \opt{} during the campaign; therefore, the same fallback penalty was applied to the measurement.

On reviewing the results, \opt{} identified the fallback encoding as the source of the problem, where a failed measurement should be treated as missing data instead of an unfavourable outcome.
It proposed a new campaign seeded with all 18 unique valid observations from the two preceding campaigns, so no physical experiment was repeated.
The revised procedure retried the same formulation up to twice and, if neither retry produced a contact-angle value, recorded the failure locally without submitting an objective value to BO-MCP.
Following operator approval, all five \ac{bo} iterations were completed without any measurement failures or retries.
This highlights a broader requirement for agentic \acp{sdl}, in which sensor data and platform state are readily accessible to the agent, enabling it to determine whether an unexpected measurement arose from the formulation or from changes in its physical environment.

Across the 27 plotted evaluations, the best result moved from $71.4^{\circ}$ for the warm start to $68.1^{\circ}$ in the clean campaign.
Five additional physical measurements were performed for implementation validation and preliminary checking and were not included in the optimization trajectory.
When the operator allocated one final experiment because of the remaining time constraint, \opt{} selected the maximum feasible ethanol concentration with SDS fixed at its best observed level, yielding $67.8^{\circ}$.
This result demonstrates how one-shot agent suggestions can complement \ac{bo} by translating the evidence accumulated during a campaign into an effective final experiment under changing operational constraints.
The campaign did not reach the $\pm 1^{\circ}$ target window.
From this boundary result, \opt{} inferred that $65^{\circ}$ was unlikely to be attainable with the two available reagents and recommended replicate measurements and a change of formulation system before further optimization.
A complete report of the session, including the optimization configuration and per-campaign data, is provided in \ac{si}, \cref{si:raise}.

Over 11 top-level turns spanning 2~h~38~min of calendar time, the campaign used 148 \ac{llm} calls: 34 in the main agent (\texttt{gpt-5.5}) and 114 across five subagent (\texttt{gpt-5.4}) sessions.
These calls consumed 7.75M input and 97k output tokens at a total cost of \$15.45.

\subsection{Multi-objective flow photochemistry with RoboChem-Flex}

We next evaluated \optima{} in continuous-flow synthetic chemistry using RoboChem-Flex.
As a test case, we revisited the platform's published photocatalytic radical trifluoromethylation benchmark, with trifluoroacetic anhydride as the \ce{CF3} source and a pyridine \textit{N}-oxide as the activator.
Whereas the original workflow delegated experimental design to a dedicated BoTorch-based optimization engine~\cite{pilonFlexibleAffordableSelfdriving2026}, here \opt{} assumed that role.
\opt{} executed the live campaign, transferred the experimental history between objectives, and increased the best measured yield from 30.0\% among the informed seed experiments to 58.8\%.

\begin{figure}[!ht]
    \centering
    \includegraphics[width=\linewidth]{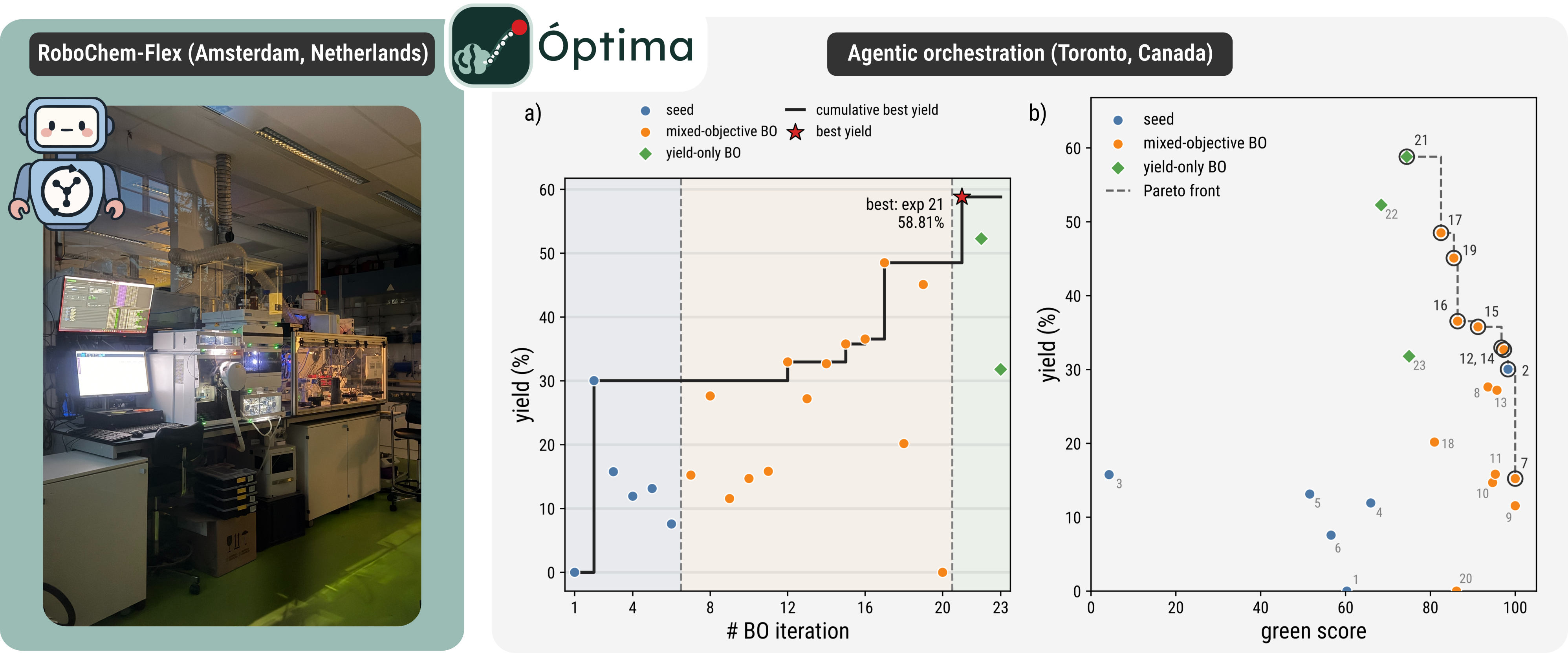}
    \caption{
        Agent-controlled photocatalytic trifluoromethylation campaign on RoboChem-Flex.
        \textbf{(a)} Measured yield across the 23 valid experiments, together with the cumulative best-so-far yield.
        Background shading indicates the six informed seed experiments, the 14 multi-objective \ac{bo} experiments, and the three yield-only \ac{bo} experiments initialized from the migrated campaign history.
        The highest measured yield, 58.8\% in experiment~21, is marked by a star.
        \textbf{(b)} The same observations in the two-dimensional objective space defined by yield and the agent-defined green score.
        The dashed staircase denotes the empirical Pareto front obtained by maximizing both objectives, and labels indicate experiment numbers.
        Open outlined markers identify non-dominated points on the Pareto front.
        Three additional runs affected by analytical failures are excluded from both panels.
        Both panels were generated by \opt{} from its exported campaign data.
    }
    \label{fig:robochemflex}
\end{figure}

\subsubsection{Campaign design under platform constraints}
The operator supplied CSV files defining the available chemistry, experimental search space, stock concentrations, and analytical constants.
From these inputs, \opt{} constructed a seven-parameter search space spanning photocatalyst and oxidant identities and loadings, \ac{tfaa} loading, residence time, and light intensity (\ac{si}, \cref{si:robochemflex:trace:s1}).

Interestingly, both the operator's process table and the platform's live capability descriptor reported a generic light-intensity range of 0--100\%.
However, \textit{RoBridge}'s capability descriptors of the mounted UFlow~\cite{massonOpensource3DPrinted2024} reactor light array supported only $\{0,~25,~50,~75,~100\}$\%.
\opt{} therefore adopted these discrete settings, preserving the hardware's capabilities.
More generally, this demonstrates the ability of the agent to bridge the gap between user intent and laboratory reality, automatically mapping abstract experimental requests to the nearest physically realizable application.

\opt{} initialized the optimization by deliberately selecting six diverse experiments covering all categorical variables, including five photocatalysts, both oxidants, and both bounds of each stoichiometric variable, while sampling multiple light intensities and residence-time regimes.
\opt{} autonomously translated each candidate's named search-space parameters into a RoboChem-Flex experimental request, mapping reagent identities to platform identifiers, converting residence time to the required units, and appending the fixed reaction and analytical settings.
Before any hardware was addressed, \opt{} emitted the experimental request as a hypothetical JSON document for operator inspection, making its interpretation of the proposed conditions auditable in advance.

\subsubsection{Multi-objective optimization and objective revision}
The operator requested optimizing two objectives in the first campaign: rewarding greener conditions alongside yield.
\opt{} defined an analytic green score from the same parameter vector.
The score combined normalized penalties for catalyst loading, TFAA and oxidant equivalents, and photonic dose based on light intensity and residence time; its full definition and normalization are provided in the \ac{si} (\cref{si:robochemflex:trace:s1}).
Notably, although the agent was not prompted with any explicit formulation of the ``green score'', the resulting metric closely matches the normalization strategy and weighting that operators have manually designed for unrelated previous experimental campaigns on RoboChem, indicating that the agent independently converged on a comparable and chemically intuitive objective~\cite{robochem_up}.

With over 20 valid observations, comprising six informed seeds and 14 \ac{bo}-selected experiments, the campaign explicitly mapped the trade-off between yield and this agent-defined resource-efficiency metric.
The observed conditions ranged from maximally frugal but poorly productive settings, yielding 15.2\% at a green score of 100, to the most productive point of this phase, yielding 48.5\% at a green score of 82.5 (\cref{fig:robochemflex}b).
When the operator subsequently narrowed the goal to yield alone, \opt{} transferred all valid observations from the multi-objective campaign into a new single-objective campaign.
Continuing from that history, the best measured yield rose to 58.8\% in experiment~21 (\texttt{R0067}), roughly double the best seed observation (30.0\%) and above the best point of the mixed-objective phase (\cref{fig:robochemflex}a; \ac{si}, \cref{tab:robochemflex-runs}).
When evaluated against both objectives, the new condition added a high-yield endpoint to the Pareto front: yield increased from 48.5\% to 58.8\%, while the green score decreased from 82.5 to 74.4 (\cref{fig:robochemflex}b).
Starting from the deliberately diverse seed experiments, \ac{bo} converged on chemistry closely related to the published RoboChem-Flex optimum, using the same tris(bipyridine)ruthenium photocatalyst family and pyridine \textit{N}-oxide activator while identifying distinct operating conditions~\cite{pilonFlexibleAffordableSelfdriving2026}.

\subsubsection{Long-horizon execution and analytical diagnosis}
Sustaining this loop required a horizon far longer than a conversational turn, with measurements arriving roughly hourly and campaigns running unattended overnight.
\opt{} ran the campaign as a monitored background task, submitting experiments one at a time and checking their status at fixed intervals.
It reported only state changes, alerts, and periodic heartbeats to the conversation while saving the complete instrument record to disk.
Only runs with a passing analysis and finite yield were added to the surrogate model.
When a run failed before analysis, \opt{} either retried it or paused the campaign for later resumption.
Operator involvement was limited to high-level oversight and necessary physical interventions.

Notably, the first seed experiment (\cref{fig:robochemflex}a, experiment~1) returned a yield of exactly zero with the analysis marked as passing, an outcome that can mean either that no product formed or that the analysis failed to detect a sufficiently well-defined peak in the expected spectral region.
Prompted by the operator to investigate, \opt{} located the platform's raw-result endpoint and mirrored the run's complete analytical record into the workspace, including the free induction decay and the processed spectrum, then analyzed the spectrum itself:
It reconstructed the chemical-shift axis from the file header, estimated the noise floor from the median absolute deviation, and confirmed from the acquisition parameters that the intended 32-scan \ce{^{19}F} protocol had in fact run.
\opt{} observed a dominant \ce{^{19}F} signal at $-76.5$~ppm and proposed a trifluoroacetyl-derived species as a possible source.
The expected product region ($-58 \pm 3$~ppm) contained only baseline structure rather than a true peak, supporting the reported zero yield as a genuine absence of detectable product.
This addresses a recurring operational bottleneck in autonomous experimentation: determining whether an anomalous analytical result reflects failed chemistry or a failed measurement.
Because the campaign continued in the background, \opt{} could perform this assessment in dialogue with the operator without pausing execution.
The agent also ran a lightweight campaign supervisor that flagged zero-yield and no-peak analyses and was configured to halt execution after five consecutive such outcomes.
Such a streak would suggest an instrument malfunction rather than a genuinely unproductive region of the search space, a distinction that required active human oversight and domain expertise in the pre-\ac{llm} era~\cite{bai2024dynamic}.

\subsubsection{Resource use relative to human-directed optimization}
The campaign covered 5~d~8~h of calendar time and 69 top-level operator turns.
The platform was occupied for 34~h of that span and \opt{} itself accounted for 2~h~16~min of \ac{llm} and tool execution; the remaining 94~h were idle, dominated by nights and by a 1~d~21~h stop for hardware preparation, without which the campaign spanned 3~d~11~h.
Across the campaign, it made 1108 \ac{llm} calls (\texttt{gpt-5.5}), 498 in the main agent plus 610 across 14 subagent sessions, consuming 114.5M input and 328k output tokens at a total cost of \$109.55.
Even with the 1M context window of \texttt{gpt-5.5}, this campaign was only made possible by the subagent architecture, which separates the context.

Benchmarked against a human-directed \ac{bo} campaign on the same transformation and platform, the agent was stopped after 23 experiments, compared with 50 for the human-directed campaign, consuming less than half the starting material.
Despite the additional inference cost, the agent campaign maintained a 13\% overall cost advantage, with savings primarily driven by the reduced number of experiments.
The agent reached 58.8\% yield, compared with 70.9\% for the full human-directed campaign and 63.9\% at matched consumable budget.
Importantly, both agent-identified optima were more mass-efficient, using substantially less reagent and achieving higher \acp{rme}; this difference was even more pronounced across the complete campaigns, where the agent returned only two zero-yield experiments compared with 27.
Full cost and mass accounting is provided in the \ac{si}, \cref{si:robochemflex:cost}, while the complete operator--agent exchange, control layer, and experimental conditions are provided in \cref{si:robochemflex:trace}.

\section{Discussion}

Across the benchmarks and case studies, candidate selection followed a flexible division of labour between \ac{bo} and the agent.
The architecture comparison demonstrated programmatic \ac{bo} for routine search, whereas the RAISE experiment showed that the agent could propose an experiment directly when the accumulated evidence and remaining budget called for scientific judgment.
\optima{} can therefore move between classical \ac{bo} search and \ac{llm}-based selection within the same framework.

Importantly, this adaptability is not limited to choosing the next experiment.
The scientific problem itself can be revised as a campaign unfolds.
Through natural-language interaction with \opt{}, the researcher can redefine the scientific question, where \opt{} translates that decision into concrete changes to the running campaign.
The researcher would otherwise need to manually stop the current execution, preserve its state, and reconstruct the campaign according to the revised specification.
In the \ac{mof} study, this involved replacing a search-space representation that produced many invalid structures; in RoboChem-Flex, it involved changing the objective while retaining the existing experimental history.
By preserving the campaign state through such changes, \opt{} allows the researcher to focus on the scientific question rather than campaign execution.

The value of this adaptability, however, depends on the evidence available to support those decisions.
The search for inverted-gap chromophores inherited the bias of its evaluator; a change in imaging conditions appeared to RAISE only as a missing value, and the RoboChem-Flex zero-yield result became interpretable only after the raw spectrum was inspected.
In each case, the numerical result alone provided an incomplete account of what had occurred.
While provenance preserves how a result was produced, access to sensor data, such as camera frames, analytical records, and platform state, enables the agent to diagnose failures and base its advice on underlying evidence rather than solely on the numerical observations typically exposed to the \ac{bo} algorithm.
As agent access expands to laboratory sensing and control, it should remain bounded by deterministic safety interlocks and explicit human authorization~\cite{leong2025steering}.
Structured safety reasoning and accountability mechanisms provide complementary safeguards for agentic \acp{sdl}~\cite{Kang2026,AspuruGuzik2026}.

Providing such context is only part of the challenge.
It must also remain usable as the campaign evolves over time.
Programmatic access to \ac{bo} state and intermediate results enabled \opt{} to repair and resume campaigns when research plans or operating conditions changed.
Persistent workspaces and executable programs maintained this continuity across model invocations and evaluators by encoding repeated tool use in code and loading intermediate results into the agent context only when needed.\footnote{Related patterns have been described for general-purpose agents: Anthropic uses \href{https://www.anthropic.com/engineering/code-execution-with-mcp}{programmatic tool calls} to keep intermediate results outside the model context and \href{https://www.anthropic.com/engineering/effective-harnesses-for-long-running-agents}{structured artifacts} to maintain continuity across context windows, and OpenAI uses a \href{https://openai.com/index/equip-responses-api-computer-environment/}{filesystem-backed computer environment} for persistent runtime state.}
For long-running execution, the background monitor further reduced the need for continuous model involvement by filtering incremental process output into compact progress events for the agent and its frontend.
\opt{} could therefore attend to the campaign only when a decision or user notification was required, while remaining available to discuss other scientific questions with the researcher as experiments continued in the background.
This event-driven supervision preserves visibility into campaign progress as autonomous experiments extend over longer time horizons.\footnote{\href{https://openai.com/index/safety-alignment-long-horizon-models/}{OpenAI's discussion of long-horizon models} similarly emphasizes trajectory-level monitoring, user visibility, and mechanisms to intervene, pause, or roll back execution.}
These mechanisms allow unforeseen events to be accommodated without discarding valid prior work, as is routinely required in research practice.

Supporting this level of adaptive orchestration introduces an additional item into the operating budget: \ac{llm} inference.
In the RoboChem-Flex study, inference cost more than the chemistry it directed, yet the complete agent campaign remained 13\% cheaper than the human-directed campaign.
\opt{} used 46\% of the experiments and 39\% of the starting material while identifying more mass-efficient conditions, although with lower final yield.
This highlights that optimization performance extends beyond the highest attainable yield: experimental efficiency, material consumption, and cost are equally relevant measures of a campaign's value~\cite{Schuurmans2026}.
Further reductions in inference cost would strengthen this cost advantage.
Reducing equipment occupancy, in turn, will require optimization strategies that account for how long an experiment ties up the instrument, not just how informative it is expected to be, for example, preferring a short, moderately informative experiment over a much longer one that offers only a marginally better expected outcome.
Long-horizon orchestration also places new demands on agent context management.
Simply retaining an ever-growing interaction history, or compressing it into an undifferentiated summary, may be insufficient when decisions depend on experimental observations and constraints established much earlier in a campaign.
Instead, campaign context could be separated into structured classes, for example, persistent scientific objectives, constraints, experimental evidence, and transient execution details.
Long-term information could then be retrieved in a structured way, while short-lived information can be summarized or branched off.

\begin{figure}
    \centering
    \includegraphics[width=1\linewidth]{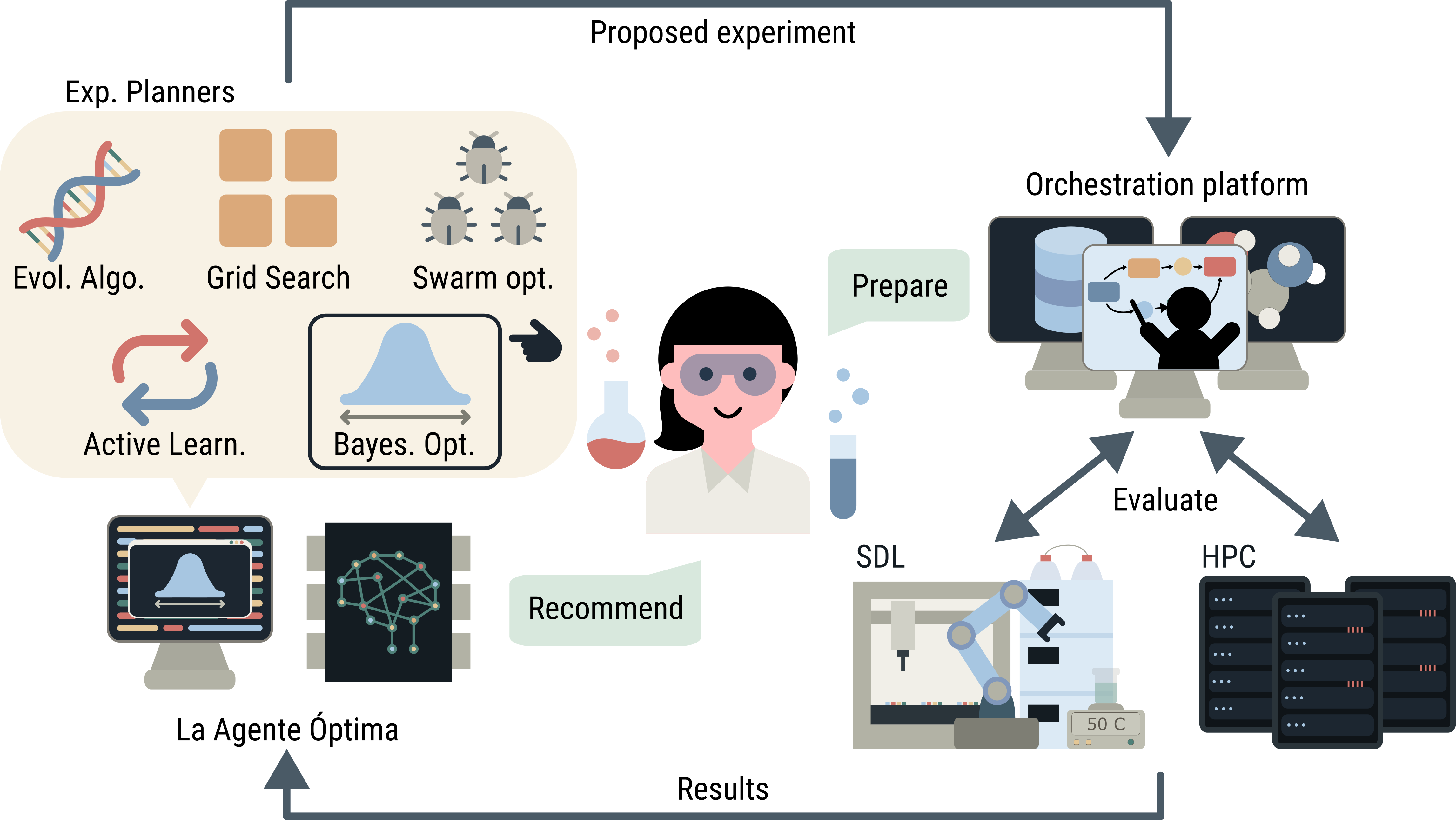}
    \caption{Conceptual vision of \opt{} as a general optimization layer for autonomous experimental campaigns.
        Based on the problem definition and available resources, \opt{} selects and configures an appropriate experimental design strategy and operates in a closed loop with the scientist, laboratory infrastructure (SDL), and computational resources (HPC), adapting its decisions as new experimental data become available.}
    \label{fig:agentic-ai-self-driving-labs}
\end{figure}

These studies show how an agent can maintain alignment between an evolving scientific question and the executable state of a running campaign.
\Cref{fig:agentic-ai-self-driving-labs} extends this principle to more general experimental orchestration, in which the user remains central throughout.
Ultimately, a researcher would specify only the campaign's goal, available resources, levers to pull, constraints, and success criteria, stating all of this in natural language rather than manually configuring an experimental design.
\opt{} would then select the experimental design method best suited to the problem, with \ac{bo} among several options.
It would define the search space, select suitable features and representations, incorporate prior data, and configure the optimizer.
\opt{} would also be able to autonomously construct physics-informed parameter representations, using properties computed with the PySCF execution graph~\cite{liIntroducingGPUAcceleration2025} or lower-cost proxies obtained from semiempirical quantum-mechanical methods~\cite{bannwarthExtendedTightbindingQuantum2021,froitzheimGxTBGeneralPurposeExtended}.
This would allow the surrogate model to exploit physically meaningful similarities between categorical candidates rather than treating them as unrelated labels.
As the campaign runs, \opt{} would analyze the resulting evidence and decide whether to continue, reformulate, or stop, adapting the strategy as the scientific question evolves.
A workflow orchestrator would translate these decisions into instructions for how, where, and when each experiment is executed.
Persistent campaign state and event-driven monitoring would record every decision made by \opt{}, including the chosen design method and the rationale behind it. This would allow researchers to review its decisions and trace how scientific choices led to actions across computational and laboratory systems.\footnote{A related interface-based separation of agent reasoning, durable session state, and execution environments is described in \href{https://www.anthropic.com/engineering/managed-agents}{Anthropic's Managed Agents architecture}.}
When multiple campaigns run concurrently, their agents could query one another for structured information about objectives, observations, decisions, and resource requirements.
This would allow each agent to retrieve relevant evidence on demand without exchanging or merging the campaigns' full contexts.
With shared abstractions for workflows, data, and provenance~\cite{Gottstein2026foundational}, agentic \acp{sdl} could extend beyond individual laboratories to form distributed research networks that coordinate scientific discovery across institutional and geographic boundaries~\cite{bai2024dynamic,strieth-kalthoffDelocalizedAsynchronousClosedloop2024,Gaidimas2026Reimagining}.

\section*{Data availability}

The raw data underlying all showcases and benchmark evaluations reported in this work are available at \url{https://github.com/the-matter-lab/La-Agente-Optima-artifacts}.
The Supporting Information covers the BO-MCP (accessible at \url{https://github.com/AccelerationConsortium/bo-mcp}) system architecture and implementation, the framework-comparison benchmark, the computational and experimental showcases, and the detailed RoboChem-Flex campaign record and resource accounting.

\section*{Acknowledgements}
We gratefully acknowledge the longstanding contributions of the Matter Lab’s current and past group members (\url{https://matter.toronto.edu}), in particular, the El Agente subgroup.
M.M. was partially supported through a collaborative partnership with Merck KGaA, Darmstadt, Germany.
J.B. acknowledges funding from the Eric and Wendy Schmidt AI in Science Postdoctoral Fellowship Program, a program by Schmidt Futures.
E.S. and T.N. acknowledge the generous funding provided by the European Innovation Council through the reaCtor project (grant No. 101099405). E.S., T.N., and S.P. acknowledge funding from the European Union through an ERC Proof of Concept Grant (RoboChem, grant No. 101246252). T.N. gratefully acknowledges funding from the Dutch Research Council (NWO) under the Talent Programme VICI (SynthBot, grant No. 20453).
A.A.-G. thanks Anders~G.~Fr{\o}seth for his generous support.
A.A.-G. also acknowledges the generous support of Natural Resources Canada and the Canada 150 Research Chairs program.
This research is part of the University of Toronto’s Acceleration Consortium, which receives funding from the CFREF-2022-00042 Canada First Research Excellence Fund, and was supported by the Defense Advanced Research Projects Agency (DARPA) under Agreement No. HR0011262E022 and the AI2050 program of Schmidt Sciences.

\clearpage

{
    \small
    \bibliography{references}
    \bibliographystyle{assets/plainnat}
}

\clearpage

\appendix
{\Huge \textbf{Supporting Information}}
\startcontents[appendices]
\section*{Contents}
\printcontents[appendices]{l}{1}{\setcounter{tocdepth}{2}}

\section{BO-MCP: system architecture and implementation details}
\label{si:bomcp}

This section expands on the BO-MCP implementation summarized in the main text: how a request reaches the optimization engine (\cref{si:bomcp:architecture}), how the backends are selected (\cref{si:bomcp:backends}), how campaigns are tracked and versioned (\cref{si:bomcp:lifecycle}), how the system stays auditable and explainable (\cref{si:bomcp:provenance}), and how molecular design spaces are represented (\cref{si:bomcp:molecular}).

\subsection{Request flow and architecture}
\label{si:bomcp:architecture}

\begin{figure}[h]
    \centering
    \includegraphics[width=\linewidth]{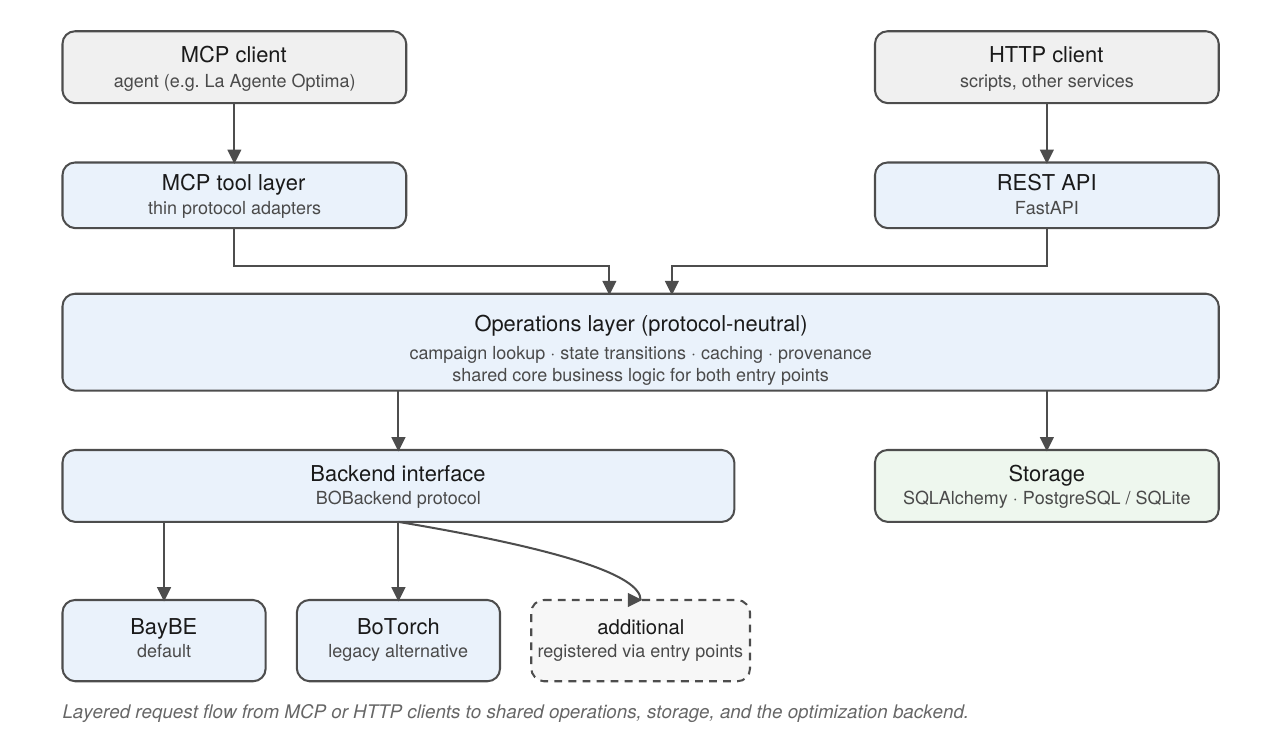}
    \caption{Layered request flow in BO-MCP.
        An MCP client (an agent) and an HTTP client reach the same protocol-neutral operations layer through separate thin protocol adapters, so both interfaces share core business semantics.
        The operations layer coordinates two parallel components: a pluggable optimization backend, BayBE by default, BoTorch as a legacy alternative, or a third-party backend registered via a Python entry point, and a SQLAlchemy storage layer backed by PostgreSQL or SQLite.}
    \label{fig:si-bomcp-architecture}
\end{figure}

BO-MCP has a layered architecture, implemented in Python using FastMCP for the MCP interface, FastAPI for the REST interface, Pydantic for data schemas and SQLAlchemy for persistence.
MCP tools and REST routes act as thin protocol adapters: they handle transport-specific concerns such as request parsing, identity, idempotency, and response formatting before invoking a shared, protocol-neutral operations layer.
That operations layer implements the core campaign-orchestration logic, including campaign lookup, lifecycle transitions, caching, and provenance persistence, and coordinates two parallel lower-level components: a pluggable optimization backend and a persistent storage layer.
Because MCP and REST invoke the same operation functions, they share core business semantics, while retaining transport-specific validation, authorization, and presentation behaviour.
Storage is a SQLAlchemy layer that supports both PostgreSQL, for shared deployments, and SQLite, for isolated single-user use.
\Cref{fig:si-bomcp-architecture} summarizes this layering.

The tool layer exposes 22 MCP tools, grouped by workflow stage in \cref{tab:si-bomcp-tools}, together with eight read-only MCP resources that expose campaign, suggestion, and event state directly by URI (for example, \code{campaign://\{campaign\_id\}}) without requiring an explicit tool call.
Rather than reproduce the full one-by-one listing here, we summarize the tools by workflow stage in \cref{tab:si-bomcp-tools}.
Exact tool schemas are exposed dynamically through MCP tool discovery, REST schemas through generated OpenAPI documentation, and resource URIs and templates through MCP resource discovery, rather than published as a fixed, static reference.

\begin{table}[h]
    \centering
    \small
    \caption{MCP tools grouped by workflow stage (22 total).
        Eight further read-only MCP resources expose campaign, suggestion, and event state directly by URI.
        Exact schemas, and the URIs, templates, and descriptions of the resources, are discoverable dynamically rather than reproduced here.}
    \label{tab:si-bomcp-tools}
    \begin{tabular}{lp{0.1\linewidth}p{0.6\linewidth}}
        \toprule
        Stage        & Tools & Purpose                                                                                                                               \\
        \midrule
        Setup        & 3     & Create a campaign, validate a specification before committing it, list backend capabilities                                           \\
        Exploration  & 5     & List and filter campaigns, results, and suggestions; compare campaigns; export a full campaign record                                 \\
        Optimization & 3     & Generate the next batch of suggestions; explain why a suggestion was made; accept, reject, or expire a suggestion                     \\
        Results      & 2     & Submit results directly, or from an uploaded file                                                                                     \\
        Diagnostics  & 4     & Model health, convergence, and Pareto/hypervolume diagnostics; server health; batch status across campaigns; poll a long-running call \\
        Lifecycle    & 4     & Pause, resume, terminate, and reopen a campaign                                                                                       \\
        Transfer     & 1     & Discover campaigns suitable as transfer-learning sources                                                                              \\
        \bottomrule
    \end{tabular}
\end{table}

The MCP server can be run over three transports: stdio, for direct integration with an agent process; streamable HTTP, for a persistent, shared server that multiple MCP clients connect to concurrently; and a legacy SSE transport retained for older clients.
The REST API does not use any of these MCP transports and is not connected to the MCP server; it runs as an independent service that, in a typical deployment, shares only the same database and the same Python operations code with the MCP server, so MCP and REST clients observe the same persisted campaign state, subject to transport-specific authorization and response formatting, rather than receiving session-local copies.
The MCP HTTP transports do not currently provide built-in per-tenant authorization; network-exposed deployments are expected to rely on loopback binding or an authenticating proxy in front of the server.

\subsection{Backend abstraction and algorithm selection}
\label{si:bomcp:backends}

The optimization engine is accessed only through a backend protocol comprising fifteen methods and properties in total: property accessors for backend identity and for supported and conditionally supported features, schema hooks for parameter and backend-specific options, validation methods for specifications and capabilities, generation methods for initial designs and suggestions, and further methods covering hypervolume computation, duplicate detection, state updates after new results, batch-diversity computation, method selection, and diagnostics.
All of these accept and return plain Python types rather than backend-specific tensors or data frames.
Two backends implement this protocol today.
BayBE~\cite{fitznerBayBEBayesianBack2025} is the default and provides native mixed search spaces (continuous, discrete, and categorical parameters together), multi-objective optimization via a Pareto objective, and the molecular-representation support described in \cref{si:bomcp:molecular}.

BoTorch is retained as a legacy backend and additionally supports trust-region (TuRBO) search for high-dimensional problems, which the standard suggestion pipeline dispatches to automatically for single-objective campaigns once the search space reaches 20 or more parameters.
Three further modules, sparse-axis-aligned-subspace \ac{bo} (SAASBO) for very high-dimensional problems, multi-fidelity optimization, and ensemble-based (RGPE) transfer learning between campaigns, are also implemented, but are not yet dispatched automatically through the standard suggestion pipeline; campaign intake requesting one of these features is rejected with a structured backend-capability error rather than silently downgraded to a simpler method.
New backends are not hard-coded: they register under a Python entry-point group and are discovered at start-up, so a campaign can request a specific backend, or leave the choice to an automatic selector that checks which registered backend supports the requested features.

\begin{table}[h]
    \centering
    \small
    \caption{Feature support by backend, as currently implemented.
        Trust-region search is dispatched automatically within the standard suggestion pipeline for single-objective campaigns once the search space is large enough; SAASBO, multi-fidelity, and transfer learning beyond the task-type mechanism are implemented on the BoTorch backend but not yet dispatched through that pipeline, and campaign intake requesting one of them directly is rejected with a structured backend-capability error rather than silently downgraded.}
    \label{tab:si-bomcp-backend-features}
    \begin{tabular}{p{0.24\linewidth}p{0.32\linewidth}p{0.36\linewidth}}
        \toprule
        Feature                                  & BayBE (default)                              & BoTorch (legacy)                                                             \\
        \midrule
        Multi-objective                          & Yes                                          & Yes                                                                          \\
        Mixed search space                       & Yes, native                                  & Yes                                                                          \\
        Molecular/substance parameters           & Yes                                          & No, rejected as unsupported                                                  \\
        Trust-region (TuRBO) search              & Not supported                                & Yes, auto-dispatched for single-objective campaigns at 20 or more parameters \\
        High-dimensional sparse \ac{bo} (SAASBO) & Not supported                                & Implemented, not yet dispatched                                              \\
        Multi-fidelity                           & Not supported                                & Implemented, not yet dispatched                                              \\
        Transfer learning                        & Supported via a declared task-type parameter & Standalone RGPE module implemented, not dispatched                           \\
        Leave-one-out cross-validation           & No                                           & Posterior downdate at fixed hyperparameters, batched-refit fallback          \\
        \bottomrule
    \end{tabular}
\end{table}

\begin{table}[h]
    \centering
    \small
    \caption{Automatic model and acquisition-function selection on the BoTorch backend, applied when no explicit configuration is supplied.
        The BayBE backend uses its own internal selection logic and is not shown here.}
    \label{tab:si-bomcp-botorch-selection}
    \begin{tabular}{p{0.34\linewidth}p{0.15\linewidth}p{0.15\linewidth}p{0.28\linewidth}}
        \toprule
        Phase / problem                                                                                                      & Model        & Acquisition & Strategy                                    \\
        \midrule
        Initial design: fewer than max(2, $n_{\text{parameters}}$ + 1, configured \code{initial\_design\_size}) observations & None         & None        & Sobol space-filling design                  \\
        Model-guided, one objective, fewer than 20 parameters                                                                & SingleTaskGP & qLogNEI     & Standard acquisition optimization           \\
        Model-guided, one objective, 20 or more parameters                                                                   & SingleTaskGP & qLogNEI     & Trust-region (TuRBO) search                 \\
        Model-guided, two or more objectives                                                                                 & ModelListGP  & qLogNEHVI   & Standard acquisition optimization, no TuRBO \\
        \bottomrule
    \end{tabular}
\end{table}

For model-guided optimization, continuous search spaces use L-BFGS-B, purely categorical spaces use discrete enumeration over the categorical pool, and mixed continuous/categorical spaces use mixed discrete/continuous optimization.
Models with categorical inputs use a mixed kernel construction, combining a continuous kernel over the continuous parameters with a Hamming kernel over the categorical ones.

\subsection{Campaign lifecycle and concurrency}
\label{si:bomcp:lifecycle}

Each campaign is a versioned record with an explicit status: created, running, paused, completed, or failed.
Only the first four are reachable in the current implementation; failed is defined in the data model and consumed defensively wherever status is read (health scoring, transfer-candidate filtering, recommended next actions), but nothing currently writes it, so a campaign cannot yet end in that state.
Four actions move a campaign between the reachable states, summarized in \cref{tab:si-bomcp-lifecycle}.

\begin{table}[h]
    \centering
    \small
    \caption{The four campaign lifecycle actions and the states from which each is valid.
        Pause, resume, and terminate additionally return a no-op success, rather than an error, when a retried call finds the campaign already at the listed result state; reopen has no such fallback and always requires a completed campaign.}
    \label{tab:si-bomcp-lifecycle}
    \begin{tabular}{lll}
        \toprule
        Action    & Valid from                  & Result    \\
        \midrule
        pause     & running                     & paused    \\
        resume    & paused                      & running   \\
        terminate & created, running, or paused & completed \\
        reopen    & completed                   & running   \\
        \bottomrule
    \end{tabular}
\end{table}

These four are the explicit lifecycle actions; a campaign also moves automatically from created to running the first time suggestion generation succeeds, without a dedicated lifecycle action being invoked.

A version counter increases on every campaign-level mutation, lifecycle transitions, suggestion generation, and result submission, and is checked before each write: if two updates race, the second one to commit is rejected rather than silently overwriting the first, and the caller can retry against the latest state.
Because a network failure can leave a caller uncertain whether its previous request actually succeeded, pause, resume, and terminate treat a retry that finds the campaign already at the intended state as a success rather than an error.
Reopen is deliberately excluded from this behaviour: a fresh, never-completed campaign and a reopened one are indistinguishable once running, so treating a mistaken reopen as a silent no-op would hide a genuine error rather than surface it.
Manual suggestion-status transitions, accepting, rejecting, or expiring a suggestion, go through a separate atomic compare-and-update on the suggestion row and do not increment the campaign version; a diagnostics response cached before such a transition may therefore continue to reflect the pre-transition suggestion state until the cache entry expires.
Every lifecycle tool additionally supports a dry-run mode that validates and reports a proposed transition without committing it.

\subsection{Provenance, diagnostics, and explainability}
\label{si:bomcp:provenance}

Every successfully dispatched MCP tool call that returns an application result, including expected structured failures, is recorded as a compact audit event (a truncated summary of inputs, a success flag, and any error code), attributed to a campaign where applicable and queryable per campaign.
Audit writes happen in a transaction separate from the underlying operation and are best-effort by default: a failed audit write does not block or roll back the operation it describes, it only increments an internal failure counter so persistent gaps can be caught by monitoring.
An optional fatal mode is available for deployments that want audit-persistence failures surfaced to the caller rather than silently tolerated; because the audit write happens after the business operation has already been committed, in its own transaction, this mode cannot roll back the operation itself if the write fails.
Instead, it replaces the tool's returned result with a structured database-error envelope, surfacing the logging failure to the caller rather than silently continuing.
The \code{events://} MCP resource exposes the 50 most recent audit events for a campaign as an operational trace, not a complete or paginated audit export.

Failures the system anticipates (an invalid state transition, a malformed specification, a request for a feature the active backend does not support, and so on) are returned as structured errors carrying a stable code, a human-readable message, and machine-actionable recovery guidance, including whether the request is safe to retry, rather than a bare exception.

Diagnostics are computed on request and cached briefly (120 seconds by default), keyed to the campaign's version, so most mutations invalidate the cache immediately.
Manual suggestion-status transitions are the one exception, as noted above: because they do not increment the campaign version, a cached diagnostics response may continue to reflect the pre-transition suggestion state until that cache entry's window expires.
Available diagnostics include the current Pareto front and hypervolume for multi-objective campaigns, with the hypervolume history accumulated incrementally as results are submitted, model health and convergence indicators, and leave-one-out cross-validation, computed exactly on the BoTorch backend at fixed, already-fitted hyperparameters via a closed-form posterior downdate, with a batched-refit procedure as a fallback, and unavailable on the default BayBE backend.
Model diagnostics are likewise backend-dependent: the BoTorch backend reports kernel lengthscales for computational input dimensions, whereas BayBE provides SHAP-based importance for either experimental parameters or their computational representations when the optional \code{shap} dependency is installed and sufficient observations are available.

Every suggestion records the generation method, the iteration, and the random seed used to produce it.
Suggestions generated by the underlying model additionally carry a fuller provenance record where available: the fitted model type, the acquisition function and its value at the suggested point, and the model's predicted mean and uncertainty; these model-derived fields are not populated for suggestions that are not model-guided, such as those from the initial design.

\subsection{Molecular and other categorical representations}
\label{si:bomcp:molecular}

For campaigns whose categorical parameters denote molecules (the fragment libraries used in the OSL case study, for example), the BayBE backend can map each category label to a SMILES string and encode it through cheminformatics descriptors rather than treating the labels as unrelated one-hot categories.
Roughly three dozen encodings are exposed, mirroring BayBE's own descriptor and fingerprint options; the default is a Mordred-derived descriptor block (about 1800 descriptors), with lighter alternatives such as ECFP fingerprints, MACCS keys, or RDKit 2D descriptors available where a smaller representation is preferred.
A related mechanism accepts a user-supplied numeric descriptor table directly, for precomputed quantum-chemical or spectroscopic features, for example, instead of computing one from a SMILES string, so a representation that already exists does not need to be re-derived.
Because the BoTorch backend has no chemistry-aware encoding path, a campaign requesting either mechanism while explicitly pinned to BoTorch is refused with a typed error rather than silently degraded to plain one-hot categories.

Search spaces built from large fragment or category libraries can be combinatorially large.
The BayBE backend estimates the size of the resulting discrete space before building it and, above a configurable budget, builds a deterministically subsampled candidate set instead of the full combinatorial product.
The base-sampling seed is derived from a canonicalized representation of the search-space parameters and constraints, so the same search-space definition regenerates the same deterministic base sample across processes and rebuilds.
Valid observed and actionable pending configurations are then unioned into that sample; points that fall outside the declared parameter value pools or violate a constraint can still be dropped, with a warning.
Because campaign state can change between rebuilds, the final candidate frame can differ even when the underlying search-space definition does not.

This implementation, including each backend, the lifecycle state machine, and the diagnostics pipeline, is covered by an automated test suite.

\newpage

\section{Framework comparison}
\label{sec:framework-comparison-si}

As outlined in \cref{sec:implementation}, we propose an architecture where a specialist subagent authors the campaign, ensures programmatic execution of the optimization loop, and BO-MCP maintains campaign configuration and accumulated observations across iterations.
Here, we evaluate whether the design choices underlying \optima{} support stronger end-to-end optimization than three complementary architectural ablations.
The architecture comparison used \texttt{GPT-5.4} throughout.
Separately, we retained the proposed subagent-based architecture and varied the \ac{llm} assigned to the \ac{bo} specialist while keeping \texttt{GPT-5.4} as the coordinating main agent.

We compared four architectures:
\begin{itemize}
    \itemsep0em
    \item \textbf{\archspecialist{} (proposed):}
          the \ac{bo} specialist authors and smoke-tests a reusable BO-MCP campaign program, which the main agent then executes and monitors.
    \item \textbf{\archmainscript{}:}
          the main agent receives equivalent scientific and scripting guidance.
          It then authors and executes the BO-MCP campaign program itself.
    \item \textbf{\archtoolloop{}:}
          the main agent conducts the optimization through repeated BO-MCP tool calls, without a reusable campaign program, subagents, Python, or shell execution.
    \item \textbf{\archlocalbo{}:}
          the main agent implements and executes Bayesian optimization, state management, and result tracking locally without BO-MCP.
          The agent has full access to its local compute environment.
\end{itemize}
These targeted end-to-end ablations probe complementary aspects of the proposed separation of reasoning, execution, and campaign-state management; they are not a full factorial decomposition.

\begin{figure}
    \centering
    \captionsetup[subfigure]{font=small,labelfont=bf}

    \begin{subfigure}{\textwidth}
        \centering
        \caption{Architecture comparison.}
        \label{fig:framework-summary-quality}
        \label{fig:framework-summary-reliability}

        \resizebox{\linewidth}{!}{%
            \begin{tabular}{lrrrrrrrr}
                \toprule
                 & \multicolumn{4}{c}{Optimization performance}
                 & \multicolumn{4}{c}{Protocol reliability and resources} \\
                \cmidrule(lr){2-5}
                \cmidrule(lr){6-9}
                Architecture
                 & \shortstack{Ackley\\final}
                 & \shortstack{Ackley\\MBSF@60}
                 & \shortstack{Shields $\geq 90\%$\\runs; mean attempt}
                 & \shortstack{Shields\\MBSF@60}
                 & \shortstack{Full-protocol\\valid}
                 & \shortstack{Mean\\cost/run}
                 & \shortstack{Mean\\time/run}
                 & \shortstack{Mean\\tokens/run}                          \\
                \midrule
                \archspecialist{}
                 & \textbf{0.770}
                 & \textbf{0.442}
                 & \textbf{3/3; 25.0}
                 & 0.865
                 & 6/6
                 & \$1.61
                 & 11.05~min
                 & 2.72M                                                  \\

                \archmainscript{}
                 & 0.557
                 & 0.265
                 & 3/3; 38.3
                 & \textbf{0.891}
                 & 6/6
                 & \$0.83
                 & 8.49~min
                 & 1.24M                                                  \\

                \archtoolloop{}
                 & 0.319
                 & 0.153
                 & 3/3; 45.3
                 & 0.824
                 & 6/6
                 & \$1.42
                 & 10.21~min
                 & 2.41M                                                  \\

                \archlocalbo{}
                 & 0.411
                 & 0.171
                 & 1/3; 26.0
                 & 0.756
                 & 6/6
                 & \$0.26
                 & 2.12~min
                 & 0.18M                                                  \\
                \bottomrule
            \end{tabular}%
        }
    \end{subfigure}

    \vspace{0.8em}

    \begin{subfigure}[t]{\linewidth}
        \centering
        \includegraphics[width=\linewidth]{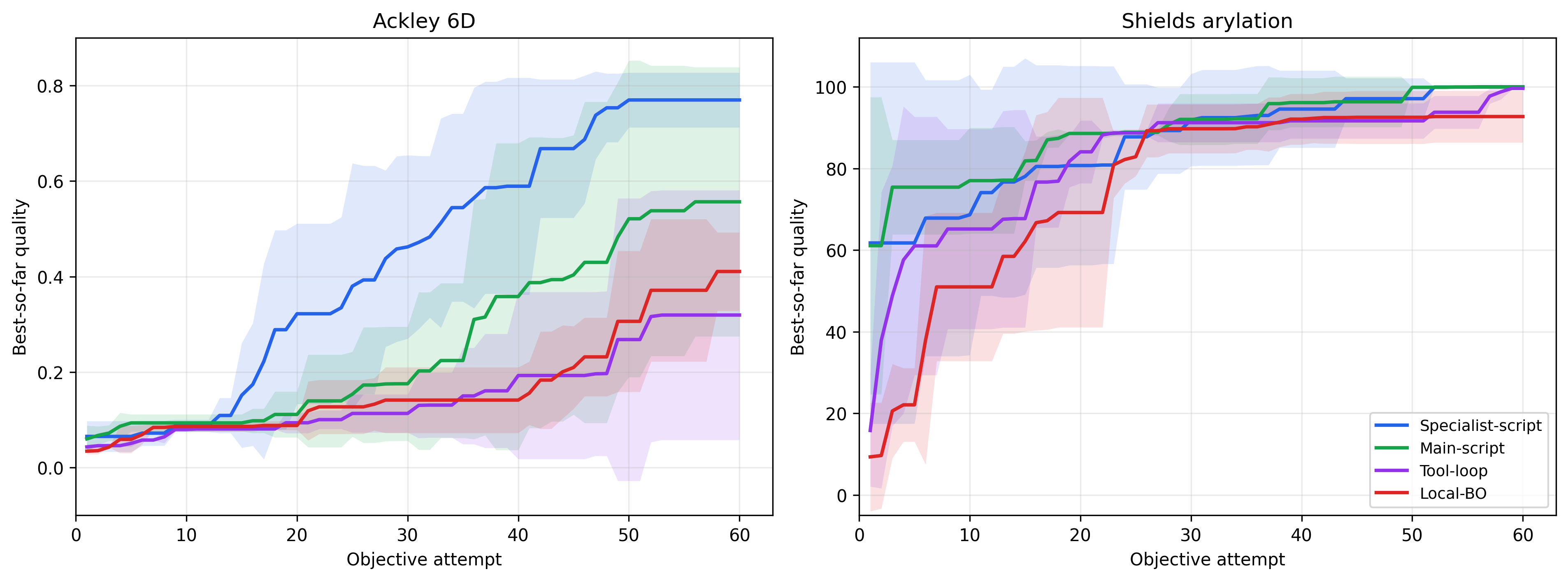}
        \caption{Optimization progress under the proposed architecture and its ablations.}
        \label{fig:framework-summary-convergence}
    \end{subfigure}

    \caption{
        \textbf{Framework architecture comparison.}
        \textbf{(a)} Architecture-level optimization performance, full-protocol reliability, and inference resources.
        For the \bencharylation{} 90\%-yield column, entries report the number of successful runs out of three and, after the semicolon, the mean attempt at which 90\% yield was first reached, calculated over successful runs only.
        Cost, time, and token values are means per requested run across the six runs for each architecture.
        \textbf{(b)} Mean best-so-far trajectories over three 60-attempt runs for each architecture--benchmark combination, with shaded regions showing the sample standard deviation.
    }
    \label{fig:framework-summary}
\end{figure}

We used two complementary single-objective benchmarks: (\textit{i}) a six-dimensional instance of the Ackley function~\cite{ackleyConnectionistMachine1987}, hereafter \benchackley{}, and (\textit{ii}) a palladium-catalyzed direct-arylation reaction-yield benchmark derived from the complete dataset reported by \citeauthor{shieldsBayesianReactionOptimization2021}, hereafter \bencharylation{}~\cite{shieldsBayesianReactionOptimization2021,fitznerBayBEBayesianBack2025}.
\benchackley{} is a deterministic, multimodal continuous optimization problem with many local optima.
The \bencharylation{} benchmark contains 1728 measured reactions spanning four bases, twelve ligands, four solvents, three concentrations, and three temperatures.
Ackley objective values were computed using the deterministic benchmark function, whereas \bencharylation{} yields were retrieved from the hidden reaction table for one selected set of conditions at a time.
Agents could not access, reconstruct, enumerate, or sort the underlying reaction table.
Across architectures, prompts varied only to implement the intended responsibility and tool boundaries.
Campaign design was left to the agent and therefore part of the evaluated task; stochastic settings such as the random seed were likewise not externally fixed.
Each architecture--benchmark combination was evaluated three times.
Every run received a global budget of exactly 60 attempted objective evaluations, including evaluations used during smoke testing, debugging, or restarted campaigns.
Prespecified run-validity and equal-budget comparability criteria are reported in the Supporting Information (\cref{sec:benchmark-validity}).
Optimization quality is reported in \cref{fig:framework-summary-quality} as final quality, defined as the best usable objective value observed within the 60-attempt horizon and reported in the native benchmark units, and as \ac{mbsf}@60, defined as the mean normalized best-so-far quality across those 60 attempts.
Specifically, if $b_t$ is the normalized best usable objective observed by attempt $t$, then
\[
    \mathrm{MBSF@60}
    =
    \frac{1}{60}\sum_{t=1}^{60} b_t.
\]
A failed attempt consumes the evaluation budget and does not improve the best-so-far curve.
For this calculation, \bencharylation{} yields were divided by 100, whereas \benchackley{} scores were already on a zero-to-one scale.
Mean best-so-far trajectories for each architecture--benchmark combination are shown in \cref{fig:framework-summary-convergence}.

\begin{figure}[!t]
    \centering
    \captionsetup[subfigure]{font=small,labelfont=bf}

    \begin{subfigure}{\textwidth}
        \centering
        \includegraphics[width=0.90\linewidth]{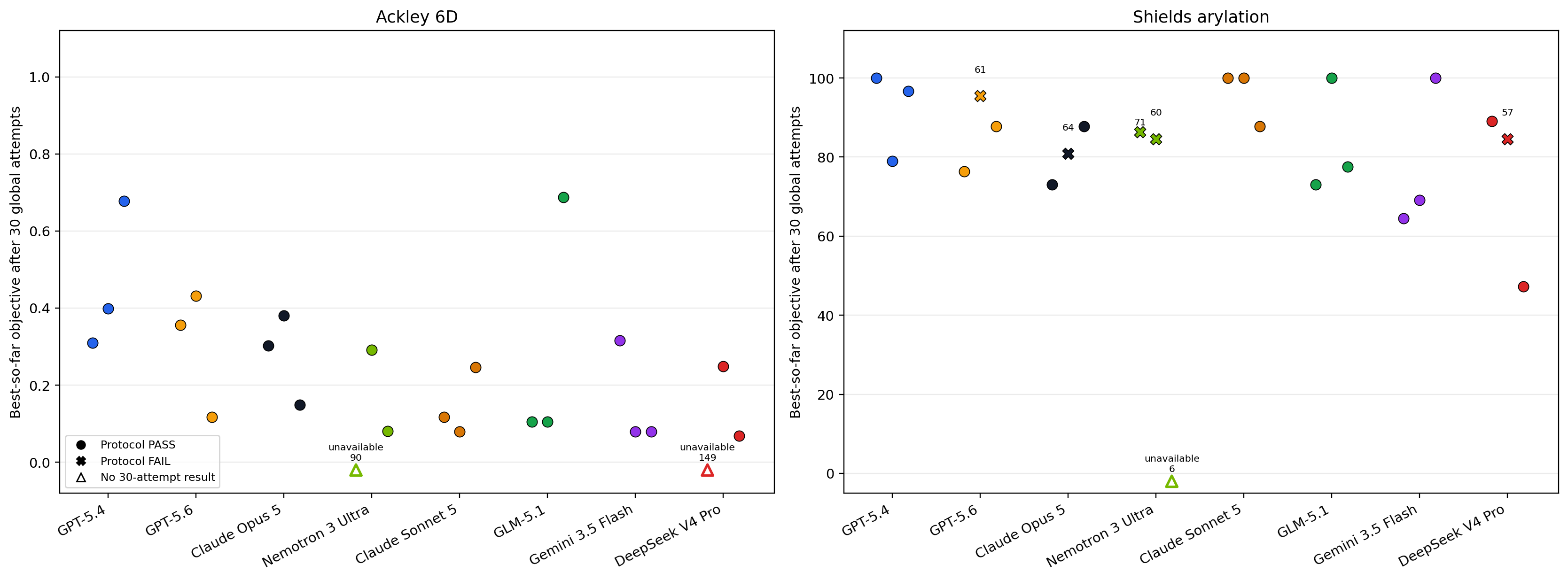}
        \caption{Individual outcomes at the 30-attempt horizon.}
        \label{fig:framework-summary-specialists}
    \end{subfigure}

    \vspace{0.6em}

    \begin{subfigure}{\textwidth}
        \centering
        \resizebox{0.54\linewidth}{!}{%
            \begin{tabular}{lcc}
                \toprule
                Specialist model
                 & \shortstack{Ackley\\Final / MBSF@60 ($n$)}
                 & \shortstack{Shields arylation\\Final / MBSF@60 ($n$)} \\
                \midrule
                \texttt{GPT-5.4}
                 & 0.770 / \textbf{0.442} (3)
                 & \textbf{100.00} / 0.865 (3)                           \\

                \texttt{GPT-5.6}
                 & 0.731 / 0.350 (3)
                 & 94.59 / 0.812 (2)                                     \\

                \texttt{Claude Opus 5}
                 & \textbf{0.772} / 0.344 (3)
                 & 94.85 / 0.785 (2)                                     \\

                \texttt{Nemotron 3 Ultra}
                 & 0.690 / 0.273 (2)
                 & 89.17 / 0.844 (1)                                     \\

                \texttt{Claude Sonnet 5}
                 & 0.422 / 0.207 (3)
                 & 96.57 / \textbf{0.907} (3)                            \\

                \texttt{GLM-5.1}
                 & 0.363 / 0.259 (3)
                 & 98.88 / 0.790 (3)                                     \\

                \texttt{Gemini 3.5 Flash}
                 & 0.397 / 0.204 (3)
                 & 92.21 / 0.637 (3)                                     \\

                \texttt{DeepSeek V4 Pro}
                 & 0.566 / 0.212 (2)
                 & 73.11 / 0.623 (2)                                     \\
                \bottomrule
            \end{tabular}%
        }
        \caption{Aggregate specialist-model optimization quality.}
        \label{fig:specialist-model-quality}
    \end{subfigure}

    \caption{
        \textbf{Specialist-model optimization quality and robustness under
            \archspecialist{}.}
        \textbf{(a)} Best-so-far objective values at the 30-attempt horizon.
        Models have the same left-to-right order in both benchmark panels,
        based on their average normalized 30-attempt best-so-far quality, final
        quality, and MBSF@60 across both benchmarks.
        Circles denote
        protocol-valid runs, filled X markers denote protocol failures, and
        open triangles identify runs without a reconstructable result at the
        30-attempt horizon.
        The triangles are offset below zero for visibility
        and do not represent objective values.
        Labels beside non-passing runs
        report the case-wide number of attempted objective evaluations.
        \textbf{(b)} Cells report mean final quality / mean MBSF@60.
        The number
        in parentheses indicates how many of the three requested runs were
        included in each mean.
        \bencharylation{} final quality is reported as
        yield (\%).
    }
    \label{fig:specialist-model-comparison}
\end{figure}

Overall, \archspecialist{} achieved the strongest performance on \benchackley{} and reached the maximum \bencharylation{} yield in all three runs, while all four architectures passed the full protocol.
Although \archmainscript{} achieved the highest mean MBSF@60 for \bencharylation{}, \archspecialist{} consistently reached the maximum yield across all three runs.
Together, these results suggest an advantage for established \ac{bo} implementations when combined with programmatic execution, while \archtoolloop{} shows that access to the optimizer service alone is not sufficient.
\archtoolloop{} is disadvantaged by the need to translate each function evaluation into a result submission to the BO-MCP backend and, conversely, to execute each function evaluation from the candidate proposed by BO-MCP; these steps must be performed explicitly at every iteration rather than programmatically.
While \archmainscript{} performs comparably to \archspecialist{} on \bencharylation{} at a noticeably lower cost, we argue that an important practical advantage of the latter is difficult to capture in static benchmarks such as those conducted here:
separating the \ac{llm} context used for script creation and campaign authoring from the domain-specific context (for example, reasoning about reaction-yield optimization) helps prevent technical implementation details from interfering with domain-level reasoning.
This context separation can also limit context-window growth during large optimization campaigns or subsequent follow-up interactions.
Detailed per-architecture results and resource interpretation are provided in the Supporting Information (\cref{sec:benchmark-detailed-results}).

We next evaluated eight specialist models under \archspecialist{}: \texttt{Claude Opus 5}, \texttt{Claude Sonnet 5}, \texttt{DeepSeek V4 Pro}, \texttt{Gemini 3.5 Flash}, \texttt{GLM-5.1}, \texttt{GPT-5.4}, \texttt{GPT-5.6}, and \texttt{Nemotron 3 Ultra}.
Optimization quality and workflow reliability were considered separately because model-level quality means include only scientifically comparable trajectories.

Across the specialist-model analysis, 48 runs for 8 different models, 2 different benchmark tasks, and 3 repetitions per combination were requested.
Of these, 41 were eligible for equal-budget analysis and 40 passed the full protocol.
The eight protocol-failing outcomes included seven failures of the global-budget criterion and one ownership-only violation.
These outcomes were reported as workflow results rather than treated as missing data; runs that were not scientifically comparable were excluded from equal-budget quality aggregates.
List-price accounting was exact for 47 runs; one \texttt{Nemotron 3 Ultra} \bencharylation{} run is reported as a lower bound because part of the \texttt{GPT-5.4} main-agent usage was not retained.
Individual outcomes and protocol failures are shown at the fixed 30-attempt horizon in \cref{fig:framework-summary-specialists}.
The early-stage snapshot makes differences among models and repeats, particularly on \bencharylation{}, easier to distinguish before trajectories approach their final values.

The specialist-model comparison summarized in \cref{fig:specialist-model-quality} shows that optimization performance is not captured by a single measure of model quality.
Across both benchmarks, \texttt{GPT-5.4} combined strong optimization performance with high robustness, passing the full protocol in all six runs, whereas models such as \texttt{DeepSeek V4 Pro} tended to produce weaker campaigns and \texttt{Nemotron 3 Ultra} yielded comparatively few scientifically comparable trajectories.
Differences between final quality and MBSF@60 further show that, even within the same \ac{bo} infrastructure, specialist models can author campaigns with distinct convergence behaviour and evaluation efficiency.
Because the specialist configures and validates the executable campaign rather than proposing candidates directly, these differences reflect how reliably and effectively each model translates a scientific task into an optimization procedure.
The changing model ordering between \benchackley{} and \bencharylation{} nevertheless indicates that performance remains task dependent rather than universally ranked.
Overall, most models produced valid campaigns under \archspecialist{}, while the observed failures show that optimization quality and operational robustness are distinct aspects of specialist performance.
One further caveat applies specifically to \texttt{GPT-5.4}: the framework was developed with \texttt{GPT-5.4} in the specialist role, and the instructions, tool descriptions, and failure modes addressed during development were shaped by its initial failures.
Its strong and robust performance may therefore partly reflect this development bias rather than a general capability advantage over the other specialists.
Given the small and unequal number of comparable trajectories, these results should be interpreted as behavioural and architectural trends rather than as a model leaderboard.

\subsection{Framework-comparison evaluation details}

For \benchackley{}, each normalized coordinate $x_i\in[0,1]$ was mapped to
$z_i=-40+80x_i$.
The conventional Ackley function was converted into the
maximization score
\[
    q_{\mathrm{Ackley}}(x)
    =
    1-\frac{f_{\mathrm{Ackley}}(z)}
    {20+e-e^{-1}},
\]
which lies on a zero-to-one scale.
The global optimum therefore corresponds
to $q_{\mathrm{Ackley}}=1$.

\subsubsection{Validity and aggregation criteria}
\label{sec:benchmark-validity}

Runs were assessed along separate validity dimensions.
\textbf{Global-budget validity} required exactly 60 attempted objective evaluations across smoke tests, debugging attempts, and restarted campaigns.
\textbf{Scientific comparability} additionally required valid objective values, the intended benchmark and backend, and a complete result-derived trajectory.

\textbf{Architecture validity} required compliance with each architecture's delegation, tool-access, execution, and script-artifact rules.
For \archspecialist{} specifically, the specialist had to author the campaign program and the main agent had to perform the production execution.
\textbf{Full-protocol validity} required the global-budget, scientific-comparability, architecture, and artifact checks all to pass.

Equal-budget quality aggregates included only globally budget-valid, scientifically comparable runs.
All requested outcomes were retained in the reliability analysis, including those excluded from quality aggregates.

\subsubsection{Detailed architecture comparison}
\label{sec:benchmark-detailed-results}

\benchackley{} clearly separated the proposed architecture from the three ablations (\cref{fig:framework-summary-quality}).
\archspecialist{} achieved the highest mean terminal quality and MBSF@60, substantially exceeding \archmainscript{}, \archlocalbo{}, and \archtoolloop{}.

\archmainscript{} retained programmatic execution and BO-MCP campaign management but removed specialist delegation.
On \benchackley{}, \archmainscript{} outperformed the other two ablations but remained below \archspecialist{}.
Because both
used BO-MCP and programmatic campaign execution, this difference is consistent with a benefit from specialist-led campaign planning.
\archtoolloop{} retained BO-MCP but required the main agent to mediate every step through tool calls.
It produced the lowest mean \benchackley{} performance while consuming nearly as many resources as \archspecialist{}, showing that access to a structured optimizer service was not sufficient to recover the performance of the proposed architecture under repeated \ac{llm}-mediated execution.
\archlocalbo{} benefited from programmatic execution but remained below \archspecialist{}; because it used an agent-authored local optimizer, this difference reflects the complete workflow change rather than an isolated campaign-management effect.

Across the observed \benchackley{} trajectories, the separation of \archspecialist{} became larger at later attempt horizons: \archmainscript{} was slightly ahead after 10 attempts, whereas \archspecialist{} led after 20 attempts and remained ahead through attempt 60.
This pattern is consistent with the intended division of responsibilities, in which the specialist determines campaign strategy while repetitive execution is delegated to a reusable program (\cref{fig:framework-summary-convergence}).

The \bencharylation{} benchmark was less discriminating because the three BO-MCP-based architectures approached the top of the finite yield landscape.
\archspecialist{} reached the maximum observed yield of 100\% in all three runs.
\archmainscript{} reached 100\%, 100\%, and 99.81\%, producing nearly identical mean terminal performance.

\archmainscript{} nevertheless had a slightly higher mean MBSF@60, 0.891 compared with 0.865 for \archspecialist{}.
The \archspecialist{} values were 0.984, 0.679, and 0.931, whereas \archmainscript{} produced 0.827, 0.915, and 0.932.
The difference was driven by one comparatively slow but valid \archspecialist{} run, which first reached 90\% yield at attempt 44 and 100\% at attempt 56.
We therefore interpret this benchmark as showing comparable sample efficiency between \archspecialist{} and \archmainscript{} with three repeats, with \archspecialist{} exhibiting perfect terminal success.
The principal performance advantage of \archspecialist{} appears on \benchackley{}, where campaign-design choices had greater opportunity to affect the search.

All 24 architecture runs passed the full protocol, so the quality comparison does not depend on omitting unsuccessful architecture runs.
\archspecialist{} used more resources than the ablations, consistent with the additional specialist interaction required for campaign construction and validation.
On \benchackley{}, this additional inference coincided with substantially stronger final quality and sample efficiency.

Taken together, these results cannot be explained by optimizer access, programmatic execution, or inference cost alone.
\archtoolloop{} retained BO-MCP but did not recover strong \benchackley{} performance, whereas \archlocalbo{} retained programmatic execution at substantially lower cost but remained below the two script-based BO-MCP architectures.
\archmainscript{} reduced resource use and performed comparably to \archspecialist{} on \bencharylation{}, but its lower \benchackley{} performance suggests that specialist-led campaign design becomes more valuable when optimizer configuration materially affects the search.
Because these are end-to-end architectural ablations with three repeats rather than a factorial experiment, the comparisons should be interpreted as evidence about complete workflows rather than isolated causal estimates for any single architectural component.

\newpage

\section{Supplementary material for the showcases}

\label{si:showcases}

\subsection{Phosphine ligand electronic tuning}
\label{si:phosphine}

This section documents a purely digital, multi-objective campaign in which \optima{} authored, executed, monitored, repaired and interpreted a finite-candidate \ac{bo} campaign over monodentate phosphine ligands, using the \grafico{} PySCF execution graph as the evaluator.
The task is a ligand-level proxy for catalyst tuning inspired by the phosphine example of \citet{laplazaGeneticOptimizationHomogeneous2022}, and was chosen because it is the compact opposite of the hardware campaigns of \cref{si:raise} and \cref{si:robochemflex:trace}: the search space is finite and enumerable, every evaluation is a deterministic quantum-chemical calculation, and the scientific interest lies entirely in whether the optimizer finds the trade-off structure of the objective set.

\subsubsection{Task definition and division of labour}
\label{si:phosphine:roles}

We tasked \opt{} to search neutral phosphines P(R$^1$)(R$^2$)(R$^3$) for an electronic profile consistent with a strong but not overly reducing donor for a hypothetical Ni-catalyzed cross-coupling.
The operator fixed the substituent library (alkyl Me, Et, iPr, tBu, Cy; aryl Ph, pMePh, pOMePh; electron-poor aryl pFPh, pClPh, pCF$_3$Ph, pCNPh), the four objective transformations and their target tolerances, the electronic-structure protocol, and the budget (8 warm-start ligands, then 10 \ac{bo} batches of 2).
Everything else --- symmetry reduction of the candidate table, the choice of warm-start ligands, the campaign intake, the evaluator, the reporting, and all plotting --- was left to the agent.
The brief explicitly forbade the agent from assembling or repairing molecular graphs during the loop and from using chemical judgment to pick candidates after the warm start.

Three components carried out the campaign.
The \emph{orchestrator} (main \opt{} agent, gpt-5.5) held the operator conversation, delegated code authorship, launched and watched the campaign through its shell and monitor tools, and performed all analysis and plotting.
A \emph{bo-pyscf-specialist} subagent (gpt-5.5) authored the campaign package --- a symmetry-reduced 364-row candidate table, the search space, intake, evaluator, reporting module, an executable entry point and a runbook (\texttt{HOW\_TO\_\allowbreak EXECUTE\_\allowbreak CAMPAIGN.md}) --- after first reading the BO-MCP OpenAPI description and validating its work with a compile check, a synthetic-evaluator end-to-end smoke test on a disposable campaign, and one real PySCF call on PH$_3$.
The \emph{BO-MCP service} owned all optimizer state, so that no \ac{bo} mathematics was performed by an \ac{llm}; the campaign script holds no local optimizer state and pauses the campaign at the end of every invocation so that a later invocation can resume it by identifier.
Ligand evaluation itself ran through the \grafico{} PySCF graph, whose internal routing agent (gpt-4.1) selected the molecular-analysis node after the single-point calculation.

\subsubsection{Campaign configuration}
\label{si:phosphine:config}

The candidate table enumerates all 364 unordered triples over the twelve substituents, each with a pre-validated SMILES string, and the \ac{bo} search space is the single categorical parameter \texttt{candidate\_id} --- so every proposal is by construction a chemically valid, already-enumerated ligand.
\Cref{tab:si-phosphine-config} lists the complete configuration.
Four raw objectives are minimized jointly: the absolute error of the HOMO energy to $-5.8$~eV (donor strength), the absolute error of the HOMO--LUMO gap to 5.0~eV (electronic stability), the molecular volume in excess of 350~\AA$^3$ (steric bulk), and the heavy-atom count (molecular complexity).
The phosphorus partial charge was recorded as an auxiliary donor descriptor with no optimization target of its own, as the brief requested.
Where the PySCF workflow does not expose a requested property directly, the agent substituted a named proxy applied identically to every ligand and recorded the proxy name in the artifacts --- an RDKit ETKDG/UFF volume and the Löwdin charge on phosphorus.

\begin{table}[htbp]
    \centering
    \small
    \caption{Configuration of the phosphine electronic-tuning campaign, as recorded in the agent-authored campaign intake and runbook.}
    \label{tab:si-phosphine-config}
    \begin{tabular}{ll}
        \toprule
        Setting                     & Value                                                                                       \\
        \midrule
        Parameters                  & \texttt{candidate\_id} (categorical, 364 symmetry-reduced ligands)                          \\
        Objectives                  & \texttt{donor\_homo\_error} $=|E_\mathrm{HOMO}+5.8~\mathrm{eV}|$ (tolerance 0.4~eV)         \\
                                    & \texttt{gap\_error} $=|E_\mathrm{gap}-5.0~\mathrm{eV}|$ (tolerance 1.5~eV)                  \\
                                    & \texttt{steric\_excess} $=\max(0, V-350)$ in \AA$^3$                                        \\
                                    & \texttt{heavy\_atom\_count}                                                                 \\
        Auxiliary descriptor        & phosphorus Löwdin charge (tracked, not optimized)                                           \\
        Scalarization / acquisition & Pareto / hypervolume improvement                                                            \\
        Batch size                  & 2 ligands per \ac{bo} suggestion request                                                    \\
        Random seed                 & 31841                                                                                       \\
        Evaluator                   & \grafico{} PySCF graph from SMILES, PBE/def2-SVP, neutral singlet,                          \\
                                    & restricted, molecular analysis only (no frequencies, no TD-DFT)                             \\
        Proxies                     & \texttt{rdkit\_ETKDG\_UFF\_ComputeMolVolume}, \texttt{pyscf\_lowdin\_atomic\_charge\_on\_P} \\
        Budget                      & 8 warm start $+$ 20 \ac{bo} (first invocation), $+$ 20 \ac{bo} (continuation)               \\
        End-of-run state            & campaign paused (resumable, not terminated)                                                 \\
        \bottomrule
    \end{tabular}
\end{table}

The agent selected the eight warm-start ligands itself and recorded a one-line rationale for each: PMe$_3$, PMe$_2$Ph, PMePh$_2$ and PPh$_3$ as an alkyl-to-aryl series, P(tBu)$_2$Ph and PCy$_3$ as bulky-alkyl probes, and P(pOMePh)$_3$ and P(pCF$_3$Ph)$_3$ as the electron-rich and electron-poor aryl boundaries.
This is a deliberate bracket of the electronic axis rather than a space-filling design.

\subsubsection{Execution and outcome}
\label{si:phosphine:outcome}

The campaign ran in two invocations against the same BO-MCP campaign (\texttt{f4e94d3d-0e06-43aa-baab-8bb15da9b843}) and the same cumulative artifact directory.
All 48 evaluations succeeded; no candidate failed, no suggestion was rejected as a duplicate, and 48 of 364 candidates (13\%) were ever computed.

After the first invocation the operator asked the agent how far to continue.
The agent recommended 5 further batches (10 ligands), arguing from its own improvement curve that the hypervolume gain was flattening while new Pareto members were still appearing, and proposed an explicit stopping rule (halt when the normalized hypervolume improves by less than ${\sim}1\%$ relative and no new ligand lands near the target region).
The operator overrode the recommendation and requested 20 further evaluations, which the agent executed.

\begin{figure}[htbp]
    \centering
    \includegraphics[width=\linewidth]{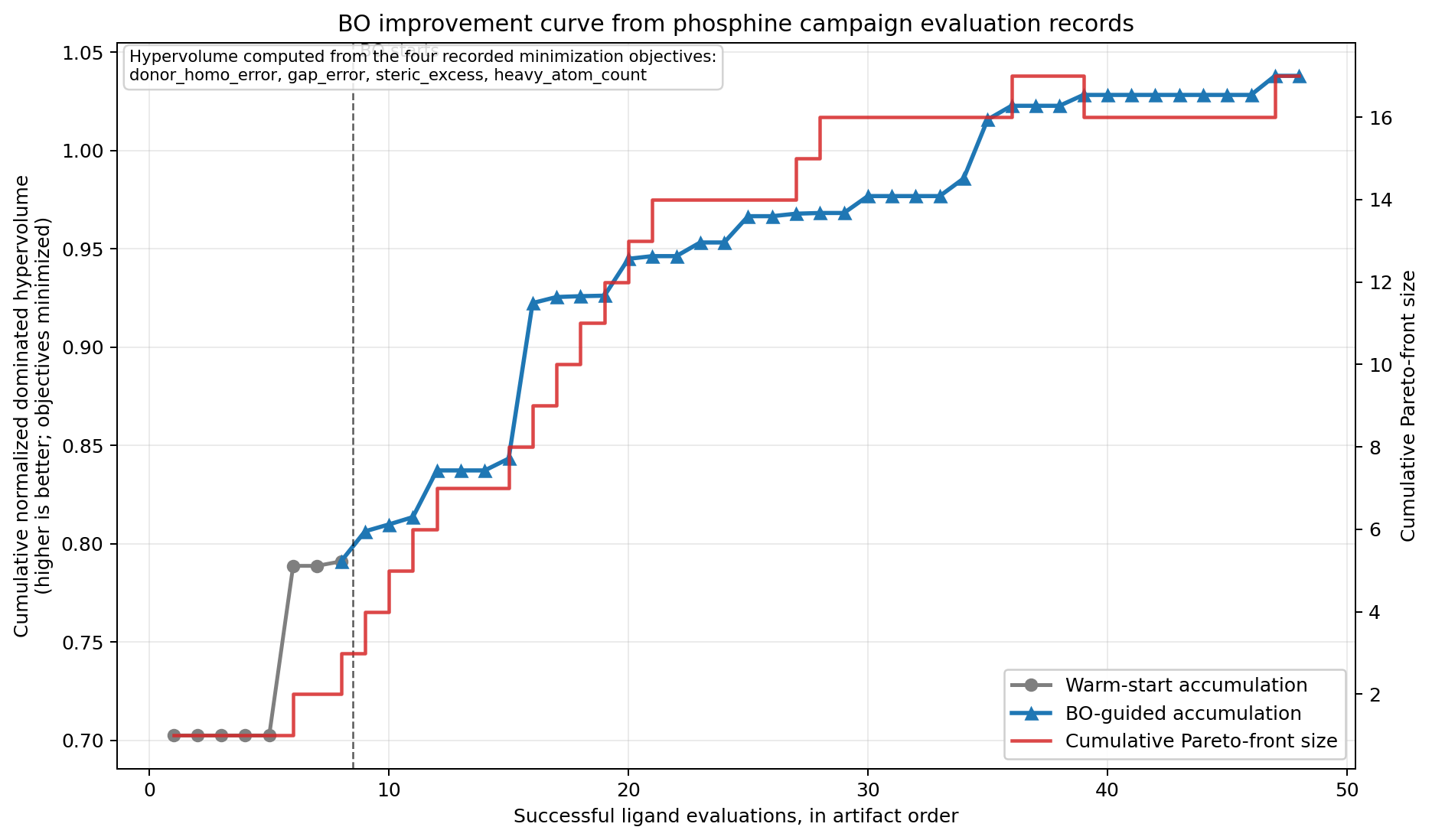}
    \caption{
        Multi-objective improvement curve for the phosphine electronic-tuning campaign, as generated by \opt{} from \texttt{evaluation\_records.jsonl} (axis labels and layout unmodified).
        Grey markers are the eight script-selected warm-start ligands, blue markers the 40 \ac{bo}-guided evaluations, and the dashed line marks the transition.
        The left axis shows the cumulative dominated hypervolume over the four minimized objectives, normalized against the range spanned by the complete record set; the right axis shows the size of the Pareto front observed so far, which is not monotone because a later ligand can dominate several earlier front members at once.
    }
    \label{fig:si-phosphine-improvement}
\end{figure}

\Cref{fig:si-phosphine-improvement} shows the improvement curve the agent produced from the evaluation records.
The normalized dominated hypervolume rises from 0.791 at the end of the warm start to 1.038 after 48 evaluations, and the observed Pareto front grows from 3 to 17 members, 14 of them discovered after the warm start.
The second invocation contributed most of the front's late growth but only ${\sim}7\%$ of the hypervolume, confirming the diminishing-returns regime the agent had diagnosed before it was asked to continue.

The chemistry the front encodes is a genuine conflict between the two electronic targets.
No ligand in the evaluated set satisfies both tolerances: within this family, pulling the HOMO down to $-5.8$~eV requires electron-poor aryl substituents, whose low-lying $\pi^*$ orbitals simultaneously collapse the gap to ${\sim}3$~eV, while the trialkyl phosphines that keep a 5--6~eV gap sit 0.8--0.9~eV too shallow.
The two warm-start boundary ligands therefore already hold the extreme positions --- P(pCF$_3$Ph)$_3$ gives the best HOMO error of the entire campaign (0.085~eV) and PCy$_3$ the best gap error (0.274~eV) --- and \ac{bo} did not beat either on its own axis.
What it did instead is the point of a multi-objective run: it found cheaper versions of the same electronics.
P(iPr)(pCF$_3$Ph)$_2$ reproduces the HOMO error of P(pCF$_3$Ph)$_3$ to within 0.03~eV while improving the gap error, removing seven heavy atoms and 55~\AA$^3$ of volume, and P(tBu)(Cy)$_2$ matches the gap error of PCy$_3$ with two fewer heavy atoms.
The resulting front spans 4 to 31 heavy atoms and is populated overwhelmingly by mixed methyl/alkyl ligands at the compact end and mixed alkyl/electron-poor-aryl ligands at the electronic end.

One objective turned out to be nearly inert.
The steric term is non-zero for exactly one of the 48 ligands (P(pCF$_3$Ph)$_3$, at 2.2~\AA$^3$ above the 350~\AA$^3$ threshold), because the operator-specified threshold lies above almost the entire reachable volume range of this substituent library.
The campaign is therefore effectively three-objective, with heavy-atom count carrying the size preference on its own.
This is a property of the brief rather than a failure of the optimizer, but it illustrates a practical hazard of specifying hinge-shaped objectives before the descriptor range is known.

\subsubsection{Failure mode and autonomous repair}
\label{si:phosphine:failures}

The campaign produced no chemical or optimizer failures, only one infrastructure defect.
The progress monitor of the first invocation went silent after four lines although the run completed normally; asked why, the agent traced this to its own evaluator, which wrapped each PySCF call in \texttt{contextlib.redirect\_stdout} --- a process-global redirect that, with two evaluator threads, captured the main thread's tagged progress lines into an in-memory buffer.
It added a lock around the redirected section, ran the continuation single-threaded, and, at the operator's prompting, had the \emph{bo-pyscf-specialist} subagent write the lesson to its persistent memory for future campaign packages.

\subsubsection{Resource usage and provenance}
\label{si:phosphine:resources}

The session spanned 38 agent turns over 3~h~11~min of wall-clock time, of which 14 were operator messages and the remainder autonomous responses to monitor events.
It comprised 128 \ac{llm} calls --- 71 by the main agent, 56 across the two \texttt{bo-pyscf-specialist} runs (both gpt-5.5), and one gpt-4.1 routing call inside the PySCF graph --- consuming 6.18M input tokens (5.62M served from cache) and 50.4k output tokens, of which 11.0k were reasoning tokens, at a total cost of \$7.08 (\$3.80 orchestrator, \$3.28 specialists).
Prompt caching accounts for a factor of 5.5 on the input side: the same 6.18M input tokens billed at the uncached rate would have cost \$30.9 rather than \$5.57.
The main agent made 33 tool calls (19 shell executions, 6 monitor queries, 2 subagent delegations with 2 waits, 2 monitor starts and 2 stops); the specialist runs made 54 in aggregate, dominated by file writes, code execution for validation, BO-MCP interface inspection and the two memory writes.
The 48 production DFT evaluations were performed by the campaign script rather than by the agent and consumed only 15.8~min of cumulative PySCF wall-clock time (median 18.8~s per ligand), yet the two invocations occupied 78~min of campaign time in total, the difference being suggestion generation and a fixed inter-batch pacing interval in the campaign script rather than computation.

The campaign workspace preserves the optimization itself --- the candidate table, the campaign package, the warm-start rationale, the per-evaluation records with descriptors, objectives and proxy names, the BO-MCP campaign export, the reports and the plotting scripts.
At the end of the session the agent collected all of these, together with the monitor logs and a SHA-256 manifest, into a single reproducibility bundle.

\newpage

\subsection{Singlet--triplet gap search over the Pollice 2021 library}
\label{si:pollice}

This section documents a single-objective, fixed-library \ac{bo} campaign in which \optima{} was asked to find the molecule with the smallest \ac{td-dft} singlet--triplet gap in the INVEST candidate set of \citet{polliceOrganicMoleculesInverted2021}.
It complements the phosphine campaign of \cref{si:phosphine} in three ways: the search space is two orders of magnitude larger (1512 molecules rather than 364), each evaluation is a conformer search followed by an excited-state calculation rather than a single ground-state job, and the underlying dataset carries published reference values, so the quality of the agent's cheap digital proxy can be measured rather than merely asserted.
It is also the campaign in which the infrastructure failed hardest, which makes it the more informative record of how the agent behaves when its evaluator stops working.

\subsubsection{Task definition and division of labour}
\label{si:pollice:roles}

The operator supplied the candidate table, the objective and the evaluation protocol in outline, and left the concrete choices to the agent.
For each molecule selected by the optimizer, \opt{} was to take \texttt{smiles\_canonical} as the only structural input, generate a small set of low-energy ground-state conformers, keep the lowest-energy one, run a fixed \ac{td-dft} single point on it, and report $\Delta E_\mathrm{ST}=E(\mathrm{S}_1)-E(\mathrm{T}_1)$ with the \ac{bo} objective defined as $-\Delta E_\mathrm{ST}$.
Everything else --- the electronic-structure method, the candidate filter, the molecular representation, the campaign package and all reporting --- was left to the agent, subject to operator review before execution.

The interactive design phase is worth recording because it shows the operator and the agent converging on a protocol rather than the agent executing a specification.
Asked which functional it would use, \opt{} proposed $\omega$B97X-D/def2-SVP on charge-transfer grounds and named PBE0/def2-SVP as the cheaper fallback; asked about molecule size, it profiled the table with RDKit and reported a median of 26 and a maximum of 85 heavy atoms; when the operator asked to drop the largest decile it derived the tie-inclusive cutoff \texttt{heavy\_atoms}~$<56$ itself, leaving 1512 of 1708 candidates.
Instructed to time one evaluation before committing to a budget, it ran a median-sized molecule end to end and estimated 2.5~min, of which the \ac{td-dft} step was inferred from file timestamps as ${\sim}30$~s; when the operator corrected this from the server logs to ${\sim}11$~min, the agent accepted the operator's number as the better source, revised the campaign estimate from 5.4~h to 6--15~h, and switched its own recommendation to the cheaper PBE0 fallback and a deliberately small budget.
That exchange, rather than the optimization, is the reason the campaign was affordable at all.

Three components then carried out the campaign, as in \cref{si:phosphine}.
The \emph{orchestrator} (main \opt{} agent, gpt-5.5) held the operator conversation, reviewed and launched the campaign, watched it through its monitor tools, and performed all diagnosis, analysis and plotting.
A \emph{bo-pyscf-specialist} subagent (gpt-5.5) authored the campaign package --- search space, intake, evaluator, campaign loop, entry point and runbook --- in a single delegated run.
The \emph{BO-MCP service} held all optimizer state, and the campaign script pauses the campaign at the end of every invocation so that a later invocation resumes it by identifier; the \grafico{} PySCF graph, with its internal gpt-4.1 routing agent, performed the excited-state calculations.

\subsubsection{Campaign configuration}
\label{si:pollice:config}

\Cref{tab:si-pollice-config} lists the configuration as recorded in the agent-authored intake.
The search variable is the single categorical parameter \texttt{molecule\_key}, so every suggestion is by construction a molecule that exists in the table.
Its BayBE custom descriptors are built by the agent from the SMILES string alone: twelve standardized RDKit 2D scalars, 32 truncated-SVD components of a 2048-bit Morgan/ECFP4 fingerprint, and a deterministic identity code that keeps otherwise indistinguishable rows separable.
Evaluation failures are never submitted as observations --- the corresponding suggestion is explicitly rejected in BO-MCP --- so the optimizer's posterior is never contaminated by an infrastructure fault.

\begin{table}[htbp]
    \centering
    \small
    \caption{Configuration of the Pollice 2021 singlet--triplet gap campaign, as recorded in the agent-authored campaign intake and runbook.}
    \label{tab:si-pollice-config}
    \begin{tabular}{ll}
        \toprule
        Setting              & Value                                                                                       \\
        \midrule
        Candidate pool       & 1708 tabulated molecules, filtered to \texttt{heavy\_atoms} $<56$ (1512 kept)               \\
        Parameter            & \texttt{molecule\_key} (categorical, 1512 levels)                                           \\
        Descriptors          & 12 standardized RDKit 2D scalars, 32 SVD components of a                                    \\
                             & 2048-bit Morgan/ECFP4 fingerprint, and an identity code                                     \\
        Objective            & \texttt{negative\_singlet\_triplet\_gap} $=-(E_{\mathrm{S}_1}-E_{\mathrm{T}_1})$, maximized \\
        Backend              & BayBE through BO-MCP                                                                        \\
        Initial design       & 5 molecules (seeded random, not curated)                                                    \\
        Batch size           & 2 molecules per \ac{bo} suggestion request                                                  \\
        Random seed          & 2021                                                                                        \\
        Conformers           & CREST/GFN2-xTB \texttt{imtd-gc} from SMILES, lowest-energy conformer only                   \\
        Electronic structure & \grafico{} PySCF graph, restricted PBE0/def2-SVP gas-phase \ac{td-dft},                     \\
                             & 5 states, neutral singlet, no solvation, no geometry optimization                           \\
        Timeouts             & 7200~s per evaluation, 5400~s per PySCF workflow                                            \\
        Budget               & 10 \ac{bo} loops, $+10$ (continuation), $+20$ requested (halted early)                      \\
        End-of-run state     & campaign paused (resumable, not terminated)                                                 \\
        \bottomrule
    \end{tabular}
\end{table}

\subsubsection{Execution and outcome}
\label{si:pollice:outcome}

The campaign ran in three invocations against the same BO-MCP campaign (\texttt{f023cf90-a1a1-470a-987d-134a38919812}) and the same cumulative artifact directory, with the operator setting each continuation budget.
Of 44 attempted evaluations, 39 succeeded and 5 failed; all 39 successes are distinct molecules, so 2.6\% of the filtered pool was ever computed.
The best molecule is a methylsulfinyl-substituted cycl[3.3.3]azine (\texttt{WGKMZGAJDYWUCE-UHFFFAOYSA-N}) with $\Delta E_\mathrm{ST}=0.223$~eV, $E(\mathrm{S}_1)=1.398$~eV, $E(\mathrm{T}_1)=1.175$~eV and an oscillator strength of $2.3\times10^{-4}$.

\begin{figure}[htbp]
    \centering
    \includegraphics[width=\linewidth]{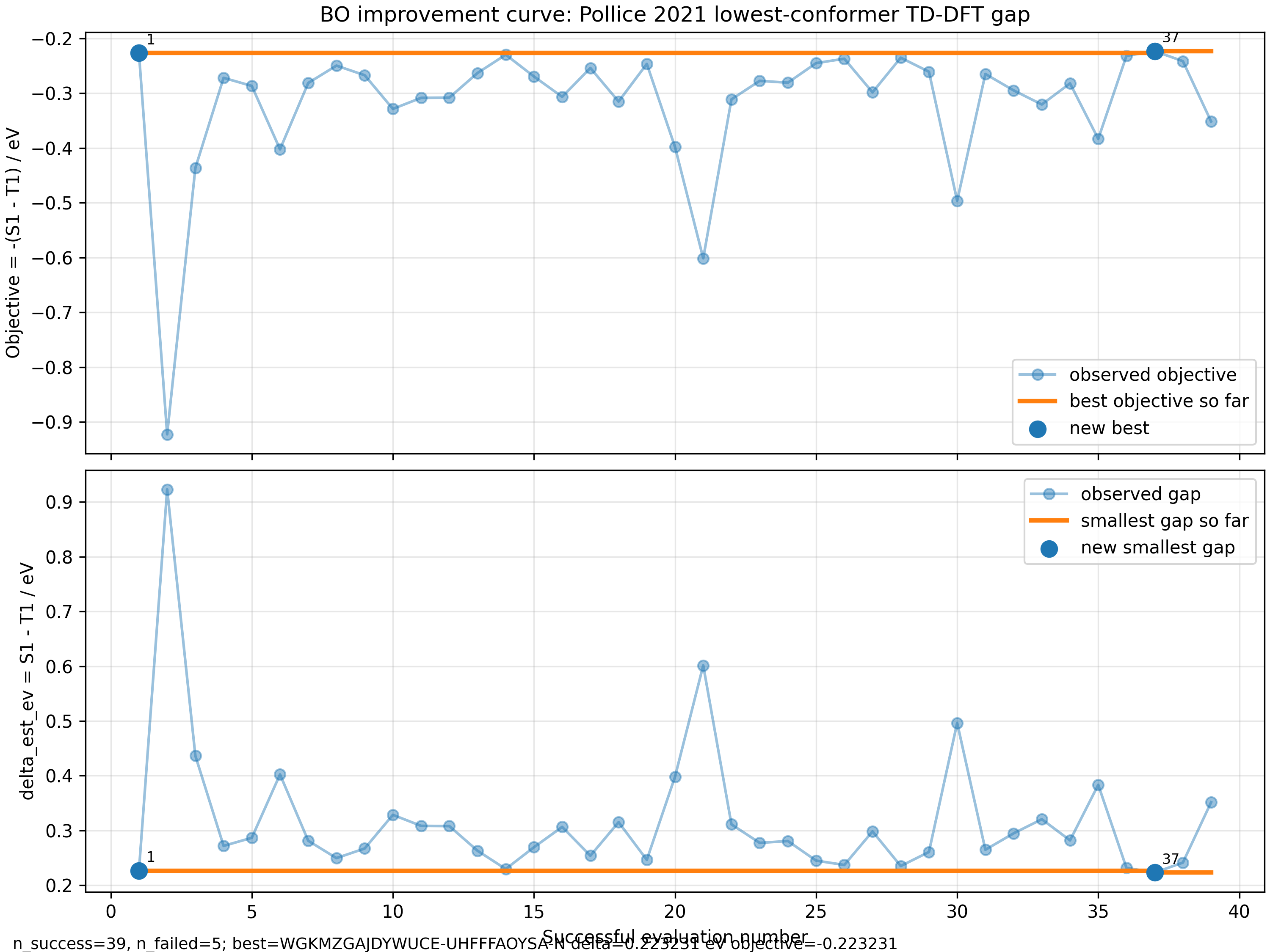}
    \caption{
        Improvement curve for the Pollice 2021 singlet--triplet gap campaign, as generated by \opt{} from \texttt{evaluation\_results.csv} (axis labels, annotations and layout unmodified).
        The two panels carry the same information with opposite sign, since the maximized objective is the negated gap; markers are the 39 successful evaluations in execution order, the step trace is the record so far, and the filled circles are the two records set.
        The five failed evaluations are omitted from the axis, as they never became observations.
    }
    \label{fig:si-pollice-improvement}
\end{figure}

\Cref{fig:si-pollice-improvement} shows the improvement curve the agent produced.
The best-so-far trace is almost flat, and honestly so: the first molecule of the random initial design was already a cyclazine at 0.227~eV, and only the 37th success improved on it, by 3.6~meV.
The optimizer's contribution is therefore not visible in the record trace but in where it spent the budget.
Of the 34 \ac{bo}-guided evaluations, 31 fall in the subset of the source table that carries at least one inverted-gap reference record, against a base rate of 42\% in the pool, and 21 of 34 landed below 0.30~eV; the mean observed gap fell from 0.429~eV over the five initial-design molecules to 0.302~eV over the \ac{bo}-guided ones.
Nine of the ten best molecules are monosubstituted cycl[3.3.3]azines and the tenth is a triazine-fused heptazine analogue, so the optimizer recovered the two scaffold families that motivate the dataset from a 1512-molecule table and 2048-bit fingerprint descriptors, without any chemical prior.

The campaign also quantifies the limits of its own evaluator, which is the reason for choosing a dataset with published reference values.
No evaluation returned an inverted gap: the smallest value reached is $+0.223$~eV, whereas the aggregated reference values shipped with the table put the same top molecules near $-0.34$~eV.
Across the 39 evaluated molecules the campaign's PBE0/def2-SVP gaps are offset from the reference values by $+0.33$~eV on average but track them well in rank (Spearman $\rho=0.79$, Pearson $r=0.89$).
This is the expected behaviour of adiabatic linear-response \ac{td-dft} for these systems, whose gap inversion requires the double-excitation character that a single-reference linear-response treatment omits, and it is a useful reminder that a cheap digital evaluator can be a good ranking device and a bad absolute one at the same time.
An agent-run campaign does not change that; what it does change is that the discrepancy is reconstructable from the preserved artifacts.

\subsubsection{Failure modes and autonomous response}
\label{si:pollice:failures}

Five evaluations failed, in two distinct modes, and neither was a chemistry failure.
One was a genuine cost overrun: a 47-heavy-atom triazine-fused system with two piperidinyl-aminophenyl arms exceeded the 7200~s per-evaluation ceiling and was killed and rejected.
Asked afterwards whether the timeout should be raised, the agent laid out the trade-off and the operator elected to keep 7200~s, so the ceiling stayed a deliberate budget rather than an accident.

The other four were an infrastructure fault, and the agent's handling of it is the interesting part.
In the third invocation, four consecutive evaluations failed inside \texttt{run\_pyscf\_workflow}.
After the third, the agent stated that this was a repeated pattern rather than a one-off and pre-committed to a stopping rule; when the fourth failed it created the campaign's \texttt{STOP} marker itself, so the loop halted before the next suggestion request, exported its artifacts and paused the campaign cleanly rather than being killed mid-batch.
When the operator later asked for a direct single-point calculation on water --- a minimal reproducer --- the same fault appeared in the foreground: \texttt{gpu4pyscf} and \texttt{cupy} were importable in the container but no CUDA device was present, so the mean-field object's \texttt{.to\_gpu()} call raised \texttt{cudaErrorNoDevice} instead of falling back.
The agent read the relevant graph source, patched \texttt{PyscfInput.pyscf\_mf} to catch the failure and continue on CPU, reran the water calculation successfully ($-76.276266$~Hartree), and flagged in the exported bundle that this patch is a host-environment fix outside the campaign package and must be reapplied or avoided on reproduction.

\subsubsection{Resource usage and provenance}
\label{si:pollice:resources}

The session comprised 24 operator messages and 89 autonomous responses to monitor events, and 34 tool calls by the main agent (21 shell executions, 3 monitor starts with 3 status queries, 3 direct PySCF workflow calls, 1 CREST call, 1 unit conversion, and 1 subagent delegation with its wait).
It consumed 191 gpt-5.5 calls --- 136 by the orchestrator and 55 in the single \emph{bo-pyscf-specialist} run --- plus 88 one-shot gpt-4.1 routing calls inside the PySCF graph, for 13.1M input tokens (12.1M served from cache) and 76.5k output tokens, of which 10.9k were reasoning tokens, at a total cost of \$13.01 (\$8.79 orchestrator, \$4.04 specialist, \$0.18 routing).
Prompt caching accounts for a factor of 5.2 on the gpt-5.5 side: the same traffic billed at the uncached rate would have cost \$67.3 rather than \$12.8.
The 44 attempted evaluations consumed 7.1~h of cumulative evaluator wall-clock time --- 4.98~h across the 39 successes, of which 3.91~h was PySCF and 1.07~h CREST, with a median of 6.5~min per molecule --- and, because the campaign script evaluates a batch of two concurrently, the three invocations occupied 6~h~15~min of campaign time in total.

At the operator's request the agent assembled a single reproduction bundle containing the input table, the campaign package, the intake payload, the per-evaluation records and BO-MCP export, the three monitor logs, the ranked results, the plotting script with its regenerated figure, and a metadata folder recording the software environment, the campaign snapshot and the out-of-package CPU-fallback patch.

\newpage

\subsection{Agent-to-agent cobalt catalyst tuning}
\label{si:cobalt}

This section documents a digital, multi-objective campaign on cationic Co(II) bisphosphine complexes in which \optima{} owned neither the molecular structures nor the optimizer state.
Molecular construction was delegated over the \ac{a2a} interface to \estructural{}, a separately deployed agent (\cref{sec:a2a}); optimizer state lived in BO-MCP; and evaluation was an unrestricted \ac{dft} geometry optimization on the \grafico{} PySCF execution graph.
The campaign is the transition-metal counterpart to the ligand-only phosphine campaign of \cref{si:phosphine}: the search space is again finite and enumerable, but each evaluation now requires a metal complex to be built from a textual ligand description before any electronic structure can be computed.

\subsubsection{Task definition and division of labour}
\label{si:cobalt:roles}

Taking the cationic Co(II) hydroformylation catalysts of \citet{hoodHighlyActiveCationic2020} as the starting point, we tasked \opt{} to search bidentate bisphosphines R$^1_2$P--linker--PR$^2_2$ for a Co(II) precursor-like complex balancing electronic activation, coordination stability, chelate geometry and steric crowding.
The operator fixed the four linkers (ethylene, propylene, 1,2-phenylene, cis-1,2-cyclohexylene), the eight substituents (Me, Et, iPr, Cy, Ph, $p$-Tol, $p$-Anisyl, $p$-CF$_3$-Ph), the four objective descriptions with their suggested proxies, the feasibility criteria, the complex to be built (\mbox{[Co(acac)(P$_2$)]$^+$}), the restriction to a modest geometry optimization without transition states, frequencies, \ac{td-dft} or catalytic-cycle calculations, and the budget (4 warm-start candidates, then 10 \ac{bo} iterations).
Everything else --- construction and symmetry reduction of the candidate table, the descriptor representation, the warm-start selection, the campaign code, the objective scoring functions, the reporting and all plotting --- was left to the agent.
The operator also required the complete candidate table to be reported and audited for duplicate unordered R$^1$/R$^2$ permutations before any calculation was allowed to start.

Four components carried out the campaign.
The \emph{orchestrator} (main \opt{} agent, gpt-5.5) held the operator conversation, delegated code authorship, launched and watched the campaign through its shell and monitor tools, and performed all analysis and plotting.
A \emph{bo-pyscf-specialist} subagent (gpt-5.5) authored the campaign package --- the 144-row candidate library, the descriptor vectors, the campaign intake, the evaluator, the reporting module, an executable entry point and a runbook (\texttt{HOW\_TO\_\allowbreak EXECUTE\_\allowbreak CAMPAIGN.md}).
The \emph{BO-MCP service} owned all optimizer state, so no \ac{bo} mathematics was performed by an \ac{llm}, and the campaign script pauses the campaign at the end of every invocation so a later invocation can resume it by identifier.
Finally, \estructural{} owned molecular construction: for every candidate the campaign script issued an \ac{a2a} task carrying only the ligand label and a natural-language connectivity description, and \estructural{} answered with an XYZ file written into the room-scoped workspace, using its own structure-generation and structure-editing tools.
Ligand evaluation then ran through the \grafico{} PySCF graph, whose internal routing agent (gpt-4.1) selected the molecular-analysis node after the geometry optimization.

The division is worth stating explicitly because no single component held the whole problem: \opt{} never manipulated atomic coordinates, \estructural{} never saw the objectives or the optimizer, and BO-MCP never saw a molecule.
The \ac{a2a} contract between the first two is a plain-text request and a filename; coordinates travel through the shared workspace and never enter an \ac{llm} context.

\subsubsection{Campaign configuration}
\label{si:cobalt:config}

The candidate table enumerates all unordered substituent pairs per linker, giving 36 candidates for each of the four linkers, 144 in total, of which 32 are symmetric (R$^1$ = R$^2$) and 112 unsymmetric, with no residual duplicate permutations --- a report the script regenerates and prints on every invocation before any calculation begins.
As in the phosphine campaign, the \ac{bo} search space is the single categorical parameter \texttt{candidate\_id}, so every proposal is by construction an already-enumerated ligand; the agent attached a nine-component custom descriptor vector to each category (linker size and rigidity, mean and difference of substituent steric bulk, aryl fraction, mean and difference of substituent electronics, a symmetry flag, and an identity code), so the surrogate model can exploit ligand similarity rather than treating the identifiers as unrelated labels.
\Cref{tab:si-cobalt-config} lists the complete configuration.

Two properties of that configuration matter for the interpretation below.
First, the electronic-activation objective is parsed from the PySCF checkpoint file rather than from the summary text, using the spin-resolved frontier orbitals, the Mulliken charge and the Mulliken spin population at cobalt, and taking the $\alpha$ SOMO for the unrestricted doublet; the per-evaluation record stores the frontier energies in both Hartree and eV together with the score components and their provenance.
Second, an evaluation counts as successful only if the workflow summary explicitly reports a completed geometry optimization and contains no failure indicator, so a converged \ac{scf} with an unconverged relaxation is treated as infeasible rather than scored.

\begin{table}[htbp]
    \centering
    \small
    \caption{Configuration of the agent-to-agent cobalt bisphosphine campaign, as recorded in the agent-authored campaign intake and runbook.}
    \label{tab:si-cobalt-config}
    \begin{tabular}{ll}
        \toprule
        Setting                     & Value                                                                          \\
        \midrule
        Parameters                  & \texttt{candidate\_id} (categorical, 144 symmetry-reduced ligands,             \\
                                    & 9 custom descriptors per category)                                             \\
        Objectives                  & \texttt{electronic\_activation} (maximize; frontier-orbital energy, Co charge, \\
                                    & Co spin population, parsed from the PySCF chkfile)                             \\
                                    & \texttt{coordination\_stability} (maximize; Co--P range, Co--P asymmetry,      \\
                                    & acac O,O-coordination, no dissociation)                                        \\
                                    & \texttt{chelate\_geometry} (maximize; P--Co--P bite angle, square-planar       \\
                                    & distortion)                                                                    \\
                                    & \texttt{steric\_crowding} (minimize; heavy atoms near Co, nonbonded contacts)  \\
        Scalarization / acquisition & Pareto / hypervolume improvement                                               \\
        Batch size                  & 1 ligand per \ac{bo} suggestion request                                        \\
        Random seed                 & 2020                                                                           \\
        Structure generation        & \estructural{} over \ac{a2a}, one task per candidate, room-scoped              \\
                                    & \texttt{context\_id}, XYZ returned through the shared workspace                \\
        Evaluator                   & \grafico{} PySCF graph from literal XYZ, charge $+1$, doublet,                 \\
                                    & unrestricted PBE/def2-SVP, geometry optimization to convergence                \\
                                    & (max.\ 200 steps, 7200~s timeout), then molecular/electronic analysis          \\
        Feasibility criterion       & explicit geometry-optimization completion, intact CoP$_2$O$_2$ core,           \\
                                    & no dissociation or severe collapse                                             \\
        Infeasibility handling      & finite hard penalty ($\mp100$ per objective) submitted as an observation       \\
        Budget                      & 4 warm start $+$ 10 \ac{bo}                                                    \\
        End-of-run state            & campaign paused (resumable, not terminated)                                    \\
        \bottomrule
    \end{tabular}
\end{table}

The agent selected the four warm-start candidates itself to bracket the design space rather than to fill it: \texttt{eth\_\_Me\_\_Me} (smallest symmetric alkyl case), \texttt{prop\_\_iPr\_\_Ph} (mixed alkyl/aryl on the flexible propylene linker), \texttt{ophen\_\_pAnisyl\_\_pCF3Ph} (rigid 1,2-phenylene with an electron-rich/electron-poor aryl contrast) and \texttt{cchex\_\_Cy\_\_pTol} (bulky cycloalkyl/aryl on the cis-cyclohexylene linker).

\subsubsection{Execution and outcome}
\label{si:cobalt:outcome}

The campaign (\texttt{62fb243b-265e-4ba4-b5a8-d97e414fce2f}) ran in a single invocation of 6~h~45~min and submitted all 14 planned observations.
Six were feasible and eight received the hard penalty.
All six feasible candidates carry the ethylene linker with Me, Et or iPr substituents; every propylene, 1,2-phenylene and cis-1,2-cyclohexylene candidate failed, as did the three ethylene candidates carrying cyclohexyl or $p$-CF$_3$-phenyl groups.
Seven of the eight failures are unconverged geometry optimizations --- the \ac{scf} converged in every one of them, so the failure is in the relaxation, not the electronic structure --- and the eighth (\texttt{eth\_\_pCF3Ph\_\_pCF3Ph}) exceeded the 7200~s workflow timeout.
Failure did not track molecular size alone: \texttt{prop\_\_Me\_\_Me} failed at 42 atoms while \texttt{eth\_\_iPr\_\_iPr} converged at 63, so the flexible and rigid linkers are harder to relax than their atom count suggests.

\begin{figure}[htbp]
    \centering
    \includegraphics[width=\linewidth]{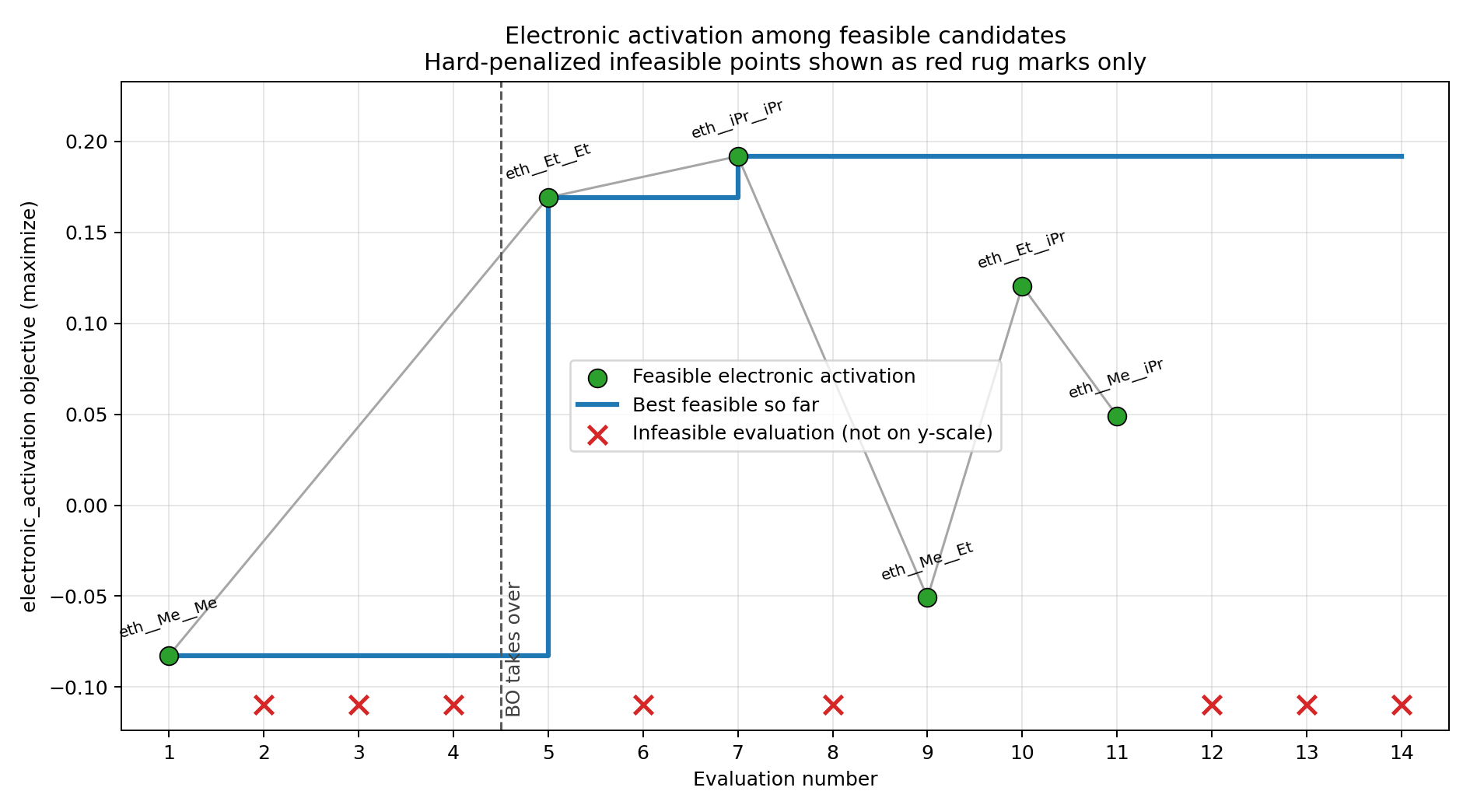}
    \caption{
        Electronic-activation objective over the course of the cobalt bisphosphine campaign, plotted by the reproducible script \opt{} wrote for this purpose (axis labels and layout unmodified).
        Green markers are the six feasible evaluations with the candidate identifier annotated, the blue step function is the best feasible value so far, and the red rug marks at the bottom are the eight hard-penalized infeasible evaluations, which are kept off the $y$-scale so that the decimal-scale differences among feasible candidates remain visible.
        The dashed line marks the transition from the four agent-selected warm-start candidates to the \ac{bo}-selected iterations.
    }
    \label{fig:si-cobalt-activation}
\end{figure}

Within the feasible subset the optimization behaved as intended.
The normalized feasible-only dominated hypervolume rises from 0.572 after the single feasible warm start to 1.000 after the sixth feasible evaluation, with four of the six feasible candidates on the observed Pareto front, and the best electronic activation improves from $-0.083$ (\texttt{eth\_\_Me\_\_Me}) to $0.192$ (\texttt{eth\_\_iPr\_\_iPr}); see \cref{fig:si-cobalt-activation}.
The front also reproduces, on a much smaller sample, the behaviour that made the phosphine campaign of \cref{si:phosphine} instructive: the best electronic activation sits at the highest steric-crowding score of the set, while \texttt{eth\_\_Et\_\_Et} reaches 0.169 --- within 12\% of it --- at the lowest crowding score, so the optimizer again found a less bulky ligand at essentially the same electronics rather than a single dominant winner.
All six feasible complexes retain the intended four-coordinate CoP$_2$O$_2$ core, with Co--P distances of 2.19--2.22~\AA{}, Co--O of 1.90--1.91~\AA{}, P--Co--P bite angles of 87--88$^\circ$ and Co--P asymmetries below 0.04~\AA{}.

The more informative outcome is how \ac{bo} responded to the feasibility structure.
Only one of the four bracketing warm starts was feasible, so after the warm start the optimizer had a single positive example, all of it on the ethylene linker.
It then spent 8 of its 10 suggestions on ethylene candidates although ethylene is only a quarter of the library, which is the correct response to a descriptor space in which one linker family carries all the observed signal, and it still probed the two remaining propylene suggestions and the bulkier ethylene substituents, both of which failed.
The campaign is therefore best read as a demonstration that the delegated loop is sound while the evaluator protocol is the binding constraint: at a 43\% feasibility rate the effective budget was 6 rather than 14 evaluations, which is too small to resolve a four-objective front over 144 candidates.
Raising the feasible fraction is a question of the geometry-optimization protocol and of the quality of the delegated starting structures, not of the optimizer or of the ligand chemistry.

\subsubsection{Resource usage and provenance}
\label{si:cobalt:resources}

The session spanned 111 agent turns across 4~d~23~h calendar time, of which 36 were operator messages and the remainder autonomous responses to monitor events; cumulative agent-run wall-clock time was 1.79~h.
It comprised 557 \ac{llm} calls with recorded token counts: 472 gpt-5.5 calls on the \opt{} side (main agent plus \texttt{bo-pyscf-specialist} delegations), 2 gpt-4.1 routing calls inside the PySCF graph, and 83 gpt-5.5 calls inside the 41 \estructural{} \ac{a2a} tasks.
Together these consumed 56.1M input and 210k output tokens at a total cost of \$53.29, split \$38.71 on the \opt{} side and \$14.58 on the \estructural{} side.
The main agent made 94 tool calls, as counted from the preserved conversation export; this excludes tool use internal to the specialist and \estructural{} conversations.
Prompt caching accounts for a factor of 6.0 on the input side: the same 56.1M input tokens billed entirely at the uncached rate would have cost \$281 rather than \$47.
The caching benefit is very unevenly distributed --- the per-call costs imply that ${\sim}96\%$ of the orchestrator's input tokens were served from cache against ${\sim}46\%$ of \estructural{}'s --- because each \ac{a2a} task is a short, fresh conversation with little prefix to reuse, whereas the orchestrator replays one long and growing history.
This is the practical cost of delegating structure generation to a separate agent: \estructural{} handled 8\% of the input tokens but 27\% of the spend.

The 14 production evaluations were performed by the campaign script rather than by the agent.
They consumed 4.87~h of cumulative GPU wall-clock time in the PySCF graph (median 17.1~min, maximum 60.6~min per candidate), against 19.3~min of cumulative \estructural{} task time for all 41 structure requests --- so structure generation cost roughly 6\% of the electronic-structure budget, and the failed relaxations, not the delegation, dominate the campaign's cost.

The campaign workspace preserves the candidate library, the agent-authored campaign package and runbook, the warm-start selection with per-candidate rationale, the per-candidate \estructural{} task responses and generated XYZ files, the PySCF consoles and optimized geometries, the per-evaluation records with descriptors and objective components, the BO-MCP campaign export, and the plotting scripts.

\newpage

\subsection{Xe/Kr separation over PORMAKE-assembled MOFs}
\label{si:mof}

This section documents a two-stage, multi-objective campaign in which \optima{} designed \acp{mof} for a Xe/Kr separation proxy, assembling each candidate with PORMAKE~\cite{Lee2021pormake} and scoring it with Zeo++~\cite{Willems2012zeopp}.
It is the only campaign in this series whose evaluator is purely geometric rather than quantum-chemical, which makes it the cleanest test of the agentic loop itself: an evaluation costs a few seconds, so nothing in the record is hidden behind expensive chemistry.
It is also the campaign in which the agent's first search-space parameterization was wrong in a way that is specific to combinatorial materials spaces, and in which the agent diagnosed and replaced that parameterization itself.

\subsubsection{Task definition and division of labour}
\label{si:mof:roles}

We tasked \opt{} to find \acp{mof} balancing Xe/Kr selectivity against capacity, representing every candidate exactly as a PORMAKE topology plus one node building block plus one edge building block.
The brief was adapted from the inverse-design study of~\citet{limFinelyTunedInverse2021}, who tuned \acp{mof} to a user-specified Xe/Kr selectivity obtained from grand-canonical Monte Carlo simulations.
Because no adsorption simulation is performed here, that objective was replaced by a geometric selectivity proxy computable with Zeo++, in the spirit of pore-limiting-diameter pre-screens used to narrow \ac{mof} libraries before adsorption calculations~\cite{zhouEfficientScreeningEnhanced2026}.
The operator fixed the nine admissible topologies (\texttt{pcu}, \texttt{dia}, \texttt{rtl}, \texttt{ths}, \texttt{bcu}, \texttt{srs}, \texttt{nbo}, \texttt{tbo}, \texttt{pts}), the qualitative objectives (a pore size suitable for Xe/Kr separation; pore volume as large as possible), the tools (PORMAKE and Zeo++) and a total budget of 30 evaluations, and explicitly left the batch size, iteration count and initial design size to the agent.
Everything else was left to the agent: the search over the PORMAKE database for usable node and edge building blocks, the functional form of both objective proxies, the campaign code, the reporting and all plotting.

The division of labour follows the phosphine campaign of \cref{si:phosphine}.
The \emph{orchestrator} (main \opt{} agent, gpt-5.5) held the operator conversation, delegated code authorship, launched and watched both campaigns through its shell and monitor tools, and performed all analysis and plotting.
A \emph{bo-pyscf-specialist} subagent (gpt-5.5), which also owns MOF work, authored the campaign package (search-space construction, campaign intake, evaluator, reporting module, an executable entry point and a runbook) and validated it with a compile check, a bounded one-evaluation \ac{bo} smoke test against a disposable campaign, and a direct PORMAKE/Zeo++ construction check before any production run.
The \emph{BO-MCP service} owned all optimizer state, so no \ac{bo} mathematics was performed by an \ac{llm}.
No \grafico{} PySCF graph was involved: the evaluator calls PORMAKE and Zeo++ directly.

\subsubsection{Campaign configuration}
\label{si:mof:config}

Rather than accepting the nine requested topologies as given, the agent's script inspects the installed PORMAKE database at runtime and excludes \texttt{rtl} and \texttt{tbo} with a tagged \texttt{[ALERT]} line, because neither can be represented by a single node building block; it then ranks the compatible node building blocks per topology by local-structure \ac{rmsd} (six retained each) and selects ten edge building blocks spread over the available edge lengths (1.14--8.47~\AA).
Both objectives are maximized: \texttt{selectivity\_proxy} is a bounded score that saturates at unity for a pore-limiting diameter in the 3.6--7.0~\AA{} window bracketing the Xe and Kr kinetic diameters and decays outside it, damped by a penalty on largest-cavity diameters above 12~\AA, while \texttt{capacity\_proxy} is the Zeo++ pore volume in cm$^3$/g.
\Cref{tab:si-mof-config} lists both campaign configurations.

\begin{table}[htbp]
    \centering
    \small
    \caption{Configuration of the two Xe/Kr MOF campaigns, as recorded in the agent-authored campaign intakes and runbooks.
        Settings shared by both campaigns are given once.}
    \label{tab:si-mof-config}
    \begin{tabular}{lll}
        \toprule
        Setting                             & First campaign                                                                                               & Refined follow-up                        \\
        \midrule
        Parameters                          & \texttt{topology} (7 categories)                                                                             & \texttt{candidate\_id} (109 categories,  \\
                                            & \texttt{node} (40), \texttt{edge} (10)                                                                       & decoding to \texttt{topology|node|edge}) \\
        Nominal / valid space               & 2800 / 420                                                                                                   & 109 / 109                                \\
        Objectives                          & \multicolumn{2}{l}{\texttt{selectivity\_proxy} (maximize; pore-diameter window, bounds $[0,1]$, weight 0.6)}                                            \\
                                            & \multicolumn{2}{l}{\texttt{capacity\_proxy} (maximize; pore volume, bounds $[0,10]$~cm$^3$/g, weight 0.4)}                                              \\
        Scalarization / acquisition         & \multicolumn{2}{l}{BayBE desirability, weighted geometric mean}                                                                                         \\
        Batch size                          & 3                                                                                                            & 5                                        \\
        Initial design size                 & 9                                                                                                            & 15 historical seed rows                  \\
        Budget (\texttt{max\_observations}) & 30                                                                                                           & $15 + 50 = 65$                           \\
        Random seed                         & \multicolumn{2}{l}{20260812}                                                                                                                            \\
        Evaluator                           & \multicolumn{2}{l}{PORMAKE \texttt{build\_by\_type} from topology/node/edge, then Zeo++ pore}                                                           \\
                                            & \multicolumn{2}{l}{diameter and pore volume; CIF written per successful candidate}                                                                      \\
        Infeasibility handling              & \multicolumn{2}{l}{zero on both objectives, submitted as an observation}                                                                                \\
        End-of-run state                    & \multicolumn{2}{l}{campaign paused (resumable, not terminated)}                                                                                         \\
        \bottomrule
    \end{tabular}
\end{table}

\subsubsection{The first campaign and its parameterization failure}
\label{si:mof:first}

The first campaign exposed a defect that only appears in this class of search space.
Representing the candidate as three independent categorical parameters, exactly as the brief phrased it, makes the nominal space $7 \times 40 \times 10 = 2800$ combinations, but a node building block can only serve a topology whose vertices have its connectivity, so only $42 \times 10 = 420$ triples, 15\% of the nominal product, are constructible at all.
BayBE has no way to know this, and 15 of the 30 evaluations were spent on incompatible topology--node pairs, which the evaluator rejected before any construction and submitted as zeros.
The optimizer nevertheless learned the constraint from those penalized observations: only one of the nine initial-design points was constructible, against 14 of the 21 \ac{bo}-selected points, and 18 of those 21 went to \texttt{pcu}, the topology carrying all observed signal.
All 15 successful candidates are \texttt{pcu}.

Asked by the operator whether it made sense to continue, \opt{} declined to simply resume.
It reported the 50\% invalid rate as the binding problem, noted that the incumbent had not improved since evaluation 12 and that all ten retained edges had already been paired with the best node, and offered three options: a refined second campaign restricted to validated compatible triples and seeded with the existing results, a short unchanged continuation as a sanity check, or a local screen around the incumbent.
It recommended the first, and the operator selected it.

\subsubsection{The refined campaign and outcome}
\label{si:mof:outcome}

The refined follow-up reuses the evaluator unchanged and replaces only the search-space concern.
The single \ac{bo} parameter is a finite \texttt{candidate\_id} that decodes to \texttt{topology|node|edge}, enumerated over the validated \texttt{pcu} family, its six compatible nodes, and 19 edges obtained by expanding the original ten with near-length neighbours from the PORMAKE database.
That gives 109 triples, every one of them constructible.
The 15 successful prior evaluations were submitted as historical seed rows, so the follow-up continued the same optimization rather than restarting it.
All 50 new evaluations succeeded; no candidate failed and no suggestion was rejected.

\begin{figure}[htbp]
    \centering
    \includegraphics[width=\linewidth]{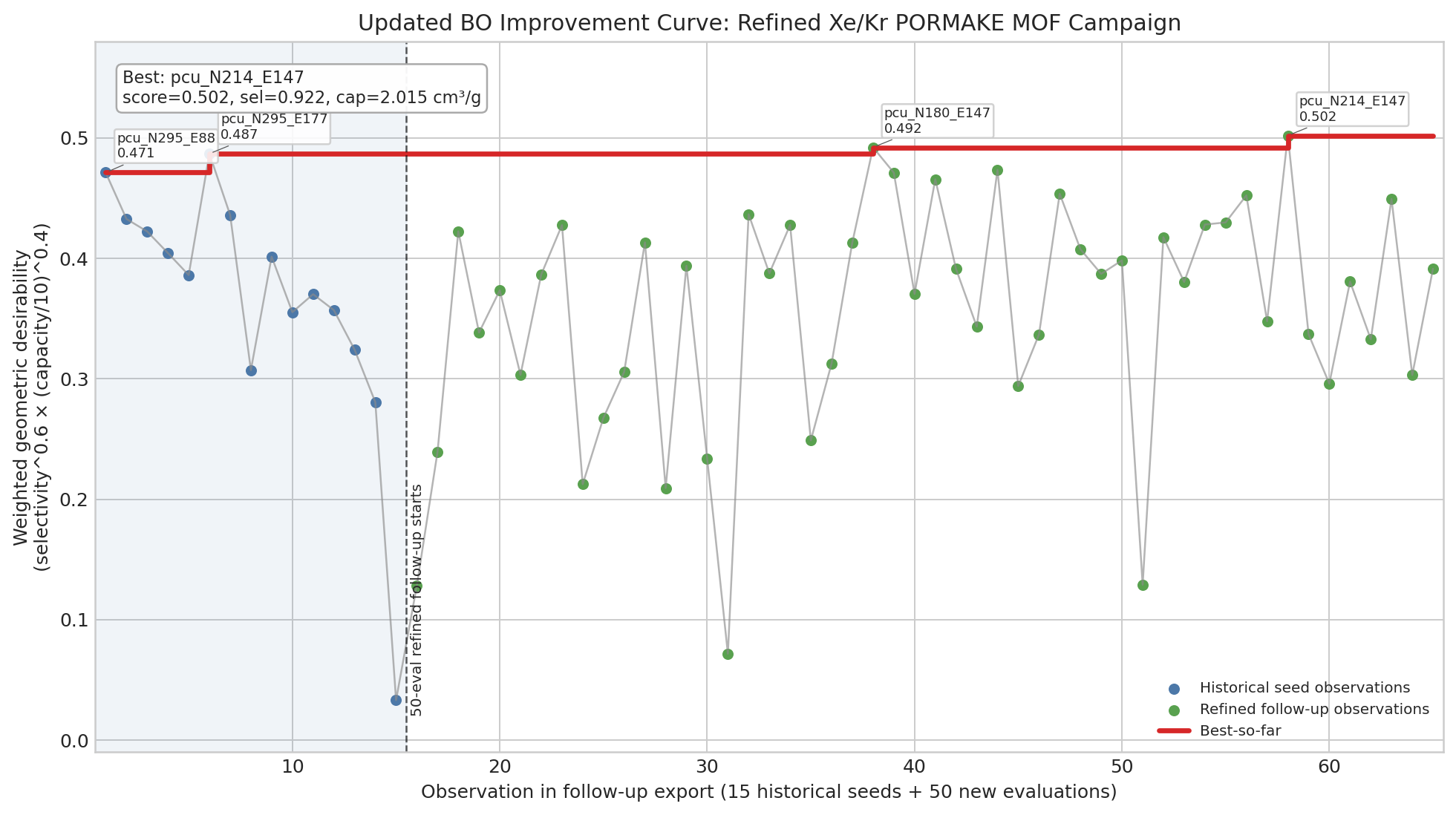}
    \caption{
        Improvement curve for the refined Xe/Kr MOF campaign, as generated by \opt{} from the BO-MCP campaign export (axis labels, annotations and layout unmodified).
        Blue markers are the 15 historical seed observations carried over from the first campaign, green markers the 50 evaluations of the refined follow-up, and the dashed line marks the transition.
        The ordinate is the weighted geometric desirability $\mathrm{selectivity}^{0.6} \times (\mathrm{capacity}/10)^{0.4}$ actually used by the BayBE backend, and the red step function is the best value observed so far.
    }
    \label{fig:si-mof-improvement}
\end{figure}

\Cref{fig:si-mof-improvement} shows the improvement curve the agent produced from the campaign export.
The best scalarized desirability rises from 0.487 (\texttt{pcu\_N295\_E177}, found in the first campaign) to 0.502 (\texttt{pcu\_N214\_E147}), and the observed Pareto front over the two raw objectives grows from 7 to 12 members, with 7 of the 12 contributed by the follow-up.
The optimizer spent its budget on the three node building blocks the first campaign had never reached (\texttt{N16}, \texttt{N180} and \texttt{N214} took 48 of the 50 evaluations) rather than revisiting \texttt{N295}, whose edge ladder was already exhausted.

The modest size of that gain is a property of the chemistry, not of the optimizer.
Across the 50 refined evaluations the edge building-block length alone explains most of the pore geometry, correlating with the pore-limiting diameter at $r = 0.93$ and with the pore volume at $r = 0.88$.
Within a fixed topology the two objectives are therefore governed by a single, nearly one-dimensional coordinate and are in direct conflict.
Every candidate reaching the maximum selectivity score carries an edge shorter than 2.2~\AA, and the resulting front is a smooth ladder from \texttt{pcu\_N295\_E88} (selectivity 1.00, 1.52~cm$^3$/g) through the balanced \texttt{pcu\_N214\_E147} (0.92, 2.02~cm$^3$/g) to \texttt{pcu\_N295\_E161} (0.005, 5.03~cm$^3$/g).
What the refined campaign bought is therefore not a qualitatively better material but a denser and better-resolved trade-off curve at no failed evaluations, which is the realistic return from a second-stage \ac{bo} run on a low-dimensional objective landscape.
It should be read as a demonstration that the agent can detect and repair a search-space specification error, not as a Xe/Kr screening result: the selectivity proxy is a geometric window score rather than an adsorption calculation, and none of the candidates was evaluated with the grand-canonical Monte Carlo simulations that a genuine adsorption-based screen requires~\cite{limFinelyTunedInverse2021}.
A published screen of this kind uses comparable pore-geometric criteria only as a first filter and then establishes performance by adsorption measurement~\cite{zhouEfficientScreeningEnhanced2026}; the campaign reported here stops at the filter stage.

\subsubsection{Resource usage and provenance}
\label{si:mof:resources}

The session spanned 78 agent turns over 1~h~20~min of wall-clock time, of which only 7 were operator messages; the remaining 93 incoming events were autonomous responses to campaign monitor output.
It comprised 162 \ac{llm} calls, 78 by the main agent and 84 across the two \texttt{bo-pyscf-specialist} runs, all gpt-5.5.
These consumed 11.6M input tokens (10.7M, or 93\%, served from cache) and 61.3k output tokens, of which 11.0k were reasoning tokens, at a total cost of \$11.44 (\$3.60 orchestrator, \$7.84 specialists).
Prompt caching accounts for a factor of 6.0 on the input side: the same 11.6M input tokens billed at the uncached rate would have cost \$57.8 rather than \$9.60.
The main agent made 25 tool calls (13 shell executions, 2 subagent delegations with 3 waits, 2 monitor starts, 3 monitor queries and 2 stops); the two specialist runs made 105 in aggregate, dominated by code execution for validation, file writes and edits, and two inspections of the BO-MCP OpenAPI description.

The cost structure is the inverse of the quantum-chemical campaigns.
The 80 production evaluations were performed by the campaign scripts rather than by the agent and consumed 2.1~min (30 evaluations) and 2.8~min (50 evaluations) of wall-clock time, roughly 3.7~s per candidate for PORMAKE construction and both Zeo++ analyses.
The entire scientific computation of this showcase therefore took under 5~min, against 80~min of session time and \$11.44 of \ac{llm} spend, or \$0.14 per evaluated MOF.
When the evaluator is this cheap, the agent, not the science, is the bottleneck and essentially the whole cost; the same orchestration overhead is negligible next to the 4.87~h of GPU time consumed by the cobalt campaign of \cref{si:cobalt}.

The campaign workspace preserves both campaign packages and runbooks, the inspected PORMAKE candidate space with its exclusion report, the refined compatible-candidate space with its seed mapping, the per-evaluation records with Zeo++ metrics and objectives, a CIF file for every successfully constructed MOF, both BO-MCP campaign exports, the monitor logs and the agent-written plotting scripts.

\newpage

\subsection{Contact angle matching with RAISE}
\label{si:raise}

This section documents a complete closed-loop optimization session in which \optima{} designed, executed, diagnosed, repaired, and interpreted a laboratory campaign on the RAISE \ac{sdl} platform~\cite{nazeriRAISESelfdrivingLaboratory2026}.
The archived execution trace spans 11 operator turns, and the campaign artifacts were preserved in the workspace throughout the session.
The following subsections describe the task and campaign implementation (\cref{si:raise:roles}), campaign chronology (\cref{si:raise:chronology}), final campaign configuration (\cref{si:raise:config}), failure handling and agent-initiated, operator-approved repair (\cref{si:raise:failures}), experimental outcome and interpretation (\cref{si:raise:outcome}), and resource usage and provenance (\cref{si:raise:resources}).

\subsubsection{Task definition and campaign implementation}
\label{si:raise:roles}

We tasked \opt{} to find an ethanol and sodium dodecyl sulfate (SDS) aqueous mixture whose static contact angle on the RAISE substrate matches a target of $65^{\circ}$, with a stopping tolerance of $\pm 1^{\circ}$ (the closed interval $[64^{\circ}, 66^{\circ}]$).
The initial search space was ethanol 0--60~v/v\% and SDS 0--1~w/v\%.
We additionally requested two warm-start formulations derived from a web search of the relevant literature and divided the campaign into small, explicitly approved increments.

For this campaign, the \emph{bo-raise-specialist} authored and revised the Python campaign package, executable entry point, and runbook (\texttt{HOW\_TO\_EXECUTE\_CAMPAIGN.md}), while \opt{} executed the operator-approved increments.
Before modifying the package, the specialist inspected the BO-MCP OpenAPI description to obtain the current service contract.
It validated the implementation with compile and command-line interface checks and, where applicable, dry runs confirming that only finite, in-bounds, non-penalty, and deduplicated historical observations were transferred.
It also used short smoke tests on disposable campaigns to verify suggestion retrieval, RAISE evaluation, and observation submission after relevant code changes.

BO-MCP generated suggestions through its BayBE backend~\cite{fitznerBayBEBayesianBack2025} and retained the campaign state.
The campaign programs held no local optimizer state, submitted measured objectives against suggestion identifiers using idempotency keys, and paused the campaign after each invocation so that it could be inspected and resumed after operator approval.
RAISE exposed \texttt{run\_raise\_experiment}, which accepts a formulation and returns its measured static contact angle.
Except for the final operator-authorized one-shot experiment (\cref{si:raise:outcome}), all measurements were initiated by campaign programs executed by \opt{}.

\subsubsection{Campaign chronology}
\label{si:raise:chronology}

The session produced three user-facing campaigns, an original, a constraint-corrected, and a clean reseeded campaign, summarized in \cref{tab:si-raise-campaigns}.
Campaign revisions were made only after operator approval, and the diagnostic turns preceding them were explicitly read-only.

\emph{Literature-informed initialization.}
Before writing code, \opt{} searched the literature and consulted the RAISE study, which reported that aqueous SDS approaches a contact-angle plateau near 70--72$^{\circ}$ and that adding ethanol lowers the contact angle further~\cite{nazeriRAISESelfdrivingLaboratory2026}.
The search also identified earlier measurements of aqueous SDS--ethanol mixtures on PTFE and PMMA surfaces~\cite{Zdziennicka2010}.
Using the RAISE trends as the quantitative basis, \opt{} selected two warm starts expected to approach the target from above: (ethanol $= 0$~v/v\%, SDS $= 0.60$~w/v\%) and (ethanol $= 15$~v/v\%, SDS $= 0.35$~w/v\%).

\emph{Original campaign.}
The two warm starts measured $71.403^{\circ}$ and $72.673^{\circ}$, and the first two \ac{bo} iterations reached $70.268^{\circ}$ at (0, 1.0).
The operator then approved ten additional iterations.
During this continuation, RAISE platform rejected the suggestion (60, 1.0) as infeasible, and the evaluator submitted a fallback observation of $180^{\circ}$ (\cref{si:raise:failures}).
The other nine evaluations remained below 20~v/v\% ethanol and did not improve on $70.268^{\circ}$.

\emph{Constraint correction.}
The operator then supplied a hardware constraint omitted from the initial brief: with two stock solutions, the platform could guarantee feasibility only up to half the concentration of each stock, limiting the 100\% ethanol stock to 50~v/v\% ethanol.
The specialist represented this constraint conservatively by reducing the ethanol upper bound, and rebuilt the campaign with the 13 valid observations from the original campaign while excluding its penalty row.
Two \ac{bo} iterations moved into the corrected high-ethanol region and improved the best observation to $68.610^{\circ}$ at (32.14, 1.0).
A five-iteration continuation added three valid observations and two $180^{\circ}$ penalty rows for feasible formulations for which RAISE returned no contact-angle value.

\emph{Diagnosis and clean reseed.}
During a read-only analysis, \opt{} identified the submission of $180^{\circ}$ for missing measurements as the source of the distorted observations.
It proposed a third campaign seeded with the valid data from both predecessors and revised the failure handling to retry a failed measurement without submitting a penalty value.
After operator approval, the specialist combined the two campaign histories.
Because the corrected campaign already contained the 13 valid observations transferred from the original campaign, merging both exports produced 13 duplicate records.
Removing these records and the three penalty rows left 18 unique valid observations for the clean campaign.
Five clean \ac{bo} iterations then completed without a measurement failure and improved the best observation to $68.108^{\circ}$ at (34.56, 0.752).

\emph{One-shot extrapolation.}
The operator then allowed exactly one further measurement due to time constraints.
\opt{} selected the maximum feasible ethanol concentration (50~v/v\%) with SDS by itself at the level of the best observed formulation (0.75~w/v\%), avoiding the high-SDS region where two evaluations had returned no contact-angle value.
The measurement returned $67.755^{\circ}$, the best result of the session, though still outside the target window.

\begin{table}[htbp]
    \centering
    \caption{Summary of the three BO-MCP campaigns in the RAISE contact-angle session.
        ``New evaluations'' includes the two warm-start measurements in the original campaign.
        ``Penalty rows'' denote failed evaluations submitted as $180^{\circ}$ observations.
        The final one-shot experiment is not included.}
    \label{tab:si-raise-campaigns}
    \begin{tabular}{lccccc}
        \toprule
        Campaign  & Search space (EtOH; SDS) & Seeded            & New evaluations & Penalty rows & Best angle       \\
        \midrule
        Original  & 0--60~v/v\%; 0--1~w/v\%  & 0 (2 warm starts) & 14              & 1            & $70.268^{\circ}$ \\
        Corrected & 0--50~v/v\%; 0--1~w/v\%  & 13                & 7               & 2            & $68.610^{\circ}$ \\
        Clean     & 0--50~v/v\%; 0--1~w/v\%  & 18                & 5               & 0            & $68.108^{\circ}$ \\
        \bottomrule
    \end{tabular}
\end{table}

\subsubsection{Final campaign configuration}
\label{si:raise:config}

\Cref{tab:si-raise-config} summarizes the final clean campaign as recorded in its intake, run context, and optimizer diagnostics.
\opt{} specified the backend, parameters, objective, batch size, and random seed while leaving the model stack at the BO-MCP defaults.
The resolved configuration used a \ac{gp} surrogate with a Mat\'ern kernel, no input warping, and qLogNoisyExpectedImprovement.
The two warm starts were submitted as ordinary observations rather than generated as an optimizer initial design, so the surrogate treated them identically to \ac{bo} observations.
The operator supplied the target, tolerance, iteration budgets, and feasibility constraint; \opt{} selected the remaining campaign settings.

\begin{table}[ht]
    \centering
    \caption{Configuration of the final clean campaign recorded in BO-MCP (campaign intake and optimizer diagnostics).}
    \label{tab:si-raise-config}
    \begin{tabularx}{\linewidth}{@{}l X@{}}
        \toprule
        Setting                         & Value                                                                                                        \\
        \midrule
        Parameters                      & Ethanol $\in [0, 50]$~v/v\% (continuous); SDS $\in [0, 1]$~w/v\% (continuous)                                \\
        Objective                       & \texttt{static\_contact\_angle}, match target $65^{\circ}$; minimize $|\theta-65^{\circ}|$                   \\
        Early stop                      & any measurement in $[64^{\circ}, 66^{\circ}]$                                                                \\
        Backend                         & BayBE (BO-MCP service)                                                                                       \\
        Surrogate                       & \ac{gp}, Mat\'ern kernel, no input warping                                                                   \\
        Recommender                     & BotorchRecommender                                                                                           \\
        Acquisition function            & qLogNoisyExpectedImprovement                                                                                 \\
        Batch size                      & 1                                                                                                            \\
        Random seed                     & 7                                                                                                            \\
        Historical seeding              & 18 finite, in-bounds, non-penalty observations; parameter tuples deduplicated after rounding to six decimals \\
        Failure handling                & retry same candidate up to 2 times, then expire suggestion; no penalty value                                 \\
        Iteration budget per invocation & 5 \ac{bo} iterations                                                                                         \\
        Hardware timeout                & 500~s per \texttt{run\_raise\_experiment} call                                                               \\
        \bottomrule
    \end{tabularx}
\end{table}

\subsubsection{Failure handling and agent-initiated repair}
\label{si:raise:failures}

The session encountered two classes of unsuccessful evaluation that required different treatment.
The first was \emph{formulation infeasibility}: the suggestion (60, 1.0) violated the two-reagent stock constraint, which no retry could resolve.
The second was a \emph{no-value evaluation}: for two feasible formulations near 30--36~v/v\% ethanol, RAISE reported ``contact angle measurement failed, retry experiment'' and returned no contact-angle value.
The measurement errors were mainly caused by a slight shift in the backlight position relative to the droplet, which the image-processing pipeline could not fully account for during contour detection.

The initial campaign package mapped both classes to the same fallback by submitting $180^{\circ}$, the theoretical non-wetting maximum, as the observed objective.
For the infeasible formulation, this penalty directed the optimizer away from an unusable point, but also distorted the surrogate along the ethanol axis: all eight subsequent suggestions remained below 20~v/v\% ethanol.
For the no-value evaluations, the synthetic observations fell inside the most promising region and biased the surrogate against those compositions without a measured contact angle.

During a read-only analysis, \opt{} distinguished the two cases and proposed the repair.
Formulation infeasibility was handled by correcting the search space, whereas a no-value evaluation was retried up to twice and then expired without submitting an objective value.
Following operator approval, all valid measurements were retained and only the three synthetic $180^{\circ}$ values were excluded.
The final campaign completed all five iterations without a failure or retry, so the expiration path was implemented but not exercised.

\subsubsection{Outcome and interpretation}
\label{si:raise:outcome}

\begin{figure}[ht]
    \centering
    \includegraphics[width=\linewidth]{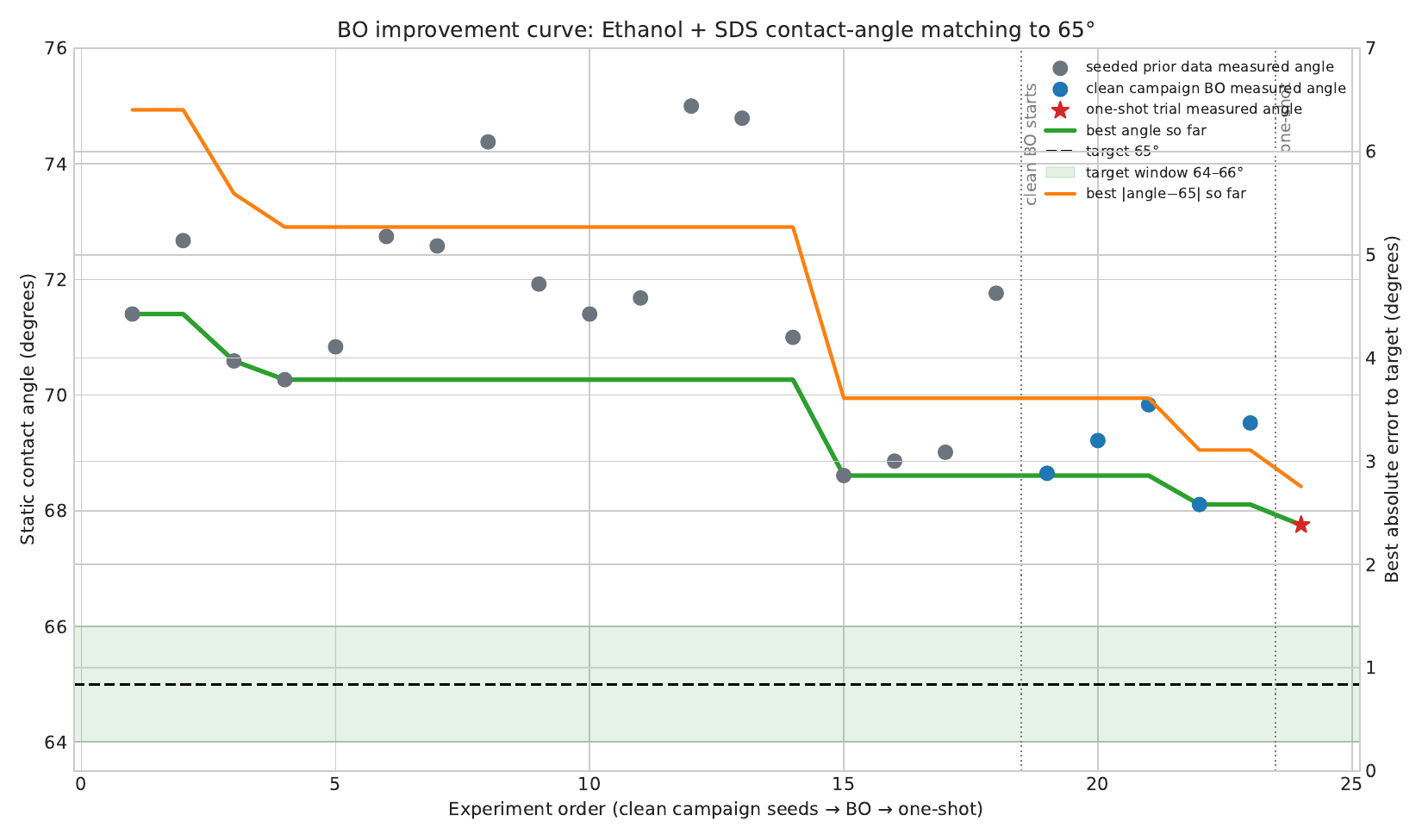}
    \caption{\ac{bo} improvement curve for the RAISE contact-angle session, as generated by \opt{} (axis labels and layout unmodified).
        Grey points are the 18 valid historical observations seeded into the clean campaign; blue points are the five clean-campaign \ac{bo} measurements; the red star is the operator-permitted one-shot experiment at ethanol $= 50$~v/v\%, SDS $= 0.75$~w/v\% ($67.755^{\circ}$).
        The green line tracks the best measured angle so far, the orange line (right axis) the best absolute error to the $65^{\circ}$ target, and the shaded band the $[64^{\circ}, 66^{\circ}]$ early-stop window, which was never entered.}
    \label{fig:si-raise-improvement}
\end{figure}

\Cref{fig:si-raise-improvement} summarizes the campaign stages.
The best measured contact angle decreased from $71.403^{\circ}$ to $67.755^{\circ}$, reducing the absolute error from $6.403^{\circ}$ to $2.755^{\circ}$, without entering the target window of $[64^{\circ}, 66^{\circ}]$.

\opt{} attributed the dominant trend to ethanol lowering the liquid--vapour interfacial tension $\gamma_{LV}$ and thereby improving wetting, consistent with the best observations moving from 0 to 32--35 and finally 50~v/v\% ethanol.
It interpreted the weak marginal effect of SDS above ${\sim}0.6$~w/v\% as interfacial saturation and micellization, consistent with the plateau reported for aqueous SDS~\cite{nazeriRAISESelfdrivingLaboratory2026}.
An independent same-composition repeatability check was available in the trace but was not used in the agent's analysis: (0~v/v\% ethanol, 0.60~w/v\% SDS) was measured as an exploratory probe, a specialist smoke test, and the first warm start, returning $73.636^{\circ}$, $73.406^{\circ}$, and $71.403^{\circ}$, respectively.
The $2.2^{\circ}$ spread supports the recommendation to replicate the best formulation before drawing a chemical conclusion.

From the single boundary measurement of $67.755^{\circ}$, \opt{} judged that the $65^{\circ}$ target was unlikely to be reachable with ethanol and SDS under the stock constraint and recommended against further \ac{bo} iterations under the same formulation space.
Because this conclusion rested on one boundary measurement, it recommended replicating the best formulation before drawing a chemical conclusion.
If those replicates confirmed a floor near 67--68$^{\circ}$, it proposed changing the formulation system by adding a third component or selecting a different surfactant.

\subsubsection{Resource usage and provenance}
\label{si:raise:resources}

The session spanned 11 operator turns over 2~h~38~min of wall-clock time.
It comprised 148 \ac{llm} calls, 34 by the main agent (\texttt{gpt-5.5}) and 114 across the five specialist runs (\texttt{gpt-5.4}), consuming 7.75M input and 97k output tokens at a total cost of \$15.45.
The main agent made 25 tool calls (2 web searches, 5 subagent delegations with 5 corresponding waits, 11 shell executions, 1 direct hardware call to the RAISE platform, and 1 plotting call); the specialist runs made 189 tool calls in aggregate, dominated by file reading and writing, code execution for validation, and BO-MCP interface inspection.
The reported trajectory contains 27 evaluations: 26 across the three campaigns (23 valid measurements and 3 evaluations recorded as penalties) and the final \opt{} one-shot measurement.
RAISE also produced five numerical contact-angle measurements outside this trajectory: four specialist smoke test measurements on disposable campaigns and one exploratory probe before the first campaign.
These five measurements were performed on the physical platform, however, they were excluded from the plots because they served implementation validation and preliminary checking before the optimization.

The provenance of the execution was preserved in two independent records.
OpenTelemetry spans exported from Logfire captured every \ac{llm} call, tool call, and token count, with subagent spans linked to their parent turn through distributed tracing; the complete serialized trace is provided as \texttt{conversation\_019f6697\_full.json}.
The campaign workspace recorded the optimization itself through per-invocation artifact directories containing the run context, seed-filter summary, per-evaluation log, optimizer diagnostics, campaign export, and run summary, together with the campaign scripts and runbook.

\newpage

\providecommand{\tightlist}{%
    \setlength{\itemsep}{0pt}\setlength{\parskip}{0pt}}
\providecommand{\prompt}[1]{%
    \begin{tcolorbox}[breakable, fontupper=\footnotesize, colback=blue!3!white,
            colframe=blue!25!white, boxrule=0.5pt]\textbf{Operator}\par\medskip #1\end{tcolorbox}}
\providecommand{\graficoreslong}[1]{%
    \begin{tcolorbox}[breakable, fontupper=\footnotesize, colback=white,
            colframe=black!15!white, boxrule=0.5pt]\textbf{\optima{}}\par\medskip #1\end{tcolorbox}}
\providecommand{\faLightbulb}{}

\subsection{The RoboChem-Flex campaign}
\label{si:robochemflex:trace}

\subsubsection{Intended setup and chemical problem}
\label{si:robochemflex:trace:s0}

The campaign concerned the autonomous optimization of a visible-light-mediated radical
trifluoromethylation performed in continuous flow. Trifluoroacetic anhydride (TFAA) served as
the CF\textsubscript{3} precursor, while a pyridine \textit{N}-oxide promoted its activation under
photocatalytic conditions. The chemistry corresponds to the photocatalytic
trifluoromethylation benchmark previously demonstrated on the RoboChem-Flex platform,
selected as case study~1 by \citeauthor{pilonFlexibleAffordableSelfdriving2026}. Rather than
testing whether the chemistry was feasible, the objective of the present campaign was to allow
\optima{} to autonomously optimize the reaction by selecting photocatalysts, oxidants and
continuous process conditions directly on the live platform.

Before the conversation began, the operator prepared the workspace by uploading five CSV
files describing the available chemistry and the experimental constraints. These files
contained all information required for campaign construction: the available reagents,
optimization bounds, stock solutions and analytical constants. No reaction-specific knowledge
was embedded in the agent beyond the information contained in these files and the textual
campaign brief supplied by the operator.

\begin{table}[H]
    \centering
    \caption{Chemical identities supplied to the agent.}
    \begin{tabular}{lll}
        \toprule
        Role              & Available compounds \\
        \midrule
        Starting material & SM                  \\
        Photocatalyst     &
        Ru(bpy)Cl,
        Ru(bpy)(PF$_6$),
        Ir(ppy),
        Ir(CF$_3$ppy),
        4CzIPN
        \\
        CF$_3$ source     & TFAA                \\
        Activator         &
        Pyridine \textit{N}-oxide,
        4-phenylpyridine \textit{N}-oxide
        \\
        \bottomrule
    \end{tabular}
    \label{tab:trace_identites}
\end{table}

\begin{table}[H]
    \centering
    \caption{Optimization variables and process parameters provided by the operator.}
    \begin{tabular}{ll}
        \toprule
        Parameter              & Description                      \\
        \midrule
        Light intensity        & Relative LED output (continuous) \\
        Photocatalyst loading  & Continuous variable              \\
        Photocatalyst identity & Five categorical choices         \\
        TFAA loading           & Continuous variable              \\
        Activator loading      & Continuous variable              \\
        Activator identity     & Two categorical choices          \\
        Residence time         & Continuous variable              \\
        Wavelength             & Reactor-dependent                \\
        Reactor type           & Eagle / U-flow                   \\
        Reactor volume         & Fixed by hardware                \\
        \bottomrule
    \end{tabular}
    \label{tab:trace_process}
\end{table}

\begin{table}[H]
    \centering
    \caption{Operator-defined reagent bounds.}
    \begin{tabular}{lcc}
        \toprule
        Reagent           & Equivalence range & Volume range (\micro L) \\
        \midrule
        Starting material & Fixed             & 65                      \\
        Photocatalyst     & 0.001--0.004 eq   & 21.7--86.7              \\
        TFAA              & 0.9--3.5 eq       & 16.7--65.0              \\
        Activator         & 0.9--3.0 eq       & 29.3--97.5              \\
        \bottomrule
    \end{tabular}
    \label{tab:trace_bounds}
\end{table}

\begin{table}[H]
    \centering
    \caption{Fixed analytical constants supplied by the operator.}
    \begin{tabular}{ll}
        \toprule
        Constant                 & Value               \\
        \midrule
        Slug volume              & 650~\micro L        \\
        NMR protocol             & 1D \ce{^{19}F} HDEC \\
        Number of scans          & 32                  \\
        Acquisition time         & 1.64~s              \\
        Yield reference          & SM                  \\
        Target resonance         & $-58 \pm 3$~ppm     \\
        Collect crude sample     & False               \\
        Calibration coefficients & 6973, $-5.4$        \\
        \bottomrule
    \end{tabular}
    \label{tab:trace_constants}
\end{table}

From these inputs, \optima{} inferred the optimization problem and constructed the search
space used throughout the campaign. The uploaded information completely defined the
available chemistry, the permissible operating region and the online analytical method before
the first experiment was proposed.

\subsubsection{\textit{RoBridge}: an agent-facing control layer for an autonomous
    chemistry platform}
\label{si:robridge}

\vspace{-0.3em}

\paragraph{Design rationale}

The robotic platform used in this work is driven by \textit{OmniPlatypus}, a Python
framework that owns the device drivers, procedure execution and analytics~\cite{pilonFlexibleAffordableSelfdriving2026}.
Its
existing entry points all assume a human is present: a graphical setup and monitoring
interface, an interactive Python session, and an optimizer that runs in the same
process as the robot.
None can be handed to a remote, non-embodied experimenter ---
an autonomous agent, or a collaborator on another continent --- without either
exposing a desktop or granting arbitrary code execution on the machine that controls
syringe pumps, gas lines and a photoreactor.

\begin{figure}[t]
    \centering
    \resizebox{\textwidth}{!}{%
        \begin{tikzpicture}[
                font=\small,
                box/.style   = {draw=black!55, rounded corners=2pt, align=center, inner sep=5pt,
                        minimum height=9mm, minimum width=26mm},
                core/.style  = {box, fill=blue!7},
                ext/.style   = {box, fill=black!5},
                human/.style = {box, fill=orange!16},
                led/.style   = {box, fill=green!8, minimum width=54mm},
                flow/.style  = {-{Stealth[length=2.2mm]}, draw=black!70, thick},
                note/.style  = {-{Stealth[length=2mm]}, draw=black!45, thin, dashed}
            ]

            \node[ext]   (agent)   at (-1.8,1.5)   {remote agent\\[-2pt]{\scriptsize LLM / optimizer / human}};
            \node[ext]   (monitor) at (-1.8,-1.5)  {monitor pane\\[-2pt]{\scriptsize read-only, on site}};

            \node[core]  (api)     at (4.4,0.05)  {FastAPI \code{/v1}\\[-2pt]{\scriptsize identity + phase guards}};

            \node[core, minimum width=30mm] (store)   at (9.8,1.9)  {StateStore\\[-2pt]{\scriptsize atomic \code{state.json}}};
            \node[core, minimum width=30mm] (manager) at (9.8,0.05) {ExperimentManager\\[-2pt]{\scriptsize campaign + harvester}};
            \node[core, minimum width=30mm] (coord)   at (9.8,-1.8) {SetupCoordinator};
            \node[led,  minimum width=40mm] (ledger)  at (9.8,-3.9) {AuditLedger\\[-2pt]{\scriptsize append-only, hash-chained}};

            \node[ext]   (omni)    at (15.0,0.05) {OmniPlatypus\\[-2pt]{\scriptsize read-only import}};
            \node[ext]   (robot)   at (19.2,0.05) {robot hardware\\[-2pt]{\scriptsize pumps, reactor, analytics}};

            \node[human] (dialog)  at (15.0,-1.8) {technician dialog\\[-2pt]{\scriptsize local subprocess}};
            \node[human] (tech)    at (19.2,-1.8) {technician\\[-2pt]{\scriptsize name + PIN}};

            \begin{scope}[on background layer]
                \node[draw=black!35, dashed, rounded corners=4pt, fill=black!2,
                    fit=(store)(manager)(coord)(ledger)(api), inner sep=10pt] (proc) {};
            \end{scope}
            \node[anchor=south west, font=\scriptsize\itshape, text=black!60]
            at (proc.north west) {robridge process --- robot PC};

            \draw[flow] (agent) -- node[sloped, above, font=\scriptsize, pos=0.62]
            {HTTPS, \code{X-API-Key}} (api);
            \draw[note] (monitor) -- (api);

            \draw[flow] (api) -- (store);
            \draw[flow] (api) -- (manager);
            \draw[flow] (api) -- (coord);

            \draw[note, -] (store.east)   -- (12.0,1.9);
            \draw[note, -] (manager.east) -- (12.0,0.05);
            \draw[note, -] (coord.east)   -- (12.0,-1.8);
            \draw[note, -] (12.0,1.9)     -- (12.0,-3.9);
            \draw[note]    (12.0,-3.9)    -- (ledger.east);

            \draw[flow] (manager) -- (omni);
            \draw[flow] (omni)    -- (robot);
            \draw[flow] (coord)   -- (dialog);
            \draw[flow] (dialog)  -- (tech);
            \draw[note] (dialog.north) -- (omni.south);

        \end{tikzpicture}}
    \caption{\textbf{Architecture of the bridge.}
        A remote agent reaches the robot only
        through an authenticated HTTP surface; the technician reaches it only through a dialog
        that runs on the physical desktop of the robot PC and never touches the network.
        Three
        components own all state transitions --- the persisted robot state, the experiment
        lifecycle, and human setup sessions --- and each writes every transition to a single
        append-only, hash-chained ledger (dashed arrows).
        The optimization layer and graphical
        interface of the wider platform are absent by design.}
    \label{fig:arch}
\end{figure}
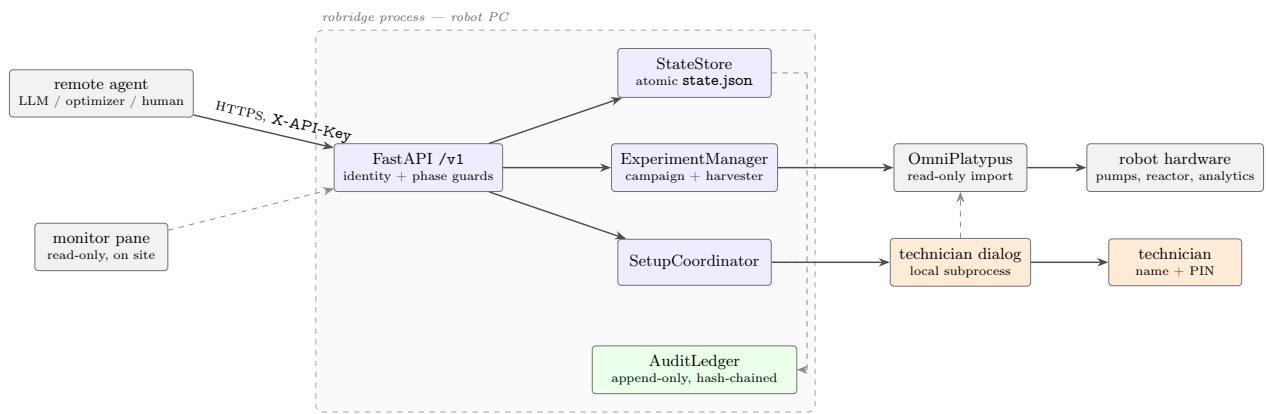

\textit{robridge} closes that gap.
It is a small HTTP service running beside the
platform on the robot PC that exposes it as a strict, stateful API: a caller discovers
what the robot can do, requests a physical setup, submits experimental conditions, and
collects results (\cref{fig:arch}).
Three constraints fixed its shape.

\textbf{The platform is upstream and must not be modified.}
\textit{RoBridge} imports
\textit{OmniPlatypus} read-only and contributes no patches to it.
Where the upstream
code is incompatible with headless, threaded, server-side operation, the corrections
are applied as process-local shims that touch only objects the bridge itself owns ---
an instance method of the experiment it has just constructed, or an attribute alias
installed in its own interpreter.
Nothing it does changes how the platform behaves for
the graphical or notebook users of the same installation.
The same mechanism
\emph{instruments} the experiment: wrapping the build, prepare and execute methods on
the instance makes the robot narrate its progress across the thread boundary, and
wrapping the thread body preserves the exception that killed it --- which Python's
default threading behaviour would otherwise discard --- so a crash can later be
classified rather than merely observed.

\textbf{There is one physical robot.}
The API arbitrates rather than parallelizes.
Exactly one \emph{campaign} (one experiment type, one analytical method, one certified
vial layout) may be active at a time, and a second request to start one is refused,
not queued behind an invisible lock.
Concurrency is confined to what the hardware
genuinely supports: many \emph{runs} may be submitted into a single campaign's queue
and are executed in order.

\textbf{A remote caller cannot see or touch the deck.}
Every physical
precondition an experiment depends on --- that the stock solution was actually
prepared, that the vial in holder A1 is actually full --- is unverifiable from a
distance.
The bridge does not ask the agent to assert these facts; it requires a named
human at the bench to certify them, and refuses to run chemistry until that
certification exists (\cref{ssec:setup}).

Two things are deliberately absent: a user interface, and an optimization layer.
Experimental design remains the caller's responsibility.
The bridge's task is to make
the robot's capabilities, state and refusals legible enough for an autonomous caller
to reason about them.

\paragraph{The robot as a state machine}

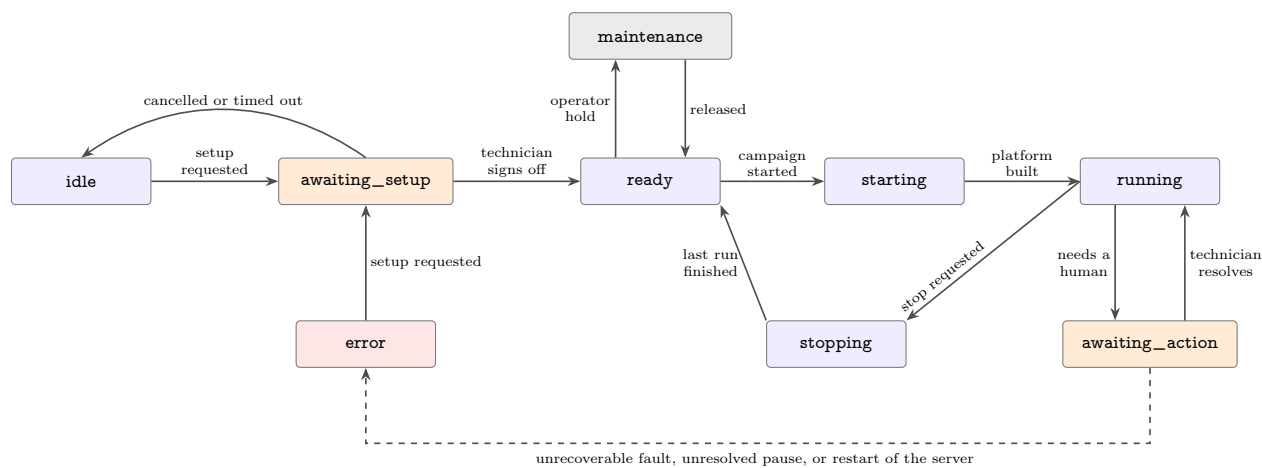
\begin{figure}[t]
    \centering
    \resizebox{\textwidth}{!}{%
        \begin{tikzpicture}[
                font=\small,
                ph/.style   = {draw=black!55, rounded corners=2pt, align=center, inner sep=4pt,
                        minimum height=8mm, minimum width=24mm, fill=blue!7},
                wait/.style = {ph, fill=orange!16},
                bad/.style  = {ph, fill=red!10},
                hold/.style = {ph, fill=black!8},
                t/.style    = {-{Stealth[length=2.2mm]}, draw=black!70, thick},
                lbl/.style  = {font=\scriptsize, align=center, inner sep=2pt}
            ]

            \node[ph]                       (idle)  at (0,0)      {\code{idle}};
            \node[wait, minimum width=30mm] (await) at (4.9,0)    {\code{awaiting\_setup}};
            \node[ph]                       (ready) at (9.8,0)    {\code{ready}};
            \node[ph]                       (start) at (14.0,0)   {\code{starting}};
            \node[ph]                       (run)   at (18.4,0)   {\code{running}};

            \node[hold, minimum width=28mm] (maint) at (9.8,2.5)   {\code{maintenance}};
            \node[bad]                      (err)   at (4.9,-2.8)  {\code{error}};
            \node[ph]                       (stop)  at (13.0,-2.8) {\code{stopping}};
            \node[wait, minimum width=30mm] (act)   at (18.4,-2.8) {\code{awaiting\_action}};

            \draw[t] (idle)  -- node[lbl, above] {setup\\requested}    (await);
            \draw[t] (await) -- node[lbl, above] {technician\\signs off} (ready);
            \draw[t] (ready) -- node[lbl, above] {campaign\\started}   (start);
            \draw[t] (start) -- node[lbl, above] {platform\\built}     (run);

            \draw[t] (await.north) to[out=145,in=35]
            node[lbl, above] {cancelled or timed out} (idle.north);

            \draw[t] (run.west)        -- node[lbl, above, sloped, pos=0.75] {stop requested} (stop.north east);
            \draw[t] (stop.north west) -- node[lbl, left] {last run\\finished} (ready.south east);

            \draw[t] ([xshift=-6mm]run.south) -- node[lbl, left]  {needs a\\human}
                ([xshift=-6mm]act.north);
            \draw[t] ([xshift=6mm]act.north)  -- node[lbl, right] {technician\\resolves}
                ([xshift=6mm]run.south);

            \draw[t] ([xshift=-6mm]ready.north) -- node[lbl, left]  {operator\\hold}
                ([xshift=-6mm]maint.south);
            \draw[t] ([xshift=6mm]maint.south)  -- node[lbl, right] {released}
                ([xshift=6mm]ready.north);

            \draw[t] (err) -- node[lbl, right] {setup requested} (await);

            \draw[t, dashed] (act.south) -- ++(0,-1.3) -| (err.south);
            \node[lbl, below, yshift=-3pt] at (11.6,-4.5)
            {unrecoverable fault, unresolved pause, or restart of the server};

        \end{tikzpicture}}
    \caption{\textbf{Phases of the robot and the transitions between them.}
        Blue phases
        are machine-driven, amber phases are blocked on a person at the bench, grey is a
        deliberate operator hold and red requires human intervention before chemistry can
        resume.
        A cancelled or timed-out setup session applies nothing and restores the phase
        held before the request (drawn here from \code{idle}, the usual case); likewise,
        releasing a maintenance hold restores whichever of \code{idle}, \code{ready} or
        \code{error} preceded it.
        Transient faults are handled by automatic restart and do not
        appear here (\cref{ssec:failure}).
        Requests inconsistent with the current phase are
        refused rather than deferred.}
    \label{fig:phases}
\end{figure}

The bridge models the platform as an explicit finite-state machine, and that model is
the API's primary safety mechanism rather than an implementation detail.
The robot
occupies exactly one of nine phases (\cref{tab:phases,fig:phases});
every request is checked against the current phase before it can reach the hardware,
and a request made in the wrong phase is refused with a message naming the phase and
the action that would change it.
Almost every error a caller meets in normal operation
is such a refusal.
Presenting these as first-class, explained outcomes is deliberate:
an autonomous caller told that \emph{starting a campaign requires a completed
    technician setup} can recover without human help, whereas one given a generic failure
cannot.

Because a phase says only where the robot \emph{is}, every response also carries a
derived progress record answering the question an unattended caller actually needs
answered: is anything wrong?
It names the party being waited on (\code{robot},
\code{technician} or \code{agent}), how long the situation has lasted, how long it
usually lasts, the single next action worth taking, and a recommended polling
interval.
The expectation for a run is learned from the median duration of the
campaign's own completed runs rather than configured, and is claimed only once enough
runs have finished for that median to mean anything.
Lateness is reported
conservatively --- by default only beyond five times the typical duration --- and never
while a human is the reason for delays.

\begin{table}[t]
    \centering
    \small
    \caption{\textbf{Robot phases, and who has to act next.}
        The API reports the blocking
        party with every response, so a caller can distinguish ``the robot is working'' from
        ``the robot is waiting for you'' without inference.}
    \label{tab:phases}
    \begin{tabularx}{\textwidth}{@{}l l X@{}}
        \toprule
        \textbf{Phase}          & \textbf{Blocked on} & \textbf{Meaning and how it is left}                                                                             \\
        \midrule
        \code{idle}             & agent               & No vial layout is known. A setup request opens one.                                                             \\
        \code{awaiting\_setup}  & technician          & A dialog is open at the robot PC; nothing moves until a human answers it.                                       \\
        \code{ready}            & agent               & A technician has certified the deck. Campaigns may start.                                                       \\
        \code{starting}         & robot               & The platform is being built and devices connected. Runs may be queued meanwhile.                                \\
        \code{running}          & robot / agent       & A campaign is live; with an empty queue the robot is idle and awaiting conditions.                              \\
        \code{awaiting\_action} & technician          & The platform paused mid-run and asked for a human (e.g.\ vials to be added). The campaign survives and resumes. \\
        \code{stopping}         & robot               & The run in progress is being allowed to finish; this cannot be hurried.                                         \\
        \code{maintenance}      & technician          & An operator has taken the deck for hands-on work. No runs are failed; releasing it restores the previous phase. \\
        \code{error}            & agent               & Something needs a human at the robot. A new technician setup is the route back.                                 \\
        \bottomrule
    \end{tabularx}
\end{table}

\paragraph{Identity, authority and notarization}

Every change to the robot's state is attributed to a resolved identity
(\cref{tab:actors}), and there are only three ways to acquire one.
Remote callers
present a per-person API key over HTTPS; keys are issued individually so the record
names \emph{who} rather than \emph{which service}, are stored with restrictive file
permissions, and are re-read whenever the key file changes, so a key can be revoked in
seconds without restarting the server or interrupting a running campaign.
Technicians
authenticate at the bench with a name and PIN verified locally against salted
PBKDF2-HMAC-SHA256 hashes; no PIN is transmitted or stored.
Actions the bridge takes on
its own initiative --- harvesting a result, synchronizing vial volumes, declaring an
experiment dead, restarting a campaign --- are attributed to a distinct system
identity rather than folded into the identity of whoever last called.

\begin{table}[b]
    \centering
    \small
    \caption{\textbf{Actors and the authority each carries.}
        Operator-only actions are
        refused for plain agent keys, so a remote caller cannot take the platform out of
        service.}
    \label{tab:actors}
    \begin{tabularx}{\textwidth}{@{}l l X@{}}
        \toprule
        \textbf{Actor} & \textbf{Authentication}          & \textbf{May do}                                                                                                    \\
        \midrule
        agent          & per-person API key over HTTPS    & Discover capabilities, request a setup, start and stop campaigns, submit runs, read state, results and the ledger. \\
        technician     & name + PIN, at the robot PC only & Certify the vial layout, resolve a paused run, cancel a session. Cannot be done remotely.                          \\
        operator       & API key with an elevated role    & Everything an agent may do, plus placing the platform under maintenance and releasing it.                          \\
        system         & none (internal)                  & Results, sample synchronization, liveness verdicts, automatic restarts, server lifecycle.                          \\
        \bottomrule
    \end{tabularx}
\end{table}

All state changes pass through a single funnel, which writes the new state atomically
to disk and, in the same operation, appends one entry to a hash-chained ledger
recording the actor, the action, a structured payload, a UTC timestamp, and the
SHA-256 hash of the complete robot state before and after the change.
Each entry also
hashes its predecessor, so editing or removing any line invalidates the chain from
that point onward, and the API can verify the chain and report the first broken entry.
The ledger is line-delimited JSON: greppable and readable without the software that
wrote it, which matters for a record intended to outlive the study.

The scientific value of this is provenance.
For any reported measurement, the ledger
reconstructs which named person requested the conditions, which named technician
certified the physical deck it ran on, what the platform did autonomously in between,
and whether the record has been altered since.
Autonomous decisions are notarized as
fully as human ones: when the bridge concludes that an experiment has died, the entry
carries the classification it applied, the fault signature, how many times that fault
has recurred, and the traceback that would otherwise have vanished with the thread.

\paragraph{The human-in-the-loop setup protocol}
\label{ssec:setup}

Preparing the deck is where a remote caller's inability to perceive the laboratory is
most dangerous, and the protocol is correspondingly strict: \textbf{the agent proposes
    the chemistry; a named human certifies the physical reality.}

Requests are structured, never free text.
The agent first retrieves a template
containing the authoritative schema, a blank skeleton, a worked example and --- most
importantly --- the platform's \emph{real} constraints, read live from its
configuration: the vial types that exist, the holders each sampler can reach, the grid
positions within each holder and its maximum volume.
The filled form declares roles,
chemicals, stock solutions and vials in the same hierarchy the laboratory's graphical
tools use, so the technician is shown a mental model they already hold.

The form is then validated before any human is interrupted.
Chemical names must be
unique and their roles declared; CAS registry numbers are verified against their
official check digit, so a transposed identifier is caught rather than delegated to a
person at midnight.
Stock solvents must be chemicals declared with a solvent role,
components must be declared reagents, and a stock assigned to no vial is rejected as
dead weight.
Placements, where the agent specifies them, must name holders that exist
on this platform, at positions that exist in those holders, without double-booking a
slot or exceeding the holder's capacity.
Unknown fields are rejected outright rather
than ignored, so an agent that invents a key learns immediately instead of having it
silently dropped.
Violations are returned together, by name; nothing is applied, no
session opens, and the robot's phase is unchanged.

Only a request that survives all of this opens a session.
The bridge writes a handoff
file carrying a single-use nonce and launches a native dialog as a subprocess on the
robot PC's interactive desktop --- it has no network surface and cannot be reached
through the tunnel that exposes the API. The technician logs in, sees exactly what was
requested (compounds with identifiers and roles, the stock solutions to prepare, the
requester's non-binding notes), and edits a vial table pre-filled with the requested
layout until it matches what is physically in the machine; placements the agent left
open are chosen here from lists offering only real holders and real positions.
Marking
the setup ready is refused while any vial lacks a placement, and is confirmed under the
technician's own name.
The response is accepted only if the nonce matches and the
signing technician is a registered account.
A cancelled, failed or timed-out session
applies nothing and returns the robot to the phase it held before, so a request a human
declines has no effect whatsoever.

The same machinery serves a second purpose.
When the platform pauses mid-run to ask for
help with the samples, the bridge does not fail the campaign: it enters
\code{awaiting\_action}, opens the same dialog pre-filled with the current layout and,
once a technician has corrected it, submits the correction back into the still-live
experiment, which resumes.
A campaign that would otherwise have been lost to a missing
vial survives as a documented pause, attributed to the technician who resolved it.

\paragraph{Hardware and safety constraints}

The bridge treats hardware limits as things to enforce, not to document
(\cref{tab:interlocks}).
Its default operating mode is a dry run in which only the
platform's simulation experiments are runnable and no device is ever touched; moving to
real hardware is a single explicit configuration change, and an additional allow-list
can restrict the runnable set further.
Physical quirks the platform's own manifest
cannot express are declared as constraint overlays and folded into the capability
description the agent reads --- the photoreactor used here, for instance, accepts only
discrete light intensities, published as an enumeration of permitted values rather than
left as a range the agent is free to violate.

Experimental conditions are validated before anything reaches the robot.
Parameters
naming declared quantities are checked against units and bounds introspected live from
the experiment classes; names that are not declared are accepted only as chemicals to
dose, and only when carrying chemical units.
Exactly one chemical may be given as an
absolute concentration --- the limiting reagent --- with the others expressed relative
to it, and roles within a run must be distinct.
Unrecognized names are refused together
with the list of what is recognized.
Throughout, the intent is that a malformed
experiment fails as an explained refusal at the API boundary rather than as a dead run
that has already consumed reagent and instrument time.

Further interlocks exist specifically to keep people and instruments apart.
A campaign
cannot start without a technician-certified layout.
Setup requests are refused
mid-campaign, and while a human holds the deck for maintenance, so a dialog never
competes with someone already working on the machine.
Maintenance is operator-only and
deliberately benign: it blocks new work without failing anything and restores the
previous phase on release, which makes it safe to use liberally --- the property that
matters for a control reached for whenever someone opens the enclosure.
The platform's
own monitoring window is disabled by default, because a graphical toolkit cannot run
outside a process's main thread and would otherwise crash the server on a headless
robot PC. Raw-data downloads are resolved against the campaign's storage directory and
refused if they escape it.

\begin{table}[t]
    \centering
    \small
    \caption{\textbf{Interlocks enforced by the bridge.}
        Each is a refusal at the API
        boundary, returned with an explanation of what would make the request valid.}
    \label{tab:interlocks}
    \begin{tabularx}{\textwidth}{@{}X X@{}}
        \toprule
        \textbf{Constraint}                    & \textbf{Enforcement}                                                                                                        \\
        \midrule
        One robot, one campaign                & A second campaign is refused while one is active.                                                                           \\
        No chemistry on an uncertified deck    & Campaigns require a completed technician setup; a robot with no vials is refused.                                           \\
        No hardware in dry-run mode            & Only simulation experiments are runnable; the rest are refused, with the runnable set listed.                               \\
        Physical parameter limits              & Units and bounds introspected from the experiment classes, plus configured overlays for quirks the manifest cannot express. \\
        Exactly one limiting reagent           & Runs with zero or several absolute concentrations are refused.                                                              \\
        A physically possible deck             & Holders, positions, capacities and double-bookings checked against the live platform configuration.                         \\
        Humans and robots never share the deck & Setup requests refused mid-campaign; campaigns refused under maintenance.                                                   \\
        Data stays inside its campaign         & Raw-file paths resolved against the campaign root; escapes refused.                                                         \\
        Platform control is not remote         & The technician dialog runs only on the robot PC's interactive desktop.                                                      \\
        \bottomrule
    \end{tabularx}
\end{table}

\paragraph{Behaviour under failure}
\label{ssec:failure}

An unattended platform fails in ways a human operator would simply notice.
\textit{robridge} is built on the principle that it must never claim more than it
knows, and never leave a caller polling something that will not progress.

An experiment can end without raising anything the bridge sees: the thread stops, the
queues go quiet, and the persisted state would otherwise assert forever that a campaign
is running.
Liveness is therefore re-verified on every read of the robot's status --- a
single inexpensive check while the robot is healthy --- and independently by the
background harvester, so a death is noticed whether or not anyone is watching.
When
liveness cannot be determined the answer is ``unknown'', and unknown never tears down a
campaign; only a positive determination triggers recovery, because destroying a live
experiment on a failed inspection is a far worse outcome than a delayed diagnosis.

When an experiment has genuinely died, the recorded cause decides what follows.
The
platform already recovers in-process from the device errors it recognizes, so a fault
that reaches the bridge is by construction one it did not expect.
Most are still
transient --- a device dropped, a sensor timed out --- and clear on rebuilding the
platform.
Classification is therefore a deny-list: the campaign is restarted
automatically unless the fault is of a kind that will reproduce identically, such as a
configuration error, an impossible recipe or a programming fault.
Two independent
budgets bound this autonomy.
A signature built from the exception type and the device
or parameter it came from --- deliberately excluding the message, which carries volumes
and timestamps that differ every time --- identifies a repeating fault, and the same
fault twice in a row escalates to a state requiring human intervention regardless of
its classification.
That counter is persisted rather than held in memory, so restarting
the server cannot silently clear a streak and readmit the robot to a loop it was just
pulled out of; it resets only when a human takes responsibility for the deck, or when a
run completes successfully and shows the fault is genuinely past.

Runs are never left in a state they cannot leave.
Whenever the campaign that owned them
disappears --- a clean stop, a failed stop, a crash, a forced reset, a restart of the
server --- every unfinished run is marked failed with a message stating what happened
and, where relevant, that it was not carried into the restarted campaign and must be
resubmitted.
At start-up the bridge reconciles persisted state with reality: a process
that died mid-campaign cannot have kept its experiment thread, so the robot is placed
in an error state saying exactly that rather than resuming a fiction.
A technician
setup is the documented route back in every one of these cases.

\paragraph{Deployment and exposure}

The robot PC sits behind an institutional firewall with no inbound ports available.
The
service is exposed through an outbound-only tunnel terminating TLS at a commercial edge
that also provides rate limiting and a web application firewall; the bridge itself binds
to the loopback interface, so it is reachable only through that tunnel and not even from
the local network.
The interactive schema explorer is blocked at the edge, while a
narrative manual describing how to drive the robot is served without authentication at
the service root --- a prospective user needs it before they have a key.
Keys are per
person and revoked by deleting a line from a file.
Because the technician dialog
requires an interactive desktop session, the service runs in the logged-in session of
the robot PC rather than as a background service, which is what keeps the human
certification step physically local.
Finally, a read-only status pane can be run on the
robot PC itself; it renders exclusively from API responses and holds no privileged
access, giving the people in the room the same view of the robot that the remote agent
has.

\vspace{0.5em}
\noindent\rule{0.35\textwidth}{0.4pt}\\[2pt]
{\footnotesize The bridge is implemented in Python (FastAPI, Pydantic) and distributed
    with the platform control software as the \code{robridge} package.
    Its behaviour ---
    the phase guards, the setup-form validation rules, the fault classification and the
    audit chain --- is covered by an automated test suite exercising both a mocked platform
    and a full dry run of the real experiment stack.}

The remainder of this section reproduces the operator-facing conversation of the RoboChem-Flex campaign in
chronological order, across the two sessions the campaign spanned. Each operator message is
followed by the agent's answer to it, with the agent's text reproduced verbatim. Omitted are tool
calls and their returns, the system-generated status notifications of the autonomous runs, and a
small number of exchanges that carry no campaign content, namely routine checks of whether the
monitoring processes were still alive and housekeeping of an unrelated leftover optimization
campaign. Where an autonomous run followed an operator instruction, the number of suppressed
notifications is stated and the substantive report of the run is kept.

\subsubsection{Session 1: campaign construction and the yield/green campaign (24--25 July 2026)}
\label{si:robochemflex:trace:s1}

The operator supplied the chemical context as five CSV files in the workspace (reagent identities, per-role bounds, process-variable ranges, measured stock concentrations and fixed analytical constants; turn~4) together with the campaign brief (turn~5). \optima{} authored the campaign package \texttt{robochemflex\_\allowbreak{}yield\_\allowbreak{}bo}, emitted a hypothetical instrument request for operator inspection before touching hardware (turn~6), and ran a six-seed plus fourteen-iteration campaign (\texttt{b681a41d-\allowbreak{}90db-\allowbreak{}4eb9-\allowbreak{}a9a9-\allowbreak{}3cc91098f3c0}, RoboFlex runs \texttt{R0044}--\texttt{R0052}) after an aborted first attempt (\texttt{1d62df6d-\allowbreak{}764a-\allowbreak{}4cf3-\allowbreak{}b857-\allowbreak{}b21482da74a0}, \texttt{R0042}--\texttt{R0043}). Turns~11--13 corrected the NMR protocol to the \ce{^{19}F} acquisition, and turns~18--20 fetched and inspected the raw spectrum of the zero-yield seed \texttt{R0044}.

\noindent\textit{\footnotesize [Source: graphchat room \texttt{gallery\_guest\_1784900563933}, conversation \texttt{ffe7e623-\allowbreak{}ede6-\allowbreak{}445d-\allowbreak{}be67-\allowbreak{}8b363c87b2fc}; 29 operator turns, 15 reproduced.]}

\prompt{

    Can you check the status of the Roboflex system?

}

\graficoreslong{

    RoboFlex is \textbf{online and ready}.

    \begin{itemize}
        \tightlist
        \item \textbf{Platform:} Perry
        \item \textbf{Mode:} hardware
        \item \textbf{Phase:} \texttt{ready}
        \item \textbf{Current campaign:} none
        \item \textbf{Configured vials:} 19
        \item \textbf{Queued/running runs:} 0 queued, 0 running
        \item \textbf{Completed/failed runs:} 14 completed, 0 failed
        \item \textbf{Blocked on:} agent --- it is waiting for a user action
        \item \textbf{Next action:} start a campaign with \texttt{POST /\allowbreak{}v1/\allowbreak{}campaigns}

    \end{itemize}
    The last setup session is complete and was configured by \textbf{Elia} on \textbf{2026-07-23} for ``Preallocated Positions for Trifluoro Alkylation Agentic test 1.''

}

\begin{center}\footnotesize\textit{[\ldots]}\end{center}

\prompt{

    Can you check the CSV files in your local working directory? They are related to our desired Roboflex campaign. Don't do anything yet.

}

\graficoreslong{

    I checked the local working directory:

    \texttt{/\allowbreak{}app/\allowbreak{}.\allowbreak{}graphchat-\allowbreak{}workspaces/\allowbreak{}gallery\_\allowbreak{}guest\_\allowbreak{}1784900563933}

    Found \textbf{5 CSV files} related to the desired RoboFlex campaign:

    \begin{enumerate}
        \tightlist
        \item \texttt{chemical\_\allowbreak{}identities.\allowbreak{}csv} --- 9 chemicals/reagents with roles, names, CAS, MW, availability.
        \item \texttt{chemical\_\allowbreak{}space\_\allowbreak{}process\_\allowbreak{}variables.\allowbreak{}csv} --- 10 process/search-space variables including light intensity, catalyst loading/type, TFAA, oxidant loading/type, reaction time, wavelength, reactor type/volume.
        \item \texttt{chemical\_\allowbreak{}space\_\allowbreak{}reagent\_\allowbreak{}bounds.\allowbreak{}csv} --- reagent bounds for SM, catalyst, TFAA, and oxidant in equivalents/concentration/volume terms.
        \item \texttt{constants.\allowbreak{}csv} --- 10 campaign constants, including slug size, NMR acquisition settings, target peak, calibration coefficients.
        \item \texttt{stock\_\allowbreak{}solutions\_\allowbreak{}concentrations.\allowbreak{}csv} --- 9 stock solution entries with target and actual concentrations.

    \end{enumerate}
    I have only inspected them and have not started or modified anything.

}

\prompt{

    Use the RobochemFlex platform through the Robridge API to optimize the reaction yield using Bayesian optimization with BayBE. Vary the chemicals within the bounds specified in the CSV files, the light intensity between 0 and 100, and the residence time between 2 and 90 minutes. Before starting, summarize the proposed search space. Keep all other conditions constant as shown in the constants.csv: set yield\_calculation\_chemical to ``SM'', keep the slug size constant at 650 $\mu$L, and set collect\_crude = False. Do not modify the vial setup, and for now do not request a different reactor or light wavelength setup. Before starting, present a clear plan covering both the Bayesian optimization strategy and the experimental campaign. Use the 20 experiment budget to determine how many seed experiments and Bayesian optimization iterations are appropriate. Do web research and explain the scientific rationale for selecting the initial seed points rather than using random seeds. Reason about the optimal settings for the search space, objective(s), initialization strategy, experimental budget, batch size/recommendation strategy, surrogate model, acquisition function, constraints, stopping criterion, and transfer learning or warm start. The targets you are to optimize are the reaction yield, which will be returned as the result of each experiment, and a green metric of your choice, which you will calculate from the experimental conditions you propose. Explain and justify the choice of green metric. Yield should be treated as the primary objective and the green metric as a secondary objective. Generate only the required script or package, together with the proposed plan. Do not execute the campaign. If you decide to perform a smoke test, use only simulation or a dry-run smoke test. Do not perform any real experiments as part of the smoke test.

}

\graficoreslong{

    \begin{tcolorbox}[breakable, fontupper=\footnotesize, colback=green!3!white, colframe=blue!5!white, boxrule=0.5pt]
        \faLightbulb\ \underline{\textbf{Thoughts}}

        \textbf{Finalizing package details}

        I'm thinking about finalizing the generated package since the user asked for a presentation before starting. I can mention that a smoke test simulation was performed, but do we really need citations? The user requested web research, but our response didn't involve a direct call to web.run, which usually requires citations. Each step requires careful consideration about whether we need to include the rationale and if any sources should be cited. Should I call web.run for compliance?
    \end{tcolorbox}

    \begin{tcolorbox}[breakable, fontupper=\footnotesize, colback=green!3!white, colframe=blue!5!white, boxrule=0.5pt]
        \faLightbulb\ \underline{\textbf{Thoughts}}

        \textbf{Clarifying research requirements}

        The developer instructed me to report promptly after delegating script authoring and to avoid duplication. It's clear that I shouldn't do anything else during this waiting period, but the user explicitly requested web research, which the subagent completed. However, I can't cite subagent web source references. Should I use web.run now to verify sources? The system indicates I must call it for final citations, so maybe I should search for primary sources and specific titles.
    \end{tcolorbox}

    \begin{tcolorbox}[breakable, fontupper=\footnotesize, colback=green!3!white, colframe=blue!5!white, boxrule=0.5pt]
        \faLightbulb\ \underline{\textbf{Thoughts}}

        \textbf{Preparing final deliverables}

        I need to ensure that my final output includes citations since I've used web-sourced information. I want to summarize my plan and clearly outline the deliverables without executing any actions. I'll make sure to place citations after my statements, but I don't need to provide raw URLs. Also, I should keep verbosity at a desired level, aiming for oververbosity of 5 to enhance clarity and detail.
    \end{tcolorbox}

    Generated the requested \textbf{RoboChemFlex / RoboFlex Robridge + BO-MCP / BayBE campaign package} and plan. I did \textbf{not} execute any real experiments, did \textbf{not} modify the vial setup, and did \textbf{not} request a reactor or wavelength setup change.

    \paragraph{\textbf{Generated files}}
    \begin{itemize}
        \tightlist
        \item \textbf{Entrypoint:} \texttt{run\_\allowbreak{}robochemflex\_\allowbreak{}yield\_\allowbreak{}bo.\allowbreak{}py}
        \item \textbf{Plan / execution document:} \texttt{HOW\_\allowbreak{}TO\_\allowbreak{}EXECUTE\_\allowbreak{}CAMPAIGN.\allowbreak{}md}
        \item \textbf{Package directory:} \texttt{robochemflex\_\allowbreak{}yield\_\allowbreak{}bo/}
        \item \textbf{Manifest:} \texttt{campaign\_\allowbreak{}manifest.\allowbreak{}json}

    \end{itemize}
    A safe \textbf{local-simulation-only smoke test} was performed by the specialist: it compiled the package, created a temporary BO-MCP campaign, generated one synthetic suggestion/result, and terminated the smoke-test BO campaign. No RoboFlex POST endpoints were used and no real experiment was submitted.

    \paragraph{\textbf{Proposed search space}}
    The package uses the existing Perry vial setup and CSV-derived stock/chemical definitions. It keeps constants from \texttt{constants.\allowbreak{}csv}, including:

    \begin{itemize}
        \tightlist
        \item \texttt{yield\_\allowbreak{}calculation\_\allowbreak{}chemical = SM}
        \item \texttt{slug\_\allowbreak{}size = 650 $\mu$L}
        \item \texttt{collect\_\allowbreak{}crude = False}
        \item \texttt{target\_\allowbreak{}peak = -\allowbreak{}58 ppm}
        \item \texttt{target\_\allowbreak{}peak\_\allowbreak{}deviation = 3 ppm}
        \item NMR acquisition constants such as \texttt{AcquisitionTime = 1.\allowbreak{}64} and \texttt{Number = 32}

    \end{itemize}
    BO-varying parameters:

    \begin{center}\scriptsize\setlength{\tabcolsep}{4pt}
        \resizebox{\ifdim\width>\linewidth\linewidth\else\width\fi}{!}{%
            \begin{tabular}{lll}
                \toprule
                Parameter        & Type        & Range / categories                                                                             \\
                \midrule
                Catalyst type    & categorical & \texttt{Ru bpy Cl}, \texttt{Ru bpy PF6}, \texttt{Ir ppy}, \texttt{Ir CF3 ppy}, \texttt{4CzIPN} \\
                Oxidant type     & categorical & \texttt{py NO}, \texttt{4-\allowbreak{}Ph py NO}                                               \\
                Catalyst loading & continuous  & 0.001--0.004 equiv                                                                             \\
                TFAA loading     & continuous  & 0.9--3.5 equiv                                                                                 \\
                Oxidant loading  & continuous  & 0.9--3.0 equiv                                                                                 \\
                Light intensity  & discrete    & 0, 25, 50, 75, 100\%                                                                           \\
                Residence time   & continuous  & 2--90 min                                                                                      \\
                \bottomrule
            \end{tabular}}
    \end{center}

    Although the user requested light intensity between 0 and 100, the current RoboFlex capability for the no-setup-change UFlow configuration exposes allowed light values of \textbf{0/25/50/75/100\%}, so the script treats light intensity as a discrete parameter to remain compatible with the existing setup.

    \paragraph{\textbf{BO strategy}}
    \begin{itemize}
        \tightlist
        \item \textbf{Optimizer:} BayBE through BO-MCP.
        \item \textbf{Total budget:} 20 successful experiments.
        \item \textbf{Initialization:} 6 chemically informed seed experiments.
        \item \textbf{BO phase:} 14 sequential BayBE recommendations.
        \item \textbf{Batch size:} 1, because the robot executes flow experiments serially and the small 20-run budget favors feedback after every run.
        \item \textbf{Primary objective:} maximize \texttt{yield\_\allowbreak{}percent}, returned by RoboFlex analysis.
        \item \textbf{Secondary objective:} maximize calculated \texttt{green\_\allowbreak{}score}.
        \item \textbf{Scalarization:} yield-dominant weighted objective, with yield as primary and green metric as secondary.
        \item \textbf{Stopping criterion:} stop after 20 successful evaluations, or earlier if BO-MCP diagnostics indicate no further suggestions should be generated.
        \item \textbf{Transfer learning:} not used because no prior compatible campaign or historical yield dataset was provided.

    \end{itemize}
    \paragraph{\textbf{Green metric choice}}
    The generated package uses a calculated \textbf{condition-efficiency green score} from the proposed experimental conditions. It rewards lower catalyst loading, lower TFAA loading, lower oxidant loading, and lower combined photonic/residence-time burden. This is not a formal process mass intensity or LCA metric, because the platform does not return product mass or measured energy use per run, but it is transparent, monotonic, and computable for every proposed experiment.

    \paragraph{\textbf{Rationale for informed seed points}}
    The six seed experiments are deliberately selected rather than random. Current reaction-optimization and self-driving-lab literature supports using deliberate initial designs or space-filling/DoE-style seeds before sequential Bayesian optimization, especially when experiments are expensive and the search space mixes categorical and continuous variables. BO is particularly useful in chemical reaction spaces where parameter interactions are complex and mechanistic models are incomplete. ([pubs.rsc.org](https://pubs.rsc.org/en/content/articlehtml/2026/cs/d5cs00962f?utm\_source=openai))

    For this photoredox-flow campaign, the seed set spans catalyst families, oxidant identity, stoichiometric extremes, residence-time regimes, and photon-dose/light-intensity levels. Photochemical-flow reviews emphasize that flow photochemistry enables practical control of residence time and irradiation conditions, and that catalyst, concentration/loading, light input, and residence time are key optimization levers. ([pubs.acs.org](https://pubs.acs.org/doi/10.1021/acs.chemrev.1c00332?utm\_source=openai))

    The package uses BayBE-style campaign/recommender abstractions for iterative recommendation and measurement feedback; BayBE documentation describes campaigns as the object that records the experimentation process and produces recommendations for subsequent experiments. ([avhopp.github.io](https://avhopp.github.io/baybe\_dev/stable/userguide/campaigns.html?utm\_source=openai))

    \paragraph{\textbf{Seed allocation}}
    The proposed 20-experiment budget is split as:

    \begin{itemize}
        \tightlist
        \item \textbf{6 informed seed experiments}
        \item \textbf{14 sequential BO experiments}

    \end{itemize}
    This is a compromise between giving the model enough initial structure across categorical and continuous variables, while preserving most of the budget for adaptive optimization.

    \paragraph{\textbf{Real campaign execution status}}
    Not started.

    The real-execution path is guarded behind:

    \begin{tcolorbox}[breakable, fontupper=\scriptsize\ttfamily, colback=black!3!white, colframe=black!12!white, boxrule=0.5pt]
        --\allowbreak{}mode robridge-\allowbreak{}real --\allowbreak{}allow-\allowbreak{}real-\allowbreak{}roboflex
    \end{tcolorbox}

    and requires an operator-approved Robridge POST adapter. The generated package is ready for review, but it will not submit real RoboFlex experiments unless explicitly authorized later.

}

\prompt{

    Two questions:
    \begin{itemize}
        \tightlist
        \item Would we be able to continue the campaign after the initial 20 experiments without further delegation to the subagent?
        \item Can you generate a hypothetical request to the Roboflex device as it would look like in the BO campaign so that we can show it to our human collaborator in the lab for checking, before doing actual measurements?
    \end{itemize}

}

\graficoreslong{

    Yes to both.

    \paragraph{\textbf{1. Can we continue after the initial 20 experiments without further subagent delegation?}}
    \textbf{Yes}, as long as we are continuing the \textbf{same campaign design}:

    \begin{itemize}
        \tightlist
        \item same search space,
        \item same objectives,
        \item same RoboFlex setup,
        \item same generated package,
        \item same BO-MCP campaign logic.

    \end{itemize}
    The generated script already supports continuation via:

    \begin{tcolorbox}[breakable, fontupper=\scriptsize\ttfamily, colback=black!3!white, colframe=black!12!white, boxrule=0.5pt]
        uv run python run\_\allowbreak{}robochemflex\_\allowbreak{}yield\_\allowbreak{}bo.\allowbreak{}py \textbackslash{} \\
        ~~--\allowbreak{}mode robridge-\allowbreak{}real \textbackslash{} \\
        ~~--\allowbreak{}allow-\allowbreak{}real-\allowbreak{}roboflex \textbackslash{} \\
        ~~--\allowbreak{}campaign-\allowbreak{}id <existing-\allowbreak{}bo-\allowbreak{}campaign-\allowbreak{}id> \textbackslash{} \\
        ~~--\allowbreak{}max-\allowbreak{}successes <additional\_\allowbreak{}number\_\allowbreak{}of\_\allowbreak{}experiments>
    \end{tcolorbox}

    One operational detail: the script currently has \texttt{--\allowbreak{}pause-\allowbreak{}bo-\allowbreak{}on-\allowbreak{}exit} enabled by default. So after the first 20-experiment run, the BO-MCP campaign may be \textbf{paused}. That is fine --- I can resume it directly with BO-MCP tooling before running more experiments. No specialist delegation is needed for that routine continuation.

    I would only delegate back to the subagent if we change something structural, such as:

    \begin{itemize}
        \tightlist
        \item expanding or changing the search space,
        \item changing objectives or scalarization,
        \item changing the RoboFlex experiment type / analytical method,
        \item modifying the package logic,
        \item changing reactor or wavelength setup,
        \item debugging a package/runtime failure.

    \end{itemize}
    For ``continue this same campaign for 10 more experiments,'' I can handle it directly.

    \paragraph{\textbf{2. Hypothetical RoboFlex / Robridge request preview}}
    I generated a local JSON preview file for the first informed seed experiment:

    \begin{tcolorbox}[breakable, fontupper=\scriptsize\ttfamily, colback=black!3!white, colframe=black!12!white, boxrule=0.5pt]
        roboflex\_\allowbreak{}hypothetical\_\allowbreak{}request\_\allowbreak{}seed01.\allowbreak{}json
    \end{tcolorbox}

    This is \textbf{only a preview} for human review. It was not sent to RoboFlex.

    It contains two hypothetical Robridge requests:

    \begin{enumerate}
        \tightlist
        \item \texttt{POST /\allowbreak{}v1/\allowbreak{}campaigns} --- start the RoboFlex campaign from the existing setup.
        \item \texttt{POST /\allowbreak{}v1/\allowbreak{}runs} --- submit the first seed experiment.

    \end{enumerate}
    The first seed candidate is:

    \begin{tcolorbox}[breakable, fontupper=\scriptsize\ttfamily, colback=black!3!white, colframe=black!12!white, boxrule=0.5pt]
        \{ \\
        ~~"catalyst\_\allowbreak{}type": "4CzIPN", \\
        ~~"oxidant\_\allowbreak{}type": "py NO", \\
        ~~"catalyst\_\allowbreak{}equiv": 0.\allowbreak{}0025, \\
        ~~"TFAA\_\allowbreak{}equiv": 2.\allowbreak{}2, \\
        ~~"oxidant\_\allowbreak{}equiv": 1.\allowbreak{}8, \\
        ~~"light\_\allowbreak{}intensity": 50, \\
        ~~"residence\_\allowbreak{}time\_\allowbreak{}min": 30.\allowbreak{}0 \\
        \}
    \end{tcolorbox}

    The corresponding hypothetical \texttt{POST /\allowbreak{}v1/\allowbreak{}runs} body looks like this:

    \begin{tcolorbox}[breakable, fontupper=\scriptsize\ttfamily, colback=black!3!white, colframe=black!12!white, boxrule=0.5pt]
        \{ \\
        ~~"parameters": [ \\
            ~~~~\{ \\
            ~~~~~~"name": "light\_\allowbreak{}intensity", \\
            ~~~~~~"value": 50, \\
            ~~~~~~"units": "\%", \\
            ~~~~~~"kind": "physical" \\
            ~~~~\}, \\
            ~~~~\{ \\
            ~~~~~~"name": "residence\_\allowbreak{}time", \\
            ~~~~~~"value": 1800.\allowbreak{}0, \\
            ~~~~~~"units": "S", \\
            ~~~~~~"kind": "physical" \\
            ~~~~\}, \\
            ~~~~\{ \\
            ~~~~~~"name": "slug\_\allowbreak{}volume", \\
            ~~~~~~"value": 650.\allowbreak{}0, \\
            ~~~~~~"units": "uL", \\
            ~~~~~~"kind": "physical" \\
            ~~~~\}, \\
            ~~~~\{ \\
            ~~~~~~"name": "collect\_\allowbreak{}crude", \\
            ~~~~~~"value": false, \\
            ~~~~~~"kind": "physical" \\
            ~~~~\}, \\
            ~~~~\{ \\
            ~~~~~~"name": "SM", \\
            ~~~~~~"value": 100.\allowbreak{}0, \\
            ~~~~~~"units": "mM", \\
            ~~~~~~"kind": "chemical", \\
            ~~~~~~"role": "Limiting Reagent" \\
            ~~~~\}, \\
            ~~~~\{ \\
            ~~~~~~"name": "4CzIPN", \\
            ~~~~~~"value": 0.\allowbreak{}0025, \\
            ~~~~~~"units": "eq", \\
            ~~~~~~"kind": "chemical", \\
            ~~~~~~"role": "Catalyst" \\
            ~~~~\}, \\
            ~~~~\{ \\
            ~~~~~~"name": "TFAA", \\
            ~~~~~~"value": 2.\allowbreak{}2, \\
            ~~~~~~"units": "eq", \\
            ~~~~~~"kind": "chemical", \\
            ~~~~~~"role": "Anhydride" \\
            ~~~~\}, \\
            ~~~~\{ \\
            ~~~~~~"name": "PyNO", \\
            ~~~~~~"value": 1.\allowbreak{}8, \\
            ~~~~~~"units": "eq", \\
            ~~~~~~"kind": "chemical", \\
            ~~~~~~"role": "Oxidant" \\
            ~~~~\}, \\
            ~~~~\{ \\
            ~~~~~~"name": "sample\_\allowbreak{}name", \\
            ~~~~~~"value": "seed01\_\allowbreak{}<bo\_\allowbreak{}campaign\_\allowbreak{}id\_\allowbreak{}prefix>" \\
            ~~~~\}, \\
            ~~~~\{ \\
            ~~~~~~"name": "target\_\allowbreak{}peak", \\
            ~~~~~~"value": -\allowbreak{}58.\allowbreak{}0, \\
            ~~~~~~"units": "ppm" \\
            ~~~~\}, \\
            ~~~~\{ \\
            ~~~~~~"name": "metric", \\
            ~~~~~~"value": "yield" \\
            ~~~~\}, \\
            ~~~~\{ \\
            ~~~~~~"name": "yield\_\allowbreak{}calculation\_\allowbreak{}chemical", \\
            ~~~~~~"value": "SM" \\
            ~~~~\}, \\
            ~~~~\{ \\
            ~~~~~~"name": "target\_\allowbreak{}peak\_\allowbreak{}deviation", \\
            ~~~~~~"value": 3.\allowbreak{}0, \\
            ~~~~~~"units": "ppm" \\
            ~~~~\}, \\
            ~~~~\{ \\
            ~~~~~~"name": "centerFrequency", \\
            ~~~~~~"value": -\allowbreak{}60.\allowbreak{}0 \\
            ~~~~\}, \\
            ~~~~\{ \\
            ~~~~~~"name": "target\_\allowbreak{}peak\_\allowbreak{}calibration\_\allowbreak{}coeff\_\allowbreak{}1", \\
            ~~~~~~"value": 6973.\allowbreak{}0 \\
            ~~~~\}, \\
            ~~~~\{ \\
            ~~~~~~"name": "target\_\allowbreak{}peak\_\allowbreak{}calibration\_\allowbreak{}coeff\_\allowbreak{}0", \\
            ~~~~~~"value": -\allowbreak{}5.\allowbreak{}4 \\
            ~~~~\}, \\
            ~~~~\{ \\
            ~~~~~~"name": "AcquisitionTime", \\
            ~~~~~~"value": "1.\allowbreak{}64", \\
            ~~~~~~"units": "S" \\
            ~~~~\}, \\
            ~~~~\{ \\
            ~~~~~~"name": "Number", \\
            ~~~~~~"value": "32" \\
            ~~~~\} \\
            ~~], \\
        ~~"note": "BO-\allowbreak{}MCP RoboChemFlex yield optimization seed01\_\allowbreak{}<bo\_\allowbreak{}campaign\_\allowbreak{}id\_\allowbreak{}prefix>" \\
        \}
    \end{tcolorbox}

    The chemical names in the request use the active Robridge setup aliases, e.g.:

    \begin{itemize}
        \tightlist
        \item \texttt{py NO} $\rightarrow$ \texttt{PyNO}
        \item \texttt{4-\allowbreak{}Ph py NO} $\rightarrow$ \texttt{4PhPyNO}
        \item \texttt{Ru bpy Cl} $\rightarrow$ \texttt{Ru-\allowbreak{}bpy-\allowbreak{}Cl}
        \item \texttt{Ru bpy PF6} $\rightarrow$ \texttt{Ru-\allowbreak{}bpy-\allowbreak{}PF}
        \item \texttt{Ir CF3 ppy} $\rightarrow$ \texttt{IrCF3ppy}

    \end{itemize}
    This preview should be suitable to send to the lab collaborator for checking before any real measurements.

}

\prompt{

    Run the campaign with the proposed 6+14 measurements; maybe we continue afterwards after discussing the results. Monitor the campaign throughout, and analyze the results after each experiment. For example, if the Roboflex NMR doesn't find any peak, or if the yield is constantly zero (even though we expect some finite number) for 5 or 6 experiments in a row, you can also stop the campaign.

    A wrong result can for example look like the following:

    "result": \{ "analytes": \{ "main": \{ "yield": 0, "yield (IS)": null, "concentration": 0, "peak integral": 0, "peak width": 0, "peak position": null, "pass": true \} \}, "yield": 0, "pass": true \}, "monitoring": [], "error": null \}

}

\vspace{0.5em}\noindent\textit{\footnotesize [Autonomous monitored execution: 21 status reports followed; the most substantive one is reproduced here:]}

\graficoreslong{

    Campaign execution is underway and I'm monitoring it.

    Current state:

    \begin{itemize}
        \tightlist
        \item \textbf{BO-MCP campaign ID:} \texttt{1d62df6d-\allowbreak{}764a-\allowbreak{}4cf3-\allowbreak{}b857-\allowbreak{}b21482da74a0}
        \item \textbf{RoboFlex campaign ID:} \texttt{robochemflex\_\allowbreak{}yield\_\allowbreak{}bo\_\allowbreak{}20260724T141938Z-\allowbreak{}20260724-\allowbreak{}161937}
        \item \textbf{RoboFlex phase:} \texttt{starting}
        \item \textbf{Progress state:} \texttt{preparing\_\allowbreak{}platform}
        \item \textbf{Blocked on:} \texttt{robot}
        \item \textbf{Overdue:} \texttt{false}
        \item \textbf{Queued runs:} 1
        \item \textbf{Running runs:} 0
        \item \textbf{Completed runs in this campaign:} 0
        \item \textbf{Failed runs:} 0

    \end{itemize}
    First submitted experiment:

    \begin{itemize}
        \tightlist
        \item \textbf{Run ID:} \texttt{R0042}
        \item \textbf{Status:} \texttt{queued}
        \item \textbf{Label:} \texttt{seed01\_\allowbreak{}1d62df6d}
        \item \textbf{Seed condition:} 4CzIPN / PyNO, 0.0025 equiv catalyst, 2.2 equiv TFAA, 1.8 equiv oxidant, 50\% light, 30 min residence time.
        \item The robot is still preparing the platform; this is expected and not overdue.

    \end{itemize}
    I will continue monitoring. After each completed experiment, the supervisor will extract the RoboFlex result, compute/record the green score, submit the result to BO-MCP, and check for repeated zero/no-peak-like NMR outcomes. If it sees \textbf{5 consecutive} suspicious zero/no-peak-like results, it will stop early.

}

\begin{center}\footnotesize\textit{[\ldots]}\end{center}

\prompt{

    What type of NMR analysis will be run? And is that documented somewhere in the files in this workspace?

}

\graficoreslong{

    The campaign is using the RoboFlex analytical method:

    \textbf{\texttt{NMR}}, i.e. RoboFlex's \texttt{NMRAnalysis} method, not \texttt{NMR\_\allowbreak{}dummy}, \texttt{NMR\_\allowbreak{}Agentic\_\allowbreak{}test}, or \texttt{UPLC}.

    For the currently submitted run \texttt{R0042}, the request includes NMR analysis parameters for \textbf{target-peak-based yield calculation}:

    \begin{tcolorbox}[breakable, fontupper=\scriptsize\ttfamily, colback=black!3!white, colframe=black!12!white, boxrule=0.5pt]
        analytical\_\allowbreak{}method: NMR \\
        metric: yield \\
        yield\_\allowbreak{}calculation\_\allowbreak{}chemical: SM \\
        target\_\allowbreak{}peak: -\allowbreak{}58 ppm \\
        target\_\allowbreak{}peak\_\allowbreak{}deviation: 3 ppm \\
        centerFrequency: -\allowbreak{}60 \\
        target\_\allowbreak{}peak\_\allowbreak{}calibration\_\allowbreak{}coeff\_\allowbreak{}1: 6973 mM/\allowbreak{}AU \\
        target\_\allowbreak{}peak\_\allowbreak{}calibration\_\allowbreak{}coeff\_\allowbreak{}0: -\allowbreak{}5.\allowbreak{}4 mM \\
        AcquisitionTime: 1.\allowbreak{}64 s \\
        Number: 32
    \end{tcolorbox}

    So the intended analysis is: \textbf{look for the product/target NMR peak near -58 ppm, within $\pm$3 ppm, convert the peak integral/area to concentration using the linear calibration coefficients, and calculate yield relative to SM.}

    \paragraph{\textbf{Is this documented in the workspace?}}
    Yes, partially in several places.

    \paragraph{\textbf{1. \texttt{constants.\allowbreak{}csv}}}
    This is the main user-facing source for the analysis constants:

    \begin{tcolorbox}[breakable, fontupper=\scriptsize\ttfamily, colback=black!3!white, colframe=black!12!white, boxrule=0.5pt]
        AcquisitionTime,\allowbreak{}1.\allowbreak{}64, \\
        Number,\allowbreak{}32, \\
        yield\_\allowbreak{}calculation\_\allowbreak{}chemical,\allowbreak{}SM, \\
        target\_\allowbreak{}peak,-\allowbreak{}58,\allowbreak{}ppm \\
        target\_\allowbreak{}peak\_\allowbreak{}deviation,\allowbreak{}3,\allowbreak{}ppm \\
        centerFrequency,-\allowbreak{}60, \\
        target\_\allowbreak{}peak\_\allowbreak{}calibration\_\allowbreak{}coeff\_\allowbreak{}1,\allowbreak{}6973, \\
        target\_\allowbreak{}peak\_\allowbreak{}calibration\_\allowbreak{}coeff\_\allowbreak{}0,-\allowbreak{}5.\allowbreak{}4,
    \end{tcolorbox}

    \paragraph{\textbf{2. \texttt{HOW\_\allowbreak{}TO\_\allowbreak{}EXECUTE\_\allowbreak{}CAMPAIGN.\allowbreak{}md}}}
    The plan document repeats these constants and states that the campaign uses:

    \begin{tcolorbox}[breakable, fontupper=\scriptsize\ttfamily, colback=black!3!white, colframe=black!12!white, boxrule=0.5pt]
        Flow Photochemical Reaction with NMR by default
    \end{tcolorbox}

    \paragraph{\textbf{3. \texttt{run\_\allowbreak{}robochemflex\_\allowbreak{}yield\_\allowbreak{}bo.\allowbreak{}py}}}
    The script default is:

    \begin{tcolorbox}[breakable, fontupper=\scriptsize\ttfamily, colback=black!3!white, colframe=black!12!white, boxrule=0.5pt]
        parser.\allowbreak{}add\_\allowbreak{}argument("--\allowbreak{}analytical-\allowbreak{}method", default=\allowbreak{}"NMR")
    \end{tcolorbox}

    So unless overridden, the RoboFlex campaign is started with analytical method \texttt{NMR}.

    \paragraph{\textbf{4. \texttt{robochemflex\_\allowbreak{}yield\_\allowbreak{}bo/\allowbreak{}space.\allowbreak{}py}}}
    This is where the actual Robridge run parameters are constructed. It sends:

    \begin{tcolorbox}[breakable, fontupper=\scriptsize\ttfamily, colback=black!3!white, colframe=black!12!white, boxrule=0.5pt]
        \{"name": "target\_\allowbreak{}peak", "value": -\allowbreak{}58.\allowbreak{}0, "units": "ppm"\} \\
        \{"name": "metric", "value": "yield"\} \\
        \{"name": "yield\_\allowbreak{}calculation\_\allowbreak{}chemical", "value": "SM"\} \\
        \{"name": "target\_\allowbreak{}peak\_\allowbreak{}deviation", "value": 3.\allowbreak{}0, "units": "ppm"\} \\
        \{"name": "centerFrequency", "value": -\allowbreak{}60.\allowbreak{}0\} \\
        \{"name": "target\_\allowbreak{}peak\_\allowbreak{}calibration\_\allowbreak{}coeff\_\allowbreak{}1", "value": 6973.\allowbreak{}0\} \\
        \{"name": "target\_\allowbreak{}peak\_\allowbreak{}calibration\_\allowbreak{}coeff\_\allowbreak{}0", "value": -\allowbreak{}5.\allowbreak{}4\} \\
        \{"name": "AcquisitionTime", "value": "1.\allowbreak{}64", "units": "S"\} \\
        \{"name": "Number", "value": "32"\}
    \end{tcolorbox}

    \paragraph{\textbf{5. \texttt{roboflex\_\allowbreak{}hypothetical\_\allowbreak{}request\_\allowbreak{}seed01.\allowbreak{}json}}}
    The preview file I generated for the lab collaborator also shows the same \texttt{NMR} method and submitted analytical parameters.

    \paragraph{\textbf{What is not fully documented in the CSV files?}}
    The CSVs document the \textbf{analysis constants}, but not the full RoboFlex NMR method definition. The full platform-side \texttt{NMR} capability comes from the live Robridge \texttt{/\allowbreak{}v1/\allowbreak{}capabilities/\allowbreak{}Flow Photochemical Reaction} endpoint.

    From that capability, \texttt{NMR} corresponds to:

    \begin{tcolorbox}[breakable, fontupper=\scriptsize\ttfamily, colback=black!3!white, colframe=black!12!white, boxrule=0.5pt]
        analysis\_\allowbreak{}class: NMRAnalysis \\
        analytical\_\allowbreak{}device: NMR \\
        requires\_\allowbreak{}human: false
    \end{tcolorbox}

    It also has additional optional/default NMR parameters such as:

    \begin{itemize}
        \tightlist
        \item \texttt{protocol}
        \item \texttt{RepetitionTime}
        \item \texttt{PulseAngle}
        \item \texttt{min\_\allowbreak{}SN\_\allowbreak{}ratio}
        \item \texttt{max\_\allowbreak{}peak\_\allowbreak{}width}
        \item \texttt{peak\_\allowbreak{}resolution}
        \item \texttt{solvent}
        \item \texttt{data\_\allowbreak{}folder}

    \end{itemize}
    Our current script does \textbf{not} explicitly set those optional parameters, so RoboFlex will use its platform defaults for them. Notably, the live capability lists the default NMR \texttt{protocol} as \textbf{\texttt{1D EXTENDED+}}, with allowed values including \textbf{\texttt{1D FLUORINE HDEC}}. Since we are targeting a peak at \textbf{-58 ppm}, that default may be worth confirming with the lab collaborator after the first result, especially if we see the ``no peak / zero yield'' pattern you warned about.

}

\vspace{0.5em}\noindent\textit{\footnotesize [Autonomous monitored execution: 3 status reports followed; the most substantive one is reproduced here:]}

\graficoreslong{

    Important update: the first run did \textbf{not} produce a valid measurement.

    \paragraph{\textbf{What happened}}
    \texttt{R0042} failed during the RoboFlex/NMR analysis step, not because of a zero-yield/no-peak result. The actual RoboFlex result says:

    \begin{tcolorbox}[breakable, fontupper=\scriptsize\ttfamily, colback=black!3!white, colframe=black!12!white, boxrule=0.5pt]
        NMR Error: Problem parsing option 'AcquisitionTime' = '1.\allowbreak{}64': \\
        AcquisitionTime : Invalid enumeration value '-\allowbreak{}1' should be one of \\
        '0.\allowbreak{}4,\allowbreak{}0.\allowbreak{}8,\allowbreak{}1.\allowbreak{}6,\allowbreak{}3.\allowbreak{}2,\allowbreak{}6.\allowbreak{}4'
    \end{tcolorbox}

    So this was an \textbf{instrument parameter validation failure}. No valid yield was obtained, and nothing useful should be submitted to BO-MCP.

    \paragraph{\textbf{Immediate action I took}}
    The old verbose monitor/supervisor submitted a duplicate queued run \texttt{R0043} before I could switch monitoring. Since \texttt{R0043} had the same invalid \texttt{AcquisitionTime = 1.\allowbreak{}64}, I immediately requested:

    \begin{tcolorbox}[breakable, fontupper=\scriptsize\ttfamily, colback=black!3!white, colframe=black!12!white, boxrule=0.5pt]
        POST /\allowbreak{}v1/\allowbreak{}campaigns/\allowbreak{}current/\allowbreak{}stop
    \end{tcolorbox}

    to prevent wasting more instrument time. RoboFlex is currently in:

    \begin{itemize}
        \tightlist
        \item \textbf{phase:} \texttt{stopping}
        \item \textbf{state:} \texttt{stopping}
        \item \textbf{queued:} 1
        \item \textbf{running:} 0
        \item \textbf{completed:} 0
        \item \textbf{failed:} 1

    \end{itemize}
    It says it is waiting for the campaign to wind down; not overdue.

    \paragraph{\textbf{Package fix}}
    I handed the runtime failure back to the RoboFlex BO specialist, who patched the package.

    Changes made:

    \begin{itemize}
        \tightlist
        \item The script now sends:

    \end{itemize}
    \begin{tcolorbox}[breakable, fontupper=\scriptsize\ttfamily, colback=black!3!white, colframe=black!12!white, boxrule=0.5pt]
        AcquisitionTime = 1.\allowbreak{}6
    \end{tcolorbox}

    instead of \texttt{1.\allowbreak{}64}.

    \begin{itemize}
        \tightlist
        \item This is documented in \texttt{HOW\_\allowbreak{}TO\_\allowbreak{}EXECUTE\_\allowbreak{}CAMPAIGN.\allowbreak{}md} as a live instrument enum deviation from \texttt{constants.\allowbreak{}csv}.
        \item Failure handling was fixed so failed RoboFlex evaluations are:
              \begin{itemize}
                  \tightlist
                  \item not submitted to BO-MCP,
                  \item not counted as successful experiments,
                  \item recorded as failures with run IDs and messages.
              \end{itemize}
        \item Duplicate hardware submission protection was added.
        \item Quiet/event-only supervision was added.

    \end{itemize}
    \paragraph{\textbf{Current monitoring}}
    I have started a \textbf{quiet stop-wait monitor}. It only reports meaningful phase/status changes or sparse heartbeat-style updates, not every 120 s routine line.

    \paragraph{\textbf{Next step}}
    I will \textbf{not resume hardware} until RoboFlex leaves \texttt{stopping} and returns to \texttt{ready}, and until \texttt{R0043} is no longer queued/running.

    Once safe, the corrected resume will intentionally retry seed 1 with the fixed NMR acquisition time and a retry suffix, using the same BO-MCP campaign:

    \begin{tcolorbox}[breakable, fontupper=\scriptsize\ttfamily, colback=black!3!white, colframe=black!12!white, boxrule=0.5pt]
        uv run python run\_\allowbreak{}robochemflex\_\allowbreak{}yield\_\allowbreak{}bo.\allowbreak{}py \textbackslash{} \\
        ~~--\allowbreak{}mode robridge-\allowbreak{}real \textbackslash{} \\
        ~~--\allowbreak{}allow-\allowbreak{}real-\allowbreak{}roboflex \textbackslash{} \\
        ~~--\allowbreak{}allow-\allowbreak{}hardware-\allowbreak{}retry \textbackslash{} \\
        ~~--\allowbreak{}retry-\allowbreak{}suffix r2 \textbackslash{} \\
        ~~--\allowbreak{}campaign-\allowbreak{}id 1d62df6d-\allowbreak{}764a-\allowbreak{}4cf3-\allowbreak{}b857-\allowbreak{}b21482da74a0 \textbackslash{} \\
        ~~--\allowbreak{}campaign-\allowbreak{}name robochemflex\_\allowbreak{}yield\_\allowbreak{}baybe\_\allowbreak{}real\_\allowbreak{}20260724T141938Z \textbackslash{} \\
        ~~--\allowbreak{}artifact-\allowbreak{}dir artifacts/\allowbreak{}real\_\allowbreak{}robochemflex\_\allowbreak{}yield\_\allowbreak{}bo\_\allowbreak{}resume\_\allowbreak{}r2 \textbackslash{} \\
        ~~--\allowbreak{}max-\allowbreak{}successes 20
    \end{tcolorbox}

    But I will wait for the platform to be ready before doing that.

}

\prompt{

    I noticed that you have requested a Proton (1D experiment)  instead for 1D FLUORINE HDEC. Is that true? Check also constants.csv that I just updated. Maybe you have to update the campaign.

}

\graficoreslong{

    Yes --- your concern is correct.

    The campaign was started with analytical method \textbf{\texttt{NMR}}, but the run payload did \textbf{not} explicitly include the NMR \texttt{protocol} parameter. According to the live RoboFlex capability, the default protocol for \texttt{NMR} is:

    \begin{tcolorbox}[breakable, fontupper=\scriptsize\ttfamily, colback=black!3!white, colframe=black!12!white, boxrule=0.5pt]
        1D EXTENDED+
    \end{tcolorbox}

    So unless RoboFlex applied some hidden lab-side default, the original request likely used the default \textbf{1D EXTENDED+} protocol rather than \textbf{1D FLUORINE HDEC}.

    That is especially relevant because our target peak is at:

    \begin{tcolorbox}[breakable, fontupper=\scriptsize\ttfamily, colback=black!3!white, colframe=black!12!white, boxrule=0.5pt]
        target\_\allowbreak{}peak = -\allowbreak{}58 ppm \\
        centerFrequency = -\allowbreak{}60
    \end{tcolorbox}

    which is consistent with wanting a fluorine NMR method.

    \paragraph{\textbf{I checked your updated \texttt{constants.\allowbreak{}csv}}}
    It now includes:

    \begin{tcolorbox}[breakable, fontupper=\scriptsize\ttfamily, colback=black!3!white, colframe=black!12!white, boxrule=0.5pt]
        Protocol,\allowbreak{}1D FLUORINE HDEC,
    \end{tcolorbox}

    The script previously did \textbf{not} read or send that field.

    \paragraph{\textbf{Package update applied}}
    I had the RoboFlex specialist patch the existing package. The campaign package now explicitly sends:

    \begin{tcolorbox}[breakable, fontupper=\scriptsize\ttfamily, colback=black!3!white, colframe=black!12!white, boxrule=0.5pt]
        \{ \\
        ~~"name": "protocol", \\
        ~~"value": "1D FLUORINE HDEC" \\
        \}
    \end{tcolorbox}

    The CSV key is \texttt{Protocol}, while the RoboFlex API parameter name is lowercase \texttt{protocol}; the package now maps that correctly.

    The patched package also still sends the fixed hardware-compatible acquisition time:

    \begin{tcolorbox}[breakable, fontupper=\scriptsize\ttfamily, colback=black!3!white, colframe=black!12!white, boxrule=0.5pt]
        \{ \\
        ~~"name": "AcquisitionTime", \\
        ~~"value": "1.\allowbreak{}6", \\
        ~~"units": "S" \\
        \}
    \end{tcolorbox}

    rather than \texttt{1.\allowbreak{}64}, because the live NMR parser rejected \texttt{1.\allowbreak{}64} and said the accepted values include \texttt{1.\allowbreak{}6}.

    \paragraph{\textbf{Files updated}}
    \begin{itemize}
        \tightlist
        \item \texttt{robochemflex\_\allowbreak{}yield\_\allowbreak{}bo/\allowbreak{}space.\allowbreak{}py}
        \item \texttt{HOW\_\allowbreak{}TO\_\allowbreak{}EXECUTE\_\allowbreak{}CAMPAIGN.\allowbreak{}md}
        \item \texttt{roboflex\_\allowbreak{}hypothetical\_\allowbreak{}request\_\allowbreak{}seed01.\allowbreak{}json}

    \end{itemize}
    The regenerated preview request now includes both:

    \begin{tcolorbox}[breakable, fontupper=\scriptsize\ttfamily, colback=black!3!white, colframe=black!12!white, boxrule=0.5pt]
        \{ \\
        ~~"name": "protocol", \\
        ~~"value": "1D FLUORINE HDEC" \\
        \}
    \end{tcolorbox}

    and:

    \begin{tcolorbox}[breakable, fontupper=\scriptsize\ttfamily, colback=black!3!white, colframe=black!12!white, boxrule=0.5pt]
        \{ \\
        ~~"name": "AcquisitionTime", \\
        ~~"value": "1.\allowbreak{}6", \\
        ~~"units": "S" \\
        \}
    \end{tcolorbox}

    \paragraph{\textbf{Validation}}
    Only local/dry validation was performed:

    \begin{itemize}
        \tightlist
        \item Python compilation passed.
        \item Generated run parameters were checked locally.
        \item A local-simulation-only BO-MCP smoke test passed.
        \item No RoboFlex mutation endpoint was called.
        \item No real hardware run was submitted.

    \end{itemize}
    \paragraph{\textbf{Current hardware status}}
    RoboFlex is still in \texttt{stopping} from the earlier invalid duplicate run cleanup. I will not resume until it returns to \texttt{ready} and there are no queued/running invalid runs.

}

\prompt{

    You overcorrected, the correction on the protocol is perfect, however the previous NMR error was due to the mismatch between AcquisitionTime and protocol. So revert the acquisition time to ``1.64''. Don't start anything, just correct the campaign.

}

\vspace{0.5em}\noindent\textit{\footnotesize [Autonomous monitored execution: 2 status reports followed; the most substantive one is reproduced here:]}

\graficoreslong{

    Corrected. I did \textbf{not} start or resume anything.

    \paragraph{\textbf{Campaign package correction}}
    The package now sends:

    \begin{tcolorbox}[breakable, fontupper=\scriptsize\ttfamily, colback=black!3!white, colframe=black!12!white, boxrule=0.5pt]
        \{ \\
        ~~"name": "protocol", \\
        ~~"value": "1D FLUORINE HDEC" \\
        \}
    \end{tcolorbox}

    and restores:

    \begin{tcolorbox}[breakable, fontupper=\scriptsize\ttfamily, colback=black!3!white, colframe=black!12!white, boxrule=0.5pt]
        \{ \\
        ~~"name": "AcquisitionTime", \\
        ~~"value": "1.\allowbreak{}64", \\
        ~~"units": "S" \\
        \}
    \end{tcolorbox}

    So the NMR settings now match the updated \texttt{constants.\allowbreak{}csv}:

    \begin{tcolorbox}[breakable, fontupper=\scriptsize\ttfamily, colback=black!3!white, colframe=black!12!white, boxrule=0.5pt]
        Protocol,\allowbreak{}1D FLUORINE HDEC, \\
        AcquisitionTime,\allowbreak{}1.\allowbreak{}64,
    \end{tcolorbox}

    The previous \texttt{1.\allowbreak{}6} override has been removed.

    \paragraph{\textbf{Files updated}}
    \begin{itemize}
        \tightlist
        \item \texttt{robochemflex\_\allowbreak{}yield\_\allowbreak{}bo/\allowbreak{}space.\allowbreak{}py}
        \item \texttt{HOW\_\allowbreak{}TO\_\allowbreak{}EXECUTE\_\allowbreak{}CAMPAIGN.\allowbreak{}md}
        \item \texttt{roboflex\_\allowbreak{}hypothetical\_\allowbreak{}request\_\allowbreak{}seed01.\allowbreak{}json}

    \end{itemize}
    Local validation passed; no real RoboFlex POSTs or measurements were performed.

    \paragraph{\textbf{Hardware status}}
    RoboFlex has now returned to:

    \begin{itemize}
        \tightlist
        \item \textbf{phase:} \texttt{ready}
        \item \textbf{state:} \texttt{ready}
        \item \textbf{queued:} 0
        \item \textbf{running:} 0
        \item \textbf{failed:} 1
        \item \textbf{completed:} 0

    \end{itemize}
    The setup is still present and the robot is idle. No campaign is currently running.

}

\begin{center}\footnotesize\textit{[\ldots]}\end{center}

\prompt{

    Can you now run the BO campaign (6+14 measurements) (maybe from scratch?).

    Monitor the campaign throughout, and analyze the results after each experiment. For example, if the Roboflex NMR doesn't find any peak, or if the yield is constantly zero (even though we expect some finite number) for 5 or 6 experiments in a row, you can also stop the campaign.

    A wrong result can for example look like the following:

    "result": \{   "analytes": \{    "main": \{     "yield": 0,     "yield (IS)": null,     "concentration": 0,     "peak integral": 0,     "peak width": 0,     "peak position": null,     "pass": true    \}   \},   "yield": 0,   "pass": true  \},  "monitoring": [],  "error": null \}

    Use the updated monitoring rules: Continue polling internally every 120--300 s. Print to chat only on meaningful state changes or alerts. Add a low-frequency heartbeat, e.g. every 30--60 min. Continue printing full per-experiment analysis after each completed experiment. Keep full detailed logs on disk for later review. That would reduce context growth from hundreds of routine lines/day to maybe 20--80 lines/day, depending on how many experiments complete and whether any alerts occur.

}

\graficoreslong{

    Started a fresh, corrected BO campaign run with the updated quiet/event-only monitoring.

    \paragraph{\textbf{What is now running}}
    \begin{itemize}
        \tightlist
        \item \textbf{Supervisor:} \texttt{robochemflex-\allowbreak{}quiet-\allowbreak{}real-\allowbreak{}bo}
        \item \textbf{Mode:} real RoboFlex hardware
        \item \textbf{Campaign style:} fresh BO-MCP campaign, 6 informed seeds + 14 BO recommendations
        \item \textbf{Budget:} 20 successful measurements
        \item \textbf{Early stop rule:} stop if 5 consecutive zero/no-peak-like NMR results are detected
        \item \textbf{Per-run timeout:} 6 h
        \item \textbf{Monitoring:} quiet/event-only
              \begin{itemize}
                  \tightlist
                  \item reports meaningful state changes, alerts, heartbeats, and full per-experiment analyses
                  \item keeps detailed logs on disk

              \end{itemize}
    \end{itemize}
    \paragraph{\textbf{Corrected NMR settings now in use}}
    The generated RoboFlex payload includes:

    \begin{tcolorbox}[breakable, fontupper=\scriptsize\ttfamily, colback=black!3!white, colframe=black!12!white, boxrule=0.5pt]
        \{ \\
        ~~"name": "protocol", \\
        ~~"value": "1D FLUORINE HDEC" \\
        \}
    \end{tcolorbox}

    and:

    \begin{tcolorbox}[breakable, fontupper=\scriptsize\ttfamily, colback=black!3!white, colframe=black!12!white, boxrule=0.5pt]
        \{ \\
        ~~"name": "AcquisitionTime", \\
        ~~"value": "1.\allowbreak{}64", \\
        ~~"units": "S" \\
        \}
    \end{tcolorbox}

    \paragraph{\textbf{Log/artifact behavior}}
    The supervisor will write detailed records under timestamped folders in:

    \begin{tcolorbox}[breakable, fontupper=\scriptsize\ttfamily, colback=black!3!white, colframe=black!12!white, boxrule=0.5pt]
        logs/ \\
        artifacts/
    \end{tcolorbox}

    I'll report only meaningful events and analyses as they arrive.

}

\vspace{0.5em}\noindent\textit{\footnotesize [Autonomous monitored execution: 6 status reports followed; the most substantive one is reproduced here:]}

\graficoreslong{

    First corrected run is submitted and waiting for the platform.

    \begin{itemize}
        \tightlist
        \item \textbf{BO-MCP campaign ID:} \texttt{b681a41d-\allowbreak{}90db-\allowbreak{}4eb9-\allowbreak{}a9a9-\allowbreak{}3cc91098f3c0}
        \item \textbf{RoboFlex campaign:} started from the existing setup; no vial setup change.
        \item \textbf{Run:} \texttt{R0044}
        \item \textbf{Label:} \texttt{seed01\_\allowbreak{}b681a41d}
        \item \textbf{Status:} queued
        \item \textbf{Platform state:} \texttt{building\_\allowbreak{}platform}
        \item \textbf{Blocked on:} robot
        \item \textbf{Overdue:} false

    \end{itemize}
    Now waiting quietly for the platform to finish building and start executing \texttt{R0044}.

}

\begin{center}\footnotesize\textit{[\ldots]}\end{center}

\prompt{

    What is the full result of the first measurement?

}

\graficoreslong{

    The full RoboFlex result for the first measurement, \texttt{R0044}, is:

    \begin{tcolorbox}[breakable, fontupper=\scriptsize\ttfamily, colback=black!3!white, colframe=black!12!white, boxrule=0.5pt]
        \{ \\
        ~~"run\_\allowbreak{}id": "R0044", \\
        ~~"status": "completed", \\
        ~~"success": true, \\
        ~~"started\_\allowbreak{}at": "2026-\allowbreak{}07-\allowbreak{}24T16:\allowbreak{}11:\allowbreak{}14.\allowbreak{}200638+00:\allowbreak{}00", \\
        ~~"finished\_\allowbreak{}at": "2026-\allowbreak{}07-\allowbreak{}24T17:\allowbreak{}08:\allowbreak{}15.\allowbreak{}466857+00:\allowbreak{}00", \\
        ~~"collection\_\allowbreak{}vial\_\allowbreak{}id": null, \\
        ~~"parameters": [ \\
            ~~~~\{ \\
            ~~~~~~"name": "light\_\allowbreak{}intensity", \\
            ~~~~~~"value": 50.\allowbreak{}0, \\
            ~~~~~~"units": "\%", \\
            ~~~~~~"type": "NumericalParameter" \\
            ~~~~\}, \\
            ~~~~\{ \\
            ~~~~~~"name": "residence\_\allowbreak{}time", \\
            ~~~~~~"value": 1800.\allowbreak{}0, \\
            ~~~~~~"units": "S", \\
            ~~~~~~"type": "NumericalParameter" \\
            ~~~~\}, \\
            ~~~~\{ \\
            ~~~~~~"name": "slug\_\allowbreak{}volume", \\
            ~~~~~~"value": 650.\allowbreak{}0, \\
            ~~~~~~"units": "uL", \\
            ~~~~~~"type": "NumericalParameter" \\
            ~~~~\}, \\
            ~~~~\{ \\
            ~~~~~~"name": "collect\_\allowbreak{}crude", \\
            ~~~~~~"value": false, \\
            ~~~~~~"type": "ExperimentalParameter" \\
            ~~~~\}, \\
            ~~~~\{ \\
            ~~~~~~"name": "SM", \\
            ~~~~~~"value": 100.\allowbreak{}0, \\
            ~~~~~~"units": "mM", \\
            ~~~~~~"kind": "chemical", \\
            ~~~~~~"sampling\_\allowbreak{}priority": 1000, \\
            ~~~~~~"role": "Limiting Reagent", \\
            ~~~~~~"type": "ChemicalParameter" \\
            ~~~~\}, \\
            ~~~~\{ \\
            ~~~~~~"name": "4CzIPN", \\
            ~~~~~~"value": 0.\allowbreak{}0025, \\
            ~~~~~~"units": "eq", \\
            ~~~~~~"kind": "chemical", \\
            ~~~~~~"sampling\_\allowbreak{}priority": 1000, \\
            ~~~~~~"role": "Catalyst", \\
            ~~~~~~"type": "ChemicalParameter" \\
            ~~~~\}, \\
            ~~~~\{ \\
            ~~~~~~"name": "TFAA", \\
            ~~~~~~"value": 2.\allowbreak{}2, \\
            ~~~~~~"units": "eq", \\
            ~~~~~~"kind": "chemical", \\
            ~~~~~~"sampling\_\allowbreak{}priority": 1000, \\
            ~~~~~~"role": "Anhydride", \\
            ~~~~~~"type": "ChemicalParameter" \\
            ~~~~\}, \\
            ~~~~\{ \\
            ~~~~~~"name": "PyNO", \\
            ~~~~~~"value": 1.\allowbreak{}8, \\
            ~~~~~~"units": "eq", \\
            ~~~~~~"kind": "chemical", \\
            ~~~~~~"sampling\_\allowbreak{}priority": 1000, \\
            ~~~~~~"role": "Oxidant", \\
            ~~~~~~"type": "ChemicalParameter" \\
            ~~~~\}, \\
            ~~~~\{ \\
            ~~~~~~"name": "sample\_\allowbreak{}name", \\
            ~~~~~~"value": "seed01\_\allowbreak{}b681a41d", \\
            ~~~~~~"units": "", \\
            ~~~~~~"type": "AnalyticalParameter" \\
            ~~~~\}, \\
            ~~~~\{ \\
            ~~~~~~"name": "target\_\allowbreak{}peak", \\
            ~~~~~~"value": -\allowbreak{}58.\allowbreak{}0, \\
            ~~~~~~"units": "ppm", \\
            ~~~~~~"type": "AnalyticalParameter" \\
            ~~~~\}, \\
            ~~~~\{ \\
            ~~~~~~"name": "metric", \\
            ~~~~~~"value": "yield", \\
            ~~~~~~"units": "", \\
            ~~~~~~"type": "AnalyticalParameter" \\
            ~~~~\}, \\
            ~~~~\{ \\
            ~~~~~~"name": "yield\_\allowbreak{}calculation\_\allowbreak{}chemical", \\
            ~~~~~~"value": "SM", \\
            ~~~~~~"units": "", \\
            ~~~~~~"type": "AnalyticalParameter" \\
            ~~~~\}, \\
            ~~~~\{ \\
            ~~~~~~"name": "target\_\allowbreak{}peak\_\allowbreak{}deviation", \\
            ~~~~~~"value": 3.\allowbreak{}0, \\
            ~~~~~~"units": "ppm", \\
            ~~~~~~"type": "AnalyticalParameter" \\
            ~~~~\}, \\
            ~~~~\{ \\
            ~~~~~~"name": "centerFrequency", \\
            ~~~~~~"value": -\allowbreak{}60.\allowbreak{}0, \\
            ~~~~~~"units": "", \\
            ~~~~~~"type": "AnalyticalParameter" \\
            ~~~~\}, \\
            ~~~~\{ \\
            ~~~~~~"name": "target\_\allowbreak{}peak\_\allowbreak{}calibration\_\allowbreak{}coeff\_\allowbreak{}1", \\
            ~~~~~~"value": 6973.\allowbreak{}0, \\
            ~~~~~~"units": "mM/\allowbreak{}AU", \\
            ~~~~~~"type": "AnalyticalParameter" \\
            ~~~~\}, \\
            ~~~~\{ \\
            ~~~~~~"name": "target\_\allowbreak{}peak\_\allowbreak{}calibration\_\allowbreak{}coeff\_\allowbreak{}0", \\
            ~~~~~~"value": -\allowbreak{}5.\allowbreak{}4, \\
            ~~~~~~"units": "mM", \\
            ~~~~~~"type": "AnalyticalParameter" \\
            ~~~~\}, \\
            ~~~~\{ \\
            ~~~~~~"name": "protocol", \\
            ~~~~~~"value": "1D FLUORINE HDEC", \\
            ~~~~~~"units": "", \\
            ~~~~~~"type": "AnalyticalParameter" \\
            ~~~~\}, \\
            ~~~~\{ \\
            ~~~~~~"name": "AcquisitionTime", \\
            ~~~~~~"value": "1.\allowbreak{}64", \\
            ~~~~~~"units": "S", \\
            ~~~~~~"type": "AnalyticalParameter" \\
            ~~~~\}, \\
            ~~~~\{ \\
            ~~~~~~"name": "Number", \\
            ~~~~~~"value": "32", \\
            ~~~~~~"units": "", \\
            ~~~~~~"type": "AnalyticalParameter" \\
            ~~~~\} \\
            ~~], \\
        ~~"result": \{ \\
        ~~~~"analytes": \{ \\
        ~~~~~~"main": \{ \\
        ~~~~~~~~"yield": 0.\allowbreak{}0, \\
        ~~~~~~~~"yield (IS)": null, \\
        ~~~~~~~~"concentration": 0.\allowbreak{}0, \\
        ~~~~~~~~"peak integral": 0.\allowbreak{}0, \\
        ~~~~~~~~"peak width": 0.\allowbreak{}0, \\
        ~~~~~~~~"peak position": null, \\
        ~~~~~~~~"pass": true \\
        ~~~~~~\} \\
        ~~~~\}, \\
        ~~~~"yield": 0.\allowbreak{}0, \\
        ~~~~"pass": true \\
        ~~\}, \\
        ~~"monitoring": [], \\
        ~~"error": null \\
        \}
    \end{tcolorbox}

    Summary: the run technically \textbf{completed successfully}, but the analytical result is the suspicious no-peak-like pattern: \texttt{yield = 0.\allowbreak{}0}, \texttt{concentration = 0.\allowbreak{}0}, \texttt{peak integral = 0.\allowbreak{}0}, \texttt{peak width = 0.\allowbreak{}0}, and \texttt{peak position = null}. This is why the monitor marked it as \texttt{bad\_\allowbreak{}zero\_\allowbreak{}or\_\allowbreak{}no\_\allowbreak{}peak=\allowbreak{}True} and set the zero/no-peak streak to 1.

}

\prompt{

    Do you have access to the raw spectrum? Don't change anything, just check if you can access it

}

\graficoreslong{

    Yes --- I can access the raw-file listing for \texttt{R0044}, and it includes the raw/processed NMR spectrum files.

    Relevant files available include:

    \begin{tcolorbox}[breakable, fontupper=\scriptsize\ttfamily, colback=black!3!white, colframe=black!12!white, boxrule=0.5pt]
        raw\_\allowbreak{}data\_\allowbreak{}analysis\textbackslash{}sample\_\allowbreak{}20260724\_\allowbreak{}185954\textbackslash{}acqu.\allowbreak{}par \\
        raw\_\allowbreak{}data\_\allowbreak{}analysis\textbackslash{}sample\_\allowbreak{}20260724\_\allowbreak{}185954\textbackslash{}data.\allowbreak{}1d \\
        raw\_\allowbreak{}data\_\allowbreak{}analysis\textbackslash{}sample\_\allowbreak{}20260724\_\allowbreak{}185954\textbackslash{}data.\allowbreak{}jpg \\
        raw\_\allowbreak{}data\_\allowbreak{}analysis\textbackslash{}sample\_\allowbreak{}20260724\_\allowbreak{}185954\textbackslash{}display.\allowbreak{}par \\
        raw\_\allowbreak{}data\_\allowbreak{}analysis\textbackslash{}sample\_\allowbreak{}20260724\_\allowbreak{}185954\textbackslash{}nmr\_\allowbreak{}fid.\allowbreak{}dx \\
        raw\_\allowbreak{}data\_\allowbreak{}analysis\textbackslash{}sample\_\allowbreak{}20260724\_\allowbreak{}185954\textbackslash{}processed\_\allowbreak{}spectrum.\allowbreak{}dx \\
        raw\_\allowbreak{}data\_\allowbreak{}analysis\textbackslash{}sample\_\allowbreak{}20260724\_\allowbreak{}185954\textbackslash{}spectrum.\allowbreak{}1d
    \end{tcolorbox}

    So yes, the raw FID and processed spectrum appear accessible through Robridge's raw results endpoint. I only listed the available files; I did not download or modify anything.

}

\prompt{

    Can we download them to the local working directory?

}

\graficoreslong{

    Yes --- I downloaded the R0044 NMR raw/processed spectrum files into the local working directory.

    Local directory:

    \begin{tcolorbox}[breakable, fontupper=\scriptsize\ttfamily, colback=black!3!white, colframe=black!12!white, boxrule=0.5pt]
        raw\_\allowbreak{}spectra/\allowbreak{}R0044/
    \end{tcolorbox}

    Downloaded files:

    \begin{tcolorbox}[breakable, fontupper=\scriptsize\ttfamily, colback=black!3!white, colframe=black!12!white, boxrule=0.5pt]
        raw\_\allowbreak{}spectra/\allowbreak{}R0044/\allowbreak{}raw\_\allowbreak{}data\_\allowbreak{}analysis/\allowbreak{}sample\_\allowbreak{}20260724\_\allowbreak{}185954/\allowbreak{}acqu.\allowbreak{}par \\
        raw\_\allowbreak{}spectra/\allowbreak{}R0044/\allowbreak{}raw\_\allowbreak{}data\_\allowbreak{}analysis/\allowbreak{}sample\_\allowbreak{}20260724\_\allowbreak{}185954/\allowbreak{}data.\allowbreak{}1d \\
        raw\_\allowbreak{}spectra/\allowbreak{}R0044/\allowbreak{}raw\_\allowbreak{}data\_\allowbreak{}analysis/\allowbreak{}sample\_\allowbreak{}20260724\_\allowbreak{}185954/\allowbreak{}data.\allowbreak{}jpg \\
        raw\_\allowbreak{}spectra/\allowbreak{}R0044/\allowbreak{}raw\_\allowbreak{}data\_\allowbreak{}analysis/\allowbreak{}sample\_\allowbreak{}20260724\_\allowbreak{}185954/\allowbreak{}display.\allowbreak{}par \\
        raw\_\allowbreak{}spectra/\allowbreak{}R0044/\allowbreak{}raw\_\allowbreak{}data\_\allowbreak{}analysis/\allowbreak{}sample\_\allowbreak{}20260724\_\allowbreak{}185954/\allowbreak{}nmr\_\allowbreak{}fid.\allowbreak{}dx \\
        raw\_\allowbreak{}spectra/\allowbreak{}R0044/\allowbreak{}raw\_\allowbreak{}data\_\allowbreak{}analysis/\allowbreak{}sample\_\allowbreak{}20260724\_\allowbreak{}185954/\allowbreak{}processed\_\allowbreak{}spectrum.\allowbreak{}dx \\
        raw\_\allowbreak{}spectra/\allowbreak{}R0044/\allowbreak{}raw\_\allowbreak{}data\_\allowbreak{}analysis/\allowbreak{}sample\_\allowbreak{}20260724\_\allowbreak{}185954/\allowbreak{}spectrum.\allowbreak{}1d
    \end{tcolorbox}

    I also downloaded the associated input/result CSVs:

    \begin{tcolorbox}[breakable, fontupper=\scriptsize\ttfamily, colback=black!3!white, colframe=black!12!white, boxrule=0.5pt]
        raw\_\allowbreak{}spectra/\allowbreak{}R0044/\allowbreak{}raw\_\allowbreak{}data/\allowbreak{}input\_\allowbreak{}runs/\allowbreak{}R0044\_\allowbreak{}2026-\allowbreak{}07-\allowbreak{}24\_\allowbreak{}181110.\allowbreak{}csv \\
        raw\_\allowbreak{}spectra/\allowbreak{}R0044/\allowbreak{}raw\_\allowbreak{}data/\allowbreak{}results/\allowbreak{}R0044\_\allowbreak{}2026-\allowbreak{}07-\allowbreak{}24\_\allowbreak{}190815.\allowbreak{}csv
    \end{tcolorbox}

    A manifest was written here:

    \begin{tcolorbox}[breakable, fontupper=\scriptsize\ttfamily, colback=black!3!white, colframe=black!12!white, boxrule=0.5pt]
        raw\_\allowbreak{}spectra/\allowbreak{}R0044/\allowbreak{}download\_\allowbreak{}manifest.\allowbreak{}json
    \end{tcolorbox}

    This was read-only with respect to RoboFlex; it did not affect the running campaign.

}

\prompt{

    Can you interpret the NMR spectrum, for example the image in raw\_spectra/R0044/raw\_data\_analysis/sample\_20260724\_185954/data.jpg? Or any of the other data (only if helpful for us). Don't install any Python packages, just do local analyses, no changes or anything that could break anything.

}

\graficoreslong{

    \begin{tcolorbox}[breakable, fontupper=\footnotesize, colback=green!3!white, colframe=blue!5!white, boxrule=0.5pt]
        \faLightbulb\ \underline{\textbf{Thoughts}}

        \textbf{Considering monitoring results}

        I noticed that when I referenced "head spectrum.1d," there were no changes. It confirms with the CSV result. I think it's important to keep an eye on monitoring and be mindful of that --- especially since the user hasn't asked to pause anything. I should mention that the campaign continues with no changes to avoid confusion. However, I want to make sure I don't make my response too long.
    \end{tcolorbox}

    Yes. I did a \textbf{read-only local interpretation} using the downloaded spectrum files, mainly:

    \begin{tcolorbox}[breakable, fontupper=\scriptsize\ttfamily, colback=black!3!white, colframe=black!12!white, boxrule=0.5pt]
        raw\_\allowbreak{}spectra/\allowbreak{}R0044/\allowbreak{}raw\_\allowbreak{}data\_\allowbreak{}analysis/\allowbreak{}sample\_\allowbreak{}20260724\_\allowbreak{}185954/\allowbreak{}processed\_\allowbreak{}spectrum.\allowbreak{}dx \\
        raw\_\allowbreak{}spectra/\allowbreak{}R0044/\allowbreak{}raw\_\allowbreak{}data\_\allowbreak{}analysis/\allowbreak{}sample\_\allowbreak{}20260724\_\allowbreak{}185954/\allowbreak{}acqu.\allowbreak{}par \\
        raw\_\allowbreak{}spectra/\allowbreak{}R0044/\allowbreak{}raw\_\allowbreak{}data\_\allowbreak{}analysis/\allowbreak{}sample\_\allowbreak{}20260724\_\allowbreak{}185954/\allowbreak{}data.\allowbreak{}jpg
    \end{tcolorbox}

    I did \textbf{not} change campaign files, stop anything, install packages, or modify RoboFlex.

    \paragraph{\textbf{Key NMR acquisition details}}
    From \texttt{acqu.\allowbreak{}par} / \texttt{processed\_\allowbreak{}spectrum.\allowbreak{}dx}:

    \begin{itemize}
        \tightlist
        \item \textbf{Nucleus:} \texttt{\textasciicircum{}19F}
        \item \textbf{Protocol:} \texttt{1D FLUORINE HDEC}
        \item \textbf{Scans / averages:} 32
        \item \textbf{AcquisitionTime:} 1.64 s
        \item \textbf{RepetitionTime:} 10 s
        \item \textbf{Center frequency:} -60 ppm
        \item \textbf{Displayed range:} approximately -88 to -32 ppm
        \item \textbf{Processed spectrum range:} approximately -103 to -17 ppm

    \end{itemize}
    So the spectrum was indeed acquired as a fluorine experiment, not a proton experiment.

    \paragraph{\textbf{Spectrum interpretation}}
    The spectrum contains one dominant \textasciicircum{}19F signal:

    \begin{tcolorbox}[breakable, fontupper=\scriptsize\ttfamily, colback=black!3!white, colframe=black!12!white, boxrule=0.5pt]
        Main peak: \textasciitilde{} -\allowbreak{}76.\allowbreak{}47 ppm \\
        Amplitude: \textasciitilde{}0.\allowbreak{}568 arbitrary units
    \end{tcolorbox}

    In the target region:

    \begin{tcolorbox}[breakable, fontupper=\scriptsize\ttfamily, colback=black!3!white, colframe=black!12!white, boxrule=0.5pt]
        Target peak: -\allowbreak{}58 ppm \\
        Allowed target window: -\allowbreak{}58 $\pm$ 3 ppm, i.\allowbreak{}e. -\allowbreak{}61 to -\allowbreak{}55 ppm
    \end{tcolorbox}

    I do \textbf{not} see a convincing target/product peak in that region.

    Numerically:

    \begin{itemize}
        \tightlist
        \item Maximum local signal in the -61 to -55 ppm target window:

    \end{itemize}
    \begin{tcolorbox}[breakable, fontupper=\scriptsize\ttfamily, colback=black!3!white, colframe=black!12!white, boxrule=0.5pt]
        \textasciitilde{} -\allowbreak{}57.\allowbreak{}57 ppm \\
        amplitude \textasciitilde{}0.\allowbreak{}0209
    \end{tcolorbox}

    \begin{itemize}
        \tightlist
        \item This is only about \textbf{3.7\% of the main -76.47 ppm peak amplitude}.
        \item In a tighter $\pm$0.2 ppm window around -58 ppm, the rough positive area is only about \textbf{1.1\% of the main peak area} around -76.47 ppm.
        \item The target-region features look like weak baseline/noise-level structure rather than a clean integrated peak.

    \end{itemize}
    \paragraph{\textbf{Interpretation of the zero-yield result}}
    The automated RoboFlex result:

    \begin{tcolorbox}[breakable, fontupper=\scriptsize\ttfamily, colback=black!3!white, colframe=black!12!white, boxrule=0.5pt]
        "yield": 0.\allowbreak{}0, \\
        "concentration": 0.\allowbreak{}0, \\
        "peak integral": 0.\allowbreak{}0, \\
        "peak width": 0.\allowbreak{}0, \\
        "peak position": null
    \end{tcolorbox}

    is consistent with the processed spectrum: the analysis did not find a valid peak near -58 ppm.

    So for \texttt{R0044}, this appears to be a \textbf{true no-detected-target-peak analytical outcome}, not the previous protocol/acquisition parser error.

    \paragraph{\textbf{Caveat}}
    The strong peak at about \textbf{-76.47 ppm} is likely from an off-target fluorinated species/reagent/byproduct/background component, but I would not assign it definitively from this spectrum alone. Given the reaction components, it could plausibly be related to trifluoroacetic species, but that should be confirmed by the lab collaborator.

}

\vspace{0.5em}\noindent\textit{\footnotesize [Autonomous monitored execution: 1 interim status report suppressed. Concluding report of the run:]}

\graficoreslong{

    \texttt{R0045} has started running.

    \begin{itemize}
        \tightlist
        \item \textbf{Run:} \texttt{R0045}
        \item \textbf{Label:} \texttt{seed02\_\allowbreak{}b681a41d}
        \item \textbf{State:} running
        \item \textbf{Current zero/no-peak streak before this result:} 1

    \end{itemize}
    I'll wait for completion and then analyze the result.

}

\begin{center}\footnotesize\textit{[\ldots]}\end{center}

\prompt{

    Can you log all the existing experiments together with the used parameters into a file?

}

\vspace{0.5em}\noindent\textit{\footnotesize [Autonomous monitored execution: 1 interim status report suppressed. Concluding report of the run:]}

\graficoreslong{

    Logged the existing RoboFlex experiments and parameters to local files.

    \paragraph{\textbf{Files created}}
    Timestamped snapshot:

    \begin{tcolorbox}[breakable, fontupper=\scriptsize\ttfamily, colback=black!3!white, colframe=black!12!white, boxrule=0.5pt]
        campaign\_\allowbreak{}logs/\allowbreak{}roboflex\_\allowbreak{}experiment\_\allowbreak{}log\_\allowbreak{}20260725T032930Z.\allowbreak{}json \\
        campaign\_\allowbreak{}logs/\allowbreak{}roboflex\_\allowbreak{}experiment\_\allowbreak{}log\_\allowbreak{}20260725T032930Z.\allowbreak{}csv
    \end{tcolorbox}

    Convenience ``latest'' copies:

    \begin{tcolorbox}[breakable, fontupper=\scriptsize\ttfamily, colback=black!3!white, colframe=black!12!white, boxrule=0.5pt]
        campaign\_\allowbreak{}logs/\allowbreak{}roboflex\_\allowbreak{}experiment\_\allowbreak{}log\_\allowbreak{}latest.\allowbreak{}json \\
        campaign\_\allowbreak{}logs/\allowbreak{}roboflex\_\allowbreak{}experiment\_\allowbreak{}log\_\allowbreak{}latest.\allowbreak{}csv
    \end{tcolorbox}

    \paragraph{\textbf{Contents}}
    The log includes all visible RoboFlex runs in the active campaign at the time of logging:

    \begin{itemize}
        \tightlist
        \item \texttt{R0044}--\texttt{R0052}
        \item run status
        \item yield/result where available
        \item green score from local BO artifacts where available
        \item full RoboFlex parameter set, including:
              \begin{itemize}
                  \tightlist
                  \item light intensity
                  \item residence time
                  \item catalyst identity/loading
                  \item TFAA loading
                  \item oxidant identity/loading
                  \item NMR protocol
                  \item AcquisitionTime
                  \item Number
                  \item target peak settings
              \end{itemize}
        \item detailed JSON result records in the \texttt{.\allowbreak{}json}

    \end{itemize}
    At the time of file creation, \texttt{R0052} was included from RoboFlex as \texttt{completed} with yield \texttt{11.\allowbreak{}56\%}, but the local BO artifact had not yet caught up with green score/submission analysis, so that row may be missing local artifact fields until the supervisor finishes processing it. The ``latest'' files can be regenerated later to capture subsequent BO-artifact updates and future runs.

}

\prompt{

    Can you plot the BO improvement curve so far?

}

\graficoreslong{

    Yes. I generated the BO improvement curve using the \textbf{analyzed/submitted results so far}.

    \paragraph{\textbf{Output files}}\mbox{}

    \begin{tcolorbox}[breakable, fontupper=\scriptsize\ttfamily, colback=black!3!white, colframe=black!12!white, boxrule=0.5pt]
        plots/\allowbreak{}bo\_\allowbreak{}improvement\_\allowbreak{}curve\_\allowbreak{}so\_\allowbreak{}far.\allowbreak{}png \\
        plots/\allowbreak{}bo\_\allowbreak{}improvement\_\allowbreak{}curve\_\allowbreak{}so\_\allowbreak{}far.\allowbreak{}pdf \\
        plots/\allowbreak{}bo\_\allowbreak{}improvement\_\allowbreak{}curve\_\allowbreak{}data.\allowbreak{}csv
    \end{tcolorbox}

    The plot includes:

    \begin{itemize}
        \tightlist
        \item measured yield per experiment,
        \item best yield so far,
        \item green score on a secondary axis,
        \item marker for the transition from seed phase to BO phase.

    \end{itemize}
    \paragraph{\textbf{Data included}}
    It currently includes \textbf{8 analyzed results}. \texttt{R0052} had completed at the RoboFlex level, but at the time of plotting it had not yet appeared in the supervisor's analyzed/submitted results file, so I did \textbf{not} include it in the official BO improvement curve yet.

    \begin{center}\scriptsize\setlength{\tabcolsep}{4pt}
        \resizebox{\ifdim\width>\linewidth\linewidth\else\width\fi}{!}{%
            \begin{tabular}{llllll}
                \toprule
                Exp. & Run   & Type & Yield \% & Best yield so far \% & Green score \\
                \midrule
                1    & R0044 & seed & 0.00     & 0.00                 & 60.31       \\
                2    & R0045 & seed & 30.04    & 30.04                & 98.30       \\
                3    & R0046 & seed & 15.77    & 30.04                & 4.26        \\
                4    & R0047 & seed & 11.93    & 30.04                & 65.95       \\
                5    & R0048 & seed & 13.14    & 30.04                & 51.57       \\
                6    & R0049 & seed & 7.57     & 30.04                & 56.59       \\
                7    & R0050 & BO   & 15.23    & 30.04                & 100.00      \\
                8    & R0051 & BO   & 27.64    & 30.04                & 93.58       \\
                \bottomrule
            \end{tabular}}
    \end{center}

    Current best yield remains \textbf{30.04\%} from \texttt{R0045}, with BO run \texttt{R0051} close behind at \textbf{27.64\%}.

}

\begin{center}\footnotesize\textit{[\ldots]}\end{center}

\subsubsection{Session 2: campaign continuation and the yield-only refocus (25--29 July 2026)}
\label{si:robochemflex:trace:s2}

The campaign continued with the nine valid measurements of the preceding session as its history. \optima{} carried the yield/green campaign to 20~observations (\texttt{ccbfc92e-\allowbreak{}c646-\allowbreak{}4943-\allowbreak{}a44d-\allowbreak{}9277f2f2d8d4}), then refocused on yield alone (\texttt{1970655b-\allowbreak{}a702-\allowbreak{}4963-\allowbreak{}874b-\allowbreak{}6973489cc89d}, later reseeded as \texttt{13bb9fef-\allowbreak{}7bd3-\allowbreak{}4db2-\allowbreak{}8e99-\allowbreak{}2e39e469dc92}), reaching the best measurement of the campaign, 58.81\%\ yield (\texttt{R0067}). The exchange below starts with the instruction that launched the first autonomous continuation; the preceding turns of the session concern the transfer of the campaign to a replacement workstation and are omitted, since they do not bear on the optimization.

\noindent\textit{\footnotesize [Source: graphchat room \texttt{gallery\_guest\_1784953742092}, conversation \texttt{513305b8-\allowbreak{}d434-\allowbreak{}42a5-\allowbreak{}a324-\allowbreak{}85f3b18d33d5}; 40 operator turns, 19 reproduced.]}

\prompt{

    Now, run the BO continuation for the (for the moment) 11 remaining measurements as discussed before autonomously.

    Monitor the campaign throughout, and analyze the results after each experiment. For example, if the Roboflex NMR doesn't find any peak, or if the yield is constantly zero (even though we expect some finite number) for 5 or 6 experiments in a row, you can also stop the campaign.

    A wrong result can for example look like the following:

    "result": \{   "analytes": \{    "main": \{     "yield": 0,     "yield (IS)": null,     "concentration": 0,     "peak integral": 0,     "peak width": 0,     "peak position": null,     "pass": true    \}   \},   "yield": 0,   "pass": true  \},  "monitoring": [],  "error": null \}

    Use the updated monitoring rules: Continue polling internally every 120--300 s. Print to chat only on meaningful state changes or alerts. Add a low-frequency heartbeat, e.g. every 30--60 min. Continue printing full per-experiment analysis after each completed experiment. Keep full detailed logs on disk for later review. That would reduce context growth from hundreds of routine lines/day to maybe 20--80 lines/day, depending on how many experiments complete and whether any alerts occur.

}

\vspace{0.5em}\noindent\textit{\footnotesize [Autonomous monitored execution: 32 interim status reports suppressed. Concluding report of the run:]}

\graficoreslong{

    The autonomous continuation stopped safely, as designed, after a failed RoboFlex run.

    \paragraph{\textbf{Stop event}}\mbox{}

    \begin{tcolorbox}[breakable, fontupper=\scriptsize\ttfamily, colback=black!3!white, colframe=black!12!white, boxrule=0.5pt]
        Measurement: \#17 \\
        Sample: bo\_\allowbreak{}a9f8598d-\allowbreak{}e \\
        RoboFlex run: R0060 \\
        Status: failed \\
        Reason: analysis result reported pass=\allowbreak{}false
    \end{tcolorbox}

    The script did the right thing:

    \begin{itemize}
        \tightlist
        \item \textbf{No BO result was submitted} for \texttt{R0060}.
        \item The BO campaign was paused.
        \item The autonomous process exited with status \texttt{1}.
        \item RoboFlex returned to:

    \end{itemize}
    \begin{tcolorbox}[breakable, fontupper=\scriptsize\ttfamily, colback=black!3!white, colframe=black!12!white, boxrule=0.5pt]
        phase: running \\
        state: awaiting\_\allowbreak{}run \\
        blocked\_\allowbreak{}on: agent
    \end{tcolorbox}

    So the robot is not actively running a queued experiment now.

    \paragraph{\textbf{R0060 result details}}
    Even though the run failed at the RoboFlex analysis/pass level, the result object contains a finite yield-like value:

    \begin{tcolorbox}[breakable, fontupper=\scriptsize\ttfamily, colback=black!3!white, colframe=black!12!white, boxrule=0.5pt]
        yield: 25.\allowbreak{}03 \\
        concentration: 25.\allowbreak{}0174 \\
        peak\_\allowbreak{}position: -\allowbreak{}57.\allowbreak{}6064 ppm \\
        peak\_\allowbreak{}integral: 0.\allowbreak{}00436217 \\
        peak\_\allowbreak{}width: 0.\allowbreak{}330695 \\
        pass: false
    \end{tcolorbox}

    The failed run was not the zero/no-peak pattern you warned about. It found a peak and finite yield, but RoboFlex marked the analysis as failed because \texttt{pass=\allowbreak{}false}.

    The result payload includes:

    \begin{tcolorbox}[breakable, fontupper=\scriptsize\ttfamily, colback=black!3!white, colframe=black!12!white, boxrule=0.5pt]
        \{ \\
        ~~"result": \{ \\
        ~~~~"analytes": \{ \\
        ~~~~~~"main": \{ \\
        ~~~~~~~~"yield": 25.\allowbreak{}027637631355404, \\
        ~~~~~~~~"yield (IS)": null, \\
        ~~~~~~~~"concentration": 25.\allowbreak{}017434056013393, \\
        ~~~~~~~~"peak integral": 0.\allowbreak{}0043621732476715035, \\
        ~~~~~~~~"peak width": 0.\allowbreak{}33069463327341, \\
        ~~~~~~~~"peak position": -\allowbreak{}57.\allowbreak{}606400749640024, \\
        ~~~~~~~~"pass": "False" \\
        ~~~~~~\} \\
        ~~~~\}, \\
        ~~~~"yield": 25.\allowbreak{}027637631355404, \\
        ~~~~"pass": false \\
        ~~\}, \\
        ~~"status": "failed", \\
        ~~"success": false, \\
        ~~"error": null \\
        \}
    \end{tcolorbox}

    The unusually large peak width may be related to the \texttt{pass=\allowbreak{}false}, but I'm not inferring more without operator/NMR criteria.

    \paragraph{\textbf{Campaign progress before stop}}
    Successfully completed and BO-submitted during this autonomous invocation:

    \begin{tcolorbox}[breakable, fontupper=\scriptsize\ttfamily, colback=black!3!white, colframe=black!12!white, boxrule=0.5pt]
        \#10 R0053 yield 14.\allowbreak{}70 green 94.\allowbreak{}66 \\
        \#11 R0054 yield 15.\allowbreak{}82 green 95.\allowbreak{}31 \\
        \#12 R0055 yield 32.\allowbreak{}96 green 96.\allowbreak{}77 \\
        \#13 R0056 yield 27.\allowbreak{}20 green 95.\allowbreak{}70 \\
        \#14 R0057 yield 32.\allowbreak{}68 green 97.\allowbreak{}33 \\
        \#15 R0058 yield 35.\allowbreak{}78 green 91.\allowbreak{}22 \\
        \#16 R0059 yield 36.\allowbreak{}55 green 86.\allowbreak{}42
    \end{tcolorbox}

    Stopped at:

    \begin{tcolorbox}[breakable, fontupper=\scriptsize\ttfamily, colback=black!3!white, colframe=black!12!white, boxrule=0.5pt]
        \#17 R0060 failed, no BO submission
    \end{tcolorbox}

    Current BO-MCP result count should therefore be:

    \begin{tcolorbox}[breakable, fontupper=\scriptsize\ttfamily, colback=black!3!white, colframe=black!12!white, boxrule=0.5pt]
        9 historical + 7 new = 16 results
    \end{tcolorbox}

    The failed suggestion for measurement \#17 is still likely pending in BO-MCP:

    \begin{tcolorbox}[breakable, fontupper=\scriptsize\ttfamily, colback=black!3!white, colframe=black!12!white, boxrule=0.5pt]
        suggestion\_\allowbreak{}id: a9f8598d-\allowbreak{}edd7-\allowbreak{}48fa-\allowbreak{}bbf6-\allowbreak{}b94ca3618912 \\
        sample: bo\_\allowbreak{}a9f8598d-\allowbreak{}e
    \end{tcolorbox}

    \paragraph{\textbf{Artifacts}}
    Autonomous run artifact directory:

    \begin{tcolorbox}[breakable, fontupper=\scriptsize\ttfamily, colback=black!3!white, colframe=black!12!white, boxrule=0.5pt]
        artifacts/\allowbreak{}recreated\_\allowbreak{}robochemflex\_\allowbreak{}yield\_\allowbreak{}bo\_\allowbreak{}20260725/\allowbreak{}autonomous\_\allowbreak{}continuation\_\allowbreak{}20260725T053043Z/
    \end{tcolorbox}

    R0060 / measurement \#17 files:

    \begin{tcolorbox}[breakable, fontupper=\scriptsize\ttfamily, colback=black!3!white, colframe=black!12!white, boxrule=0.5pt]
        artifacts/\allowbreak{}recreated\_\allowbreak{}robochemflex\_\allowbreak{}yield\_\allowbreak{}bo\_\allowbreak{}20260725/\allowbreak{}autonomous\_\allowbreak{}continuation\_\allowbreak{}20260725T053043Z/\allowbreak{}measurement17/
    \end{tcolorbox}

    Important files there:

    \begin{tcolorbox}[breakable, fontupper=\scriptsize\ttfamily, colback=black!3!white, colframe=black!12!white, boxrule=0.5pt]
        candidate.\allowbreak{}json \\
        suggestion.\allowbreak{}json \\
        roboflex\_\allowbreak{}request.\allowbreak{}json \\
        roboflex\_\allowbreak{}final\_\allowbreak{}run\_\allowbreak{}record.\allowbreak{}json \\
        roboflex\_\allowbreak{}result.\allowbreak{}json \\
        run\_\allowbreak{}poll\_\allowbreak{}trail.\allowbreak{}jsonl
    \end{tcolorbox}

    Run stdout log:

    \begin{tcolorbox}[breakable, fontupper=\scriptsize\ttfamily, colback=black!3!white, colframe=black!12!white, boxrule=0.5pt]
        logs/\allowbreak{}autonomous\_\allowbreak{}continuation\_\allowbreak{}20260725T053042Z/\allowbreak{}autonomous\_\allowbreak{}stdout.\allowbreak{}log
    \end{tcolorbox}

    \paragraph{\textbf{Best result so far}}
    Best yield so far in the autonomous continuation:

    \begin{tcolorbox}[breakable, fontupper=\scriptsize\ttfamily, colback=black!3!white, colframe=black!12!white, boxrule=0.5pt]
        Measurement \#16 / R0059 \\
        yield: 36.\allowbreak{}55\% \\
        green\_\allowbreak{}score: 86.\allowbreak{}42
    \end{tcolorbox}

    Best green among high-yield continuation points appears to be:

    \begin{tcolorbox}[breakable, fontupper=\scriptsize\ttfamily, colback=black!3!white, colframe=black!12!white, boxrule=0.5pt]
        Measurement \#14 / R0057 \\
        yield: 32.\allowbreak{}68\% \\
        green\_\allowbreak{}score: 97.\allowbreak{}33
    \end{tcolorbox}

    \paragraph{\textbf{Recommended next decision}}
    We should \textbf{not automatically continue} until we decide how to treat \texttt{R0060}.

    Options:

    \begin{enumerate}
        \tightlist
        \item \textbf{Treat R0060 as invalid} because RoboFlex marked \texttt{pass=\allowbreak{}false}.
              \begin{itemize}
                  \tightlist
                  \item Then reject/expire the pending BO suggestion if appropriate.
                  \item Continue with a fresh BO suggestion for measurement \#17 replacement.

              \end{itemize}
        \item \textbf{Manually accept R0060 as valid despite \texttt{pass=\allowbreak{}false}} if an operator/NMR expert confirms the peak/yield should be trusted.
              \begin{itemize}
                  \tightlist
                  \item Then submit it to BO-MCP manually with the measured yield and green score.
                  \item Continue from 17 BO results.

              \end{itemize}
        \item \textbf{Rerun the same suggestion} if the failure is suspected to be analytical/instrumental rather than chemistry.
              \begin{itemize}
                  \tightlist
                  \item That would consume another RoboFlex run at the same conditions.

              \end{itemize}
    \end{enumerate}
    Given the script's safety rules and the explicit \texttt{pass=\allowbreak{}false}, it correctly stopped and avoided contaminating the BO model.

}

\prompt{

    Follow suggestion 3 (Rerun the same suggestion if the failure is suspected to be analytical/instrumental rather than chemistry.) and continue the BO campaign from there.

}

\graficoreslong{

    Started the retry/continuation workflow.

    It is using the failed measurement \#17 suggestion again, but with a new non-duplicate sample name:

    \begin{tcolorbox}[breakable, fontupper=\scriptsize\ttfamily, colback=black!3!white, colframe=black!12!white, boxrule=0.5pt]
        original failed sample: bo\_\allowbreak{}a9f8598d-\allowbreak{}e \\
        retry sample: bo\_\allowbreak{}a9f8598d-\allowbreak{}e\_\allowbreak{}r2 \\
        failed run being retried: R0060 \\
        suggestion\_\allowbreak{}id: a9f8598d-\allowbreak{}edd7-\allowbreak{}48fa-\allowbreak{}bbf6-\allowbreak{}b94ca3618912
    \end{tcolorbox}

    The script is configured to:

    \begin{itemize}
        \tightlist
        \item resume the paused BO campaign if needed,
        \item submit the exact same experimental conditions as the failed measurement \#17,
        \item submit the BO result only if the retry passes,
        \item then continue autonomously until the campaign reaches 20 total BO results or 4 valid new measurements complete,
        \item keep the same monitoring rules:
              \begin{itemize}
                  \tightlist
                  \item polling every 180 s,
                  \item heartbeat every 1800 s,
                  \item stop on zero/no-peak streak limit 5,
                  \item meaningful stdout only,
                  \item detailed artifacts on disk.

              \end{itemize}
    \end{itemize}
    Monitor:

    \begin{tcolorbox}[breakable, fontupper=\scriptsize\ttfamily, colback=black!3!white, colframe=black!12!white, boxrule=0.5pt]
        mon\_\allowbreak{}3 / roboflex\_\allowbreak{}bo\_\allowbreak{}failed17\_\allowbreak{}retry\_\allowbreak{}continue
    \end{tcolorbox}

    I'll continue monitoring and report meaningful state changes, alerts, heartbeats, and full per-experiment analyses.

}

\vspace{0.5em}\noindent\textit{\footnotesize [Autonomous monitored execution: 2 status reports followed; the most substantive one is reproduced here:]}

\graficoreslong{

    The retry continuation has begun correctly.

    Key state:

    \begin{tcolorbox}[breakable, fontupper=\scriptsize\ttfamily, colback=black!3!white, colframe=black!12!white, boxrule=0.5pt]
        BO campaign: ccbfc92e-\allowbreak{}c646-\allowbreak{}4943-\allowbreak{}a44d-\allowbreak{}9277f2f2d8d4 \\
        Previous BO state: paused \\
        Action: BO campaign resumed \\
        Retry measurement: \#17 \\
        Retry sample: bo\_\allowbreak{}a9f8598d-\allowbreak{}e\_\allowbreak{}r2 \\
        Original failed suggestion reused: a9f8598d-\allowbreak{}edd7-\allowbreak{}48fa-\allowbreak{}bbf6-\allowbreak{}b94ca3618912
    \end{tcolorbox}

    The script is now submitting the retry of measurement \#17 to RoboFlex. No result yet.

}

\prompt{

    Explain how the scoring of the two different objectives work. What is the BO backend optimizing for effectively?

}

\graficoreslong{

    \begin{tcolorbox}[breakable, fontupper=\footnotesize, colback=green!3!white, colframe=blue!5!white, boxrule=0.5pt]
        \faLightbulb\ \underline{\textbf{Thoughts}}

        \textbf{Considering weighted geometric mean}

        I need to mention the geometric mean weighted by objective desirabilities like d\_y (yield) and d\_g (green). I'm thinking about handling these as d\_y\textasciicircum{}0.8 * d\_g\textasciicircum{}0.2 for a normalized score. It seems important to clarify expectations since BO-MCP suggests optimization improves scalarized desirability, not directly yielding or hitting the Pareto front. The green score is determined by inputs, and I want to ensure the user knows the ongoing monitor is active and can provide explanations.
    \end{tcolorbox}

    The BO campaign has \textbf{two maximize objectives}, but they are not optimized as a Pareto front. They are scalarized into one effective score using \textbf{desirability scalarization} with a \textbf{geometric-mean scalarizer}.

    So, effectively, the BO backend is trying to find conditions that give \textbf{high yield}, while also preferring \textbf{greener/lower-burden conditions}, with yield weighted more strongly.

    \paragraph{\textbf{Objective 1: \texttt{yield\_\allowbreak{}percent}}}
    This is the experimental yield extracted from the RoboFlex/NMR result.

    From the package logic:

    \begin{tcolorbox}[breakable, fontupper=\scriptsize\ttfamily, colback=black!3!white, colframe=black!12!white, boxrule=0.5pt]
        objective\_\allowbreak{}values(candidate, yield\_\allowbreak{}percent)
    \end{tcolorbox}

    returns:

    \begin{tcolorbox}[breakable, fontupper=\scriptsize\ttfamily, colback=black!3!white, colframe=black!12!white, boxrule=0.5pt]
        \{ \\
        ~~"yield\_\allowbreak{}percent": yield\_\allowbreak{}percent, \\
        ~~"green\_\allowbreak{}score": green\_\allowbreak{}score(candidate) \\
        \}
    \end{tcolorbox}

    The yield is clipped to the range:

    \begin{tcolorbox}[breakable, fontupper=\scriptsize\ttfamily, colback=black!3!white, colframe=black!12!white, boxrule=0.5pt]
        0--\allowbreak{}100 \%
    \end{tcolorbox}

    The BO objective definition is:

    \begin{tcolorbox}[breakable, fontupper=\scriptsize\ttfamily, colback=black!3!white, colframe=black!12!white, boxrule=0.5pt]
        \{ \\
        ~~"name": "yield\_\allowbreak{}percent", \\
        ~~"direction": "maximize", \\
        ~~"weight": 0.\allowbreak{}8, \\
        ~~"normalization\_\allowbreak{}bounds": [0.\allowbreak{}0, 100.\allowbreak{}0] \\
        \}
    \end{tcolorbox}

    So a 35\% yield corresponds roughly to a normalized desirability of:

    \begin{tcolorbox}[breakable, fontupper=\scriptsize\ttfamily, colback=black!3!white, colframe=black!12!white, boxrule=0.5pt]
        35 / 100 = 0.\allowbreak{}35
    \end{tcolorbox}

    for that objective.

    This is the \textbf{primary objective}.

    \paragraph{\textbf{Objective 2: \texttt{green\_\allowbreak{}score}}}
    The \texttt{green\_\allowbreak{}score} is not measured experimentally. It is computed deterministically from the candidate conditions.

    The code is:

    \begin{tcolorbox}[breakable, fontupper=\scriptsize\ttfamily, colback=black!3!white, colframe=black!12!white, boxrule=0.5pt]
        cat = (catalyst\_\allowbreak{}equiv - 0.\allowbreak{}001) / (0.\allowbreak{}004 - 0.\allowbreak{}001) \\
        tfaa = (TFAA\_\allowbreak{}equiv - 0.\allowbreak{}9) / (3.\allowbreak{}5 - 0.\allowbreak{}9) \\
        oxidant = (oxidant\_\allowbreak{}equiv - 0.\allowbreak{}9) / (3.\allowbreak{}0 - 0.\allowbreak{}9) \\
        ~ \\
        photonic = (light\_\allowbreak{}intensity / 100.\allowbreak{}0) * ( \\
        ~~~~(residence\_\allowbreak{}time\_\allowbreak{}min - 2.\allowbreak{}0) / (90.\allowbreak{}0 - 2.\allowbreak{}0) \\
        ) \\
        ~ \\
        penalty = 0.\allowbreak{}25 * cat + 0.\allowbreak{}25 * tfaa + 0.\allowbreak{}25 * oxidant + 0.\allowbreak{}25 * photonic \\
        ~ \\
        green\_\allowbreak{}score = 100 * (1 - penalty)
    \end{tcolorbox}

    clipped to:

    \begin{tcolorbox}[breakable, fontupper=\scriptsize\ttfamily, colback=black!3!white, colframe=black!12!white, boxrule=0.5pt]
        0--\allowbreak{}100
    \end{tcolorbox}

    So \texttt{green\_\allowbreak{}score} rewards:

    \begin{itemize}
        \tightlist
        \item lower catalyst loading,
        \item lower TFAA loading,
        \item lower oxidant loading,
        \item lower combined light/residence-time burden.

    \end{itemize}
    The four green-score penalty terms are equally weighted:

    \begin{tcolorbox}[breakable, fontupper=\scriptsize\ttfamily, colback=black!3!white, colframe=black!12!white, boxrule=0.5pt]
        25\% catalyst loading penalty \\
        25\% TFAA loading penalty \\
        25\% oxidant loading penalty \\
        25\% photonic burden penalty
    \end{tcolorbox}

    The photonic burden is not just light intensity and not just residence time; it is their product-like combination:

    \begin{tcolorbox}[breakable, fontupper=\scriptsize\ttfamily, colback=black!3!white, colframe=black!12!white, boxrule=0.5pt]
        (light\_\allowbreak{}intensity / 100) $\times$ normalized residence time
    \end{tcolorbox}

    So, for example:

    \begin{itemize}
        \tightlist
        \item long residence time with light off is not heavily penalized by the photonic term,
        \item high light intensity with very short residence time is only modestly penalized,
        \item high light intensity plus long residence time is strongly penalized.

    \end{itemize}
    The BO objective definition is:

    \begin{tcolorbox}[breakable, fontupper=\scriptsize\ttfamily, colback=black!3!white, colframe=black!12!white, boxrule=0.5pt]
        \{ \\
        ~~"name": "green\_\allowbreak{}score", \\
        ~~"direction": "maximize", \\
        ~~"weight": 0.\allowbreak{}2, \\
        ~~"normalization\_\allowbreak{}bounds": [0.\allowbreak{}0, 100.\allowbreak{}0] \\
        \}
    \end{tcolorbox}

    So this is the \textbf{secondary objective}.

    \paragraph{\textbf{Effective scalar objective}}
    The BO-MCP intake says:

    \begin{tcolorbox}[breakable, fontupper=\scriptsize\ttfamily, colback=black!3!white, colframe=black!12!white, boxrule=0.5pt]
        \{ \\
        ~~"scalarization": "desirability", \\
        ~~"scalarizer": "geom\_\allowbreak{}mean", \\
        ~~"acquisition\_\allowbreak{}method": "expected\_\allowbreak{}improvement" \\
        \}
    \end{tcolorbox}

    With both objectives normalized to \texttt{[0, 100]}, a useful way to think about the effective score is approximately:

    \begin{tcolorbox}[breakable, fontupper=\scriptsize\ttfamily, colback=black!3!white, colframe=black!12!white, boxrule=0.5pt]
        desirability $\approx$ (yield\_\allowbreak{}percent / 100)\textasciicircum{}0.\allowbreak{}8 $\times$ (green\_\allowbreak{}score / 100)\textasciicircum{}0.\allowbreak{}2
    \end{tcolorbox}

    That is, a \textbf{weighted geometric mean} of normalized yield and normalized green score.

    Because it is a geometric mean, a very poor value in either objective hurts the combined score, but the \texttt{0.\allowbreak{}8 / 0.\allowbreak{}2} weights make yield much more important.

    \paragraph{\textbf{What is the BO backend optimizing for?}}
    The backend is using BayBE with expected improvement. So it is effectively optimizing:

    \begin{tcolorbox}[breakable, fontupper=\scriptsize\ttfamily, colback=black!3!white, colframe=black!12!white, boxrule=0.5pt]
        expected improvement in the scalarized desirability score
    \end{tcolorbox}

    not raw yield alone.

    In plain language:

    > The optimizer mostly seeks higher yield, but if two conditions have similar predicted yields, it will prefer the one with lower catalyst/TFAA/oxidant burden and lower light-time burden.

    \paragraph{\textbf{Example intuition}}
    Suppose two experiments have similar yields:

    \begin{center}\scriptsize\setlength{\tabcolsep}{4pt}
        \resizebox{\ifdim\width>\linewidth\linewidth\else\width\fi}{!}{%
            \begin{tabular}{lll}
                \toprule
                Yield & Green score & Interpretation                        \\
                \midrule
                36\%  & 86          & higher yield but less green           \\
                33\%  & 97          & slightly lower yield but much greener \\
                \bottomrule
            \end{tabular}}
    \end{center}

    Because yield has weight \texttt{0.\allowbreak{}8}, the 36\% yield point may still score better. But the greener 33\% point remains competitive because green contributes through the geometric scalarization.

    Approximate desirability:

    \begin{tcolorbox}[breakable, fontupper=\scriptsize\ttfamily, colback=black!3!white, colframe=black!12!white, boxrule=0.5pt]
        36\% yield, 86 green: \\
        (0.\allowbreak{}36)\textasciicircum{}0.\allowbreak{}8 $\times$ (0.\allowbreak{}86)\textasciicircum{}0.\allowbreak{}2 $\approx$ 0.\allowbreak{}427 \\
        ~ \\
        33\% yield, 97 green: \\
        (0.\allowbreak{}33)\textasciicircum{}0.\allowbreak{}8 $\times$ (0.\allowbreak{}97)\textasciicircum{}0.\allowbreak{}2 $\approx$ 0.\allowbreak{}409
    \end{tcolorbox}

    So the optimizer would slightly prefer the 36\% yield case, but not overwhelmingly. If the yield difference were smaller, the greener condition could win.

    \paragraph{\textbf{Important consequence}}
    The campaign is \textbf{not purely maximizing yield}.

    It is optimizing a yield-dominant desirability function:

    \begin{tcolorbox}[breakable, fontupper=\scriptsize\ttfamily, colback=black!3!white, colframe=black!12!white, boxrule=0.5pt]
        80\% emphasis on yield \\
        20\% emphasis on green score
    \end{tcolorbox}

    So the backend may intentionally suggest lower-reagent, shorter-residence, or lower-light conditions if the model believes they preserve yield reasonably well.

}

\vspace{0.5em}\noindent\textit{\footnotesize [Autonomous monitored execution: 10 status reports followed; the most substantive one is reproduced here:]}

\graficoreslong{

    The retry-continuation stopped safely because of a \textbf{BO-MCP read timeout while generating the next suggestion}, not because of a RoboFlex/hardware failure.

    \paragraph{\textbf{What happened}}
    The workflow had successfully completed and submitted:

    \begin{tcolorbox}[breakable, fontupper=\scriptsize\ttfamily, colback=black!3!white, colframe=black!12!white, boxrule=0.5pt]
        \#17 retry / R0061 / bo\_\allowbreak{}a9f8598d-\allowbreak{}e\_\allowbreak{}r2 \\
        yield: 48.\allowbreak{}51 \\
        green: 82.\allowbreak{}50
    \end{tcolorbox}

    and:

    \begin{tcolorbox}[breakable, fontupper=\scriptsize\ttfamily, colback=black!3!white, colframe=black!12!white, boxrule=0.5pt]
        \#18 / R0062 / bo\_\allowbreak{}bb5d0b0c-\allowbreak{}c \\
        yield: 20.\allowbreak{}18 \\
        green: 80.\allowbreak{}96
    \end{tcolorbox}

    Then, while trying to generate the next BO suggestion for measurement \#19, the BO-MCP request timed out:

    \begin{tcolorbox}[breakable, fontupper=\scriptsize\ttfamily, colback=black!3!white, colframe=black!12!white, boxrule=0.5pt]
        requests.\allowbreak{}exceptions.\allowbreak{}ReadTimeout: HTTPConnectionPool(host=\allowbreak{}'api', port=\allowbreak{}8000): Read timed out. (read timeout=\allowbreak{}240.\allowbreak{}0)
    \end{tcolorbox}

    The script responded safely:

    \begin{itemize}
        \tightlist
        \item no measurement \#19 was submitted to RoboFlex,
        \item BO campaign was paused,
        \item detailed artifacts were exported,
        \item process exited nonzero.

    \end{itemize}
    \paragraph{\textbf{Current checked state}}
    I checked live/read-only state after the stop:

    \paragraph{\textbf{RoboFlex}}\mbox{}

    \begin{tcolorbox}[breakable, fontupper=\scriptsize\ttfamily, colback=black!3!white, colframe=black!12!white, boxrule=0.5pt]
        mode: hardware \\
        phase: running \\
        progress.\allowbreak{}state: awaiting\_\allowbreak{}run \\
        queue\_\allowbreak{}depth: 0 \\
        active\_\allowbreak{}run\_\allowbreak{}ids: [] \\
        runs\_\allowbreak{}completed: 18 \\
        runs\_\allowbreak{}failed: 1
    \end{tcolorbox}

    So RoboFlex is idle and waiting for the next run.

    \paragraph{\textbf{BO-MCP}}\mbox{}

    \begin{tcolorbox}[breakable, fontupper=\scriptsize\ttfamily, colback=black!3!white, colframe=black!12!white, boxrule=0.5pt]
        campaign status: paused \\
        BO results: 18 \\
        pending suggestions: none
    \end{tcolorbox}

    So the timeout did \textbf{not} leave a pending measurement \#19 suggestion behind.

    \paragraph{\textbf{Resume action}}
    I patched/added a safer ``resume from current 18 results'' path that handles exactly this case, including timeout recovery around \texttt{generate\_\allowbreak{}suggestions}.

    New files:

    \begin{tcolorbox}[breakable, fontupper=\scriptsize\ttfamily, colback=black!3!white, colframe=black!12!white, boxrule=0.5pt]
        continue\_\allowbreak{}robochemflex\_\allowbreak{}yield\_\allowbreak{}bo\_\allowbreak{}from18.\allowbreak{}py \\
        robochemflex\_\allowbreak{}yield\_\allowbreak{}bo/\allowbreak{}current\_\allowbreak{}resume.\allowbreak{}py
    \end{tcolorbox}

    The new script starts only if:

    \begin{tcolorbox}[breakable, fontupper=\scriptsize\ttfamily, colback=black!3!white, colframe=black!12!white, boxrule=0.5pt]
        BO results =\allowbreak{}= 18 \\
        pending suggestions =\allowbreak{}= 0
    \end{tcolorbox}

    and then continues to:

    \begin{tcolorbox}[breakable, fontupper=\scriptsize\ttfamily, colback=black!3!white, colframe=black!12!white, boxrule=0.5pt]
        target\_\allowbreak{}total\_\allowbreak{}results = 20 \\
        max\_\allowbreak{}new\_\allowbreak{}measurements = 2
    \end{tcolorbox}

    It also blocks accidentally reusing the old failed \#17 suggestion.

    I have now started this resume workflow under monitor:

    \begin{tcolorbox}[breakable, fontupper=\scriptsize\ttfamily, colback=black!3!white, colframe=black!12!white, boxrule=0.5pt]
        mon\_\allowbreak{}4 / roboflex\_\allowbreak{}bo\_\allowbreak{}resume\_\allowbreak{}from18\_\allowbreak{}to20
    \end{tcolorbox}

    It will attempt measurements \#19 and \#20 with the same monitoring rules:

    \begin{tcolorbox}[breakable, fontupper=\scriptsize\ttfamily, colback=black!3!white, colframe=black!12!white, boxrule=0.5pt]
        poll: 180 s \\
        heartbeat: 1800 s \\
        zero/\allowbreak{}no-\allowbreak{}peak streak limit: 5 \\
        quiet meaningful stdout \\
        full artifacts on disk
    \end{tcolorbox}

    I'll report the next meaningful state change, alert, heartbeat, or completed-experiment analysis.

}

\begin{center}\footnotesize\textit{[\ldots]}\end{center}

\prompt{

    Can you continue the BO campaign from here, using your recommended patched command with monitoring?

}

\graficoreslong{

    Started the patched continuation with monitoring.

    Monitor:

    \begin{tcolorbox}[breakable, fontupper=\scriptsize\ttfamily, colback=black!3!white, colframe=black!12!white, boxrule=0.5pt]
        mon\_\allowbreak{}5 / roboflex\_\allowbreak{}bo\_\allowbreak{}resume\_\allowbreak{}from18\_\allowbreak{}to20\_\allowbreak{}patched
    \end{tcolorbox}

    It is using the patched command with:

    \begin{tcolorbox}[breakable, fontupper=\scriptsize\ttfamily, colback=black!3!white, colframe=black!12!white, boxrule=0.5pt]
        BO results expected at start: 18 \\
        target total results: 20 \\
        max new measurements: 2 \\
        BO generate timeout: 1200 s \\
        BO generate retries: 2 \\
        poll interval: 180 s \\
        heartbeat: 1800 s \\
        zero/\allowbreak{}no-\allowbreak{}peak streak limit: 5
    \end{tcolorbox}

    Logs will be written under a timestamped directory like:

    \begin{tcolorbox}[breakable, fontupper=\scriptsize\ttfamily, colback=black!3!white, colframe=black!12!white, boxrule=0.5pt]
        logs/\allowbreak{}resume\_\allowbreak{}from18\_\allowbreak{}to20\_\allowbreak{}patched\_\allowbreak{}<UTC>/\allowbreak{}stdout.\allowbreak{}log
    \end{tcolorbox}

    I'll report meaningful state changes, alerts, heartbeats, and full per-experiment analyses as they occur.

}

\vspace{0.5em}\noindent\textit{\footnotesize [Autonomous monitored execution: 5 interim status reports suppressed. Concluding report of the run:]}

\graficoreslong{

    The patched continuation stopped safely after measurement \#19 failed RoboFlex analysis.

    \paragraph{\textbf{Stop event}}\mbox{}

    \begin{tcolorbox}[breakable, fontupper=\scriptsize\ttfamily, colback=black!3!white, colframe=black!12!white, boxrule=0.5pt]
        Measurement: \#19 \\
        Sample: bo\_\allowbreak{}5c5570bd-\allowbreak{}d \\
        RoboFlex run: R0063 \\
        Status: failed \\
        Reason: analysis result reported pass=\allowbreak{}false
    \end{tcolorbox}

    As designed:

    \begin{itemize}
        \tightlist
        \item \textbf{No BO result was submitted} for \texttt{R0063}.
        \item The BO campaign was paused.
        \item No measurement \#20 was attempted.
        \item RoboFlex returned to idle/awaiting-run state.

    \end{itemize}
    \paragraph{\textbf{Measurement \#19 failed-result details}}
    The run did produce a finite NMR-derived yield-like value, but RoboFlex marked the result as failed:

    \begin{tcolorbox}[breakable, fontupper=\scriptsize\ttfamily, colback=black!3!white, colframe=black!12!white, boxrule=0.5pt]
        yield: 27.\allowbreak{}04 \\
        concentration: 27.\allowbreak{}0339 \\
        peak\_\allowbreak{}position: -\allowbreak{}57.\allowbreak{}6746 ppm \\
        peak\_\allowbreak{}integral: 0.\allowbreak{}00465136 \\
        peak\_\allowbreak{}width: 0.\allowbreak{}278203 \\
        pass: false \\
        status: failed \\
        success: false \\
        error: null
    \end{tcolorbox}

    So this is similar to the earlier \texttt{R0060} case:

    \begin{itemize}
        \tightlist
        \item not a no-peak / zero-yield failure,
        \item a finite peak/yield was found,
        \item but analysis \texttt{pass=\allowbreak{}false}, likely due to analysis quality criteria such as peak shape/width/position.

    \end{itemize}
    The script therefore correctly avoided adding it to BO-MCP.

    \paragraph{\textbf{Current campaign status}}
    Current valid BO results remain:

    \begin{tcolorbox}[breakable, fontupper=\scriptsize\ttfamily, colback=black!3!white, colframe=black!12!white, boxrule=0.5pt]
        18
    \end{tcolorbox}

    The latest valid results are:

    \begin{tcolorbox}[breakable, fontupper=\scriptsize\ttfamily, colback=black!3!white, colframe=black!12!white, boxrule=0.5pt]
        \#17 retry / R0061 / bo\_\allowbreak{}a9f8598d-\allowbreak{}e\_\allowbreak{}r2 \\
        yield: 48.\allowbreak{}51 \\
        green\_\allowbreak{}score: 82.\allowbreak{}50 \\
        ~ \\
        \#18 / R0062 / bo\_\allowbreak{}bb5d0b0c-\allowbreak{}c \\
        yield: 20.\allowbreak{}18 \\
        green\_\allowbreak{}score: 80.\allowbreak{}96
    \end{tcolorbox}

    Failed and not BO-submitted:

    \begin{tcolorbox}[breakable, fontupper=\scriptsize\ttfamily, colback=black!3!white, colframe=black!12!white, boxrule=0.5pt]
        \#19 / R0063 / bo\_\allowbreak{}5c5570bd-\allowbreak{}d \\
        yield-\allowbreak{}like value: 27.\allowbreak{}04 \\
        pass: false \\
        not submitted to BO-\allowbreak{}MCP
    \end{tcolorbox}

    \paragraph{\textbf{Artifacts}}
    Latest resume artifact directory:

    \begin{tcolorbox}[breakable, fontupper=\scriptsize\ttfamily, colback=black!3!white, colframe=black!12!white, boxrule=0.5pt]
        artifacts/\allowbreak{}recreated\_\allowbreak{}robochemflex\_\allowbreak{}yield\_\allowbreak{}bo\_\allowbreak{}20260725/\allowbreak{}resume\_\allowbreak{}from18\_\allowbreak{}to20\_\allowbreak{}20260725T191436Z/
    \end{tcolorbox}

    Measurement \#19 artifacts:

    \begin{tcolorbox}[breakable, fontupper=\scriptsize\ttfamily, colback=black!3!white, colframe=black!12!white, boxrule=0.5pt]
        artifacts/\allowbreak{}recreated\_\allowbreak{}robochemflex\_\allowbreak{}yield\_\allowbreak{}bo\_\allowbreak{}20260725/\allowbreak{}resume\_\allowbreak{}from18\_\allowbreak{}to20\_\allowbreak{}20260725T191436Z/\allowbreak{}measurement19/
    \end{tcolorbox}

    Log:

    \begin{tcolorbox}[breakable, fontupper=\scriptsize\ttfamily, colback=black!3!white, colframe=black!12!white, boxrule=0.5pt]
        logs/\allowbreak{}resume\_\allowbreak{}from18\_\allowbreak{}to20\_\allowbreak{}patched\_\allowbreak{}20260725T191435Z/\allowbreak{}stdout.\allowbreak{}log
    \end{tcolorbox}

    \paragraph{\textbf{Recommended next decision}}
    This is now the second BO-suggested point in the same \textasciitilde{}16--17 min residence-time region that produced a \texttt{pass=\allowbreak{}false} result:

    \begin{itemize}
        \tightlist
        \item \texttt{R0060}, measurement \#17 original:
              \begin{itemize}
                  \tightlist
                  \item residence time \textasciitilde{}16.43 min
                  \item yield-like value \textasciitilde{}25.03
                  \item pass=false
              \end{itemize}
        \item \texttt{R0063}, measurement \#19:
              \begin{itemize}
                  \tightlist
                  \item residence time \textasciitilde{}16.81 min
                  \item yield-like value \textasciitilde{}27.04
                  \item pass=false

              \end{itemize}
    \end{itemize}
    But the rerun of measurement \#17 passed and gave a strong yield:

    \begin{itemize}
        \tightlist
        \item \texttt{R0061}, same conditions as failed R0060:
              \begin{itemize}
                  \tightlist
                  \item yield 48.51
                  \item pass=true

              \end{itemize}
    \end{itemize}
    So we have two possible interpretations:

    \begin{enumerate}
        \tightlist
        \item These are intermittent analytical/NMR pass-fail issues, and we should rerun \texttt{R0063} once, analogous to how we handled \texttt{R0060}.

        \item The pass=false criteria are indicating a systematic quality issue in this region, and we should stop or ask an operator to inspect the raw NMR/analysis criteria before continuing.

    \end{enumerate}
    Given that rerunning \texttt{R0060} was successful and informative, a reasonable next step would be to \textbf{rerun the same \#19 suggestion once} with a new sample name suffix, e.g.:

    \begin{tcolorbox}[breakable, fontupper=\scriptsize\ttfamily, colback=black!3!white, colframe=black!12!white, boxrule=0.5pt]
        bo\_\allowbreak{}5c5570bd-\allowbreak{}d\_\allowbreak{}r2
    \end{tcolorbox}

    and only submit to BO-MCP if the rerun passes.

    I have not done that yet.

}

\prompt{

    Can we understand better what the failure actually means? Does the failure directly come from the Roboflex device, or was it interpretation on our side as well?

}

\graficoreslong{

    Yes --- we can understand the failure more clearly now.

    \paragraph{\textbf{Short answer}}
    The failure was \textbf{reported directly by RoboFlex/Robridge}, not invented by our BO script.

    However, the wording:

    \begin{tcolorbox}[breakable, fontupper=\scriptsize\ttfamily, colback=black!3!white, colframe=black!12!white, boxrule=0.5pt]
        analysis result reported pass=\allowbreak{}false
    \end{tcolorbox}

    is from \textbf{our script's interpretation layer}. Our script looked at the RoboFlex result payload, saw that RoboFlex itself marked the run/result as failed and \texttt{pass=\allowbreak{}false}, and then stopped safely.

    So:

    \begin{tcolorbox}[breakable, fontupper=\scriptsize\ttfamily, colback=black!3!white, colframe=black!12!white, boxrule=0.5pt]
        RoboFlex/\allowbreak{}API source of failure: yes \\
        Our interpretation: only the decision to stop and not submit to BO-\allowbreak{}MCP
    \end{tcolorbox}

    \paragraph{\textbf{Evidence from R0063}}
    RoboFlex returned the final run record as:

    \begin{tcolorbox}[breakable, fontupper=\scriptsize\ttfamily, colback=black!3!white, colframe=black!12!white, boxrule=0.5pt]
        \{ \\
        ~~"run\_\allowbreak{}id": "R0063", \\
        ~~"status": "failed", \\
        ~~"success": false, \\
        ~~"error": null \\
        \}
    \end{tcolorbox}

    That \texttt{status: failed} and \texttt{success: false} are directly from RoboFlex/Robridge.

    The result payload also contains:

    \begin{tcolorbox}[breakable, fontupper=\scriptsize\ttfamily, colback=black!3!white, colframe=black!12!white, boxrule=0.5pt]
        \{ \\
        ~~"result": \{ \\
        ~~~~"analytes": \{ \\
        ~~~~~~"main": \{ \\
        ~~~~~~~~"yield": 27.\allowbreak{}04493570345765, \\
        ~~~~~~~~"concentration": 27.\allowbreak{}033909691209317, \\
        ~~~~~~~~"peak integral": 0.\allowbreak{}004651356617124526, \\
        ~~~~~~~~"peak width": 0.\allowbreak{}27820342164270784, \\
        ~~~~~~~~"peak position": -\allowbreak{}57.\allowbreak{}67463932475993, \\
        ~~~~~~~~"pass": "False" \\
        ~~~~~~\} \\
        ~~~~\}, \\
        ~~~~"yield": 27.\allowbreak{}04493570345765, \\
        ~~~~"pass": false \\
        ~~\}, \\
        ~~"status": "failed", \\
        ~~"success": false, \\
        ~~"error": null \\
        \}
    \end{tcolorbox}

    The top-level \texttt{result.\allowbreak{}pass: false} is also from RoboFlex/Robridge.

    \paragraph{\textbf{What our code did}}
    Our code checks:

    \begin{tcolorbox}[breakable, fontupper=\scriptsize\ttfamily, colback=black!3!white, colframe=black!12!white, boxrule=0.5pt]
        if record.\allowbreak{}get("status") =\allowbreak{}= "failed" or record.\allowbreak{}get("success") is False: \\
        ~~~~return record.\allowbreak{}get("error") or \_\allowbreak{}result\_\allowbreak{}failure(payload) or "RoboFlex run failed"
    \end{tcolorbox}

    Then \texttt{\_\allowbreak{}result\_\allowbreak{}failure(payload)} checks:

    \begin{tcolorbox}[breakable, fontupper=\scriptsize\ttfamily, colback=black!3!white, colframe=black!12!white, boxrule=0.5pt]
        if payload.\allowbreak{}get("pass") is False: \\
        ~~~~return "analysis result reported pass=\allowbreak{}false"
    \end{tcolorbox}

    So the phrase:

    \begin{tcolorbox}[breakable, fontupper=\scriptsize\ttfamily, colback=black!3!white, colframe=black!12!white, boxrule=0.5pt]
        analysis result reported pass=\allowbreak{}false
    \end{tcolorbox}

    is our generated explanatory message, but it is based on RoboFlex's own payload:

    \begin{tcolorbox}[breakable, fontupper=\scriptsize\ttfamily, colback=black!3!white, colframe=black!12!white, boxrule=0.5pt]
        status = failed \\
        success = false \\
        result.\allowbreak{}pass = false
    \end{tcolorbox}

    \paragraph{\textbf{What kind of failure was it?}}
    This does \textbf{not} look like a physical device crash.

    Reasons:

    \begin{itemize}
        \tightlist
        \item RoboFlex campaign remained healthy:
    \end{itemize}
    \begin{tcolorbox}[breakable, fontupper=\scriptsize\ttfamily, colback=black!3!white, colframe=black!12!white, boxrule=0.5pt]
        ~~phase=\allowbreak{}running \\
        ~~state=\allowbreak{}awaiting\_\allowbreak{}run \\
        ~~blocked\_\allowbreak{}on=\allowbreak{}agent
    \end{tcolorbox}
    \begin{itemize}
        \tightlist
        \item The run produced a result object.
        \item \texttt{error} was \texttt{null}.
        \item There was a detected peak and finite calculated yield.
        \item Raw files/logs exist for the run.

    \end{itemize}
    So it looks more like a \textbf{RoboFlex analysis/QC failure}, most likely from the NMR analysis layer, not from the pump/reactor/device layer and not from our BO layer.

    \paragraph{\textbf{Why might \texttt{pass=\allowbreak{}false} happen despite a finite yield?}}
    The API result does not include a \texttt{failure\_\allowbreak{}message}, so we cannot say with certainty from the returned JSON alone.

    But compared with successful runs, the suspicious fields are:

    \begin{tcolorbox}[breakable, fontupper=\scriptsize\ttfamily, colback=black!3!white, colframe=black!12!white, boxrule=0.5pt]
        R0063 peak\_\allowbreak{}width: 0.\allowbreak{}2782 \\
        R0063 peak\_\allowbreak{}position: -\allowbreak{}57.\allowbreak{}6746 ppm \\
        R0063 yield: 27.\allowbreak{}04
    \end{tcolorbox}

    The peak was found, but the analysis marked \texttt{pass=\allowbreak{}false}. That usually suggests a quality-control criterion failed, for example:

    \begin{itemize}
        \tightlist
        \item peak too broad,
        \item peak shape not acceptable,
        \item integration not reliable,
        \item peak assignment too marginal,
        \item analysis confidence/fit failed,
        \item some internal NMR analysis criterion not exposed in the API response.

    \end{itemize}
    The key point is: \textbf{RoboFlex did not say ``no peak''}. It said, effectively, ``I calculated something, but the analysis did not pass QC.''

    \paragraph{\textbf{Comparison with the earlier failed R0060}}
    R0060 had the same pattern:

    \begin{tcolorbox}[breakable, fontupper=\scriptsize\ttfamily, colback=black!3!white, colframe=black!12!white, boxrule=0.5pt]
        status: failed \\
        success: false \\
        error: null \\
        result.\allowbreak{}pass: false \\
        finite yield-\allowbreak{}like value: \textasciitilde{}25.\allowbreak{}03 \\
        peak found: yes
    \end{tcolorbox}

    Then the rerun of the same conditions, R0061, passed and gave:

    \begin{tcolorbox}[breakable, fontupper=\scriptsize\ttfamily, colback=black!3!white, colframe=black!12!white, boxrule=0.5pt]
        yield: 48.\allowbreak{}51 \\
        pass: true
    \end{tcolorbox}

    That supports the idea that these failures may be intermittent analytical/NMR/QC failures rather than definitely bad chemistry.

    \paragraph{\textbf{Raw data availability}}
    I also checked the raw-file listing for R0063. RoboFlex has raw/result/log files available, including:

    \begin{tcolorbox}[breakable, fontupper=\scriptsize\ttfamily, colback=black!3!white, colframe=black!12!white, boxrule=0.5pt]
        raw\_\allowbreak{}data/\allowbreak{}results/\allowbreak{}R0063\_\allowbreak{}2026-\allowbreak{}07-\allowbreak{}25\_\allowbreak{}220136.\allowbreak{}csv \\
        raw\_\allowbreak{}data/\allowbreak{}input\_\allowbreak{}runs/\allowbreak{}R0063\_\allowbreak{}2026-\allowbreak{}07-\allowbreak{}25\_\allowbreak{}212034.\allowbreak{}csv \\
        raw\_\allowbreak{}data\_\allowbreak{}analysis/\allowbreak{}sample\_\allowbreak{}20260725\_\allowbreak{}215315/\allowbreak{}data.\allowbreak{}jpg \\
        raw\_\allowbreak{}data\_\allowbreak{}analysis/\allowbreak{}sample\_\allowbreak{}20260725\_\allowbreak{}215315/\allowbreak{}processed\_\allowbreak{}spectrum.\allowbreak{}dx \\
        raw\_\allowbreak{}data\_\allowbreak{}analysis/\allowbreak{}sample\_\allowbreak{}20260725\_\allowbreak{}215315/\allowbreak{}spectrum.\allowbreak{}1d \\
        raw\_\allowbreak{}data\_\allowbreak{}analysis/\allowbreak{}sample\_\allowbreak{}20260725\_\allowbreak{}215315/\allowbreak{}nmr\_\allowbreak{}fid.\allowbreak{}dx
    \end{tcolorbox}

    So if we want to diagnose it further, the next step would be to inspect the NMR analysis outputs / spectrum image / processed spectrum for R0063 and compare against passed runs such as R0061 or R0059.

    \paragraph{\textbf{Bottom line}}
    \begin{itemize}
        \tightlist
        \item The \textbf{failure flag directly came from RoboFlex/Robridge}:
    \end{itemize}
    \begin{tcolorbox}[breakable, fontupper=\scriptsize\ttfamily, colback=black!3!white, colframe=black!12!white, boxrule=0.5pt]
        ~~status=\allowbreak{}failed, success=\allowbreak{}false, result.\allowbreak{}pass=\allowbreak{}false
    \end{tcolorbox}

    \begin{itemize}
        \tightlist
        \item Our script did \textbf{not} decide the run was bad based on yield or peak shape by itself.
        \item Our script only enforced the rule: > if RoboFlex says the run/result failed, do not submit it to BO-MCP.

        \item The failure appears to be an \textbf{analysis/QC failure}, not a hardware crash and not a BO-side failure.
    \end{itemize}

}

\begin{center}\footnotesize\textit{[\ldots]}\end{center}

\prompt{

    Can you fetch the NMR data of the retry, and investigate, why exactly it failed?

}

\graficoreslong{

    \begin{tcolorbox}[breakable, fontupper=\footnotesize, colback=green!3!white, colframe=blue!5!white, boxrule=0.5pt]
        \faLightbulb\ \underline{\textbf{Thoughts}}

        \textbf{Evaluating peak analysis}

        I'm considering whether parsing is needed. I could compute near -57.68 broadly, but I think I should log enough details and maybe compare line widths to a threshold. Creating a small plot could be useful, but it's not absolutely necessary. If I analyze the peaks, the final result needs to be concise.

        I have a target peak at -58, but several widths exceed the max threshold. It's close, but I might need to decide if we should manually accept this. Given the width exceeds QC, I'd recommend not doing that.
    \end{tcolorbox}

    \begin{tcolorbox}[breakable, fontupper=\footnotesize, colback=green!3!white, colframe=blue!5!white, boxrule=0.5pt]
        \faLightbulb\ \underline{\textbf{Thoughts}}

        \textbf{Considering peak acceptance}

        I'm thinking about the operator's opinion on the peak being acceptable, especially since I've had two consistent failures with broad peaks. I wonder if I could use the failed values, but I probably shouldn't act on that unless I'm specifically asked. Also, I need to point out that I don't need to cite anything from the web, and my final report won't include any citations. I will include that the relevant files were saved, though!
    \end{tcolorbox}

    I fetched and inspected the RoboFlex/NMR data for the retry run:

    \begin{tcolorbox}[breakable, fontupper=\scriptsize\ttfamily, colback=black!3!white, colframe=black!12!white, boxrule=0.5pt]
        R0064 \\
        sample: bo\_\allowbreak{}5c5570bd-\allowbreak{}d\_\allowbreak{}r2
    \end{tcolorbox}

    The relevant downloaded investigation artifacts are saved here:

    \begin{tcolorbox}[breakable, fontupper=\scriptsize\ttfamily, colback=black!3!white, colframe=black!12!white, boxrule=0.5pt]
        artifacts/\allowbreak{}recreated\_\allowbreak{}robochemflex\_\allowbreak{}yield\_\allowbreak{}bo\_\allowbreak{}20260725/\allowbreak{}nmr\_\allowbreak{}investigation\_\allowbreak{}R0064/
    \end{tcolorbox}

    Key files saved there include:

    \begin{tcolorbox}[breakable, fontupper=\scriptsize\ttfamily, colback=black!3!white, colframe=black!12!white, boxrule=0.5pt]
        raw\_\allowbreak{}data\_\allowbreak{}\_\allowbreak{}results\_\allowbreak{}\_\allowbreak{}R0064\_\allowbreak{}2026-\allowbreak{}07-\allowbreak{}25\_\allowbreak{}231336.\allowbreak{}csv \\
        raw\_\allowbreak{}data\_\allowbreak{}\_\allowbreak{}results\_\allowbreak{}\_\allowbreak{}R0063\_\allowbreak{}2026-\allowbreak{}07-\allowbreak{}25\_\allowbreak{}220136.\allowbreak{}csv \\
        raw\_\allowbreak{}data\_\allowbreak{}analysis\_\allowbreak{}\_\allowbreak{}sample\_\allowbreak{}20260725\_\allowbreak{}230515\_\allowbreak{}\_\allowbreak{}data.\allowbreak{}jpg \\
        raw\_\allowbreak{}data\_\allowbreak{}analysis\_\allowbreak{}\_\allowbreak{}sample\_\allowbreak{}20260725\_\allowbreak{}230515\_\allowbreak{}\_\allowbreak{}processed\_\allowbreak{}spectrum.\allowbreak{}dx \\
        raw\_\allowbreak{}data\_\allowbreak{}analysis\_\allowbreak{}\_\allowbreak{}sample\_\allowbreak{}20260725\_\allowbreak{}230515\_\allowbreak{}\_\allowbreak{}spectrum.\allowbreak{}1d \\
        raw\_\allowbreak{}data\_\allowbreak{}analysis\_\allowbreak{}\_\allowbreak{}sample\_\allowbreak{}20260725\_\allowbreak{}230515\_\allowbreak{}\_\allowbreak{}nmr\_\allowbreak{}fid.\allowbreak{}dx \\
        NMRAnalysis\_\allowbreak{}relevant\_\allowbreak{}chunks.\allowbreak{}txt
    \end{tcolorbox}

    \paragraph{\textbf{Short conclusion}}
    The retry did \textbf{not} fail because no peak was found. It failed because the NMR analysis found a finite product/target peak, but the peak failed the NMR analysis QC criterion --- specifically, the peak was \textbf{too broad} relative to the configured maximum peak width.

    The configured analysis threshold was:

    \begin{tcolorbox}[breakable, fontupper=\scriptsize\ttfamily, colback=black!3!white, colframe=black!12!white, boxrule=0.5pt]
        max\_\allowbreak{}peak\_\allowbreak{}width: 0.\allowbreak{}2 ppm
    \end{tcolorbox}

    For the retry R0064, the target peak width was:

    \begin{tcolorbox}[breakable, fontupper=\scriptsize\ttfamily, colback=black!3!white, colframe=black!12!white, boxrule=0.5pt]
        peak width: 0.\allowbreak{}2152 ppm
    \end{tcolorbox}

    So it exceeded the allowed maximum:

    \begin{tcolorbox}[breakable, fontupper=\scriptsize\ttfamily, colback=black!3!white, colframe=black!12!white, boxrule=0.5pt]
        0.\allowbreak{}2152 ppm > 0.\allowbreak{}2 ppm
    \end{tcolorbox}

    That is why \texttt{pass=\allowbreak{}False}.

    \paragraph{\textbf{R0064 retry result}}
    From the raw result CSV:

    \begin{tcolorbox}[breakable, fontupper=\scriptsize\ttfamily, colback=black!3!white, colframe=black!12!white, boxrule=0.5pt]
        run\_\allowbreak{}id: R0064 \\
        sample: bo\_\allowbreak{}5c5570bd-\allowbreak{}d\_\allowbreak{}r2 \\
        success: False \\
        {}[results] yield: 25.\allowbreak{}6907 \\
        {}[results] pass: False
    \end{tcolorbox}

    Main analyte details:

    \begin{tcolorbox}[breakable, fontupper=\scriptsize\ttfamily, colback=black!3!white, colframe=black!12!white, boxrule=0.5pt]
        yield: 25.\allowbreak{}6907 \\
        concentration: 25.\allowbreak{}6802 \\
        peak integral: 0.\allowbreak{}00445723 \\
        peak width: 0.\allowbreak{}215214 ppm \\
        peak position: -\allowbreak{}57.\allowbreak{}6851 ppm \\
        pass: False
    \end{tcolorbox}

    So the NMR did find a target-region peak and produced a finite yield-like estimate.

    \paragraph{\textbf{Why exactly did it fail?}}
    The NMR analysis log for R0064 records the analytical parameters, including:

    \begin{tcolorbox}[breakable, fontupper=\scriptsize\ttfamily, colback=black!3!white, colframe=black!12!white, boxrule=0.5pt]
        target\_\allowbreak{}peak: -\allowbreak{}58.\allowbreak{}0 ppm \\
        target\_\allowbreak{}peak\_\allowbreak{}deviation: 3.\allowbreak{}0 ppm \\
        min\_\allowbreak{}SN\_\allowbreak{}ratio: 2 \\
        max\_\allowbreak{}peak\_\allowbreak{}width: 0.\allowbreak{}2 ppm \\
        peak\_\allowbreak{}resolution: 0.\allowbreak{}01 ppm
    \end{tcolorbox}

    Then it reports the extrapolated peaks:

    \begin{tcolorbox}[breakable, fontupper=\scriptsize\ttfamily, colback=black!3!white, colframe=black!12!white, boxrule=0.5pt]
        height      shift       FWHM      integral \\
        0.\allowbreak{}030840   -\allowbreak{}57.\allowbreak{}685138   0.\allowbreak{}215214  0.\allowbreak{}004457 \\
        0.\allowbreak{}019650   -\allowbreak{}76.\allowbreak{}450746   0.\allowbreak{}073488  0.\allowbreak{}001246 \\
        0.\allowbreak{}141624   -\allowbreak{}76.\allowbreak{}765693   0.\allowbreak{}199467  0.\allowbreak{}021218
    \end{tcolorbox}

    The target peak is the one near \texttt{-\allowbreak{}57.\allowbreak{}685 ppm}, close to the requested \texttt{-\allowbreak{}58 ppm}.

    It is inside the allowed target peak window:

    \begin{tcolorbox}[breakable, fontupper=\scriptsize\ttfamily, colback=black!3!white, colframe=black!12!white, boxrule=0.5pt]
        abs(-\allowbreak{}57.\allowbreak{}685 - -\allowbreak{}58.\allowbreak{}0) = 0.\allowbreak{}315 ppm
    \end{tcolorbox}

    which is well within:

    \begin{tcolorbox}[breakable, fontupper=\scriptsize\ttfamily, colback=black!3!white, colframe=black!12!white, boxrule=0.5pt]
        target\_\allowbreak{}peak\_\allowbreak{}deviation = 3.\allowbreak{}0 ppm
    \end{tcolorbox}

    But its FWHM/peak width is:

    \begin{tcolorbox}[breakable, fontupper=\scriptsize\ttfamily, colback=black!3!white, colframe=black!12!white, boxrule=0.5pt]
        0.\allowbreak{}215214 ppm
    \end{tcolorbox}

    which is above:

    \begin{tcolorbox}[breakable, fontupper=\scriptsize\ttfamily, colback=black!3!white, colframe=black!12!white, boxrule=0.5pt]
        max\_\allowbreak{}peak\_\allowbreak{}width = 0.\allowbreak{}2 ppm
    \end{tcolorbox}

    So the likely exact reason for failure is:

    > \textbf{target peak detected, but peak width exceeded the NMR QC maximum.}

    \paragraph{\textbf{Comparison to original failed R0063}}
    The original run of the same suggestion, R0063, had:

    \begin{tcolorbox}[breakable, fontupper=\scriptsize\ttfamily, colback=black!3!white, colframe=black!12!white, boxrule=0.5pt]
        yield: 27.\allowbreak{}0449 \\
        concentration: 27.\allowbreak{}0339 \\
        peak integral: 0.\allowbreak{}00465136 \\
        peak width: 0.\allowbreak{}278203 ppm \\
        peak position: -\allowbreak{}57.\allowbreak{}6746 ppm \\
        pass: False
    \end{tcolorbox}

    That also exceeded the same width threshold:

    \begin{tcolorbox}[breakable, fontupper=\scriptsize\ttfamily, colback=black!3!white, colframe=black!12!white, boxrule=0.5pt]
        0.\allowbreak{}2782 ppm > 0.\allowbreak{}2 ppm
    \end{tcolorbox}

    So both attempts failed for the same apparent reason: \textbf{broad target peak}.

    \paragraph{\textbf{Comparison to nearby successful/failed runs}}
    Relevant examples from the NMR analysis log:

    \begin{center}\scriptsize\setlength{\tabcolsep}{4pt}
        \resizebox{\ifdim\width>\linewidth\linewidth\else\width\fi}{!}{%
            \begin{tabular}{llll}
                \toprule
                Run/sample                                                               & Yield & Peak width & Pass  \\
                \midrule
                \texttt{bo\_\allowbreak{}2584b3ce-\allowbreak{}3}                        & 36.55 & 0.1050     & True  \\
                \texttt{bo\_\allowbreak{}a9f8598d-\allowbreak{}e} original               & 25.03 & 0.3307     & False \\
                \texttt{bo\_\allowbreak{}a9f8598d-\allowbreak{}e\_\allowbreak{}r2} retry & 48.51 & 0.1050     & True  \\
                \texttt{bo\_\allowbreak{}bb5d0b0c-\allowbreak{}c}                        & 20.18 & 0.1732     & True  \\
                \texttt{bo\_\allowbreak{}5c5570bd-\allowbreak{}d} original               & 27.04 & 0.2782     & False \\
                \texttt{bo\_\allowbreak{}5c5570bd-\allowbreak{}d\_\allowbreak{}r2} retry & 25.69 & 0.2152     & False \\
                \bottomrule
            \end{tabular}}
    \end{center}

    This strongly supports that the pass/fail is governed by the target peak quality, especially width. Successful runs are below \texttt{0.\allowbreak{}2 ppm}; failed ones are above \texttt{0.\allowbreak{}2 ppm}.

    \paragraph{\textbf{Interpretation}}
    This is an \textbf{NMR analysis/QC failure}, not a no-peak failure and not an obvious RoboFlex hardware failure.

    The product/target peak exists, and the yield-like value is finite, but the peak is broader than the QC threshold. Since this happened twice for the same condition, I would treat this particular BO suggestion as \textbf{unreliable under the current analytical method}.

    \paragraph{\textbf{Recommendation}}
    I would \textbf{not retry this exact same condition again automatically}.

    We already did:

    \begin{tcolorbox}[breakable, fontupper=\scriptsize\ttfamily, colback=black!3!white, colframe=black!12!white, boxrule=0.5pt]
        R0063 original: pass=\allowbreak{}False, width=\allowbreak{}0.\allowbreak{}2782 ppm \\
        R0064 retry:    pass=\allowbreak{}False, width=\allowbreak{}0.\allowbreak{}2152 ppm
    \end{tcolorbox}

    The retry improved the peak width but still failed the \texttt{0.\allowbreak{}2 ppm} threshold. A third attempt might pass by chance, but that would be drifting away from a robust measurement policy.

    I recommend either:

    \begin{enumerate}
        \tightlist
        \item \textbf{Reject/expire this BO suggestion and continue with a new BO suggestion}, or
        \item \textbf{Have an NMR/operator review the spectra manually} and decide whether the \texttt{\textasciitilde{}25--\allowbreak{}27\%} yield should be accepted despite the width QC failure.

    \end{enumerate}
    For autonomous BO, I would prefer option 1: do not feed this failed-QC value into BO-MCP unless a human explicitly overrides the NMR QC.

}

\begin{center}\footnotesize\textit{[\ldots]}\end{center}

\prompt{

    Can we try one more time?

}

\graficoreslong{

    Yes. Given the two consistent finite yield estimates and the fact that the retry was only slightly over the peak-width threshold, I've started \textbf{one final controlled retry} of the same \#19 suggestion.

    This is the \textbf{third total attempt} for that BO suggestion:

    \begin{center}\scriptsize\setlength{\tabcolsep}{4pt}
        \resizebox{\ifdim\width>\linewidth\linewidth\else\width\fi}{!}{%
            \begin{tabular}{llll}
                \toprule
                Attempt           & Sample                                                             & Run            & Outcome                                        \\
                \midrule
                Original          & \texttt{bo\_\allowbreak{}5c5570bd-\allowbreak{}d}                  & \texttt{R0063} & failed, width \texttt{0.\allowbreak{}2782 ppm} \\
                Retry \#1         & \texttt{bo\_\allowbreak{}5c5570bd-\allowbreak{}d\_\allowbreak{}r2} & \texttt{R0064} & failed, width \texttt{0.\allowbreak{}2152 ppm} \\
                Retry \#2 / final & \texttt{bo\_\allowbreak{}5c5570bd-\allowbreak{}d\_\allowbreak{}r3} & pending        & now started                                    \\
                \bottomrule
            \end{tabular}}
    \end{center}

    Monitor:

    \begin{tcolorbox}[breakable, fontupper=\scriptsize\ttfamily, colback=black!3!white, colframe=black!12!white, boxrule=0.5pt]
        mon\_\allowbreak{}7 / roboflex\_\allowbreak{}bo\_\allowbreak{}failed19\_\allowbreak{}retry\_\allowbreak{}r3\_\allowbreak{}continue
    \end{tcolorbox}

    The script will:

    \begin{itemize}
        \tightlist
        \item reuse the same pending BO suggestion/conditions,
        \item use new sample name \texttt{bo\_\allowbreak{}5c5570bd-\allowbreak{}d\_\allowbreak{}r3},
        \item submit the BO result only if RoboFlex reports \texttt{pass=\allowbreak{}true},
        \item stop if this third attempt also fails,
        \item and if it passes, continue to measurement \#20.

    \end{itemize}
    I'll report meaningful state changes, alerts, heartbeat, and final analysis.

}

\vspace{0.5em}\noindent\textit{\footnotesize [Autonomous monitored execution: 9 interim status reports suppressed. Concluding report of the run:]}

\graficoreslong{

    The campaign continuation reached the planned target and exited successfully.

    \paragraph{\textbf{Final run status}}
    Monitor:

    \begin{tcolorbox}[breakable, fontupper=\scriptsize\ttfamily, colback=black!3!white, colframe=black!12!white, boxrule=0.5pt]
        mon\_\allowbreak{}7 / roboflex\_\allowbreak{}bo\_\allowbreak{}failed19\_\allowbreak{}retry\_\allowbreak{}r3\_\allowbreak{}continue
    \end{tcolorbox}

    exited with:

    \begin{tcolorbox}[breakable, fontupper=\scriptsize\ttfamily, colback=black!3!white, colframe=black!12!white, boxrule=0.5pt]
        EXIT\_\allowbreak{}STATUS=\allowbreak{}0
    \end{tcolorbox}

    \paragraph{\textbf{Measurement \#19 final retry}}
    The final retry of \#19 succeeded:

    \begin{tcolorbox}[breakable, fontupper=\scriptsize\ttfamily, colback=black!3!white, colframe=black!12!white, boxrule=0.5pt]
        Sample: bo\_\allowbreak{}5c5570bd-\allowbreak{}d\_\allowbreak{}r3 \\
        RoboFlex run: R0065 \\
        yield\_\allowbreak{}percent: 45.\allowbreak{}09 \\
        green\_\allowbreak{}score: 85.\allowbreak{}53 \\
        pass: true \\
        peak\_\allowbreak{}found: true \\
        peak\_\allowbreak{}width: 0.\allowbreak{}1102 ppm
    \end{tcolorbox}

    This result was submitted to BO-MCP.

    \paragraph{\textbf{Measurement \#20}}
    Measurement \#20 also completed and was submitted to BO-MCP:

    \begin{tcolorbox}[breakable, fontupper=\scriptsize\ttfamily, colback=black!3!white, colframe=black!12!white, boxrule=0.5pt]
        Sample: bo\_\allowbreak{}f20766bf-\allowbreak{}c \\
        RoboFlex run: R0066 \\
        yield\_\allowbreak{}percent: 0.\allowbreak{}00 \\
        green\_\allowbreak{}score: 86.\allowbreak{}16 \\
        pass: true \\
        peak\_\allowbreak{}found: false \\
        peak\_\allowbreak{}position: None \\
        peak\_\allowbreak{}integral: 0.\allowbreak{}0 \\
        peak\_\allowbreak{}width: 0.\allowbreak{}0 \\
        concentration: 0.\allowbreak{}0
    \end{tcolorbox}

    This triggered a zero/no-peak alert:

    \begin{tcolorbox}[breakable, fontupper=\scriptsize\ttfamily, colback=black!3!white, colframe=black!12!white, boxrule=0.5pt]
        ALERT zero/\allowbreak{}no-\allowbreak{}peak \\
        ALERT streak: 1 consecutive zero-\allowbreak{}yield/\allowbreak{}no-\allowbreak{}peak completed experiment(s)
    \end{tcolorbox}

    But since this was only one zero/no-peak result and we had reached the planned target of 20 valid BO results, the script submitted it as a valid pass=true zero-yield result, exported artifacts, paused the BO campaign, and exited successfully.

    \paragraph{\textbf{Final campaign state}}
    Expected valid BO results:

    \begin{tcolorbox}[breakable, fontupper=\scriptsize\ttfamily, colback=black!3!white, colframe=black!12!white, boxrule=0.5pt]
        20
    \end{tcolorbox}

    BO campaign was paused at the end.

    RoboFlex returned to:

    \begin{tcolorbox}[breakable, fontupper=\scriptsize\ttfamily, colback=black!3!white, colframe=black!12!white, boxrule=0.5pt]
        phase: running \\
        state: awaiting\_\allowbreak{}run \\
        blocked\_\allowbreak{}on: agent
    \end{tcolorbox}

    \paragraph{\textbf{Final artifact directory}}\mbox{}

    \begin{tcolorbox}[breakable, fontupper=\scriptsize\ttfamily, colback=black!3!white, colframe=black!12!white, boxrule=0.5pt]
        artifacts/\allowbreak{}recreated\_\allowbreak{}robochemflex\_\allowbreak{}yield\_\allowbreak{}bo\_\allowbreak{}20260725/\allowbreak{}failed\_\allowbreak{}measurement\_\allowbreak{}retry\_\allowbreak{}continuation\_\allowbreak{}20260726T184638Z
    \end{tcolorbox}

    Final log:

    \begin{tcolorbox}[breakable, fontupper=\scriptsize\ttfamily, colback=black!3!white, colframe=black!12!white, boxrule=0.5pt]
        logs/\allowbreak{}failed19\_\allowbreak{}retry\_\allowbreak{}r3\_\allowbreak{}continuation\_\allowbreak{}20260726T184636Z/\allowbreak{}stdout.\allowbreak{}log
    \end{tcolorbox}

    \paragraph{\textbf{Summary of notable outcomes}}
    Best yield observed during the continuation:

    \begin{tcolorbox}[breakable, fontupper=\scriptsize\ttfamily, colback=black!3!white, colframe=black!12!white, boxrule=0.5pt]
        Measurement \#17 retry / R0061 / bo\_\allowbreak{}a9f8598d-\allowbreak{}e\_\allowbreak{}r2 \\
        yield: 48.\allowbreak{}51\% \\
        green\_\allowbreak{}score: 82.\allowbreak{}50
    \end{tcolorbox}

    Strong final retry for \#19:

    \begin{tcolorbox}[breakable, fontupper=\scriptsize\ttfamily, colback=black!3!white, colframe=black!12!white, boxrule=0.5pt]
        Measurement \#19 final retry / R0065 / bo\_\allowbreak{}5c5570bd-\allowbreak{}d\_\allowbreak{}r3 \\
        yield: 45.\allowbreak{}09\% \\
        green\_\allowbreak{}score: 85.\allowbreak{}53
    \end{tcolorbox}

    Measurement \#20 appears to be a genuine pass=true zero/no-peak result:

    \begin{tcolorbox}[breakable, fontupper=\scriptsize\ttfamily, colback=black!3!white, colframe=black!12!white, boxrule=0.5pt]
        yield: 0.\allowbreak{}00\% \\
        pass: true \\
        peak\_\allowbreak{}found: false
    \end{tcolorbox}

    The planned 20-result BO campaign is now complete and paused.

}

\prompt{

    Can you plot the following? For each of the plot, generate a script in the workspace for generating the figure, which I can reuse later. The script should rely on logged data in the workspace (or exported campaign CSVs/JSONs (whatever feels reasonable)), not on hard-coded numbers in the script.

    \begin{itemize}
        \tightlist
        \item the BO improvement curve over the 20 experiments, visually separating the initial 6 seed experiments from the later 20 experiments.
        \item a Pareto front, where we plot all the "optimal" results in some way
    \end{itemize}

}

\graficoreslong{

    \begin{tcolorbox}[breakable, fontupper=\footnotesize, colback=green!3!white, colframe=blue!5!white, boxrule=0.5pt]
        \faLightbulb\ \underline{\textbf{Thoughts}}

        \textbf{Reviewing details for clarity}

        I'm considering if I should include images in the final output for clarity. I need to mention script paths and outputs, as well as source auto-discovery. It's important to note that the first six seeds are separate from the later BO-selected 14, even though the user mentioned a total of 20. I want to make sure I highlight the Pareto "optimal" nondominated aspect to maximize both yield and sustainability. I might keep it concise without raw markdown links since there's no web involved.
    \end{tcolorbox}

    Done. I created reusable plotting scripts and generated the figures from the logged/exported BO campaign data.

    \paragraph{\textbf{Scripts created}}
    \paragraph{\textbf{1. BO improvement curve}}\mbox{}

    \begin{tcolorbox}[breakable, fontupper=\scriptsize\ttfamily, colback=black!3!white, colframe=black!12!white, boxrule=0.5pt]
        scripts/\allowbreak{}plot\_\allowbreak{}bo\_\allowbreak{}improvement\_\allowbreak{}curve.\allowbreak{}py
    \end{tcolorbox}

    This script auto-discovers the most complete/latest \texttt{bo\_\allowbreak{}campaign\_\allowbreak{}export.\allowbreak{}csv} under:

    \begin{tcolorbox}[breakable, fontupper=\scriptsize\ttfamily, colback=black!3!white, colframe=black!12!white, boxrule=0.5pt]
        artifacts/\allowbreak{}recreated\_\allowbreak{}robochemflex\_\allowbreak{}yield\_\allowbreak{}bo\_\allowbreak{}20260725/
    \end{tcolorbox}

    It plots:

    \begin{itemize}
        \tightlist
        \item per-experiment yield,
        \item cumulative best yield,
        \item scalarized desirability,
        \item cumulative best scalarized desirability,
        \item visually separated seed region vs BO-selected region.

    \end{itemize}
    By default it treats the first 6 experiments as seeds:

    \begin{tcolorbox}[breakable, fontupper=\scriptsize\ttfamily, colback=black!3!white, colframe=black!12!white, boxrule=0.5pt]
        uv run python scripts/\allowbreak{}plot\_\allowbreak{}bo\_\allowbreak{}improvement\_\allowbreak{}curve.\allowbreak{}py
    \end{tcolorbox}

    You can also pass an explicit export:

    \begin{tcolorbox}[breakable, fontupper=\scriptsize\ttfamily, colback=black!3!white, colframe=black!12!white, boxrule=0.5pt]
        uv run python scripts/\allowbreak{}plot\_\allowbreak{}bo\_\allowbreak{}improvement\_\allowbreak{}curve.\allowbreak{}py \textbackslash{} \\
        ~~--\allowbreak{}input artifacts/\allowbreak{}recreated\_\allowbreak{}robochemflex\_\allowbreak{}yield\_\allowbreak{}bo\_\allowbreak{}20260725/\allowbreak{}failed\_\allowbreak{}measurement\_\allowbreak{}retry\_\allowbreak{}continuation\_\allowbreak{}20260726T184638Z/\allowbreak{}bo\_\allowbreak{}campaign\_\allowbreak{}export.\allowbreak{}csv
    \end{tcolorbox}

    Generated outputs:

    \begin{tcolorbox}[breakable, fontupper=\scriptsize\ttfamily, colback=black!3!white, colframe=black!12!white, boxrule=0.5pt]
        plots/\allowbreak{}bo\_\allowbreak{}improvement\_\allowbreak{}curve.\allowbreak{}png \\
        plots/\allowbreak{}bo\_\allowbreak{}improvement\_\allowbreak{}curve.\allowbreak{}svg \\
        plots/\allowbreak{}bo\_\allowbreak{}improvement\_\allowbreak{}curve\_\allowbreak{}data.\allowbreak{}csv
    \end{tcolorbox}

    \medskip

    \paragraph{\textbf{2. Pareto front}}\mbox{}

    \begin{tcolorbox}[breakable, fontupper=\scriptsize\ttfamily, colback=black!3!white, colframe=black!12!white, boxrule=0.5pt]
        scripts/\allowbreak{}plot\_\allowbreak{}pareto\_\allowbreak{}front.\allowbreak{}py
    \end{tcolorbox}

    This script also auto-discovers the latest/most complete BO export and computes the nondominated Pareto front for simultaneous maximization of:

    \begin{tcolorbox}[breakable, fontupper=\scriptsize\ttfamily, colback=black!3!white, colframe=black!12!white, boxrule=0.5pt]
        yield\_\allowbreak{}percent \\
        green\_\allowbreak{}score
    \end{tcolorbox}

    Run with:

    \begin{tcolorbox}[breakable, fontupper=\scriptsize\ttfamily, colback=black!3!white, colframe=black!12!white, boxrule=0.5pt]
        uv run python scripts/\allowbreak{}plot\_\allowbreak{}pareto\_\allowbreak{}front.\allowbreak{}py
    \end{tcolorbox}

    Generated outputs:

    \begin{tcolorbox}[breakable, fontupper=\scriptsize\ttfamily, colback=black!3!white, colframe=black!12!white, boxrule=0.5pt]
        plots/\allowbreak{}pareto\_\allowbreak{}front\_\allowbreak{}yield\_\allowbreak{}green.\allowbreak{}png \\
        plots/\allowbreak{}pareto\_\allowbreak{}front\_\allowbreak{}yield\_\allowbreak{}green.\allowbreak{}svg \\
        plots/\allowbreak{}pareto\_\allowbreak{}front\_\allowbreak{}points.\allowbreak{}csv \\
        plots/\allowbreak{}pareto\_\allowbreak{}front\_\allowbreak{}all\_\allowbreak{}points.\allowbreak{}csv
    \end{tcolorbox}

    \paragraph{\textbf{Data source used}}
    Both scripts used the final 20-result campaign export:

    \begin{tcolorbox}[breakable, fontupper=\scriptsize\ttfamily, colback=black!3!white, colframe=black!12!white, boxrule=0.5pt]
        artifacts/\allowbreak{}recreated\_\allowbreak{}robochemflex\_\allowbreak{}yield\_\allowbreak{}bo\_\allowbreak{}20260725/\allowbreak{}failed\_\allowbreak{}measurement\_\allowbreak{}retry\_\allowbreak{}continuation\_\allowbreak{}20260726T184638Z/\allowbreak{}bo\_\allowbreak{}campaign\_\allowbreak{}export.\allowbreak{}csv
    \end{tcolorbox}

    The export contains 20 rows.

    \paragraph{\textbf{Note on seed vs BO split}}
    I interpreted the ``initial 6 seed experiments'' as:

    \begin{tcolorbox}[breakable, fontupper=\scriptsize\ttfamily, colback=black!3!white, colframe=black!12!white, boxrule=0.5pt]
        experiments 1--\allowbreak{}6 = seed / informed initial design \\
        experiments 7--\allowbreak{}20 = BO-\allowbreak{}selected experiments
    \end{tcolorbox}

    The plot marks that boundary with a vertical dashed line after experiment 6.

    \paragraph{\textbf{Pareto-optimal points}}
    The Pareto script identified the nondominated observations and wrote them to:

    \begin{tcolorbox}[breakable, fontupper=\scriptsize\ttfamily, colback=black!3!white, colframe=black!12!white, boxrule=0.5pt]
        plots/\allowbreak{}pareto\_\allowbreak{}front\_\allowbreak{}points.\allowbreak{}csv
    \end{tcolorbox}

    These are the experiments where no other result has both higher/equal yield and higher/equal green score, with at least one strictly better objective.

}

\prompt{

    Can you chemically interpret the results?

}

\graficoreslong{

    \begin{tcolorbox}[breakable, fontupper=\footnotesize, colback=green!3!white, colframe=blue!5!white, boxrule=0.5pt]
        \faLightbulb\ \underline{\textbf{Thoughts}}

        \textbf{Analyzing chemical yields and optimization}

        I'm looking at optimal yield ranges: 0.00116-0.00126 and noting that 0.00159 is lower. I need to find a balance between high yield and environmental impact. High yields might require more time, but they can still be respectable regarding green considerations. I should mention that an objective scoring system may enhance desirability, factoring in chemical optimums. It might be good to suggest some recommendations for experiments and verification of NMR peak widths as part of the investigation.
    \end{tcolorbox}

    Yes. Chemically, the campaign tells a fairly coherent story: the optimizer discovered that the productive region is not ``more of everything,'' but a fairly specific \textbf{Ru-bpy-Cl / PyNO / moderate TFAA / moderate PyNO / 50\% light / \textasciitilde{}16--18 min residence-time} regime.

    \paragraph{\textbf{1. Strong catalyst/oxidant identity preference}}
    The most important qualitative result is that the successful region is overwhelmingly:

    \begin{tcolorbox}[breakable, fontupper=\scriptsize\ttfamily, colback=black!3!white, colframe=black!12!white, boxrule=0.5pt]
        catalyst\_\allowbreak{}type: Ru bpy Cl \\
        oxidant\_\allowbreak{}type: py NO
    \end{tcolorbox}

    The initial seed experiments tested several catalyst/oxidant combinations:

    \begin{itemize}
        \tightlist
        \item \texttt{4CzIPN}
        \item \texttt{Ir ppy}
        \item \texttt{Ir CF3 ppy}
        \item \texttt{Ru bpy PF6}
        \item \texttt{Ru bpy Cl}
        \item \texttt{py NO}
        \item \texttt{4-\allowbreak{}Ph py NO}

    \end{itemize}
    The non-\texttt{Ru bpy Cl} / non-\texttt{py NO} seed points were mostly poor:

    \begin{center}\scriptsize\setlength{\tabcolsep}{4pt}
        \resizebox{\ifdim\width>\linewidth\linewidth\else\width\fi}{!}{%
            \begin{tabular}{lll}
                \toprule
                Catalyst            & Oxidant                          & Yield                        \\
                \midrule
                \texttt{4CzIPN}     & \texttt{py NO}                   & \texttt{0.\allowbreak{}0\%}  \\
                \texttt{Ir CF3 ppy} & \texttt{4-\allowbreak{}Ph py NO} & \texttt{15.\allowbreak{}8\%} \\
                \texttt{Ir ppy}     & \texttt{py NO}                   & \texttt{11.\allowbreak{}9\%} \\
                \texttt{Ru bpy PF6} & \texttt{4-\allowbreak{}Ph py NO} & \texttt{13.\allowbreak{}1\%} \\
                \texttt{4CzIPN}     & \texttt{4-\allowbreak{}Ph py NO} & \texttt{7.\allowbreak{}6\%}  \\
                \bottomrule
            \end{tabular}}
    \end{center}

    The best seed was already:

    \begin{tcolorbox}[breakable, fontupper=\scriptsize\ttfamily, colback=black!3!white, colframe=black!12!white, boxrule=0.5pt]
        Ru bpy Cl / py NO $\rightarrow$ 30.\allowbreak{}0\%
    \end{tcolorbox}

    After that, BO repeatedly exploited \texttt{Ru bpy Cl / py NO}, which is chemically sensible: the data strongly indicate this pairing is the productive photocatalyst/oxidant combination for this substrate/system.

    \textbf{Interpretation:} \texttt{Ru bpy Cl} likely provides the right excited-state redox balance / photoredox lifetime / compatibility with this TFAA--pyridine-N-oxide chemistry. The organic photocatalyst and Ir catalysts did not perform as well under the tested conditions. \texttt{py NO} also outperformed \texttt{4-\allowbreak{}Ph py NO} in the observed campaign, although that comparison is somewhat confounded because most \texttt{4-\allowbreak{}Ph py NO} observations were in seed points with other nonoptimal settings.

    \medskip

    \paragraph{\textbf{2. Productive region: not minimum reagent, but moderate reagent excess}}
    The best yields are clustered around:

    \begin{tcolorbox}[breakable, fontupper=\scriptsize\ttfamily, colback=black!3!white, colframe=black!12!white, boxrule=0.5pt]
        catalyst\_\allowbreak{}equiv: \textasciitilde{}0.\allowbreak{}00115--\allowbreak{}0.\allowbreak{}00126 \\
        TFAA\_\allowbreak{}equiv:     \textasciitilde{}1.\allowbreak{}55--\allowbreak{}1.\allowbreak{}62 \\
        oxidant\_\allowbreak{}equiv:  \textasciitilde{}1.\allowbreak{}30--\allowbreak{}1.\allowbreak{}44 \\
        light:          50\% \\
        residence:      \textasciitilde{}16--\allowbreak{}17 min
    \end{tcolorbox}

    The top two valid results were:

    \begin{center}\scriptsize\setlength{\tabcolsep}{4pt}
        \resizebox{\ifdim\width>\linewidth\linewidth\else\width\fi}{!}{%
            \begin{tabular}{llllllll}
                \toprule
                Experiment       & Yield                         & Green                       & Catalyst                      & TFAA                        & Oxidant                     & Light         & Residence                       \\
                \midrule
                \#17 retry       & \texttt{48.\allowbreak{}51\%} & \texttt{82.\allowbreak{}50} & \texttt{0.\allowbreak{}00126} & \texttt{1.\allowbreak{}616} & \texttt{1.\allowbreak{}441} & \texttt{50\%} & \texttt{16.\allowbreak{}43 min} \\
                \#19 final retry & \texttt{45.\allowbreak{}09\%} & \texttt{85.\allowbreak{}53} & \texttt{0.\allowbreak{}00116} & \texttt{1.\allowbreak{}556} & \texttt{1.\allowbreak{}300} & \texttt{50\%} & \texttt{16.\allowbreak{}81 min} \\
                \bottomrule
            \end{tabular}}
    \end{center}

    This is a meaningful optimum-like cluster.

    \textbf{Interpretation:} The reaction benefits from moving above the minimum TFAA and oxidant loadings, but not from going to the upper bounds. Moderate TFAA/oxidant probably promotes the desired activation/functionalization chemistry, while excessive oxidant or reagent may lead to side reactions, quenching, decomposition, or broader/unreliable NMR signals.

    \medskip

    \paragraph{\textbf{3. Residence time has a clear ``too short / useful middle / maybe not longer'' behavior}}
    Very short residence times, especially \texttt{2 min}, were mostly low-yield despite high green scores:

    \begin{center}\scriptsize\setlength{\tabcolsep}{4pt}
        \resizebox{\ifdim\width>\linewidth\linewidth\else\width\fi}{!}{%
            \begin{tabular}{llll}
                \toprule
                Experiment & Residence                     & Yield                         & Green                       \\
                \midrule
                \#7        & \texttt{2.\allowbreak{}0 min} & \texttt{15.\allowbreak{}23\%} & \texttt{100.\allowbreak{}0} \\
                \#9        & \texttt{2.\allowbreak{}0 min} & \texttt{11.\allowbreak{}56\%} & \texttt{100.\allowbreak{}0} \\
                \#10       & \texttt{2.\allowbreak{}0 min} & \texttt{14.\allowbreak{}70\%} & \texttt{94.\allowbreak{}66} \\
                \#11       & \texttt{2.\allowbreak{}0 min} & \texttt{15.\allowbreak{}82\%} & \texttt{95.\allowbreak{}31} \\
                \bottomrule
            \end{tabular}}
    \end{center}

    The productive region emerged around:

    \begin{tcolorbox}[breakable, fontupper=\scriptsize\ttfamily, colback=black!3!white, colframe=black!12!white, boxrule=0.5pt]
        \textasciitilde{}15--\allowbreak{}21 min residence time
    \end{tcolorbox}

    Examples:

    \begin{center}\scriptsize\setlength{\tabcolsep}{4pt}
        \resizebox{\ifdim\width>\linewidth\linewidth\else\width\fi}{!}{%
            \begin{tabular}{lll}
                \toprule
                Experiment & Residence                       & Yield                         \\
                \midrule
                \#12       & \texttt{15.\allowbreak{}39 min} & \texttt{32.\allowbreak{}96\%} \\
                \#14       & \texttt{20.\allowbreak{}79 min} & \texttt{32.\allowbreak{}68\%} \\
                \#15       & \texttt{17.\allowbreak{}89 min} & \texttt{35.\allowbreak{}78\%} \\
                \#16       & \texttt{16.\allowbreak{}40 min} & \texttt{36.\allowbreak{}55\%} \\
                \#17       & \texttt{16.\allowbreak{}43 min} & \texttt{48.\allowbreak{}51\%} \\
                \#19       & \texttt{16.\allowbreak{}81 min} & \texttt{45.\allowbreak{}09\%} \\
                \bottomrule
            \end{tabular}}
    \end{center}

    \textbf{Interpretation:} The chemistry needs more than a few minutes to develop. Around 16--18 min seems to be a good kinetic compromise. Longer residence is not necessarily better; the seed at 90 min was poor, and the model did not push toward very long residence times once it found the productive window.

    \medskip

    \paragraph{\textbf{4. Light intensity: 50\% seems best in the productive region}}
    The best points all used:

    \begin{tcolorbox}[breakable, fontupper=\scriptsize\ttfamily, colback=black!3!white, colframe=black!12!white, boxrule=0.5pt]
        light\_\allowbreak{}intensity = 50\%
    \end{tcolorbox}

    High light at very short residence or moderate residence was not obviously better:

    \begin{center}\scriptsize\setlength{\tabcolsep}{4pt}
        \resizebox{\ifdim\width>\linewidth\linewidth\else\width\fi}{!}{%
            \begin{tabular}{llll}
                \toprule
                Experiment & Light          & Residence                       & Yield                         \\
                \midrule
                \#13       & \texttt{100\%} & \texttt{17.\allowbreak{}15 min} & \texttt{27.\allowbreak{}20\%} \\
                \#12       & \texttt{75\%}  & \texttt{15.\allowbreak{}39 min} & \texttt{32.\allowbreak{}96\%} \\
                \#14       & \texttt{50\%}  & \texttt{20.\allowbreak{}79 min} & \texttt{32.\allowbreak{}68\%} \\
                \#17       & \texttt{50\%}  & \texttt{16.\allowbreak{}43 min} & \texttt{48.\allowbreak{}51\%} \\
                \#19       & \texttt{50\%}  & \texttt{16.\allowbreak{}81 min} & \texttt{45.\allowbreak{}09\%} \\
                \bottomrule
            \end{tabular}}
    \end{center}

    The final experiment \#20 used lower light:

    \begin{tcolorbox}[breakable, fontupper=\scriptsize\ttfamily, colback=black!3!white, colframe=black!12!white, boxrule=0.5pt]
        25\% light, 19.\allowbreak{}39 min $\rightarrow$ 0\% yield, no peak
    \end{tcolorbox}

    \textbf{Interpretation:} There may be a threshold photon flux required. \texttt{25\%} appears too low for those conditions, while \texttt{50\%} is sufficient and perhaps avoids issues associated with higher intensity, such as overoxidation, catalyst degradation, side reactions, or poorer selectivity. \texttt{100\%} did not improve yield in the explored region.

    \medskip

    \paragraph{\textbf{5. The green/yield tradeoff is real but not severe}}
    The highest green-score points are low-reagent, short-time conditions. But those are not the highest yielding.

    For example:

    \begin{center}\scriptsize\setlength{\tabcolsep}{4pt}
        \resizebox{\ifdim\width>\linewidth\linewidth\else\width\fi}{!}{%
            \begin{tabular}{llll}
                \toprule
                Experiment & Yield                         & Green                       & Comment                                \\
                \midrule
                \#7        & \texttt{15.\allowbreak{}23\%} & \texttt{100.\allowbreak{}0} & very green, low yield                  \\
                \#2        & \texttt{30.\allowbreak{}04\%} & \texttt{98.\allowbreak{}30} & excellent green/yield compromise       \\
                \#14       & \texttt{32.\allowbreak{}68\%} & \texttt{97.\allowbreak{}33} & strong Pareto point                    \\
                \#17       & \texttt{48.\allowbreak{}51\%} & \texttt{82.\allowbreak{}50} & best yield, less green                 \\
                \#19       & \texttt{45.\allowbreak{}09\%} & \texttt{85.\allowbreak{}53} & high yield, slightly greener than \#17 \\
                \bottomrule
            \end{tabular}}
    \end{center}

    The best chemical point depends on whether we prioritize absolute yield or green/yield balance.

    \paragraph{\textbf{If maximizing yield:}}
    Best point is:

    \begin{tcolorbox}[breakable, fontupper=\scriptsize\ttfamily, colback=black!3!white, colframe=black!12!white, boxrule=0.5pt]
        \#17 / R0061 \\
        Ru bpy Cl \\
        py NO \\
        catalyst\_\allowbreak{}equiv $\approx$ 0.\allowbreak{}00126 \\
        TFAA\_\allowbreak{}equiv $\approx$ 1.\allowbreak{}616 \\
        oxidant\_\allowbreak{}equiv $\approx$ 1.\allowbreak{}441 \\
        light = 50\% \\
        residence $\approx$ 16.\allowbreak{}43 min \\
        yield = 48.\allowbreak{}51\% \\
        green = 82.\allowbreak{}50
    \end{tcolorbox}

    \paragraph{\textbf{If seeking a greener high-yield compromise:}}
    A very attractive point is:

    \begin{tcolorbox}[breakable, fontupper=\scriptsize\ttfamily, colback=black!3!white, colframe=black!12!white, boxrule=0.5pt]
        \#19 / R0065 \\
        Ru bpy Cl \\
        py NO \\
        catalyst\_\allowbreak{}equiv $\approx$ 0.\allowbreak{}00116 \\
        TFAA\_\allowbreak{}equiv $\approx$ 1.\allowbreak{}556 \\
        oxidant\_\allowbreak{}equiv $\approx$ 1.\allowbreak{}300 \\
        light = 50\% \\
        residence $\approx$ 16.\allowbreak{}81 min \\
        yield = 45.\allowbreak{}09\% \\
        green = 85.\allowbreak{}53
    \end{tcolorbox}

    This loses only \textasciitilde{}3.4 percentage points yield compared with the best point, but improves the green score.

    \medskip

    \paragraph{\textbf{6. The NMR QC issue is localized but important}}
    Several high-yield-like points around the productive region initially failed NMR QC because of broad target peaks:

    \begin{tcolorbox}[breakable, fontupper=\scriptsize\ttfamily, colback=black!3!white, colframe=black!12!white, boxrule=0.5pt]
        R0060: pass=\allowbreak{}false, width=\allowbreak{}0.\allowbreak{}3307 ppm, yield-\allowbreak{}like \textasciitilde{}25\% \\
        R0063: pass=\allowbreak{}false, width=\allowbreak{}0.\allowbreak{}2782 ppm, yield-\allowbreak{}like \textasciitilde{}27\% \\
        R0064: pass=\allowbreak{}false, width=\allowbreak{}0.\allowbreak{}2152 ppm, yield-\allowbreak{}like \textasciitilde{}25.\allowbreak{}7\%
    \end{tcolorbox}

    But reruns eventually passed:

    \begin{tcolorbox}[breakable, fontupper=\scriptsize\ttfamily, colback=black!3!white, colframe=black!12!white, boxrule=0.5pt]
        R0061: pass=\allowbreak{}true, width=\allowbreak{}0.\allowbreak{}1050 ppm, yield=\allowbreak{}48.\allowbreak{}51\% \\
        R0065: pass=\allowbreak{}true, width=\allowbreak{}0.\allowbreak{}1102 ppm, yield=\allowbreak{}45.\allowbreak{}09\%
    \end{tcolorbox}

    \textbf{Interpretation:} The chemistry in this region is productive, but the NMR/analysis can intermittently produce broad peaks. This may reflect:

    \begin{itemize}
        \tightlist
        \item sample heterogeneity,
        \item mixing/slug quality variation,
        \item transient NMR shimming/line broadening,
        \item overlapping species near the target peak,
        \item or chemical mixtures that sometimes broaden the target resonance.

    \end{itemize}
    Importantly, once the peak passed QC, the high yields were real and strong.

    I would not treat these as pure device failures, but I also would not blindly accept failed-QC values without inspection. The ``retry once, maybe twice if finite and reproducible'' policy was reasonable here.

    \medskip

    \paragraph{\textbf{7. Experiment \#20: likely a real negative control-like result}}
    Experiment \#20 gave:

    \begin{tcolorbox}[breakable, fontupper=\scriptsize\ttfamily, colback=black!3!white, colframe=black!12!white, boxrule=0.5pt]
        yield = 0 \\
        pass = true \\
        peak\_\allowbreak{}found = false \\
        light = 25\% \\
        residence = 19.\allowbreak{}39 min \\
        TFAA $\approx$ 1.\allowbreak{}40 \\
        oxidant $\approx$ 1.\allowbreak{}40 \\
        catalyst $\approx$ 0.\allowbreak{}00121
    \end{tcolorbox}

    This is interesting because it is chemically near the productive reagent/residence region, but with lower light intensity:

    \begin{tcolorbox}[breakable, fontupper=\scriptsize\ttfamily, colback=black!3!white, colframe=black!12!white, boxrule=0.5pt]
        25\% light
    \end{tcolorbox}

    \textbf{Interpretation:} This suggests that dropping to 25\% light may fall below the photon flux needed for the productive photoredox cycle. It is not just a green improvement; it may turn off the reaction under these conditions.

    \medskip

    \paragraph{\textbf{8. Overall chemical picture}}
    The campaign suggests the following reaction logic:

    \begin{enumerate}
        \tightlist
        \item \textbf{Correct photocatalyst/oxidant identity is essential.}
              \begin{itemize}
                  \tightlist
                  \item \texttt{Ru bpy Cl / py NO} is clearly favored.

              \end{itemize}
        \item \textbf{The reaction needs moderate residence time.}
              \begin{itemize}
                  \tightlist
                  \item \texttt{2 min} is too short.
                  \item \texttt{\textasciitilde{}16--\allowbreak{}18 min} is optimal in the explored region.

              \end{itemize}
        \item \textbf{The reaction needs sufficient but not excessive light.}
              \begin{itemize}
                  \tightlist
                  \item \texttt{50\%} appears optimal.
                  \item \texttt{25\%} may be too low.
                  \item \texttt{100\%} is not obviously beneficial.

              \end{itemize}
        \item \textbf{Moderate TFAA and oxidant excess improve yield.}
              \begin{itemize}
                  \tightlist
                  \item Best points use TFAA around \texttt{1.\allowbreak{}55--\allowbreak{}1.\allowbreak{}62 equiv}.
                  \item Best points use PyNO around \texttt{1.\allowbreak{}30--\allowbreak{}1.\allowbreak{}44 equiv}.

              \end{itemize}
        \item \textbf{Catalyst loading can remain low.}
              \begin{itemize}
                  \tightlist
                  \item Best points are near \texttt{0.\allowbreak{}0012 equiv}, only slightly above the lower bound.
                  \item Increasing catalyst much further does not obviously help.

              \end{itemize}
        \item \textbf{The best yield region is still reasonably green.}
              \begin{itemize}
                  \tightlist
                  \item The top-yield points do not require extreme reagent loadings or extreme light/residence time.

              \end{itemize}
    \end{enumerate}
    \medskip

    \paragraph{\textbf{Recommended next chemistry steps}}
    If continuing scientifically, I would not simply keep broad BO over the full original space. I would switch to a focused follow-up study around the discovered region:

    \begin{tcolorbox}[breakable, fontupper=\scriptsize\ttfamily, colback=black!3!white, colframe=black!12!white, boxrule=0.5pt]
        catalyst\_\allowbreak{}type = Ru bpy Cl \\
        oxidant\_\allowbreak{}type = py NO \\
        light\_\allowbreak{}intensity = 50\%, maybe compare 40/\allowbreak{}50/\allowbreak{}60 if available \\
        residence\_\allowbreak{}time\_\allowbreak{}min = 14--\allowbreak{}22 \\
        TFAA\_\allowbreak{}equiv = 1.\allowbreak{}3--\allowbreak{}1.\allowbreak{}8 \\
        oxidant\_\allowbreak{}equiv = 1.\allowbreak{}1--\allowbreak{}1.\allowbreak{}6 \\
        catalyst\_\allowbreak{}equiv = 0.\allowbreak{}0010--\allowbreak{}0.\allowbreak{}0015
    \end{tcolorbox}

    I would also include replicates of:

    \begin{tcolorbox}[breakable, fontupper=\scriptsize\ttfamily, colback=black!3!white, colframe=black!12!white, boxrule=0.5pt]
        \#17 condition: best yield \\
        \#19 condition: strong yield/\allowbreak{}green compromise \\
        \#14 condition: greener \textasciitilde{}33\% yield point
    \end{tcolorbox}

    And I would specifically monitor NMR peak width in this region, because the best chemistry appears close to a zone where the NMR target peak can broaden intermittently.

}

\prompt{

    Can you now use all the existing (valid) data points for setting up a new campaign that purely optimizes on yield? Experiment-wise, we just want to continue the series, but the BO objective should now be purely yield, not the mixed yield-green score. Also, we allow now 3 retries for these somehow stochastic NMR failures per suggestion before bugging out. Check with me again before actually starting the campaign.

}

\graficoreslong{

    Yes. I prepared the yield-only campaign workflow, but I have \textbf{not} created the new BO campaign and have \textbf{not} started any RoboFlex experiment.

    \paragraph{\textbf{Prepared files}}
    New package:

    \begin{tcolorbox}[breakable, fontupper=\scriptsize\ttfamily, colback=black!3!white, colframe=black!12!white, boxrule=0.5pt]
        robochemflex\_\allowbreak{}yield\_\allowbreak{}only\_\allowbreak{}bo/
    \end{tcolorbox}

    Key scripts:

    \begin{tcolorbox}[breakable, fontupper=\scriptsize\ttfamily, colback=black!3!white, colframe=black!12!white, boxrule=0.5pt]
        recreate\_\allowbreak{}robochemflex\_\allowbreak{}yield\_\allowbreak{}only\_\allowbreak{}bo.\allowbreak{}py \\
        continue\_\allowbreak{}robochemflex\_\allowbreak{}yield\_\allowbreak{}only\_\allowbreak{}bo.\allowbreak{}py
    \end{tcolorbox}

    Execution notes:

    \begin{tcolorbox}[breakable, fontupper=\scriptsize\ttfamily, colback=black!3!white, colframe=black!12!white, boxrule=0.5pt]
        HOW\_\allowbreak{}TO\_\allowbreak{}EXECUTE\_\allowbreak{}YIELD\_\allowbreak{}ONLY\_\allowbreak{}CAMPAIGN.\allowbreak{}md
    \end{tcolorbox}

    Manifest:

    \begin{tcolorbox}[breakable, fontupper=\scriptsize\ttfamily, colback=black!3!white, colframe=black!12!white, boxrule=0.5pt]
        campaign\_\allowbreak{}manifest\_\allowbreak{}yield\_\allowbreak{}only.\allowbreak{}json
    \end{tcolorbox}

    \paragraph{\textbf{What this new workflow does}}
    It uses the same RoboFlex/search-space/request mapping as before, but changes the BO objective to \textbf{yield only}.

    The new BO objective is:

    \begin{tcolorbox}[breakable, fontupper=\scriptsize\ttfamily, colback=black!3!white, colframe=black!12!white, boxrule=0.5pt]
        yield\_\allowbreak{}percent \\
        direction: maximize \\
        normalization bounds: [0, 100]
    \end{tcolorbox}

    \texttt{green\_\allowbreak{}score} is \textbf{not} a BO objective in the new campaign. It may still be saved as audit/provenance metadata, but the optimizer will not optimize it.

    \paragraph{\textbf{Historical data used}}
    The yield-only campaign will seed from the final valid 20-result export:

    \begin{tcolorbox}[breakable, fontupper=\scriptsize\ttfamily, colback=black!3!white, colframe=black!12!white, boxrule=0.5pt]
        artifacts/\allowbreak{}recreated\_\allowbreak{}robochemflex\_\allowbreak{}yield\_\allowbreak{}bo\_\allowbreak{}20260725/\allowbreak{}failed\_\allowbreak{}measurement\_\allowbreak{}retry\_\allowbreak{}continuation\_\allowbreak{}20260726T184638Z/\allowbreak{}bo\_\allowbreak{}campaign\_\allowbreak{}export.\allowbreak{}csv
    \end{tcolorbox}

    This includes the 20 valid BO-submitted results and excludes failed NMR attempts such as:

    \begin{tcolorbox}[breakable, fontupper=\scriptsize\ttfamily, colback=black!3!white, colframe=black!12!white, boxrule=0.5pt]
        R0060 \\
        R0063 \\
        R0064
    \end{tcolorbox}

    while including their successful valid retries where applicable:

    \begin{tcolorbox}[breakable, fontupper=\scriptsize\ttfamily, colback=black!3!white, colframe=black!12!white, boxrule=0.5pt]
        R0061 \\
        R0065
    \end{tcolorbox}

    \paragraph{\textbf{Retry policy prepared}}
    The continuation script implements your updated rule:

    > NMR-only / QC-style stochastic failures can be retried up to 3 times per BO suggestion before stopping.

    More specifically:

    \begin{itemize}
        \tightlist
        \item initial run + up to 3 retries;
        \item retries only when the platform/device looks healthy and the failure is analytical/QC-like;
        \item no BO result is submitted unless a retry passes;
        \item true device/platform failures, no-evidence/no-peak failures, duplicate sample names, request-equivalence failures, or retry exhaustion still stop safely.

    \end{itemize}
    \paragraph{\textbf{Validation performed}}
    Safe validation only:

    \begin{tcolorbox}[breakable, fontupper=\scriptsize\ttfamily, colback=black!3!white, colframe=black!12!white, boxrule=0.5pt]
        uv run python -\allowbreak{}m py\_\allowbreak{}compile recreate\_\allowbreak{}robochemflex\_\allowbreak{}yield\_\allowbreak{}only\_\allowbreak{}bo.\allowbreak{}py continue\_\allowbreak{}robochemflex\_\allowbreak{}yield\_\allowbreak{}only\_\allowbreak{}bo.\allowbreak{}py robochemflex\_\allowbreak{}yield\_\allowbreak{}only\_\allowbreak{}bo/\allowbreak{}*.\allowbreak{}py
    \end{tcolorbox}

    Dry-run/intake validation:

    \begin{tcolorbox}[breakable, fontupper=\scriptsize\ttfamily, colback=black!3!white, colframe=black!12!white, boxrule=0.5pt]
        uv run python recreate\_\allowbreak{}robochemflex\_\allowbreak{}yield\_\allowbreak{}only\_\allowbreak{}bo.\allowbreak{}py --\allowbreak{}dry-\allowbreak{}run --\allowbreak{}validate-\allowbreak{}intake
    \end{tcolorbox}

    Result:

    \begin{tcolorbox}[breakable, fontupper=\scriptsize\ttfamily, colback=black!3!white, colframe=black!12!white, boxrule=0.5pt]
        Prepared 20 seed results \\
        BO-\allowbreak{}MCP intake validation: valid=\allowbreak{}True \\
        No campaign created \\
        No hardware contacted
    \end{tcolorbox}

    Read-only continuation preflight also confirmed:

    \begin{tcolorbox}[breakable, fontupper=\scriptsize\ttfamily, colback=black!3!white, colframe=black!12!white, boxrule=0.5pt]
        Mixed-\allowbreak{}objective source BO campaign: paused, 20 results, no pending suggestions \\
        RoboFlex: hardware, running, awaiting\_\allowbreak{}run \\
        Active RoboFlex campaign matches original series
    \end{tcolorbox}

    \paragraph{\textbf{Commands for later}}
    \paragraph{\textbf{Dry-run only}}\mbox{}

    \begin{tcolorbox}[breakable, fontupper=\scriptsize\ttfamily, colback=black!3!white, colframe=black!12!white, boxrule=0.5pt]
        uv run python recreate\_\allowbreak{}robochemflex\_\allowbreak{}yield\_\allowbreak{}only\_\allowbreak{}bo.\allowbreak{}py --\allowbreak{}dry-\allowbreak{}run
    \end{tcolorbox}

    \paragraph{\textbf{Dry-run with BO intake validation}}\mbox{}

    \begin{tcolorbox}[breakable, fontupper=\scriptsize\ttfamily, colback=black!3!white, colframe=black!12!white, boxrule=0.5pt]
        uv run python recreate\_\allowbreak{}robochemflex\_\allowbreak{}yield\_\allowbreak{}only\_\allowbreak{}bo.\allowbreak{}py --\allowbreak{}dry-\allowbreak{}run --\allowbreak{}validate-\allowbreak{}intake
    \end{tcolorbox}

    \paragraph{\textbf{Create and seed the new yield-only BO campaign}}
    This touches BO-MCP but \textbf{not} RoboFlex:

    \begin{tcolorbox}[breakable, fontupper=\scriptsize\ttfamily, colback=black!3!white, colframe=black!12!white, boxrule=0.5pt]
        uv run python recreate\_\allowbreak{}robochemflex\_\allowbreak{}yield\_\allowbreak{}only\_\allowbreak{}bo.\allowbreak{}py \textbackslash{} \\
        ~~--\allowbreak{}execute-\allowbreak{}create-\allowbreak{}seed \textbackslash{} \\
        ~~--\allowbreak{}confirm-\allowbreak{}create-\allowbreak{}seed \textbackslash{} \\
        ~~--\allowbreak{}validate-\allowbreak{}intake
    \end{tcolorbox}

    The new campaign ID would then be written to:

    \begin{tcolorbox}[breakable, fontupper=\scriptsize\ttfamily, colback=black!3!white, colframe=black!12!white, boxrule=0.5pt]
        <artifact\_\allowbreak{}dir>/\allowbreak{}bo\_\allowbreak{}campaign\_\allowbreak{}id.\allowbreak{}txt
    \end{tcolorbox}

    \paragraph{\textbf{Later hardware continuation}}
    After the new campaign ID exists:

    \begin{tcolorbox}[breakable, fontupper=\scriptsize\ttfamily, colback=black!3!white, colframe=black!12!white, boxrule=0.5pt]
        uv run python continue\_\allowbreak{}robochemflex\_\allowbreak{}yield\_\allowbreak{}only\_\allowbreak{}bo.\allowbreak{}py \textbackslash{} \\
        ~~--\allowbreak{}campaign-\allowbreak{}id <NEW\_\allowbreak{}YIELD\_\allowbreak{}ONLY\_\allowbreak{}CAMPAIGN\_\allowbreak{}ID> \textbackslash{} \\
        ~~--\allowbreak{}execute \textbackslash{} \\
        ~~--\allowbreak{}confirm-\allowbreak{}autonomous-\allowbreak{}hardware \textbackslash{} \\
        ~~--\allowbreak{}max-\allowbreak{}new-\allowbreak{}measurements 1
    \end{tcolorbox}

    \paragraph{\textbf{Waiting for your confirmation}}
    Before I do anything else, please confirm which step you want next:

    \begin{enumerate}
        \tightlist
        \item \textbf{Create and seed the new yield-only BO-MCP campaign} using the 20 valid results, but do not touch RoboFlex.
        \item Only do another dry-run/preflight.
        \item Do nothing further for now.
    \end{enumerate}

}

\prompt{

    How many BO iterations (number of measurements) would you recommend?

}

\graficoreslong{

    For the \textbf{new yield-only campaign}, I would recommend a \textbf{small focused continuation first}, not a large open-ended run.

    \paragraph{\textbf{Recommendation}}
    I recommend:

    \begin{tcolorbox}[breakable, fontupper=\scriptsize\ttfamily, colback=black!3!white, colframe=black!12!white, boxrule=0.5pt]
        5 additional valid measurements
    \end{tcolorbox}

    as the next yield-only BO block.

    That would take us from:

    \begin{tcolorbox}[breakable, fontupper=\scriptsize\ttfamily, colback=black!3!white, colframe=black!12!white, boxrule=0.5pt]
        20 valid measurements $\rightarrow$ 25 valid measurements
    \end{tcolorbox}

    Then we should pause and reassess.

    \paragraph{\textbf{Why 5, not 10--20 immediately?}}
    The current data already show a fairly clear productive region:

    \begin{tcolorbox}[breakable, fontupper=\scriptsize\ttfamily, colback=black!3!white, colframe=black!12!white, boxrule=0.5pt]
        Ru bpy Cl \\
        py NO \\
        light $\approx$ 50\% \\
        residence $\approx$ 16--\allowbreak{}18 min \\
        TFAA $\approx$ 1.\allowbreak{}55--\allowbreak{}1.\allowbreak{}62 equiv \\
        oxidant $\approx$ 1.\allowbreak{}3--\allowbreak{}1.\allowbreak{}45 equiv \\
        catalyst $\approx$ 0.\allowbreak{}0011--\allowbreak{}0.\allowbreak{}0013 equiv
    \end{tcolorbox}

    The best yields are:

    \begin{tcolorbox}[breakable, fontupper=\scriptsize\ttfamily, colback=black!3!white, colframe=black!12!white, boxrule=0.5pt]
        48.\allowbreak{}51\% \\
        45.\allowbreak{}09\% \\
        36.\allowbreak{}55\% \\
        35.\allowbreak{}78\% \\
        32--\allowbreak{}33\%
    \end{tcolorbox}

    So the optimizer is no longer searching blindly. A yield-only objective will likely focus even more aggressively around the high-yield region.

    A block of 5 lets us answer:

    \begin{enumerate}
        \tightlist
        \item Does yield-only BO push beyond \textasciitilde{}50\%?
        \item Does it repeatedly suggest the same high-yield region?
        \item Are NMR peak-width failures frequent in this region?
        \item Is the apparent optimum robust or noisy?
        \item Does removing green-score pressure improve yield materially?

    \end{enumerate}
    \paragraph{\textbf{What I would not recommend yet}}
    I would not immediately run another 20 measurements under yield-only BO.

    Reasons:

    \begin{itemize}
        \tightlist
        \item The high-yield region is already narrow.
        \item Several nearby points produced NMR QC failures before passing on retry.
        \item The device/reagent budget may be better spent after checking whether the model converges.
        \item Yield-only BO may become exploitative and repeatedly probe a small region.

    \end{itemize}
    \paragraph{\textbf{Practical plan}}
    I would configure the new yield-only campaign as:

    \begin{tcolorbox}[breakable, fontupper=\scriptsize\ttfamily, colback=black!3!white, colframe=black!12!white, boxrule=0.5pt]
        seed with: 20 valid historical results \\
        run: 5 new valid measurements \\
        retry policy: up to 3 NMR/\allowbreak{}QC retries per suggestion \\
        pause after: 25 total valid BO results
    \end{tcolorbox}

    After those 5, we review:

    \begin{itemize}
        \tightlist
        \item best yield,
        \item replicate consistency,
        \item whether suggestions cluster,
        \item NMR pass/fail rate,
        \item whether the model is improving or plateauing.

    \end{itemize}
    \paragraph{\textbf{If the first 5 are promising}}
    If we see clear improvement, e.g.:

    \begin{tcolorbox}[breakable, fontupper=\scriptsize\ttfamily, colback=black!3!white, colframe=black!12!white, boxrule=0.5pt]
        best yield improves from 48.\allowbreak{}5\% to >55--\allowbreak{}60\%
    \end{tcolorbox}

    or the model identifies a new local region, then I would run another block of:

    \begin{tcolorbox}[breakable, fontupper=\scriptsize\ttfamily, colback=black!3!white, colframe=black!12!white, boxrule=0.5pt]
        5 additional measurements
    \end{tcolorbox}

    So my recommended strategy is:

    \begin{tcolorbox}[breakable, fontupper=\scriptsize\ttfamily, colback=black!3!white, colframe=black!12!white, boxrule=0.5pt]
        20 existing $\rightarrow$ +5 yield-\allowbreak{}only $\rightarrow$ review $\rightarrow$ optional +5
    \end{tcolorbox}

    rather than committing immediately to a large campaign.

}

\prompt{

    I have time until tomorrow morning 8:30 am EDT -- how many experiments can we likely conduct until then?

}

\graficoreslong{

    \begin{tcolorbox}[breakable, fontupper=\footnotesize, colback=green!3!white, colframe=blue!5!white, boxrule=0.5pt]
        \faLightbulb\ \underline{\textbf{Thoughts}}

        \textbf{Calculating experiment durations}

        I need to find the duration until July 27, 8:30 EDT, which is about 10 hours and 51 minutes. To estimate the duration per experiment, I'll look at recent run timestamps. For example, R0065 took around 42 minutes total, while R0066 took about 54 minutes. Previous experiments ranged from 39 to 60 minutes. I should also consider that BO generation can take around 5-6 minutes. On average, productive conditions bring durations to around 45-55 minutes.
    \end{tcolorbox}

    \begin{tcolorbox}[breakable, fontupper=\footnotesize, colback=green!3!white, colframe=blue!5!white, boxrule=0.5pt]
        \faLightbulb\ \underline{\textbf{Thoughts}}

        \textbf{Estimating NMR experiment retries}

        I need to consider that NMR failures could add an extra 45-55 minutes for each retry, and with up to 3 retries per suggestion, that could lead to a worst-case scenario. If everything goes smoothly, I expect about 11-13 experiments in approximately 10 hours and 50 minutes, but with retries, I think with 8-10 valid cases is more realistic. From now, which is 9:38 PM EDT, to 8:30 AM EDT is about 10 hours and 51 minutes. I'd recommend starting with 5 experiments to finish around 2-3 AM for a review.
    \end{tcolorbox}

    Current EDT time is about \textbf{9:38 PM on July 26, 2026}, so until \textbf{8:30 AM EDT tomorrow} we have roughly:

    \begin{tcolorbox}[breakable, fontupper=\scriptsize\ttfamily, colback=black!3!white, colframe=black!12!white, boxrule=0.5pt]
        10 h 50 min
    \end{tcolorbox}

    Based on the RoboFlex runs we just observed, a typical successful experiment in the productive region takes about:

    \begin{tcolorbox}[breakable, fontupper=\scriptsize\ttfamily, colback=black!3!white, colframe=black!12!white, boxrule=0.5pt]
        40--\allowbreak{}60 min per submitted run
    \end{tcolorbox}

    including reaction residence time, sample handling, NMR acquisition/analysis, and polling overhead.

    \paragraph{\textbf{Practical estimate}}
    \paragraph{\textbf{If everything passes cleanly}}
    With \textasciitilde{}10.8 hours available:

    \begin{tcolorbox}[breakable, fontupper=\scriptsize\ttfamily, colback=black!3!white, colframe=black!12!white, boxrule=0.5pt]
        10.\allowbreak{}8 h / \textasciitilde{}0.\allowbreak{}8--\allowbreak{}1.\allowbreak{}0 h per experiment $\approx$ 10--\allowbreak{}13 experiments
    \end{tcolorbox}

    So an optimistic upper bound is around:

    \begin{tcolorbox}[breakable, fontupper=\scriptsize\ttfamily, colback=black!3!white, colframe=black!12!white, boxrule=0.5pt]
        12 valid experiments
    \end{tcolorbox}

    \paragraph{\textbf{More realistic, allowing for BO generation time and occasional NMR retries}}
    Because we now know that NMR QC failures can happen in this region, and because a retry costs essentially another experiment slot, I would plan for:

    \begin{tcolorbox}[breakable, fontupper=\scriptsize\ttfamily, colback=black!3!white, colframe=black!12!white, boxrule=0.5pt]
        8--\allowbreak{}10 valid experiments
    \end{tcolorbox}

    before 8:30 AM EDT.

    \paragraph{\textbf{Conservative estimate with multiple NMR/QC failures}}
    If several suggestions need retries:

    \begin{tcolorbox}[breakable, fontupper=\scriptsize\ttfamily, colback=black!3!white, colframe=black!12!white, boxrule=0.5pt]
        5--\allowbreak{}8 valid experiments
    \end{tcolorbox}

    would be more realistic.

    \paragraph{\textbf{My recommendation}}
    For overnight operation, I would not run an unbounded yield-only campaign. I would set:

    \begin{tcolorbox}[breakable, fontupper=\scriptsize\ttfamily, colback=black!3!white, colframe=black!12!white, boxrule=0.5pt]
        max new valid measurements: 8
    \end{tcolorbox}

    with:

    \begin{tcolorbox}[breakable, fontupper=\scriptsize\ttfamily, colback=black!3!white, colframe=black!12!white, boxrule=0.5pt]
        up to 3 NMR/\allowbreak{}QC retries per suggestion
    \end{tcolorbox}

    and stop conditions for:

    \begin{itemize}
        \tightlist
        \item repeated no-peak / zero-yield results,
        \item repeated NMR QC failures for one suggestion,
        \item RoboFlex/device errors,
        \item BO generation failures,
        \item or reaching 8 valid new measurements.

    \end{itemize}
    That is a good balance: it likely fits before 8:30 AM even with one or two retries, while still giving the yield-only campaign enough room to exploit the high-yield region.

    If you want a more aggressive overnight push, we could set:

    \begin{tcolorbox}[breakable, fontupper=\scriptsize\ttfamily, colback=black!3!white, colframe=black!12!white, boxrule=0.5pt]
        max new valid measurements: 10
    \end{tcolorbox}

    but then there is a higher chance it runs close to or beyond 8:30 AM if retries occur.

    \paragraph{\textbf{Suggested setting}}
    I recommend:

    \begin{tcolorbox}[breakable, fontupper=\scriptsize\ttfamily, colback=black!3!white, colframe=black!12!white, boxrule=0.5pt]
        yield-\allowbreak{}only campaign seeded with 20 valid results \\
        run up to 8 new valid measurements overnight \\
        retry NMR/\allowbreak{}QC failures up to 3 times per suggestion \\
        pause automatically when done
    \end{tcolorbox}

    Expected outcome by morning:

    \begin{tcolorbox}[breakable, fontupper=\scriptsize\ttfamily, colback=black!3!white, colframe=black!12!white, boxrule=0.5pt]
        20 $\rightarrow$ \textasciitilde{}28 valid BO results
    \end{tcolorbox}

    with a plausible range of:

    \begin{tcolorbox}[breakable, fontupper=\scriptsize\ttfamily, colback=black!3!white, colframe=black!12!white, boxrule=0.5pt]
        25--\allowbreak{}30 valid results total
    \end{tcolorbox}

    depending on retries and run durations.

}

\prompt{

    Then let's do 8. Start.

}

\graficoreslong{

    Started.

    \paragraph{\textbf{New yield-only campaign}}
    Created and seeded a new yield-only BO-MCP campaign with the 20 valid historical results:

    \begin{tcolorbox}[breakable, fontupper=\scriptsize\ttfamily, colback=black!3!white, colframe=black!12!white, boxrule=0.5pt]
        1970655b-\allowbreak{}a702-\allowbreak{}4963-\allowbreak{}874b-\allowbreak{}6973489cc89d
    \end{tcolorbox}

    Seed/create log:

    \begin{tcolorbox}[breakable, fontupper=\scriptsize\ttfamily, colback=black!3!white, colframe=black!12!white, boxrule=0.5pt]
        logs/\allowbreak{}yield\_\allowbreak{}only\_\allowbreak{}create\_\allowbreak{}seed\_\allowbreak{}20260727T013935Z/\allowbreak{}stdout.\allowbreak{}log
    \end{tcolorbox}

    Seed artifact directory:

    \begin{tcolorbox}[breakable, fontupper=\scriptsize\ttfamily, colback=black!3!white, colframe=black!12!white, boxrule=0.5pt]
        artifacts/\allowbreak{}yield\_\allowbreak{}only\_\allowbreak{}recreation\_\allowbreak{}20260727T013936Z/
    \end{tcolorbox}

    \paragraph{\textbf{Overnight yield-only continuation}}
    Started monitor:

    \begin{tcolorbox}[breakable, fontupper=\scriptsize\ttfamily, colback=black!3!white, colframe=black!12!white, boxrule=0.5pt]
        mon\_\allowbreak{}8 / roboflex\_\allowbreak{}yield\_\allowbreak{}only\_\allowbreak{}bo\_\allowbreak{}8\_\allowbreak{}overnight
    \end{tcolorbox}

    Configuration:

    \begin{tcolorbox}[breakable, fontupper=\scriptsize\ttfamily, colback=black!3!white, colframe=black!12!white, boxrule=0.5pt]
        max new valid measurements: 8 \\
        objective: yield\_\allowbreak{}percent only \\
        max NMR/\allowbreak{}QC retries per suggestion: 3 \\
        BO generate timeout: 1200 s \\
        poll interval: 180 s \\
        heartbeat interval: 1800 s \\
        RoboFlex campaign: existing active campaign, no new RoboFlex campaign created
    \end{tcolorbox}

    Log:

    \begin{tcolorbox}[breakable, fontupper=\scriptsize\ttfamily, colback=black!3!white, colframe=black!12!white, boxrule=0.5pt]
        logs/\allowbreak{}yield\_\allowbreak{}only\_\allowbreak{}bo\_\allowbreak{}8\_\allowbreak{}<timestamp>/\allowbreak{}stdout.\allowbreak{}log
    \end{tcolorbox}

    The script will pause the BO campaign on exit. I'll monitor and report meaningful state changes, alerts, heartbeats, and completed-experiment analyses.

}

\vspace{0.5em}\noindent\textit{\footnotesize [Autonomous monitored execution: 6 status reports followed; the most substantive one is reproduced here:]}

\graficoreslong{

    Yield-only measurement \#21 completed successfully and was submitted to the yield-only BO campaign.

    \paragraph{\textbf{Measurement \#21 result}}\mbox{}

    \begin{tcolorbox}[breakable, fontupper=\scriptsize\ttfamily, colback=black!3!white, colframe=black!12!white, boxrule=0.5pt]
        Sample: bo\_\allowbreak{}cc26e7f1-\allowbreak{}b \\
        RoboFlex run: R0067 \\
        Status: completed
    \end{tcolorbox}

    Yield-only result:

    \begin{tcolorbox}[breakable, fontupper=\scriptsize\ttfamily, colback=black!3!white, colframe=black!12!white, boxrule=0.5pt]
        yield\_\allowbreak{}percent: 58.\allowbreak{}81\%
    \end{tcolorbox}

    This is a new best yield compared with the previous best from the 20-result campaign:

    \begin{tcolorbox}[breakable, fontupper=\scriptsize\ttfamily, colback=black!3!white, colframe=black!12!white, boxrule=0.5pt]
        previous best: 48.\allowbreak{}51\% \\
        new best: 58.\allowbreak{}81\%
    \end{tcolorbox}

    Current yield-only campaign progress:

    \begin{tcolorbox}[breakable, fontupper=\scriptsize\ttfamily, colback=black!3!white, colframe=black!12!white, boxrule=0.5pt]
        Seeded historical results: 20 \\
        New yield-\allowbreak{}only completed: 1 \\
        Total yield-\allowbreak{}only BO results: 21 \\
        Remaining planned overnight measurements: 7
    \end{tcolorbox}

    The script has moved on to generate the next yield-only BO suggestion.

}

\prompt{

    Which parameter has changed the most, leading to this strong increase in yield?

}

\graficoreslong{

    The biggest chemically meaningful change in the first yield-only suggestion, \#21, was that \textbf{TFAA loading and catalyst loading were both increased}, while the campaign kept the same successful identity/light region:

    \begin{tcolorbox}[breakable, fontupper=\scriptsize\ttfamily, colback=black!3!white, colframe=black!12!white, boxrule=0.5pt]
        Ru bpy Cl / py NO / 50\% light / \textasciitilde{}15--\allowbreak{}17 min residence
    \end{tcolorbox}

    \paragraph{\textbf{Comparison to previous best}}
    Previous best valid point, experiment \#17:

    \begin{tcolorbox}[breakable, fontupper=\scriptsize\ttfamily, colback=black!3!white, colframe=black!12!white, boxrule=0.5pt]
        yield: 48.\allowbreak{}51\% \\
        catalyst\_\allowbreak{}equiv: 0.\allowbreak{}001255 \\
        TFAA\_\allowbreak{}equiv:     1.\allowbreak{}616 \\
        oxidant\_\allowbreak{}equiv:  1.\allowbreak{}441 \\
        light:          50\% \\
        residence:      16.\allowbreak{}43 min
    \end{tcolorbox}

    New yield-only point, experiment \#21:

    \begin{tcolorbox}[breakable, fontupper=\scriptsize\ttfamily, colback=black!3!white, colframe=black!12!white, boxrule=0.5pt]
        yield: 58.\allowbreak{}81\% \\
        catalyst\_\allowbreak{}equiv: 0.\allowbreak{}001892 \\
        TFAA\_\allowbreak{}equiv:     1.\allowbreak{}997 \\
        oxidant\_\allowbreak{}equiv:  1.\allowbreak{}392 \\
        light:          50\% \\
        residence:      14.\allowbreak{}17 min
    \end{tcolorbox}

    \paragraph{\textbf{Parameter changes}}\mbox{}

    \begin{center}\scriptsize\setlength{\tabcolsep}{4pt}
        \resizebox{\ifdim\width>\linewidth\linewidth\else\width\fi}{!}{%
            \begin{tabular}{llll}
                \toprule
                Parameter                                               & Previous best \#17             & New \#21                       & Change           \\
                \midrule
                \texttt{catalyst\_\allowbreak{}type}                    & \texttt{Ru bpy Cl}             & \texttt{Ru bpy Cl}             & same             \\
                \texttt{oxidant\_\allowbreak{}type}                     & \texttt{py NO}                 & \texttt{py NO}                 & same             \\
                \texttt{light\_\allowbreak{}intensity}                  & \texttt{50}                    & \texttt{50}                    & same             \\
                \texttt{residence\_\allowbreak{}time\_\allowbreak{}min} & \texttt{16.\allowbreak{}43}    & \texttt{14.\allowbreak{}17}    & slightly shorter \\
                \texttt{oxidant\_\allowbreak{}equiv}                    & \texttt{1.\allowbreak{}441}    & \texttt{1.\allowbreak{}392}    & slightly lower   \\
                \texttt{TFAA\_\allowbreak{}equiv}                       & \texttt{1.\allowbreak{}616}    & \texttt{1.\allowbreak{}997}    & \textbf{higher}  \\
                \texttt{catalyst\_\allowbreak{}equiv}                   & \texttt{0.\allowbreak{}001255} & \texttt{0.\allowbreak{}001892} & \textbf{higher}  \\
                \bottomrule
            \end{tabular}}
    \end{center}

    Relative changes:

    \begin{tcolorbox}[breakable, fontupper=\scriptsize\ttfamily, colback=black!3!white, colframe=black!12!white, boxrule=0.5pt]
        TFAA:     +24\% \\
        catalyst: +51\% \\
        oxidant:  -\allowbreak{}3\% \\
        residence: -\allowbreak{}14\%
    \end{tcolorbox}

    So the clearest shift is:

    > \textbf{yield-only BO moved away from the greener low-catalyst / lower-TFAA region and increased catalyst + TFAA loading.}

    \paragraph{\textbf{Chemical interpretation}}
    The previous mixed objective penalized higher catalyst and reagent loading through \texttt{green\_\allowbreak{}score}, so it tended to prefer lower-burden points when yield was comparable.

    Once we removed the green objective, the optimizer was free to increase reagent/catalyst loading if that was predicted to improve yield. The first yield-only suggestion did exactly that:

    \begin{itemize}
        \tightlist
        \item kept the proven productive \texttt{Ru bpy Cl / py NO} combination,
        \item kept \texttt{50\%} light,
        \item stayed near the productive residence-time window,
        \item increased \texttt{TFAA\_\allowbreak{}equiv} substantially,
        \item increased \texttt{catalyst\_\allowbreak{}equiv} substantially.

    \end{itemize}
    The yield increase from \textasciitilde{}48.5\% to \textasciitilde{}58.8\% suggests that the earlier green-optimized campaign may have been under-dosing either:

    \begin{enumerate}
        \tightlist
        \item the photoredox catalyst,
        \item TFAA,
        \item or both.

    \end{enumerate}
    \paragraph{\textbf{Which parameter ``caused'' the increase?}}
    We cannot assign causality from one BO suggestion alone because both catalyst and TFAA changed together. But the strongest candidate is:

    \begin{tcolorbox}[breakable, fontupper=\scriptsize\ttfamily, colback=black!3!white, colframe=black!12!white, boxrule=0.5pt]
        TFAA\_\allowbreak{}equiv increase
    \end{tcolorbox}

    with a likely contribution from:

    \begin{tcolorbox}[breakable, fontupper=\scriptsize\ttfamily, colback=black!3!white, colframe=black!12!white, boxrule=0.5pt]
        catalyst\_\allowbreak{}equiv increase
    \end{tcolorbox}

    Why TFAA is especially plausible:

    \begin{itemize}
        \tightlist
        \item The best previous points already had moderate TFAA around \texttt{1.\allowbreak{}55--\allowbreak{}1.\allowbreak{}62}.
        \item The new point pushed TFAA closer to \texttt{2.\allowbreak{}0}.
        \item TFAA is chemically central to activation/acylation/dehydrating or trifluoroacetylating steps in this reaction manifold.
        \item Oxidant did not increase; it slightly decreased.
        \item Light did not change.

    \end{itemize}
    So my interpretation is:

    > The yield-only optimizer discovered that pushing TFAA upward, together with a somewhat higher Ru catalyst loading, improves conversion/yield while keeping light and residence time in the established productive region.

    \paragraph{\textbf{What to watch next}}
    Measurement \#22 is already probing this further:

    \begin{tcolorbox}[breakable, fontupper=\scriptsize\ttfamily, colback=black!3!white, colframe=black!12!white, boxrule=0.5pt]
        TFAA\_\allowbreak{}equiv:     2.\allowbreak{}302 \\
        catalyst\_\allowbreak{}equiv: 0.\allowbreak{}002245 \\
        oxidant\_\allowbreak{}equiv:  1.\allowbreak{}355 \\
        light:          50\% \\
        residence:      21.\allowbreak{}55 min
    \end{tcolorbox}

    So \#22 is testing whether \textbf{even more catalyst/TFAA and longer residence} continue improving yield, or whether \#21 was near the optimum.

}

\vspace{0.5em}\noindent\textit{\footnotesize [Autonomous monitored execution: 3 interim status reports suppressed. Concluding report of the run:]}

\graficoreslong{

    The yield-only overnight run stopped safely after measurement \#22 failed with a \textbf{non-retryable RoboFlex/analysis error}.

    \paragraph{\textbf{What happened}}
    Measurement \#21 succeeded strongly:

    \begin{tcolorbox}[breakable, fontupper=\scriptsize\ttfamily, colback=black!3!white, colframe=black!12!white, boxrule=0.5pt]
        R0067 / bo\_\allowbreak{}cc26e7f1-\allowbreak{}b \\
        yield: 58.\allowbreak{}81\%
    \end{tcolorbox}

    Measurement \#22 was then submitted:

    \begin{tcolorbox}[breakable, fontupper=\scriptsize\ttfamily, colback=black!3!white, colframe=black!12!white, boxrule=0.5pt]
        R0068 / bo\_\allowbreak{}31d5114e-\allowbreak{}3
    \end{tcolorbox}

    but failed with:

    \begin{tcolorbox}[breakable, fontupper=\scriptsize\ttfamily, colback=black!3!white, colframe=black!12!white, boxrule=0.5pt]
        zero-\allowbreak{}size array to reduction operation minimum which has no identity
    \end{tcolorbox}

    The result payload for R0068 has:

    \begin{tcolorbox}[breakable, fontupper=\scriptsize\ttfamily, colback=black!3!white, colframe=black!12!white, boxrule=0.5pt]
        \{ \\
        ~~"status": "failed", \\
        ~~"success": false, \\
        ~~"result": null, \\
        ~~"error": "zero-\allowbreak{}size array to reduction operation minimum which has no identity" \\
        \}
    \end{tcolorbox}

    So this is different from the previous \texttt{pass=\allowbreak{}false} NMR QC failures.

    \paragraph{\textbf{Why it did not retry automatically}}
    The retry policy allows retries for NMR/QC-style failures when there is finite yield/peak-like evidence.

    For R0068:

    \begin{tcolorbox}[breakable, fontupper=\scriptsize\ttfamily, colback=black!3!white, colframe=black!12!white, boxrule=0.5pt]
        result: null \\
        no yield-\allowbreak{}like value \\
        no peak evidence \\
        explicit error string
    \end{tcolorbox}

    Therefore the script classified it as \textbf{not safely retryable/submittable} and stopped without submitting a BO result.

    That was the correct safe behavior under the policy.

    \paragraph{\textbf{Conditions for failed \#22}}
    BO candidate for \#22:

    \begin{tcolorbox}[breakable, fontupper=\scriptsize\ttfamily, colback=black!3!white, colframe=black!12!white, boxrule=0.5pt]
        \{ \\
        ~~"catalyst\_\allowbreak{}type": "Ru bpy Cl", \\
        ~~"oxidant\_\allowbreak{}type": "py NO", \\
        ~~"catalyst\_\allowbreak{}equiv": 0.\allowbreak{}0022452457086006112, \\
        ~~"TFAA\_\allowbreak{}equiv": 2.\allowbreak{}301668795918735, \\
        ~~"oxidant\_\allowbreak{}equiv": 1.\allowbreak{}355256776264335, \\
        ~~"light\_\allowbreak{}intensity": 50, \\
        ~~"residence\_\allowbreak{}time\_\allowbreak{}min": 21.\allowbreak{}55044544824578 \\
        \}
    \end{tcolorbox}

    This was a more aggressive yield-only point than \#21:

    \begin{itemize}
        \tightlist
        \item higher catalyst,
        \item higher TFAA,
        \item longer residence time.

    \end{itemize}
    \paragraph{\textbf{Current state}}
    The script paused the yield-only BO campaign and exited with status \texttt{1}.

    Valid yield-only campaign results should now be:

    \begin{tcolorbox}[breakable, fontupper=\scriptsize\ttfamily, colback=black!3!white, colframe=black!12!white, boxrule=0.5pt]
        20 seeded historical + 1 new valid = 21
    \end{tcolorbox}

    No BO result was submitted for R0068.

    RoboFlex returned to idle/awaiting-run state according to the final record.

    \paragraph{\textbf{Artifacts}}
    Continuation artifact directory:

    \begin{tcolorbox}[breakable, fontupper=\scriptsize\ttfamily, colback=black!3!white, colframe=black!12!white, boxrule=0.5pt]
        artifacts/\allowbreak{}yield\_\allowbreak{}only\_\allowbreak{}robochemflex\_\allowbreak{}bo/\allowbreak{}continuation\_\allowbreak{}20260727T014003Z/
    \end{tcolorbox}

    Failed \#22 artifacts:

    \begin{tcolorbox}[breakable, fontupper=\scriptsize\ttfamily, colback=black!3!white, colframe=black!12!white, boxrule=0.5pt]
        artifacts/\allowbreak{}yield\_\allowbreak{}only\_\allowbreak{}robochemflex\_\allowbreak{}bo/\allowbreak{}continuation\_\allowbreak{}20260727T014003Z/\allowbreak{}measurement22\_\allowbreak{}attempt1/
    \end{tcolorbox}

    Log:

    \begin{tcolorbox}[breakable, fontupper=\scriptsize\ttfamily, colback=black!3!white, colframe=black!12!white, boxrule=0.5pt]
        logs/\allowbreak{}yield\_\allowbreak{}only\_\allowbreak{}bo\_\allowbreak{}8\_\allowbreak{}20260727T014001Z/\allowbreak{}stdout.\allowbreak{}log
    \end{tcolorbox}

    \paragraph{\textbf{Recommendation}}
    I would not immediately retry R0068 automatically, because it has \texttt{result=\allowbreak{}null} rather than a finite failed-QC result.

    Next best step is to inspect the RoboFlex/NMR logs for R0068 to see whether this was:

    \begin{enumerate}
        \tightlist
        \item an analysis pipeline edge case,
        \item no peaks / empty data in the target region,
        \item a bad spectrum/file,
        \item or a chemistry/mixture issue causing the analysis to crash.

    \end{enumerate}
    If it turns out to be an analysis pipeline failure with otherwise valid raw data, we could decide whether to rerun or skip/reject the suggestion.

}

\begin{center}\footnotesize\textit{[\ldots]}\end{center}

\prompt{

    Generate the prioritized recommended plots.

}

\graficoreslong{

    Generated the prioritized search-space coverage plots and reusable scripts.

    \paragraph{\textbf{Scripts created}}
    Shared utility:

    \begin{tcolorbox}[breakable, fontupper=\scriptsize\ttfamily, colback=black!3!white, colframe=black!12!white, boxrule=0.5pt]
        scripts/\allowbreak{}search\_\allowbreak{}space\_\allowbreak{}plot\_\allowbreak{}utils.\allowbreak{}py
    \end{tcolorbox}

    Figure scripts:

    \begin{tcolorbox}[breakable, fontupper=\scriptsize\ttfamily, colback=black!3!white, colframe=black!12!white, boxrule=0.5pt]
        scripts/\allowbreak{}plot\_\allowbreak{}search\_\allowbreak{}space\_\allowbreak{}coverage\_\allowbreak{}parallel.\allowbreak{}py \\
        scripts/\allowbreak{}plot\_\allowbreak{}search\_\allowbreak{}space\_\allowbreak{}pairwise.\allowbreak{}py \\
        scripts/\allowbreak{}plot\_\allowbreak{}search\_\allowbreak{}space\_\allowbreak{}focused\_\allowbreak{}map.\allowbreak{}py
    \end{tcolorbox}

    These scripts read from campaign exports/artifacts in the workspace, not hard-coded result values.

    By default they auto-discover the latest/most complete export, currently:

    \begin{tcolorbox}[breakable, fontupper=\scriptsize\ttfamily, colback=black!3!white, colframe=black!12!white, boxrule=0.5pt]
        artifacts/\allowbreak{}yield\_\allowbreak{}only\_\allowbreak{}robochemflex\_\allowbreak{}bo/\allowbreak{}continuation\_\allowbreak{}20260727T014003Z/\allowbreak{}bo\_\allowbreak{}campaign\_\allowbreak{}export.\allowbreak{}csv
    \end{tcolorbox}

    They also include failed/unsubmitted tested attempts when available, shown separately as gray/dashed or gray ``x'' markers. That is useful for coverage because failed attempts still correspond to actually tested coordinates.

    \paragraph{\textbf{Generated plots}}
    Output directory:

    \begin{tcolorbox}[breakable, fontupper=\scriptsize\ttfamily, colback=black!3!white, colframe=black!12!white, boxrule=0.5pt]
        plots/\allowbreak{}search\_\allowbreak{}space\_\allowbreak{}coverage/
    \end{tcolorbox}

    \paragraph{\textbf{1. Parallel-coordinates coverage plot}}\mbox{}

    \begin{tcolorbox}[breakable, fontupper=\scriptsize\ttfamily, colback=black!3!white, colframe=black!12!white, boxrule=0.5pt]
        plots/\allowbreak{}search\_\allowbreak{}space\_\allowbreak{}coverage/\allowbreak{}search\_\allowbreak{}space\_\allowbreak{}parallel\_\allowbreak{}coordinates.\allowbreak{}png \\
        plots/\allowbreak{}search\_\allowbreak{}space\_\allowbreak{}coverage/\allowbreak{}search\_\allowbreak{}space\_\allowbreak{}parallel\_\allowbreak{}coordinates.\allowbreak{}svg
    \end{tcolorbox}

    This is the best hyperdimensional overview. It shows all search-space dimensions:

    \begin{tcolorbox}[breakable, fontupper=\scriptsize\ttfamily, colback=black!3!white, colframe=black!12!white, boxrule=0.5pt]
        catalyst\_\allowbreak{}type \\
        oxidant\_\allowbreak{}type \\
        catalyst\_\allowbreak{}equiv \\
        TFAA\_\allowbreak{}equiv \\
        oxidant\_\allowbreak{}equiv \\
        light\_\allowbreak{}intensity \\
        residence\_\allowbreak{}time\_\allowbreak{}min
    \end{tcolorbox}

    Color encodes yield for valid BO results. Failed/unsubmitted attempts are shown as gray dashed lines.

    \medskip

    \paragraph{\textbf{2. Pairwise numeric projection matrix}}\mbox{}

    \begin{tcolorbox}[breakable, fontupper=\scriptsize\ttfamily, colback=black!3!white, colframe=black!12!white, boxrule=0.5pt]
        plots/\allowbreak{}search\_\allowbreak{}space\_\allowbreak{}coverage/\allowbreak{}search\_\allowbreak{}space\_\allowbreak{}pairwise\_\allowbreak{}numeric.\allowbreak{}png \\
        plots/\allowbreak{}search\_\allowbreak{}space\_\allowbreak{}coverage/\allowbreak{}search\_\allowbreak{}space\_\allowbreak{}pairwise\_\allowbreak{}numeric.\allowbreak{}svg
    \end{tcolorbox}

    This shows pairwise coverage among the numeric/discrete dimensions:

    \begin{tcolorbox}[breakable, fontupper=\scriptsize\ttfamily, colback=black!3!white, colframe=black!12!white, boxrule=0.5pt]
        catalyst\_\allowbreak{}equiv \\
        TFAA\_\allowbreak{}equiv \\
        oxidant\_\allowbreak{}equiv \\
        light\_\allowbreak{}intensity \\
        residence\_\allowbreak{}time\_\allowbreak{}min
    \end{tcolorbox}

    Valid points are colored by yield; failed/unsubmitted attempts are gray x markers.

    \medskip

    \paragraph{\textbf{3. Focused chemical map}}
    All catalyst/oxidant identities:

    \begin{tcolorbox}[breakable, fontupper=\scriptsize\ttfamily, colback=black!3!white, colframe=black!12!white, boxrule=0.5pt]
        plots/\allowbreak{}search\_\allowbreak{}space\_\allowbreak{}coverage/\allowbreak{}search\_\allowbreak{}space\_\allowbreak{}focused\_\allowbreak{}residence\_\allowbreak{}vs\_\allowbreak{}TFAA.\allowbreak{}png \\
        plots/\allowbreak{}search\_\allowbreak{}space\_\allowbreak{}coverage/\allowbreak{}search\_\allowbreak{}space\_\allowbreak{}focused\_\allowbreak{}residence\_\allowbreak{}vs\_\allowbreak{}TFAA.\allowbreak{}svg
    \end{tcolorbox}

    Filtered to the main productive region, \texttt{Ru bpy Cl / py NO}:

    \begin{tcolorbox}[breakable, fontupper=\scriptsize\ttfamily, colback=black!3!white, colframe=black!12!white, boxrule=0.5pt]
        plots/\allowbreak{}search\_\allowbreak{}space\_\allowbreak{}coverage/\allowbreak{}search\_\allowbreak{}space\_\allowbreak{}focused\_\allowbreak{}residence\_\allowbreak{}vs\_\allowbreak{}TFAA\_\allowbreak{}main\_\allowbreak{}region.\allowbreak{}png \\
        plots/\allowbreak{}search\_\allowbreak{}space\_\allowbreak{}coverage/\allowbreak{}search\_\allowbreak{}space\_\allowbreak{}focused\_\allowbreak{}residence\_\allowbreak{}vs\_\allowbreak{}TFAA\_\allowbreak{}main\_\allowbreak{}region.\allowbreak{}svg
    \end{tcolorbox}

    These plot:

    \begin{tcolorbox}[breakable, fontupper=\scriptsize\ttfamily, colback=black!3!white, colframe=black!12!white, boxrule=0.5pt]
        x = residence\_\allowbreak{}time\_\allowbreak{}min \\
        y = TFAA\_\allowbreak{}equiv \\
        color = yield\_\allowbreak{}percent \\
        marker shape = light\_\allowbreak{}intensity \\
        marker size = catalyst\_\allowbreak{}equiv
    \end{tcolorbox}

    This should be the most chemically interpretable view of how the search has moved through the productive region.

    \paragraph{\textbf{Data table written}}
    The combined plotting dataset was saved here:

    \begin{tcolorbox}[breakable, fontupper=\scriptsize\ttfamily, colback=black!3!white, colframe=black!12!white, boxrule=0.5pt]
        plots/\allowbreak{}search\_\allowbreak{}space\_\allowbreak{}coverage/\allowbreak{}search\_\allowbreak{}space\_\allowbreak{}coverage\_\allowbreak{}points.\allowbreak{}csv
    \end{tcolorbox}

    It contains both valid BO results and failed/unsubmitted tested attempts, with status/stage labels.

    \paragraph{\textbf{Reuse commands}}
    Regenerate all:

    \begin{tcolorbox}[breakable, fontupper=\scriptsize\ttfamily, colback=black!3!white, colframe=black!12!white, boxrule=0.5pt]
        uv run python scripts/\allowbreak{}plot\_\allowbreak{}search\_\allowbreak{}space\_\allowbreak{}coverage\_\allowbreak{}parallel.\allowbreak{}py \\
        uv run python scripts/\allowbreak{}plot\_\allowbreak{}search\_\allowbreak{}space\_\allowbreak{}pairwise.\allowbreak{}py \\
        uv run python scripts/\allowbreak{}plot\_\allowbreak{}search\_\allowbreak{}space\_\allowbreak{}focused\_\allowbreak{}map.\allowbreak{}py \\
        uv run python scripts/\allowbreak{}plot\_\allowbreak{}search\_\allowbreak{}space\_\allowbreak{}focused\_\allowbreak{}map.\allowbreak{}py --\allowbreak{}filter-\allowbreak{}main-\allowbreak{}region
    \end{tcolorbox}

    Use a specific export:

    \begin{tcolorbox}[breakable, fontupper=\scriptsize\ttfamily, colback=black!3!white, colframe=black!12!white, boxrule=0.5pt]
        uv run python scripts/\allowbreak{}plot\_\allowbreak{}search\_\allowbreak{}space\_\allowbreak{}coverage\_\allowbreak{}parallel.\allowbreak{}py \textbackslash{} \\
        ~~--\allowbreak{}input path/\allowbreak{}to/\allowbreak{}bo\_\allowbreak{}campaign\_\allowbreak{}export.\allowbreak{}csv
    \end{tcolorbox}

    Disable failed/unsubmitted attempts if you only want valid BO points:

    \begin{tcolorbox}[breakable, fontupper=\scriptsize\ttfamily, colback=black!3!white, colframe=black!12!white, boxrule=0.5pt]
        uv run python scripts/\allowbreak{}plot\_\allowbreak{}search\_\allowbreak{}space\_\allowbreak{}pairwise.\allowbreak{}py --\allowbreak{}no-\allowbreak{}include-\allowbreak{}failed-\allowbreak{}tested
    \end{tcolorbox}

}

\prompt{

    Regarding the error in measurement \#22: Take the observed error as another stochastic NMR failure (as long as the machine is still "running") and include this type of error also in the retry-policy of maximum three retries. Thus, continue the campaign by retrying \#22 (up to three retries in total) and then continue with the remaining measurements, applying the same new retry policy.

}

\graficoreslong{

    Started the patched yield-only continuation with the new retry policy for measurement \#22.

    Monitor:

    \begin{tcolorbox}[breakable, fontupper=\scriptsize\ttfamily, colback=black!3!white, colframe=black!12!white, boxrule=0.5pt]
        mon\_\allowbreak{}9 / roboflex\_\allowbreak{}yield\_\allowbreak{}only\_\allowbreak{}retry22\_\allowbreak{}continue
    \end{tcolorbox}

    \paragraph{\textbf{What changed}}
    The prior \#22 error:

    \begin{tcolorbox}[breakable, fontupper=\scriptsize\ttfamily, colback=black!3!white, colframe=black!12!white, boxrule=0.5pt]
        zero-\allowbreak{}size array to reduction operation minimum which has no identity
    \end{tcolorbox}

    is now treated as a retryable stochastic NMR/analysis failure \textbf{if} RoboFlex remains healthy:

    \begin{tcolorbox}[breakable, fontupper=\scriptsize\ttfamily, colback=black!3!white, colframe=black!12!white, boxrule=0.5pt]
        mode=\allowbreak{}hardware \\
        phase=\allowbreak{}running \\
        progress.\allowbreak{}state=\allowbreak{}awaiting\_\allowbreak{}run \\
        no queued/\allowbreak{}running/\allowbreak{}active runs \\
        expected RoboFlex campaign matches
    \end{tcolorbox}

    The workflow is pinned to the pending \#22 suggestion:

    \begin{tcolorbox}[breakable, fontupper=\scriptsize\ttfamily, colback=black!3!white, colframe=black!12!white, boxrule=0.5pt]
        suggestion\_\allowbreak{}id: 31d5114e-\allowbreak{}3796-\allowbreak{}4b21-\allowbreak{}9d5f-\allowbreak{}d7e9a31378b0 \\
        original sample: bo\_\allowbreak{}31d5114e-\allowbreak{}3 \\
        failed run: R0068
    \end{tcolorbox}

    Retry samples will be:

    \begin{tcolorbox}[breakable, fontupper=\scriptsize\ttfamily, colback=black!3!white, colframe=black!12!white, boxrule=0.5pt]
        bo\_\allowbreak{}31d5114e-\allowbreak{}3\_\allowbreak{}r2 \\
        bo\_\allowbreak{}31d5114e-\allowbreak{}3\_\allowbreak{}r3 \\
        bo\_\allowbreak{}31d5114e-\allowbreak{}3\_\allowbreak{}r4
    \end{tcolorbox}

    BO result submission still happens only if a retry passes with finite yield.

    If \#22 succeeds, the script will continue the yield-only campaign toward:

    \begin{tcolorbox}[breakable, fontupper=\scriptsize\ttfamily, colback=black!3!white, colframe=black!12!white, boxrule=0.5pt]
        28 total valid BO results
    \end{tcolorbox}

    i.e. 8 valid yield-only measurements beyond the original 20, with \#21 already complete.

}

\vspace{0.5em}\noindent\textit{\footnotesize [Autonomous monitored execution: 28 status reports followed; the most substantive one is reproduced here:]}

\graficoreslong{

    The yield-only continuation stopped safely after measurement \#27 failed before analysis.

    \paragraph{\textbf{Stop event}}\mbox{}

    \begin{tcolorbox}[breakable, fontupper=\scriptsize\ttfamily, colback=black!3!white, colframe=black!12!white, boxrule=0.5pt]
        Measurement: \#27 \\
        Sample: bo\_\allowbreak{}86b39c20-\allowbreak{}a \\
        RoboFlex run: R0074 \\
        Status: failed \\
        Reason: RoboFlex run failed before analysis
    \end{tcolorbox}

    This is different from the retryable NMR-analysis failures. The platform also entered:

    \begin{tcolorbox}[breakable, fontupper=\scriptsize\ttfamily, colback=black!3!white, colframe=black!12!white, boxrule=0.5pt]
        phase: stopping \\
        state: stopping
    \end{tcolorbox}

    before the run failure, so the script correctly treated this as \textbf{not safely retryable}.

    \paragraph{\textbf{Actions taken by the script}}
    \begin{itemize}
        \tightlist
        \item No BO result was submitted for R0074.
        \item Yield-only BO campaign was paused.
        \item The script exited with status \texttt{1}.
        \item Detailed artifacts were written.

    \end{itemize}
    \paragraph{\textbf{Current yield-only campaign progress}}
    Valid yield-only results completed after the 20 seeded historical rows:

    \begin{tcolorbox}[breakable, fontupper=\scriptsize\ttfamily, colback=black!3!white, colframe=black!12!white, boxrule=0.5pt]
        \#21: 58.\allowbreak{}81\% \\
        \#22: 54.\allowbreak{}50\%  (successful retry of prior analysis error) \\
        \#23: 0.\allowbreak{}00\% \\
        \#24: 0.\allowbreak{}00\% \\
        \#25: 45.\allowbreak{}79\% \\
        \#26: 0.\allowbreak{}00\%
    \end{tcolorbox}

    So current valid BO result count should be:

    \begin{tcolorbox}[breakable, fontupper=\scriptsize\ttfamily, colback=black!3!white, colframe=black!12!white, boxrule=0.5pt]
        20 seeded + 6 new valid = 26
    \end{tcolorbox}

    Failed / not submitted:

    \begin{tcolorbox}[breakable, fontupper=\scriptsize\ttfamily, colback=black!3!white, colframe=black!12!white, boxrule=0.5pt]
        \#27 / R0074 / bo\_\allowbreak{}86b39c20-\allowbreak{}a
    \end{tcolorbox}

    \paragraph{\textbf{Artifact directory}}\mbox{}

    \begin{tcolorbox}[breakable, fontupper=\scriptsize\ttfamily, colback=black!3!white, colframe=black!12!white, boxrule=0.5pt]
        artifacts/\allowbreak{}yield\_\allowbreak{}only\_\allowbreak{}robochemflex\_\allowbreak{}bo/\allowbreak{}continuation\_\allowbreak{}20260727T133832Z
    \end{tcolorbox}

    Log:

    \begin{tcolorbox}[breakable, fontupper=\scriptsize\ttfamily, colback=black!3!white, colframe=black!12!white, boxrule=0.5pt]
        logs/\allowbreak{}yield\_\allowbreak{}only\_\allowbreak{}retry22\_\allowbreak{}20260727T133831Z/\allowbreak{}stdout.\allowbreak{}log
    \end{tcolorbox}

    \paragraph{\textbf{Recommendation}}
    Since the platform entered \texttt{stopping}, I recommend \textbf{checking RoboFlex status before doing anything else}. This may indicate the active campaign is winding down or has been stopped by the platform/operator. We should not submit more runs until the robot state is confirmed.

}

\begin{center}\footnotesize\textit{[\ldots]}\end{center}

\prompt{

    The technician has prepared the robot for the continuation of the campaign, disregard experiments from R0068 onwards (when failures and questionable 0.00 yield results began -- the technician said we should discard experiments starting with R0068) and continue the yield only campaign.

}

\graficoreslong{

    Understood. I created a \textbf{clean yield-only BO campaign} that discards everything from \texttt{R0068} onward, then started continuation against the technician-prepared RoboFlex campaign.

    \paragraph{\textbf{Clean yield-only BO campaign}}
    New clean campaign:

    \begin{tcolorbox}[breakable, fontupper=\scriptsize\ttfamily, colback=black!3!white, colframe=black!12!white, boxrule=0.5pt]
        13bb9fef-\allowbreak{}7bd3-\allowbreak{}4db2-\allowbreak{}8e99-\allowbreak{}2e39e469dc92
    \end{tcolorbox}

    Seeded with:

    \begin{tcolorbox}[breakable, fontupper=\scriptsize\ttfamily, colback=black!3!white, colframe=black!12!white, boxrule=0.5pt]
        20 original valid historical results \\
        + R0067 / bo\_\allowbreak{}cc26e7f1-\allowbreak{}b / 58.\allowbreak{}81\% \\
        = 21 clean valid results
    \end{tcolorbox}

    Explicitly excluded from the clean campaign:

    \begin{tcolorbox}[breakable, fontupper=\scriptsize\ttfamily, colback=black!3!white, colframe=black!12!white, boxrule=0.5pt]
        R0068 \\
        R0069 \\
        R0070 \\
        R0071 \\
        R0072 \\
        R0073 \\
        R0074
    \end{tcolorbox}

    Create/seed artifacts:

    \begin{tcolorbox}[breakable, fontupper=\scriptsize\ttfamily, colback=black!3!white, colframe=black!12!white, boxrule=0.5pt]
        artifacts/\allowbreak{}yield\_\allowbreak{}only\_\allowbreak{}clean21\_\allowbreak{}recreation\_\allowbreak{}20260729T155858Z/
    \end{tcolorbox}

    \paragraph{\textbf{RoboFlex status}}
    The technician has prepared a new active RoboFlex campaign:

    \begin{tcolorbox}[breakable, fontupper=\scriptsize\ttfamily, colback=black!3!white, colframe=black!12!white, boxrule=0.5pt]
        roboflex\_\allowbreak{}yield\_\allowbreak{}BO\_\allowbreak{}fresh\_\allowbreak{}refilled\_\allowbreak{}stocks-\allowbreak{}20260728-\allowbreak{}112201
    \end{tcolorbox}

    Read-only preflight confirmed:

    \begin{tcolorbox}[breakable, fontupper=\scriptsize\ttfamily, colback=black!3!white, colframe=black!12!white, boxrule=0.5pt]
        mode: hardware \\
        phase: running \\
        state: awaiting\_\allowbreak{}run \\
        queue\_\allowbreak{}depth: 0 \\
        active\_\allowbreak{}run\_\allowbreak{}ids: []
    \end{tcolorbox}

    \paragraph{\textbf{Continuation started}}
    Monitor:

    \begin{tcolorbox}[breakable, fontupper=\scriptsize\ttfamily, colback=black!3!white, colframe=black!12!white, boxrule=0.5pt]
        mon\_\allowbreak{}10 / roboflex\_\allowbreak{}clean\_\allowbreak{}yield\_\allowbreak{}only\_\allowbreak{}7\_\allowbreak{}continue
    \end{tcolorbox}

    Configuration:

    \begin{tcolorbox}[breakable, fontupper=\scriptsize\ttfamily, colback=black!3!white, colframe=black!12!white, boxrule=0.5pt]
        objective: yield only \\
        starting valid results: 21 \\
        target valid results: 28 \\
        max new valid measurements: 7 \\
        max NMR/\allowbreak{}analysis retries per suggestion: 3 \\
        RoboFlex campaign: roboflex\_\allowbreak{}yield\_\allowbreak{}BO\_\allowbreak{}fresh\_\allowbreak{}refilled\_\allowbreak{}stocks-\allowbreak{}20260728-\allowbreak{}112201
    \end{tcolorbox}

    The script will continue from the clean campaign and ignore the discarded \texttt{R0068+} data.

}

\vspace{0.5em}\noindent\textit{\footnotesize [Autonomous monitored execution: 41 status reports followed; the most substantive one is reproduced here:]}

\graficoreslong{

    Third transient polling alert for \texttt{R0075}:

    \begin{tcolorbox}[breakable, fontupper=\scriptsize\ttfamily, colback=black!3!white, colframe=black!12!white, boxrule=0.5pt]
        RoboFlex run: R0075 \\
        Issue: repeated transient run-\allowbreak{}poll errors
    \end{tcolorbox}

    I also attempted a read-only RoboFlex \texttt{/\allowbreak{}v1/\allowbreak{}status} check to distinguish a run-specific polling issue from a broader API/connectivity issue, and it failed with:

    \begin{tcolorbox}[breakable, fontupper=\scriptsize\ttfamily, colback=black!3!white, colframe=black!12!white, boxrule=0.5pt]
        HTTP Error 530
    \end{tcolorbox}

    So this now looks like a \textbf{RoboFlex API/connectivity/server-side availability issue}, not necessarily a chemistry/run failure.

    The running monitor is still alive and will keep retrying according to its polling logic. I'll continue monitoring for either:

    \begin{itemize}
        \tightlist
        \item API recovery,
        \item run completion/failure,
        \item platform state change,
        \item or the script stopping safely.
    \end{itemize}

}

\prompt{

    Can you stop the polling?

}

\graficoreslong{

    Stopped the polling monitor:

    \begin{tcolorbox}[breakable, fontupper=\scriptsize\ttfamily, colback=black!3!white, colframe=black!12!white, boxrule=0.5pt]
        mon\_\allowbreak{}10 / roboflex\_\allowbreak{}clean\_\allowbreak{}yield\_\allowbreak{}only\_\allowbreak{}7\_\allowbreak{}continue
    \end{tcolorbox}

    This stops our local monitoring process. Since the RoboFlex/API state was unclear due to repeated polling errors, I did \textbf{not} submit anything further and did \textbf{not} mark \texttt{R0080} as completed or failed in BO-MCP.

}

\prompt{

    Can you log all the results to a file (export the campaign)? Also, generate a new BO improvement plot containing only the successful measurements.

}

\graficoreslong{

    Done.

    \paragraph{\textbf{Campaign export}}
    I exported the current clean yield-only BO campaign results to:

    \begin{tcolorbox}[breakable, fontupper=\scriptsize\ttfamily, colback=black!3!white, colframe=black!12!white, boxrule=0.5pt]
        artifacts/\allowbreak{}yield\_\allowbreak{}only\_\allowbreak{}clean21\_\allowbreak{}export\_\allowbreak{}20260729T212630Z/
    \end{tcolorbox}

    Files:

    \begin{tcolorbox}[breakable, fontupper=\scriptsize\ttfamily, colback=black!3!white, colframe=black!12!white, boxrule=0.5pt]
        bo\_\allowbreak{}campaign\_\allowbreak{}export.\allowbreak{}csv \\
        bo\_\allowbreak{}results.\allowbreak{}json \\
        bo\_\allowbreak{}campaign\_\allowbreak{}status.\allowbreak{}json \\
        bo\_\allowbreak{}campaign\_\allowbreak{}id.\allowbreak{}txt \\
        bo\_\allowbreak{}campaign\_\allowbreak{}export.\allowbreak{}content\_\allowbreak{}type.\allowbreak{}txt
    \end{tcolorbox}

    The export contains:

    \begin{tcolorbox}[breakable, fontupper=\scriptsize\ttfamily, colback=black!3!white, colframe=black!12!white, boxrule=0.5pt]
        26 successful/\allowbreak{}submitted BO results
    \end{tcolorbox}

    Campaign ID:

    \begin{tcolorbox}[breakable, fontupper=\scriptsize\ttfamily, colback=black!3!white, colframe=black!12!white, boxrule=0.5pt]
        13bb9fef-\allowbreak{}7bd3-\allowbreak{}4db2-\allowbreak{}8e99-\allowbreak{}2e39e469dc92
    \end{tcolorbox}

    Note: \texttt{R0080} is not included because it was still unresolved/not submitted when we stopped polling.

    \paragraph{\textbf{New successful-measurements-only improvement plot}}
    Generated:

    \begin{tcolorbox}[breakable, fontupper=\scriptsize\ttfamily, colback=black!3!white, colframe=black!12!white, boxrule=0.5pt]
        plots/\allowbreak{}yield\_\allowbreak{}only\_\allowbreak{}clean/\allowbreak{}yield\_\allowbreak{}only\_\allowbreak{}successful\_\allowbreak{}improvement\_\allowbreak{}curve.\allowbreak{}png \\
        plots/\allowbreak{}yield\_\allowbreak{}only\_\allowbreak{}clean/\allowbreak{}yield\_\allowbreak{}only\_\allowbreak{}successful\_\allowbreak{}improvement\_\allowbreak{}curve.\allowbreak{}svg \\
        plots/\allowbreak{}yield\_\allowbreak{}only\_\allowbreak{}clean/\allowbreak{}yield\_\allowbreak{}only\_\allowbreak{}successful\_\allowbreak{}improvement\_\allowbreak{}curve\_\allowbreak{}data.\allowbreak{}csv
    \end{tcolorbox}

    Reusable script:

    \begin{tcolorbox}[breakable, fontupper=\scriptsize\ttfamily, colback=black!3!white, colframe=black!12!white, boxrule=0.5pt]
        scripts/\allowbreak{}plot\_\allowbreak{}yield\_\allowbreak{}only\_\allowbreak{}successful\_\allowbreak{}improvement.\allowbreak{}py
    \end{tcolorbox}

    Regenerate with:

    \begin{tcolorbox}[breakable, fontupper=\scriptsize\ttfamily, colback=black!3!white, colframe=black!12!white, boxrule=0.5pt]
        uv run python scripts/\allowbreak{}plot\_\allowbreak{}yield\_\allowbreak{}only\_\allowbreak{}successful\_\allowbreak{}improvement.\allowbreak{}py \textbackslash{} \\
        ~~--\allowbreak{}input artifacts/\allowbreak{}yield\_\allowbreak{}only\_\allowbreak{}clean21\_\allowbreak{}export\_\allowbreak{}20260729T212630Z/\allowbreak{}bo\_\allowbreak{}campaign\_\allowbreak{}export.\allowbreak{}csv
    \end{tcolorbox}

    The plot includes only successful/submitted BO measurements and visually separates:

    \begin{tcolorbox}[breakable, fontupper=\scriptsize\ttfamily, colback=black!3!white, colframe=black!12!white, boxrule=0.5pt]
        experiments 1--\allowbreak{}6: seed \\
        experiments 7--\allowbreak{}20: previous mixed-\allowbreak{}objective BO data retained as clean history \\
        experiments 21+: clean yield-\allowbreak{}only BO continuation
    \end{tcolorbox}

}

\begin{center}\footnotesize\textit{[\ldots]}\end{center}

\subsubsection{Executed experiments}
\label{si:robochemflex:trace:runs}

\Cref{tab:robochemflex-runs} lists the 23 experiments that make up the final campaign record, in the order in which they appear in \cref{fig:robochemflex}, with the conditions as executed by the platform and the material each of them consumed. Seven further runs were submitted to the platform during the campaign but do not enter that record: three returned no usable \ce{^{19}F} NMR analysis, three were superseded when the yield-only campaign was reseeded, and one was submitted while the platform became unreachable and never returned a result.

\begin{table}[htbp]
    \centering
    \caption{The 23 experiments of the RoboChem-Flex campaign shown in \cref{fig:robochemflex}, in figure order, with the conditions as executed by the platform and the material consumed.
        Every experiment used one 650~\textmu{}L slug at 100~mM substrate, i.e.\ 65~\textmu{}mol (13.60~mg) of limiting reagent; all other amounts follow from the equivalences.
        Experiments 1--21 were carried into the reseeded yield-only campaign as history, 22--23 were measured in it.
        Photocatalyst loadings are given in units of $10^{-3}$~equiv.}
    \label{tab:robochemflex-runs}
    \scriptsize\setlength{\tabcolsep}{3.5pt}
    \begin{tabular}{rlllrrrrrrrr}
        \toprule
        \# & Run            & Photocatalyst                                                    & Oxidant              & \multicolumn{1}{c}{cat.}          & \multicolumn{1}{c}{TFAA} & \multicolumn{1}{c}{ox.} & \multicolumn{1}{c}{light} & \multicolumn{1}{c}{$t_R$} & \multicolumn{1}{c}{yield} & \multicolumn{1}{c}{$G$} & \multicolumn{1}{c}{cat.}       \\
           &                &                                                                  &                      & \multicolumn{1}{c}{$10^{-3}$\,eq} & \multicolumn{1}{c}{eq}   & \multicolumn{1}{c}{eq}  & \multicolumn{1}{c}{\%}    & \multicolumn{1}{c}{min}   & \multicolumn{1}{c}{\%}    &                         & \multicolumn{1}{c}{\textmu{}g} \\
        \midrule
        1  & \texttt{R0044} & 4CzIPN                                                           & py\,\textit{N}O      & 2.50                              & 2.200                    & 1.800                   & 50                        & 30.00                     & 0.00                      & 60.31                   & 124                            \\
        2  & \texttt{R0045} & [Ru(bpy)\textsubscript{3}]Cl\textsubscript{2}                    & py\,\textit{N}O      & 1.00                              & 0.900                    & 0.900                   & 75                        & 10.00                     & 30.04                     & 98.30                   & 49                             \\
        3  & \texttt{R0046} & Ir(CF\textsubscript{3}ppy)\textsubscript{3}                      & 4-Ph-py\,\textit{N}O & 4.00                              & 3.500                    & 3.000                   & 100                       & 75.00                     & 15.77                     & 4.26                    & 292                            \\
        4  & \texttt{R0047} & Ir(ppy)\textsubscript{3}                                         & py\,\textit{N}O      & 2.00                              & 1.400                    & 2.400                   & 25                        & 45.00                     & 11.93                     & 65.95                   & 85                             \\
        5  & \texttt{R0048} & [Ru(bpy)\textsubscript{3}](PF\textsubscript{6})\textsubscript{2} & 4-Ph-py\,\textit{N}O & 3.50                              & 3.000                    & 1.200                   & 75                        & 20.00                     & 13.14                     & 51.57                   & 196                            \\
        6  & \texttt{R0049} & 4CzIPN                                                           & 4-Ph-py\,\textit{N}O & 1.50                              & 1.700                    & 2.500                   & 50                        & 90.00                     & 7.57                      & 56.59                   & 74                             \\
        7  & \texttt{R0050} & [Ru(bpy)\textsubscript{3}]Cl\textsubscript{2}                    & py\,\textit{N}O      & 1.00                              & 0.900                    & 0.900                   & 100                       & 2.00                      & 15.23                     & 100.00                  & 49                             \\
        8  & \texttt{R0051} & [Ru(bpy)\textsubscript{3}]Cl\textsubscript{2}                    & py\,\textit{N}O      & 1.00                              & 0.900                    & 0.900                   & 75                        & 32.15                     & 27.64                     & 93.58                   & 49                             \\
        9  & \texttt{R0052} & [Ru(bpy)\textsubscript{3}]Cl\textsubscript{2}                    & py\,\textit{N}O      & 1.00                              & 0.900                    & 0.900                   & 50                        & 2.00                      & 11.56                     & 100.00                  & 49                             \\
        10 & \texttt{R0053} & [Ru(bpy)\textsubscript{3}]Cl\textsubscript{2}                    & py\,\textit{N}O      & 1.00                              & 0.900                    & 1.348                   & 75                        & 2.00                      & 14.70                     & 94.66                   & 49                             \\
        11 & \texttt{R0054} & [Ru(bpy)\textsubscript{3}]Cl\textsubscript{2}                    & py\,\textit{N}O      & 1.00                              & 1.388                    & 0.900                   & 75                        & 2.00                      & 15.82                     & 95.31                   & 49                             \\
        12 & \texttt{R0055} & [Ru(bpy)\textsubscript{3}]Cl\textsubscript{2}                    & py\,\textit{N}O      & 1.00                              & 0.939                    & 0.900                   & 75                        & 15.38                     & 32.96                     & 96.77                   & 49                             \\
        13 & \texttt{R0056} & [Ru(bpy)\textsubscript{3}]Cl\textsubscript{2}                    & py\,\textit{N}O      & 1.00                              & 0.900                    & 0.900                   & 100                       & 17.15                     & 27.20                     & 95.70                   & 49                             \\
        14 & \texttt{R0057} & [Ru(bpy)\textsubscript{3}]Cl\textsubscript{2}                    & py\,\textit{N}O      & 1.00                              & 0.900                    & 0.900                   & 50                        & 20.79                     & 32.68                     & 97.33                   & 49                             \\
        15 & \texttt{R0058} & [Ru(bpy)\textsubscript{3}]Cl\textsubscript{2}                    & py\,\textit{N}O      & 1.00                              & 1.578                    & 0.900                   & 50                        & 17.89                     & 35.78                     & 91.22                   & 49                             \\
        16 & \texttt{R0059} & [Ru(bpy)\textsubscript{3}]Cl\textsubscript{2}                    & py\,\textit{N}O      & 1.59                              & 1.589                    & 0.900                   & 50                        & 16.40                     & 36.55                     & 86.42                   & 77                             \\
        17 & \texttt{R0061} & [Ru(bpy)\textsubscript{3}]Cl\textsubscript{2}                    & py\,\textit{N}O      & 1.26                              & 1.616                    & 1.441                   & 50                        & 16.43                     & 48.51                     & 82.50                   & 61                             \\
        18 & \texttt{R0062} & [Ru(bpy)\textsubscript{3}]Cl\textsubscript{2}                    & py\,\textit{N}O      & 1.03                              & 1.577                    & 1.750                   & 50                        & 17.01                     & 20.18                     & 80.96                   & 50                             \\
        19 & \texttt{R0065} & [Ru(bpy)\textsubscript{3}]Cl\textsubscript{2}                    & py\,\textit{N}O      & 1.16                              & 1.556                    & 1.300                   & 50                        & 16.80                     & 45.09                     & 85.53                   & 56                             \\
        20 & \texttt{R0066} & [Ru(bpy)\textsubscript{3}]Cl\textsubscript{2}                    & py\,\textit{N}O      & 1.21                              & 1.404                    & 1.401                   & 25                        & 19.39                     & 0.00                      & 86.16                   & 59                             \\
        21 & \texttt{R0067} & [Ru(bpy)\textsubscript{3}]Cl\textsubscript{2}                    & py\,\textit{N}O      & 1.89                              & 1.997                    & 1.392                   & 50                        & 14.17                     & \textbf{58.81}            & 74.43                   & 92                             \\
        22 & \texttt{R0075} & [Ru(bpy)\textsubscript{3}]Cl\textsubscript{2}                    & py\,\textit{N}O      & 2.22                              & 2.314                    & 1.362                   & 50                        & 18.68                     & 52.28                     & 68.41                   & 108                            \\
        23 & \texttt{R0076} & [Ru(bpy)\textsubscript{3}]Cl\textsubscript{2}                    & py\,\textit{N}O      & 2.21                              & 1.690                    & 1.403                   & 50                        & 11.69                     & 31.80                     & 74.98                   & 107                            \\
        \midrule
        \multicolumn{12}{l}{Total consumed: 313~mg substrate, 1.87~mg photocatalyst, 488~mg TFAA, 227~mg oxidant}                                                                                                                                                                                                                                             \\
        \bottomrule
    \end{tabular}

    \vspace{0.4em}
    {\footnotesize The 23 experiments together occupied 18.2~h of reactor and \ce{^{19}F}~NMR time.
        $G$ is the agent-defined green score described in \cref{si:robochemflex:trace:s1}.
        The green score of experiment~9 (\texttt{R0052}) is recomputed from that definition, its original submission having been lost while BO-MCP was unavailable.}
\end{table}

\subsection{Cost accounting of the agent-directed campaign}
\label{si:robochemflex:cost}

\subsubsection{What is counted, and how}

We compare two optimization campaigns performed on the same transformation and
the same hardware.
\textbf{Campaign~A} is a human-directed BO campaign: the
search space, initialization design and campaign termination were chosen by an
operator, who supervised the platform throughout.
\textbf{Campaign~B} is the
agent-directed campaign described in the main text, in which \optima{}
constructed, executed and monitored the campaign; the campaign was stopped by
the operators after 23 experiments.

Three quantities are reported: the cost of the complete campaign, the cost of an
individual experiment, and the mass efficiency of the optimum conditions and of
the search as a whole.
Every experiment is charged its full reagent and solvent
load from its recorded conditions, with no recovery, recycling or bulk discount;
reagent costs use research-scale list prices.
Operator time is excluded because
it was not measured for either campaign.
The assumptions underlying the
analysis are collected in \cref{tab:si:costassump}.

Two accounting boundaries require explicit definition because they are
deliberate choices that influence the resulting metrics.

\emph{Carrier and cleaning solvent are excluded from the mass accounting.}
Each
experiment consumes 25~mL of acetonitrile outside the reaction itself: 10~mL as
carrier solvent to transport and separate the 0.65~mL reaction slug, and 15~mL
to clean the fluidic lines between experiments.
Neither contributes to the
reaction.
Both are requirements of operating the segmented-flow platform in
serial-screening mode rather than of the chemistry itself, and neither would be
required in the same form in a scaled process.
Together they account for
approximately 97\% of the total mass consumed by an experiment; including them
would therefore predominantly report the material requirements of the delivery
system rather than those of the reaction.
They are retained in the cost
accounting, where they represent a real expense, and are reported separately in
\Cref{tab:si:pmi-campaign} for completeness.

\emph{No scale-up was performed and no purification is accounted for.}
All
figures describe the reaction as executed at the 0.65~mL slug scale, with yields
determined in-line.
Work-up, quench, extraction and chromatographic purification
are excluded.
The reported mass intensities therefore represent a lower bound on
those of a complete process and are not directly comparable with literature PMI
values calculated from isolated, purified product.
They nevertheless provide a
consistent comparison between the two campaigns, which share the same accounting
boundary.

\begin{table}[htbp]
    \centering
    \caption{Assumptions underlying the cost and mass model.}
    \label{tab:si:costassump}
    \begin{tabular}{lll}
        \toprule
        Quantity                               & Value       & Basis                         \\
        \midrule
        Reaction slug volume                   & 0.65~mL     & per experiment                \\
        Carrier solvent                        & 10~mL       & per experiment; cost only     \\
        Cleaning solvent                       & 15~mL       & per experiment; cost only     \\
        Reaction solvent (MeCN)                & €196/L      & delivered                     \\
        Carrier/cleaning solvent (MeCN)        & €120/L      & delivered                     \\
        Solvent density (MeCN, 20~$^{\circ}$C) & 0.786~g/mL  & mass basis for PMI            \\
        Product molecular weight               & 261~g/mol   & ---                           \\
        Reagent recovery, bulk discount        & none        & conservative assumption       \\
        LED module draw at 100\%               & 100~W       & electrical, not photon output \\
        Platform base load                     & 500~W       & pumps, analytics, chiller, PC \\
        Electricity price                      & €0.2046/kWh & see \cref{si:cost:energy}     \\
        Operator time                          & excluded    & not measured                  \\
        Work-up and purification               & excluded    & not performed                 \\
        \bottomrule
    \end{tabular}
\end{table}

\subsubsection{Cost of the campaign and of an experiment}

\begin{table}[htbp]
    \centering
    \caption{Cost of each campaign and of an individual experiment within it.
        Inference is converted at €0.92/\$.}
    \label{tab:si:search}
    \begin{tabular}{lrrl}
        \toprule
                                          & Campaign~A       & Campaign~B       & Unit  \\
                                          & (human-directed) & (agent-directed) &       \\
        \midrule
        Experiments executed              & 50               & 23               & runs  \\
        Experiments with non-zero yield   & 23               & 21               & runs  \\
        Starting material consumed        & 3.80             & 1.50             & mmol  \\
        Best yield found                  & 70.9             & 58.8             & \%    \\
        \midrule
        Reagents                          & 73.19            & 26.07            & €     \\
        Solvent (reaction + auxiliary)    & 156.37           & 71.93            & €     \\
        \textbf{Consumables subtotal}     & \textbf{229.56}  & \textbf{98.00}   & €     \\
        Electricity                       & 2.96             & 3.79             & €     \\
        LLM inference                     & ---              & 100.79           & €     \\
        \textbf{Total campaign cost}      & \textbf{232.51}  & \textbf{202.58}  & €     \\
        \midrule
        Cost per experiment, consumables  & 4.59             & 4.26             & €/run \\
        Cost per experiment, fully loaded & 4.65             & 8.81             & €/run \\
        \bottomrule
    \end{tabular}
\end{table}

On consumables, the agent-directed campaign cost 43\% of the human-directed
campaign, while using 46\% of the experiments and 39\% of the starting
material.
Twenty-one of its 23 experiments returned non-zero yield, compared
with 23 of 50 in Campaign~A. The consumable cost per experiment was similar,
€4.26 against €4.59, showing that the saving derives primarily from the smaller
number of experiments rather than from cheaper individual runs.
Inference forms
the largest additional cost of the agent-directed campaign: at \$109.55, model
calls cost more than the chemistry they directed.
Of the campaign's 114.5M input
tokens, 105.1M were served from cache.
Including inference raises the
fully loaded cost to €8.81 per experiment, but Campaign~B remains 13\% cheaper
overall.
We therefore report the fully loaded campaign cost alongside the
consumables-only figure.

The campaigns were not run to the same length, with Campaign~A receiving
approximately 2.2 times as many experiments.
At the agent's consumable spend of
€98.00, Campaign~A had completed 18 experiments and reached a best yield of
63.9\%, compared with 58.8\% for Campaign~B after 23 experiments.
Truncated
instead at 23 experiments, Campaign~A had spent €115.73 and had reached the same
63.9\% yield.
Its best result within those first 23 experiments occurred at run
10 and originated from the initialization design rather than from an optimizer
proposal; the campaign did not improve upon it until run 29.
We therefore do not
claim that either strategy searches more effectively than the other: a single
pair of campaigns cannot support such a conclusion.

\subsubsection{Mass efficiency}

Mass efficiency is reported at two boundaries because the resulting metrics
answer different questions and rank the optimum conditions differently.
Definitions and boundary conventions follow the CHEM21 metrics
toolkit~\cite{greenmetrics}, which sets out PMI, RME and atom economy and
structures their application according to the stage of research.
The figures
reported here correspond to its early-stage treatment, in which solvent recovery
and downstream processing are not yet assessed.

\emph{Process mass intensity} (PMI) is defined as the total mass of material
charged to the reaction divided by the mass of product formed:
\begin{equation}
    \mathrm{PMI}
    =
    \frac{m_{\mathrm{reactants}} + m_{\mathrm{reaction\,solvent}}}
    {m_{\mathrm{product}}}.
    \label{eq:si:pmi}
\end{equation}
PMI is the conventional high-level mass metric and generally includes reaction
solvent~\cite{greenmetrics}.
Carrier and cleaning solvent are excluded from
\Cref{eq:si:pmi} according to the boundary defined above.
The limitation of PMI
in the present optimum-level comparison is that all three reported conditions
use a substrate concentration of 100~mM and therefore charge an identical
0.511~g of reaction solvent, accounting for 87--92\% of the input mass.
Consequently, solvent-inclusive PMI is dominated by yield and contains
comparatively little information about differences in reagent stoichiometry.

\emph{Reaction mass efficiency} (RME) is calculated here as the mass of product
formed divided by the total mass of substrate, photocatalyst and stoichiometric
reagents charged, excluding solvent:
\begin{equation}
    \mathrm{RME}
    =
    100
    \frac{m_{\mathrm{product}}}
    {m_{\mathrm{reactants}}}.
    \label{eq:si:rme}
\end{equation}
The corresponding solvent-free PMI is
\begin{equation}
    \mathrm{PMI}_{\mathrm{solvent\text{-}free}}
    =
    \frac{m_{\mathrm{reactants}}}
    {m_{\mathrm{product}}}
    =
    \frac{100}{\mathrm{RME}},
    \label{eq:si:pmi-solventfree}
\end{equation}
where RME is expressed in percent in the final equality.
Removing the dominant
solvent contribution exposes the trade-off between conversion and reagent
stoichiometry.
Campaign~B was fixed at 100~mM throughout, whereas Campaign~A
explored 100--200~mM but reached its optimum at 100~mM.
The solvent-free metrics
therefore provide a direct view of the mass-economy dimension explicitly
included in the agent's multi-objective optimization.

For completeness, we additionally report a mole-based material efficiency,
defined as
\begin{equation}
    \eta_{\mathrm{mol}}
    =
    100
    \frac{n_{\mathrm{product}}}
    {\sum_i n_i},
    \label{eq:si:molar-efficiency}
\end{equation}
where the denominator is the total amount, in moles, of substrate,
photocatalyst and stoichiometric reagents charged.
This quantity is useful as a
molecular-weight-independent view of material use, but is not treated as a
headline metric because it assigns equal weight to one mole of a light reagent
and one mole of a heavy reagent.
RME has the clearer precedent in the
green-chemistry literature.

Neither mass boundary is sufficient on its own.
We therefore report both.
\Cref{tab:si:pmi-opt} compares the individual optimum conditions, while
\Cref{tab:si:pmi-campaign} applies the same accounting to all material consumed
over each search.
The latter quantities describe the material efficiency of the
optimization campaign rather than that of an individual reaction condition.

\begin{table}[tbp]
    \centering
    \caption{Mass efficiency of the optimum conditions, at experiment scale.
        Campaign~B is given at both candidate optima: run 17, returned under the
        multi-objective phase, and run 21, the highest yield reached after the
        yield-only refocus.
        Lower PMI is better; higher RME is better.}
    \label{tab:si:pmi-opt}
    \begin{tabular}{lrrrl}
        \toprule
                                    & Campaign~A    & \multicolumn{2}{c}{Campaign~B} & Unit                   \\
        \cmidrule(lr){3-4}
                                    & run 29        & run 17                         & run 21        &        \\
        \midrule
        Yield                       & 70.9          & 48.5                           & 58.8          & \%     \\
        TFAA loading                & 3.5           & 1.6                            & 2.0           & equiv. \\
        N-oxide loading             & 2.4           & 1.4                            & 1.4           & equiv. \\
        Total charged               & 6.9           & 4.1                            & 4.4           & equiv. \\
        \addlinespace
        Substrate                   & 0.0136        & 0.0136                         & 0.0136        & g      \\
        Photocatalyst               & 0.0001        & 0.0001                         & 0.0001        & g      \\
        TFAA                        & 0.0500        & 0.0231                         & 0.0286        & g      \\
        N-oxide                     & 0.0149        & 0.0089                         & 0.0086        & g      \\
        Reactant mass               & 0.0786        & 0.0456                         & 0.0508        & g      \\
        Reaction solvent            & 0.5109        & 0.5109                         & 0.5109        & g      \\
        Product formed              & 0.01204       & 0.00823                        & 0.00998       & g      \\
        \addlinespace
        \textbf{PMI, incl. solvent} & \textbf{49.0} & \textbf{67.6}                  & \textbf{56.3} & kg/kg  \\
        \textbf{RME}                & \textbf{15.3} & \textbf{18.0}                  & \textbf{19.6} & \%     \\
        \textbf{PMI, solvent-free}  & \textbf{6.53} & \textbf{5.55}                  & \textbf{5.09} & kg/kg  \\
        Molar efficiency            & 10.3          & 12.0                           & 13.4          & \%     \\
        Solvent share of input mass & 86.7          & 91.8                           & 91.0          & \%     \\
        \addlinespace
        Cost of that experiment     & 4.16          & 4.10                           & 4.11          & €      \\
        Cost per mmol product       & 90.3          & 130.0                          & 107.6         & €/mmol \\
        \bottomrule
    \end{tabular}
\end{table}

\begin{table}[tbp]
    \centering
    \caption{Mass efficiency of each campaign as a whole.
        Every experiment,
        successful or not, is charged against the product formed during the search;
        these figures therefore describe the search rather than the chemistry.
        Carrier
        and cleaning solvent are excluded from the metrics above the final rule and
        given below it for reference.}
    \label{tab:si:pmi-campaign}
    \begin{tabular}{lrrl}
        \toprule
                                     & Campaign~A    & Campaign~B    & Unit  \\
        \midrule
        Reactant mass charged        & 4.18          & 1.05          & g     \\
        Reaction solvent charged     & 25.55         & 11.75         & g     \\
        Product formed               & 0.1505        & 0.0993        & g     \\
        Input mass per experiment    & 0.594         & 0.557         & g/run \\
        \addlinespace
        \textbf{PMI, incl. solvent}  & \textbf{197}  & \textbf{129}  & kg/kg \\
        \textbf{RME}                 & \textbf{3.60} & \textbf{9.44} & \%    \\
        \textbf{PMI, solvent-free}   & \textbf{27.8} & \textbf{10.6} & kg/kg \\
        Molar efficiency             & 2.49          & 6.50          & \%    \\
        \midrule
        Non-reaction solvent charged & 982.5         & 452.0         & g     \\
        PMI including it             & 6725          & 4681          & kg/kg \\
        \bottomrule
    \end{tabular}
\end{table}

Two sets of conditions are reported for Campaign~B. Run 17 is the optimum
returned during the multi-objective phase, in which reagent economy was
considered alongside yield; run 21 is the highest yield reached after the
campaign was refocused on yield alone.
Reporting only one would incompletely
describe the campaign: the former represents the conditions selected by its
stated multi-objective function, whereas the latter represents the highest yield
subsequently achieved on the platform.

At the solvent-inclusive boundary, the human-directed optimum has the lowest
PMI, 49.0 compared with 56.3 and 67.6 for the two agent-derived conditions.
Because the reaction solvent mass is identical and dominates the total mass
input at all three optima, this ordering is primarily determined by yield.

At the solvent-free boundary, the ordering reverses and both agent-derived
conditions are more mass-efficient.
The agent converged on leaner stoichiometry:
1.6--2.0 equivalents of TFAA compared with 3.5, and 1.4 equivalents of N-oxide
compared with 2.4, corresponding to total reagent loadings of 4.1 and 4.4
equivalents against 6.9.
This reduction in reagent mass outweighs the lower
conversion.
RME increases from 15.3\% for the human-directed optimum to 18.0\%
at run 17 and 19.6\% at run 21, while solvent-free PMI decreases from 6.53 to
5.55 and 5.09, respectively.
Molar efficiency gives the same ordering, at
10.3\%, 12.0\% and 13.4\%.
The agent therefore identified conditions requiring
less reagent mass per unit of product formed despite their lower yield.

The effects are amplified at campaign level.
Campaign~B reaches an RME of
9.44\%, compared with 3.60\% for Campaign~A, and a solvent-free PMI of 10.6
compared with 27.8.
These values reflect both the leaner conditions explored and
the smaller fraction of experiments that produced no detectable product:
Campaign~A contained 27 zero-yield experiments compared with two in Campaign~B.
These campaign-level quantities describe the material efficiency of the search,
rather than that of an individual reaction condition, and are reported as such.

None of the three optimum conditions is mass-efficient in absolute terms.
RME
remains below 20\% and solvent-inclusive PMI at or above approximately 50, while
dilution --- a major contributor to PMI --- was not an optimization variable in
Campaign~B and was only narrowly varied in Campaign~A. The purpose of this
comparison is therefore to compare resource use during two optimization
campaigns on the same chemistry, rather than to claim that either outcome
constitutes a green process.

\subsubsection{Electricity}
\label{si:cost:energy}

Electricity is charged at €0.2046/kWh, the Eurostat figure for non-household
consumers in the Netherlands in the annual-consumption band 500--2000~MWh for
the first half of 2025, excluding VAT and other recoverable taxes and levies
(Eurostat online data code \texttt{nrg\_pc\_205})~\cite{eurostat_nrg_pc_205}.
The EU-wide non-household average over the same period was €0.1902/kWh, so the
result is insensitive to the precise tariff used~\cite{eurostat_elecprices}.

The LED module is charged according to its rated electrical draw, scaled by the
recorded intensity set-point, over a lamp-on window taken as three times the
residence time to account for slug transit and purge.
The platform base load
covers the pumps, in-line analytics, chiller and control computer and is charged
for every hour for which the platform is occupied.
Campaign~A occupancy is
reconstructed from acquisition timestamps, giving a median cycle time of
32.8~min across 50 experiments, whereas Campaign~B occupancy of 34~h is measured
directly from the agent trace.
Electricity contributes less than 2\% of the
fully loaded cost of either campaign at the adopted tariff and is marginally
higher for Campaign~B because the platform was occupied for longer.
It is
included to make the accounting boundary explicit rather than because it
materially affects the comparison.

\subsubsection{Limitations}

This comparison comprises two campaigns rather than a controlled benchmark.
They were performed on the same platform at different times but were not matched
in search space, initialization design or campaign length, and the
human-directed campaign was not blinded to prior knowledge of the chemistry.
Platform occupancy was higher for Campaign~B, at approximately 89~min per
experiment compared with 33~min for Campaign~A. These values are not strictly
equivalent measurements: Campaign~A occupancy is reconstructed from acquisition
timestamps, whereas Campaign~B includes agent-driven setup and verification
captured in the execution trace.

The campaigns also converged on different reaction conditions.
Campaign~A
selected Ru(bpy)$_3$(PF$_6$)$_2$ at full lamp intensity and a two-minute
residence time, whereas Campaign~B selected
Ru(bpy)$_3$Cl$_2\cdot$6H$_2$O at half intensity and substantially longer
residence times.
Their per-experiment costs and mass intensities therefore
describe the conditions reached by each search and should not be interpreted as
interchangeable measurements of a single process.
Yields are in-line
determinations on unpurified reaction mixtures, so all PMI values exclude
isolation losses and would increase on an isolated-product basis.
Reagent costs
use research-scale list prices, while inference pricing corresponds to one model
at one point in time and is the most rapidly changing component of the cost
analysis.
The figures therefore provide a resource accounting of the two
optimization campaigns rather than a controlled benchmark of autonomous against
human-directed BO.

\end{document}